\documentclass[journal]{IEEEtran}
\usepackage[svgnames]{xcolor}
\usepackage{graphicx}

\usepackage{times}                     % we use Times as the main font
\usepackage{tabu}                      % only used for the table example
\usepackage{booktabs}                  % only used for the table example
\usepackage{lipsum}                    % used to generate placeholder text
\usepackage{mwe}                       % used to generate placeholder figures
\usepackage[normalem]{ulem}
\usepackage{pgf}
\usepackage{tikz}
\usepackage{subcaption}
\usepackage{mathptmx}                  % use matching math font
\usepackage{amsfonts}
\usepackage{amsmath}                   % for align environment and advanced math
\usepackage{amsthm}                    % for theorem and proof environments
\usepackage{xr-hyper}
\usepackage{algorithm}
\usepackage{algpseudocode}
\usepackage{setspace}
\usepackage[hidelinks]{hyperref}
\usepackage{cleveref}

\graphicspath{{figures/}{pictures/}{images/}{./}} % where to search for the images

\DeclareGraphicsExtensions{.pdf,.png}

\Crefname{section}{sec.}{sec.}

\newtheorem{theorem}{Theorem}
\newtheorem{lemma}{Lemma}
\newtheorem{proposition}{Proposition}
\newtheorem{corollary}{Corollary}

\usepackage{amssymb}

\definecolor{ClusterFlowColor}{HTML}{1565C0}   % strong blue
\definecolor{HMDColor}{HTML}{E65100}      % dark orange
\definecolor{BlobColor}{HTML}{2E7D32}     % green (bubble chart)

\newcommand{\vcflow}[1]{%
  \texorpdfstring{{\color{ClusterFlowColor}\raisebox{-0.1ex}{\includegraphics[height=0.75em]{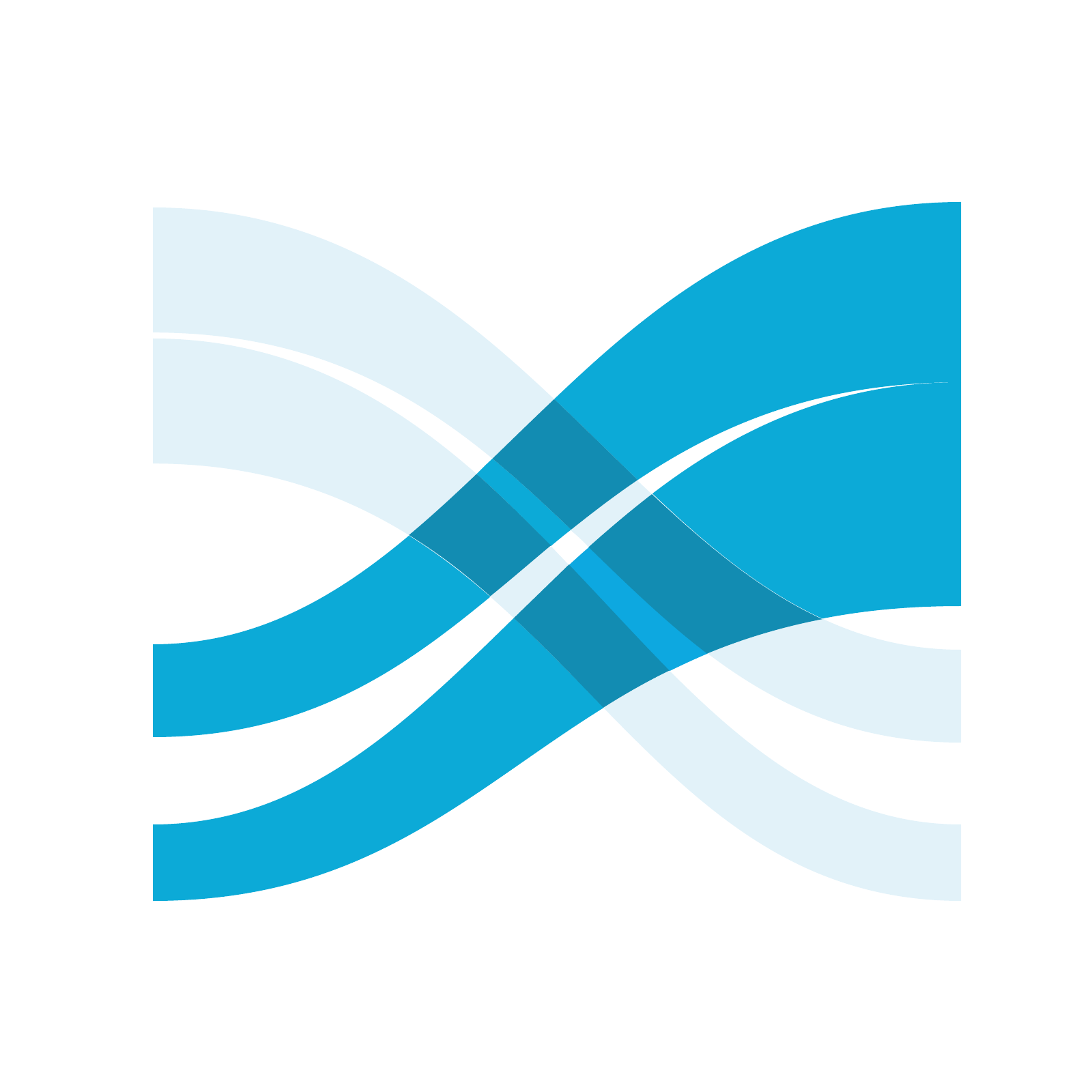}}\,#1}}{#1}}
\newcommand{\vhmd}[1]{%
  \texorpdfstring{{\color{HMDColor}\raisebox{-0.1ex}{\includegraphics[height=0.75em]{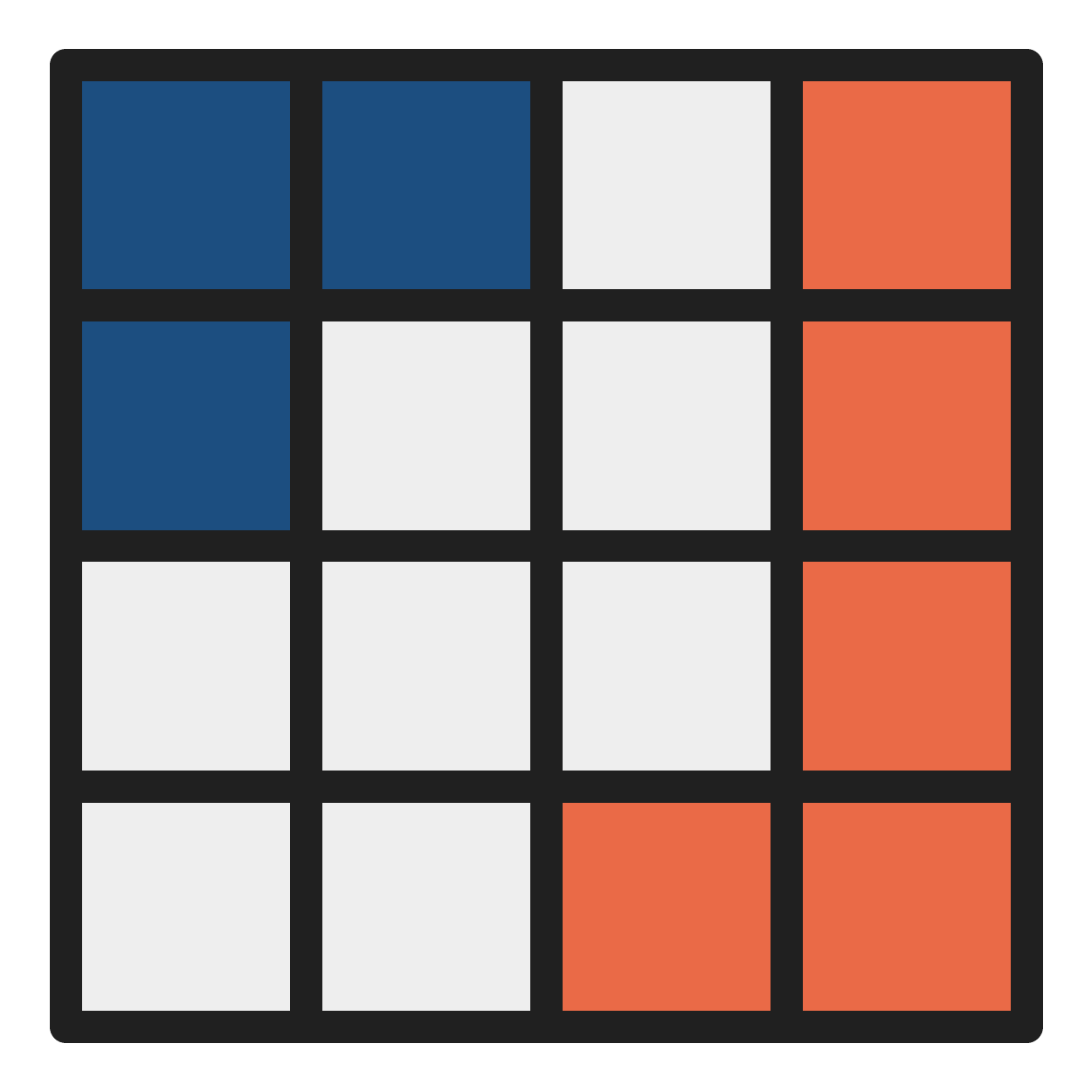}}\,#1}}{#1}}
\newcommand{\vblob}[1]{%
  \texorpdfstring{{\color{BlobColor}\raisebox{-0.1ex}{\includegraphics[height=0.75em]{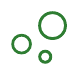}}\,#1}}{#1}}
\usepackage{scalerel}
\usepackage{tikz}
\usetikzlibrary{svg.path}

\definecolor{orcidlogocol}{HTML}{A6CE39}
\tikzset{
  orcidlogo/.pic={
    \fill[orcidlogocol] svg{M256,128c0,70.7-57.3,128-128,128C57.3,256,0,198.7,0,128C0,57.3,57.3,0,128,0C198.7,0,256,57.3,256,128z};
    \fill[white] svg{M86.3,186.2H70.9V79.1h15.4v48.4V186.2z}
                 svg{M108.9,79.1h41.6c39.6,0,57,28.3,57,53.6c0,27.5-21.5,53.6-56.8,53.6h-41.8V79.1z M124.3,172.4h24.5c34.9,0,42.9-26.5,42.9-39.7c0-21.5-13.7-39.7-43.7-39.7h-23.7V172.4z}
                 svg{M88.7,56.8c0,5.5-4.5,10.1-10.1,10.1c-5.6,0-10.1-4.6-10.1-10.1c0-5.6,4.5-10.1,10.1-10.1C84.2,46.7,88.7,51.3,88.7,56.8z};
  }
}

\newcommand\orcidicon[1]{\href{https://orcid.org/#1}{\mbox{\scalerel*{
\begin{tikzpicture}[yscale=-1,transform shape]
\pic{orcidlogo};
\end{tikzpicture}
}{|}}}}

\newcommand{\appRef}[1]{appendix~\ref{#1}}

\begin{document}

\title{HOLE: Homological Observation of Latent Embeddings for Neural Network Interpretability}

\author{
  Sudhanva Manjunath Athreya \orcidicon{0000-0003-3637-1084} and
  Paul Rosen \orcidicon{0000-0002-0873-9518}
\thanks{Sudhanva Manjunath Athreya is with University of Utah. E-mail: sud.athreya@utah.edu}
\thanks{Paul Rosen is with University of Utah. E-mail: paul.rosen@utah.edu}
}

\markboth{Journal of \LaTeX\ Class Files,~Vol.~14, No.~8, August~2021}%
{Shell \MakeLowercase{\textit{et al.}}: A Sample Article Using IEEEtran.cls for IEEE Journals}

\maketitle

\begin{abstract}

  Deep learning models have achieved remarkable success across various domains, yet their learned representations and decision-making processes remain largely opaque and hard to interpret. 
  This work introduces HOLE (Homological Observation of Latent Embeddings), a method for analyzing and interpreting discriminative neural networks through persistent homology. 
  HOLE extracts topological features from intermediate activations and presents them using a suite of visualization techniques, including \vcflow{cluster flow diagrams}, \vblob{blob graphs}, and \vhmd{heatmap dendrograms}. These tools facilitate the examination of representation structure and quality across layers.
  We evaluate HOLE using a range of discriminative models, focusing on representation quality, interpretability across layers, and robustness to input perturbations and model compression. 
  The results indicate that topological analysis reveals patterns associated with class separation, feature disentanglement, and model robustness, providing a complementary perspective for understanding and improving deep learning systems.
  % \PR{The focus here is on deep learning, but you only demonstrate your approach on a subset of model types. You should narrow the focus on the text or work with more diverse models.}
\end{abstract}

\begin{figure*}[!t]
  \centering
  \begin{tikzpicture}
    \node[anchor=south west, inner sep=0] (img) at (0,0)
      {\includegraphics[width=0.9\linewidth]{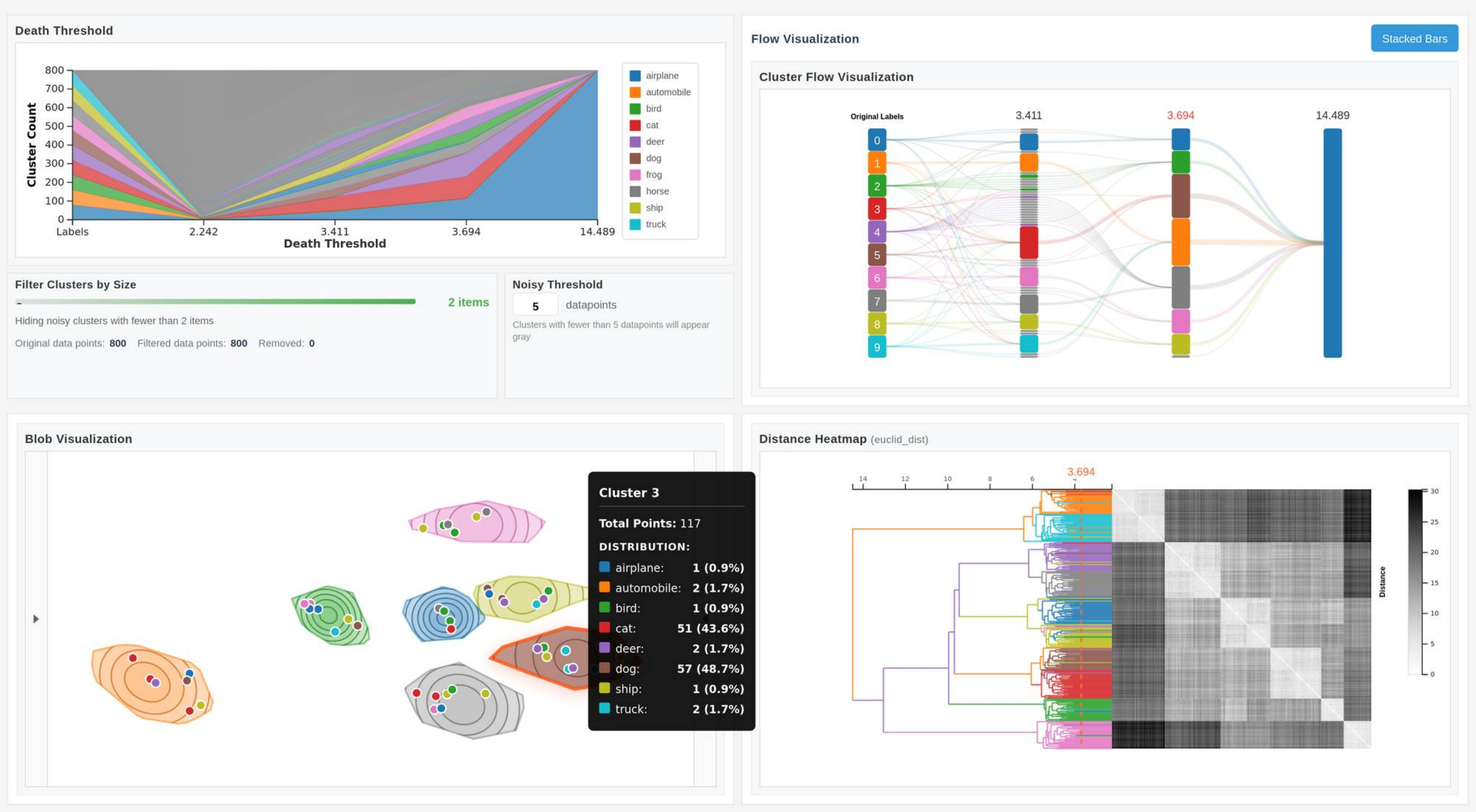}};
    \begin{scope}[x={(img.south east)}, y={(img.north west)},
                  every node/.style={font=\sffamily\bfseries\small, rounded corners=2pt,
                                     inner sep=3pt, fill=white, fill opacity=0.85, text opacity=1}]
      % Top-left: Stacked area
      \node[anchor=north west] at (0.02, 0.97) {\vcflow{Stacked Area Chart}};
      % Top-right: Sankey
      \node[anchor=north east] at (0.98, 0.97) {\vcflow{Sankey Diagram}};
      % Bottom-left: Blob
      \node[anchor=south west] at (0.02, 0.03) {\vblob{Blob Graph}};
      % Bottom-right: Dendrogram
      \node[anchor=south east] at (0.98, 0.03) {\vhmd{Heatmap Dendrogram}};
    \end{scope}
  \end{tikzpicture}
  \caption{HOLE dashboard for topological analysis of neural network activations: (top-left) \vcflow{cluster flow stacked area chart} with interactive threshold slider, (top-right) \vcflow{cluster flow Sankey diagram} for cluster and threshold selection, (bottom-left) \vblob{blob graph} of spatial cluster organization, and (bottom-right) \vhmd{heatmap dendrogram} of hierarchical structure.}
  % \PR{this image needs the full screen banner removed and annotations added. you might do the contour version of the blob graph. i think they look better.}
  \label{fig:teaser}
\end{figure*}

\setstretch{1}
%% \section{Introduction} %for journal use above \firstsection{..} instead
\section{Introduction} 
% This template is for papers of VGTC-sponsored conferences which are \emph{\textbf{not}} published in a special issue of TVCG.

Deep learning models have recently gained in popularity~\cite{goodfellow2016deep, lecun2015deep}, and have demonstrated remarkable predictive performance on a wide range of complex tasks. 
Discriminative models, with architectures trained to map inputs to class labels, have become the workhorse of computer vision, powering image classification~\cite{krizhevsky2012imagenet} through convolutional networks~\cite{he2016deep} and, more recently, transformers~\cite{dosovitskiy2020image}.
The same transformer-based paradigm has proven equally dominant in natural language processing, where models such as BERT~\cite{devlin2018bert} are used in discriminative tasks including sentiment analysis, text classification, and named entity recognition (NER)~\cite{tjong2003introduction}.
Despite their predictive accuracy, these models are considered to be black boxes, which present challenges, including difficulty in debugging model failures, a lack of trust, potential for perpetuating bias, and regulatory compliance issues~\cite{molnar2020interpretable, rudin2019stop}. 
A fundamental problem underlying these issues is the lack of interpretability and explainability, i.e., we cannot understand which features the model learned, how it processes information, or why it makes specific decisions, making it impossible to debug, trust, or verify these systems effectively.
This lack of interpretability has profound implications, as there is a rise in the deployment of AI models in critical sectors such as 
% healthcare~\cite{rajkomar2018scalable, topol2019high} and 
healthcare~\cite{topol2019high} and 
% finance~\cite{khandani2010consumer, lopez2013machine}, 
finance~\cite{lopez2013machine}, 
where transparency, fairness, accountability, and ethical considerations are important~\cite{barocas2016big, mehrabi2021survey}. 

Multiple factors contribute to the difficulty in understanding these models. 
First, their complex architectures often involve millions of parameters, resulting in high-dimensional internal states. 
Second, the use of non-linear activation functions results in complex decision boundaries that are not intuitive. 
Third, over-parameterization in many architectures poses challenges, as multiple configurations can yield similar classification performance yet differ internally. 
Finally, the learned representations are often distributed across multiple layers rather than localized and generally do not correspond to human-interpretable concepts, limiting transparency. 
% \PR{do you feel like you address all of these challenges? it would be great to be really clear that your approach addresses each of them. }

% Without a clear understanding of how these models reach their conclusions, verifying their reliability and ensuring equitable outcomes becomes a difficult task.

To address the interpretability challenges of discriminative models, we consider applying persistent homology (PH) to their intermediate activation spaces.
PH directly addresses the four difficulties identified above.
(1)~It operates natively on high-dimensional point clouds, capturing the intrinsic topological structure of activation spaces by tracking connected components (clusters) across multiple scales, without performing lossy dimension reduction.
(2)~The multi-scale filtration it produces characterizes how class clusters form, merge, and separate, offering evidence about the structure of complex decision boundaries induced by non-linear activations.
(3)~Because topology is invariant to continuous deformations of the data, PH can expose structural differences between over-parameterized models that achieve similar accuracy yet organize their representations differently.
(4)~By applying this analysis layer by layer, PH can trace how distributed representations evolve across the network, revealing at which depth class-discriminative structure emerges and how it is transformed.
Moreover, a significant advantage of PH is its model-agnostic nature and robustness to noise: because it operates solely on activation vectors extracted from any layer (convolutional, self-attention, fully connected, or residual), it is applicable to any architecture that produces such activations and extends naturally across domains from vision to language.
As a result, PH provides a principled way to analyze the internal complexity of discriminative models beyond conventional feature-space visualization or attribution methods.
% Crucially, we apply this framework in a deliberately narrow way. We track only the $0$-dimensional structure of the activation space, its connected components, because our goal is not to catalog arbitrary topological complexity but to characterize how class-discriminative clusters form, persist, and merge across layers. This targeted use of persistent homology, following the birth and death of clusters rather than loops or voids, keeps the tool aligned with the concrete question we set out to answer: at which layers, and at which scales, does a network separate its classes, and how robust is that separation to input noise and model compression?
% \PR{we don't use H1... you should be careful to discuss it here.}
% \PR{the latter part of this paragraph should directly address how it overcomes the challenges of the prior paragraph.}

% \PR{again, focus up to now on deep learning broadly risks over scoping your contribution. need to be narrower.}

In this work, we present HOLE (Homological Observation of Latent Embeddings), an approach for interpreting and analyzing discriminative neural networks using PH. 
% Crucially, we apply this framework in a deliberately narrow way. 
We track the $0$-dimensional structure of the activation space (connected components), because our goal is not to catalog arbitrary topological complexity but to characterize how class-discriminative clusters form, persist, and merge across layers. %An open-source Python library that implements the method accompanies this paper.
% \PR{the contribution is the approach, not the library. the library is a demonstration of the approach.}
%
The topological data produced by HOLE is made interpretable through a suite of coordinated visualization techniques (\cref{fig:teaser}, \cref{subsec:vis-design}): \vcflow{cluster flow diagrams} that trace how class clusters form, merge, and split across the filtration, with a stacked area chart view that supports selecting an informative filtration range; \vblob{blob graphs} that show the spatial layout of class clusters; and \vhmd{heatmap dendrograms} that reveal inter-class distance structure.
% \PR{cite the section like you do with the evaluation}
% \PR{not sure I'd bring up the distance metrics here, as they no longer play a big role. focus on the visualizations instead.}
% \PR{you don't mention the cluster flow. This is an important widget for selecting the right filtration range.}
These techniques combine in an interactive dashboard that links these views to support rapid, coordinated exploration across models and layers.
We evaluate HOLE across three applications: learned representation analysis~(\cref{sec:app1}), robustness to input noise~(\cref{sec:app2}), and the structural impact of model compression~(\cref{sec:app3}).
% \PR{you might want to cross reference the sections for these applications. that makes it clear that you mean evaluation here. you also say demonstate. you might replace that with evaluate.}
By examining the stability of topological features under these conditions, HOLE provides insights into classifier behavior beyond traditional accuracy metrics, offering a more holistic understanding of representational quality and potential failure modes.
% \PR{this paragraph needs an update.}
% , following the birth and death of clusters rather than loops or voids, keeps the tool aligned with the concrete question we set out to 
This targeted use of PH
enables answering: at which layers, and at which scales, does a network separate its classes, and how robust is that separation to input noise and model compression?

\section{Related Work}\label{sec:related}

The growing demand for interpretable machine learning, combined with recent advances in computational topology, has motivated a range of approaches.

% The growing demand for interpretable machine learning, combined with recent advances in computational topology, has motivated a range of approaches for understanding neural network representations~\cite{purvine2023experimental}.
% \PR{this only talks about topology, not ML interpretability.}

% \subsection{Scalar Field Topology}
% Understanding the connectivity of level sets is central to volume data analysis.  Early work focused on robust surface extraction, most notably \emph{marching cubes}~\cite{lorensen1987marching}, but soon evolved towards exact, topology–aware structures.  
% The merge and split trees capture the evolution of connected components; their unification, the \emph{contour tree}, can be computed efficiently even in higher dimensions~\cite{carr2003contour}. %
% Subsequent research introduced multiresolution Morse–Smale complexes that enable semantics–preserving simplification and feature–oriented segmentation~\cite{edelsbrunner2001hierarchical}.  
% Recent efforts concentrate on scalability, for example, through the multi–threaded \emph{contour forests} algorithm~\cite{gueunet2016contour} and the open–source Topology ToolKit (TTK). \todo{is this relevant?}

\subsection{Machine Learning Interpretability}
The growing complexity of machine learning models has sparked significant interest in interpretability and explainability methods.
At the local level, methods such as LIME~\cite{ribeiro2016lime} and SHAP~\cite{lundberg2017shap} explain individual predictions by fitting surrogate models or assigning game-theoretic feature importance, while concept-based approaches like network dissection~\cite{bau2017network} and TCAV~\cite{kim2018interpretability} test whether learned neurons align with human-interpretable concepts.
At the global level, activation maximization~\cite{erhan2009visualizing} and feature visualization~\cite{olah2017feature} synthesize inputs that reveal preferred stimuli for neurons or layers.
These methods operate at the level of individual predictions, features, or neurons; they do not reveal how a model's internal representations are globally organized: how class-relevant structure emerges, persists, or degrades across layers.

\paragraph{Saliency Maps and Gradient-Based Methods} Saliency maps highlight the input regions that most strongly influence a model's decision by computing gradients of the output with respect to the input.
Grad-CAM~\cite{selvaraju2017grad} operates at the feature-map level, producing coarse localization maps that show which image regions drive specific predictions.
Subsequent work refined gradient-based attribution through path integration to address saturation~\cite{sundararajan2017axiomatic}, input-noise averaging for sharper maps~\cite{smilkov2017smoothgrad}, layer-wise relevance decomposition~\cite{bach2015pixel}, and systematic occlusion to measure regional importance~\cite{zeiler2014visualizing}.
While saliency maps excel at providing intuitive visual explanations, they are fundamentally input-level attribution methods: they identify \emph{which} input regions matter for a prediction, but do not characterize \emph{how} the model internally organizes its representations, leaving the topological structure of the activation space outside their scope.
\paragraph{Interactive Visualization Systems} The visualization community has also developed interactive tools that complement traditional neural network interpretability methods; this broad space of visual analytics for explainable deep learning has itself been surveyed in detail~\cite{larosa2023sota}.
For CNNs, systems such as Summit~\cite{hohman2020summit}, ActiVis~\cite{kahng2018activis}, CNNVis~\cite{liu2017towards}, and NeuroCartography~\cite{park2022neurocartography} use coordinated multi-view designs to let users explore and cluster neuron activations, attribution graphs, and learned concepts across layers; Bluff~\cite{das2020bluff} contrasts the activation pathways of benign and adversarial inputs, while more educational or design-oriented tools support direct manipulation of small networks~\cite{smilkov2017directmanipulation} and step-by-step walkthroughs of convolutional operations~\cite{wang2020cnn}.
% Most closely related in goal is ChannelExplorer~\cite{zaman2025channelexplorer}, which also targets class separability in activation space but through a different object and method: it collapses each \emph{activation channel} to a scalar per input image and explores the resulting per-image channel vectors with a UMAP scatterplot, a Jaccard-similarity matrix over the most-activated channel sets, an activation heatmap, and a per-layer confusion hierarchy. 
ChannelExplorer~\cite{zaman2025channelexplorer} examines class separability by summarizing activation channels for individual images and visualizing the resulting channel vectors through dimensionality reduction, set similarity, heatmaps, and layer-wise confusion hierarchies.

% HOLE instead treats a layer's activations as a full-dimensional \emph{point cloud} of samples and reads its multi-scale $H_0$ persistent-homology structure. 
% The two differ in three concrete respects: HOLE's unit of analysis is the sample point cloud rather than the per-channel summary; it derives structure from a distance filtration rather than from dimensionality reduction and set overlap; and it tracks how class clusters form, merge, and split \emph{across} layers through the coordinated cluster-flow workflow, whereas ChannelExplorer analyzes each layer independently. The methods are complementary, contrasting channel-level attribution with sample-level cluster topology.
Others visualize how samples or classes traverse the network: Deep Dive with Flows~\cite{halnaut2020deepdive} traces individual samples through successive layers with a Sankey-and-parallel-coordinates design, Alsallakh et al.~\cite{alsallakh2018cnnhierarchy} derive a class hierarchy from the confusion matrix, showing that early layers separate coarse class groups and later layers separate individual classes, while further systems target specific analysis goals such as per-instance difficulty~\cite{meng2025difficultyaware,vandenelzen2026instancebased} and the behavior of reinforcement-learning agents~\cite{mishra2022whywhynot}.
For sequence and transformer models, a parallel line of tools visualizes hidden-state and attention behavior: RNNVis~\cite{ming2017rnnvis} and Seq2Seq-Vis~\cite{strobelt2019seq2seqvis} inspect recurrent hidden states, exBERT~\cite{hoover2020exbert}, SANVis~\cite{park2019sanvis}, BertViz~\cite{vig2019bertviz}, and interactive attention explorers~\cite{dong2020attentionexplorer} expose per-layer, per-head self-attention, and Rogers et al.~\cite{rogers2020bertology} survey what BERT learns across its layers.
Closest to our work, TopoBERT~\cite{rathore2023topobert} applies the Mapper algorithm to BERT embeddings, demonstrating that topological summaries can reveal how fine-tuning reorganizes word representations.
However, these systems summarize individual neurons or attention heads, or trace the raw flow of samples and output-confusion statistics; none read the multi-scale topological structure of a layer's activation point cloud, making it difficult to answer higher-level questions such as whether classes are well-separated at a given layer or how cluster structure forms and merges with depth. HOLE complements these tools by providing that layer-level, multi-scale view through PH.
% \PR{this paragraph still feels a little enumerative}
% \PR{need an explicit gap for HOLE to fill.}

In summary, existing interpretability methods leave a common gap: local explanation techniques identify important features or input regions but not the global geometry of learned representations; saliency methods are confined to input-level attribution and cannot characterize activation-space topology; and interactive visualization systems focus on neuron- or attention-level summaries without multi-scale topological analysis.
HOLE addresses this gap by applying PH to intermediate activation spaces, revealing how class-relevant structure emerges, persists, and degrades across layers---providing a structural, layer-wise perspective that complements existing feature-level and attribution-based approaches.
% \PR{this is good but should be aligned with the gaps enumerated in each of the prior paragraphs.}
% two key techniques: activation aggregation to discover important neurons and neuron-influence aggregation to identify relationships between neurons. 
% These techniques combine to create attribution graphs that reveal crucial neuron associations across network layers. 
% \todo{this section feels thin. there must be more that is related to what you're doing, particularly from the vis community. hint, look at ``Summit: Scaling Deep Learning Interpretability by Visualizing Activation and Attribution Summarizations'' and related papers.}

\subsection{Topological Data Analysis and Visualization}
\label{sec:topological-analysis}

% \PR{i moved this after interpretability. i think this will flow better, but the text might still need some updates.}

Recent advances in topological data analysis (TDA)~\cite{singh2007topological} have demonstrated that studying the shape and topological structure of data in high-dimensional spaces provides powerful insights into underlying patterns.
Specifically, \textit{persistent homology} (PH) has emerged as a powerful mathematical framework for quantifying the persistence of topological features across multiple scales~\cite{edelsbrunner2002persistence,zomorodian2005computing}. 
The output of PH, called a persistence diagram, is a collection of birth-death pairs for topological features and is stable under perturbations of the input~\cite{cohen2007stability}, which opened the door to applications in noisy settings.  

% \todo[inline]{maybe move these to related work?}

\subsubsection{Topological Visualizations}
Two standard visualizations summarize this information (\cref{fig:persistence_viz}): the \emph{persistence diagram}, a scatterplot with each feature drawn as a point; and the \emph{persistence barcode}~\cite{ghrist2008barcodes}, which draws each feature as a horizontal bar spanning its lifetime.
Vector-space representations such as persistence landscapes~\cite{bubenik2015landscapes} have been proposed to transform persistence diagrams into a function-based representation in a vector space, making them usable in machine learning and statistical analysis.
Other approaches include persistence images, which create a fixed-size vector representation from diagrams, and specialized kernels~\cite{carriere2017sliced}.
These methods, along with scalable software libraries (e.g.,~GUDHI~\cite{maria2014gudhi}), provide practical access to topological methods.

\begin{figure}[!htb]
    \centering
    \resizebox{\linewidth}{!}{\input{figures/persistence_diagram.pgf}}
    \caption{Example (left) persistence diagram and (right) barcode.}
    \label{fig:persistence_viz}
\end{figure}

These representations occupy different points in a broad design space, and the appropriate choice depends on the analytical goal. One axis is which homology dimension is studied ($H_0$ connected components, $H_1$ loops, or higher). A second is how the persistence information is summarized: a full diagram, reduced to scalar descriptors such as mean death time, total persistence, Betti curves, or persistence entropy, or vectorized into landscapes, images, and kernels for downstream learning. A third is how it is used: as a static plot, as a feature fed to a model, or as an interactive visualization. Much prior work applying PH to neural networks favors scalar or vectorized summaries because the goal is prediction or comparison; persistence statistics, for example, have been used to estimate test accuracy and the generalization gap~\cite{gutierrez2021phgeneralization,ballester2024generalizationgap}. Such summaries are powerful for those purposes, but they deliberately discard the correspondence between a topological feature and the labeled data points that compose it. HOLE occupies a complementary point in this space: it retains the full multi-scale $H_0$ structure and keeps it tied to individual, labeled probe-set samples through coordinated visualizations, because our interpretability goal is to judge which clusters correspond to which classes and where in the network that structure emerges, questions that scalar persistence descriptors cannot answer.

% \subsection{Persistent Homology}
% \marginpar{call this tda vis --> move the other paragraph from topo and interpretability here}
%

% A particularly influential method is \textit{Mapper}~\cite{singh2007topological}, which constructs simplified representations of high-dimensional data by creating nerve complexes of overlapping neighborhoods.
% Unlike persistent homology, which focuses on quantifying the lifetime of topological features across scales, Mapper addresses the complementary problem of \emph{shape summarization}---creating interpretable, graph-based visualizations that preserve the topological structure of the data while allowing interactive exploration.
% \PR{candidate to cut. we don't do anything with mapper besides the description of TopoBert in the prior section.}

% Mapper has been successfully applied to diverse domains including biological data analysis~\cite{lum2013extracting}, materials science, and social network analysis, demonstrating its versatility as a topological lens for exploratory data analysis and knowledge discovery in high-dimensional datasets. \todo{is this relevant?}

\subsubsection{Topology and Deep Learning}
% Topology‐aware visualization techniques—\emph{topological landscapes}, flexible isosurfaces, or Lagrangian coherent structures—translate abstract invariants into interactive exploration tools.  
The convergence of topological data analysis and deep learning has proceeded along three threads.
The first integrates persistence directly into training: differentiable persistence layers~\cite{hofer2017deep} and permutation-invariant architectures like PersLay~\cite{carriere2020perslay} allow networks to consume topological descriptors, while topology-aware autoencoders~\cite{moor2020topoae} use persistence as a regularizer to preserve latent-space structure.
The second uses persistence as an analytic lens on trained networks, quantifying complexity from weight matrices~\cite{rieck2019neural,watanabe2021topologicalmeasurement}, characterizing the topology of activation spaces~\cite{purvine2023experimental,gebhart2019characterizing} and how it simplifies from layer to layer~\cite{naitzat2020topology}, linking persistence to intrinsic dimension~\cite{birdal2021intrinsic}, summarizing per-layer activations with persistence landscapes~\cite{wheeler2021activationlandscapes}, and building interactive tools that explore the shape of activations, most directly TopoAct~\cite{rathore2021topoact}, which applies the Mapper algorithm to a layer's activation point cloud.
The third connects topology to generalization, showing that persistence statistics can estimate test performance without a held-out set~\cite{gutierrez2021phgeneralization} and predict the generalization gap~\cite{ballester2024generalizationgap}.

% \paragraph{Topological methods as visualization}
PH has also been used as a visualization primitive in its own right, beyond neural networks: for instance, to visually detect and communicate structural change in time-varying graphs by tracking their evolving connectivity~\cite{hajij2018visual}. Within neural-network interpretability specifically, topological visualization has so far been dominated by the Mapper-based tools discussed above (TopoAct and TopoBERT), which build a nerve graph of a single layer's activations, while the wider effort to combine PH, machine learning, and deep networks is now reaching applied domains~\cite{arfi2024promises}. Against this backdrop HOLE's contribution to the topology-for-interpretability space is specific: it reads the $H_0$ cluster structure directly from the persistence barcode rather than from a Mapper nerve, it makes that structure comparable \emph{across layers} through the coordinated cluster-flow workflow rather than exploring one layer at a time, and it keeps the structure label-aware through the purity and homogeneity metrics, filling a niche that existing topological-visualization tools leave open.

HOLE occupies a distinct niche: where prior work uses persistence to train better networks, quantify weight-level complexity, or predict generalization, HOLE applies PH \emph{post hoc} to intermediate activation spaces for visual interpretability.
It does not modify the network or require retraining; instead, it provides a global, layer-by-layer reading of a model's internal organization through coordinated visualizations (\vcflow{cluster flow diagrams}, \vblob{blob graphs}, and \vhmd{heatmap dendrograms}) that reveal where class-discriminative structure emerges, persists, or degrades across layers. Unlike TopoAct~\cite{rathore2021topoact}, the closest such tool, which builds a Mapper graph of a single layer to expose branch and loop structure, HOLE reads the $H_0$ connected-component structure across all layers and ties it to class labels through the coordinated cluster-flow workflow.

% \input{sections/background.tex}
% \externaldocument{related}

\section{Background}
% \marginpar{should i add a short summary here before the subsections?} \todo{yes, usually there is a short paragraph that describes the overall structure of the section.}
Our approach involves observing the behavior of neural networks through the lens of persistent homology (PH). 
PH is an interesting tool for this application because it can \textit{summarize multi-scale structures in complex data, independent of the dimensionality, while being robust to certain types of noise}~\cite{carlsson2009topology, edelsbrunner2010computational, cohen2007stability}.
These properties are particularly relevant for neural network activation spaces, which are high-dimensional, exhibit complex non-linear structure arising from successive layer transformations, and may contain noise from training stochasticity or input perturbations (empirically verified by repeating our analysis across 10 probe-set samples, see \appRef{sec:stability_analysis}).
% \PR{might be nice to relate the prior statement to the complexity of activation space.}
% \PR{need a citation to this claim.} 
We then construct several visualizations on top of PH to reveal important structures in the neural network.

% %In this section, the complete methodology behind HOLE is presented.
% The basics of neural networks and activation data are presented in \cref{subsec:dnn-structure}. 
% The core mathematical framework of persistent homology, construction of VR-complex, and its application to neural activation data is presented in \cref{subsec:persistent-homology}. 
% A critical component in our methodology is the usage of multiple distance metrics -- Euclidean, cosine, Mahalanobis, and relative neighborhood normalization, each designed to reveal different geometric and semantic properties of the learned representations (\cref{subsec:distance-metrics}). 
% Our library comes with a suite of visualization approaches like dendrogram heatmaps, Sankey-based diagrams, and blob graphs, which translate the abstract topological data into interpretable insights for practitioners (\cref{subsec:visualization}). 
% Together, these components form a comprehensive framework for analyzing the internal structure and evolution of neural network representations across layers, distance scales, and architectural variations.

\subsection{Deep Neural Network Structure and Data}
\label{subsec:dnn-structure}

Deep Neural Networks (DNNs) are computational graphs composed of layers and connections. 
A network consists of interconnected nodes called neurons, organized in a series of layers.
Each layer outputs a weighted sum of its inputs, which is then followed by a nonlinear function known as an activation function.
These activation functions (e.g., Sigmoid, Tanh, and ReLU) introduce non-linearity to the networks.
Without an activation function, the linear operations in the neural network can be collapsed into a single linear transformation. 
% Thus, the mappings from the input to the output space become linear. 
% To overcome this limitation, 
Instead, activation functions are placed after each linear transformation in the model.
The outputs of these activation functions are considered ``features'' that a particular layer has learned. 
In discriminative models, the features create decision boundaries that separate different classes in the activation space.
Thus, the learned representations can be found in these activations.

% \todo{add a schematic of the neural network --> show the hooks part in the same figure --> and the vectors + point cloud part in the same figure}

\begin{figure}[!b]
    \centering
    \includegraphics[width=\linewidth]{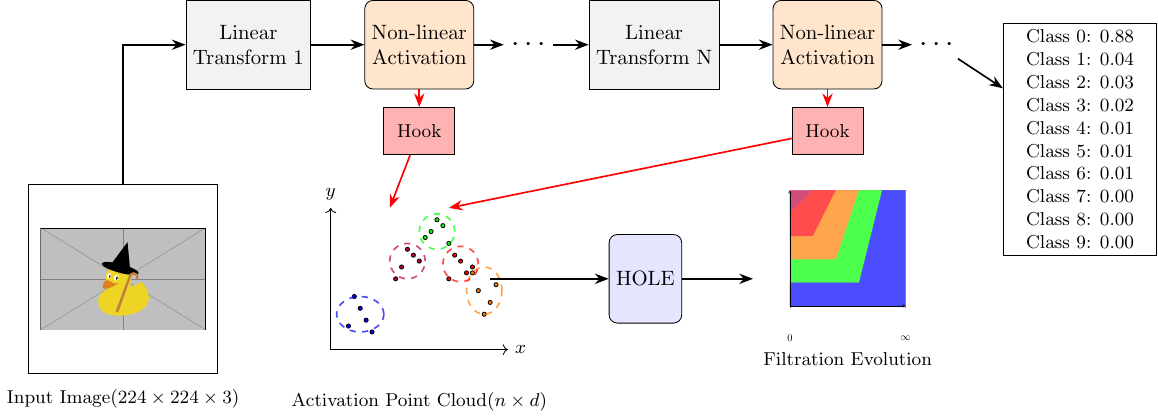}
    \caption{During inference, neural network activations are extracted via forward hooks. These activations form point clouds that are passed to the HOLE PH pipeline and visualized.
    % \PR{image quality could be improved. Some text is too small, but overall, it could look "fancier".}
    }
    \label{fig:hole_schematic}
\end{figure}

% \begin{figure}[t]
%     \centering
%     \includegraphics[width=\linewidth]{figures/ViT.pdf}
%     \caption{Architecture of a Vision Transformer (ViT) model showing the flow of data through the network. The input image is divided into patches, which are linearly projected and combined with positional embeddings. These embeddings are processed through multiple transformer encoder layers, and finally classified using an MLP head. The dimensions at each stage are shown in blue.}
%     \label{fig:vit_architecture}
% \end{figure}

% \textcolor{red}{this needs to get more specific to the background needed for understanding methods. probably ok for the time being.}

% \subsubsection{DNN Activation Data}
% \label{subsec:activation-data}

% \marginpar{belongs in the experiments section}

To understand the internal data representations of these models, we extract intermediate activations from the model as it processes input samples. 
% We utilize multiple variations of the same input data, by introducing controlled variations, such as image augmentations, noise injection, or class-balanced sampling, we ensure that the extracted activations capture a wide range of the network's representational capacity and robustness.
% \PR{again focus on DNNs. you should start with DNNs but then say, we focus on CNNs which... your discussion section will need to cover how this extends to DNNs in general.}
%
We capture these activations using \emph{forward hooks}, callback functions registered on a network layer that are automatically invoked during the forward pass, exposing that layer's inputs and outputs without modifying the model's architecture or source code. In HOLE, forward hooks are placed on pre-selected non-linear layers during inference, recording the outputs (i.e., activations) of these layers (\cref{fig:hole_schematic}).
During inference, as each batch of inputs is passed through the network, the forward hooks capture the activations at the specified layers. 
These activations are high-dimensional embeddings, which we reshape as needed to form vectors in $\mathbb{R}^d$, where $d$ corresponds to the number of output units in the layer. The activations act as a point cloud for the downstream PH analysis.
We place hooks after non-linear activations rather than before them because the activation function is what introduces the non-linear structure that makes representations class-discriminative; pre-activation outputs are affine transformations of the previous layer and therefore carry less topological information about learned class boundaries.
% \PR{is there some value in observing the network before the non-linear activations? If yes, why haven't you. If no, you should state why you only look post-activation.}
% \todo{add information to the illustrative figure for the prior subsection to highlight where the hooks go and what the vector looks like.}
% As illustrated in Figure~\ref{fig:hole_schematic}, the forward hooks are placed at non-linear layers to capture activations, which are then flattened into high-dimensional vectors that form the input point clouds for our persistent homology analysis.

% \marginpar{add a figure here -- data distribution, network hook functions}

\subsection{Persistent Homology (PH)}
\label{subsec:persistent-homology}
% \marginpar{section feels repetitive, change it later.}

% \marginpar{add citations?}
% We employ PH as our primary tool to analyze how the topological features of neural network activation point clouds evolve across the filtration.

% Intuitively, PH asks a single question at many scales at once. If we treat two activation vectors as connected whenever their distance falls below a threshold $\epsilon$, and then slowly increase $\epsilon$, how do groups of points form and merge? At a very small $\epsilon$ every activation is its own group; as $\epsilon$ grows, nearby points (typically same-class activations) join first, then whole classes merge, until eventually all points collapse into a single group. PH records the precise thresholds at which each group is born and dies, and the \emph{lifespan} of a group, its persistence, indicates how pronounced that structure is: a class that remains a distinct group across a wide range of thresholds is a robust, well-separated cluster, whereas groups that appear and vanish almost immediately are noise. This is what makes the tool useful for interpretability. Rather than committing to a single similarity threshold or a fixed number of clusters, it summarizes the entire multi-scale organization of an activation space, and does so in a way that is stable to small perturbations of the data and independent of the ambient dimension.

Intuitively, PH tries to understand the evolution of the topological structure of discrete points across multiple scales, as opposed to committing to a single similarity threshold or fixed number of clusters. To understand PH, we first describe homology, which extracts topological features at a single scale.

\subsubsection{Homology}
Given a finite point cloud $X = \{x_1, \ldots, x_n\} \subset \mathbb{R}^d$ of neural network activations, homology characterizes the topological structure of the data at a fixed scale: $H_0$ counts connected components, $H_1$ counts loops (1-dimensional holes), $H_2$ counts voids, and so on.
To construct a topological space from $X$ at a given radius $\epsilon$, we use the \textit{Vietoris--Rips (VR) complex}~\cite{edelsbrunner2002persistence}: any subset of points whose pairwise distances are all $\leq \epsilon$ forms a simplex.
Computing homology at a single $\epsilon$ yields a snapshot of the topological structure at that scale.
% \PR{this is a PH description. homology looks at a single $\epsilon$.}
In HOLE, we restrict our analysis to $H_0$ (connected components), as class-discriminative structure in activation spaces is primarily reflected in how clusters of points form and merge, rather than in higher-dimensional features such as loops or voids.
This restriction is a deliberate scoping decision, not a limitation of PH. Higher-dimensional features carry information: $H_1$ loops, for instance, have recently been shown to provide useful diagnostics of latent-space geometry in large language models~\cite{fay2026shape}. Our goal, however, is to characterize how a network organizes and \emph{separates classes} in activation space, which is fundamentally a question of how point clusters form, persist, and merge, exactly the structure that $H_0$ captures. A loop or void would instead describe the shape of a single class's point cloud, e.g., whether its activations lie on a ring, which is orthogonal to whether that class is \emph{separated} from the others.
%: a class can be annular or convex and remain equally distinguishable. Our analysis targets precisely this inter-class organization, how distinct classes separate, stay homogeneous, and merge across scales, so it is a question of connectivity ($H_0$) rather than of intra-class geometry ($H_1$ or $H_2$). Restricting to $H_0$ additionally admits an efficient near-linear union-find computation, keeping the analysis tractable for the large activation point clouds of modern networks. Extending HOLE to loops and voids ($H_1$, $H_2$) for complementary interpretability signals is a natural direction for future work.
% Additionally, restricting to $H_0$ admits an efficient near-linear union-find computation.
% Formal definitions of the simplicial complex, chain complex, boundary operators, homology groups, and VR complex are provided in appendix \cref{sec:appendix-ph-formalism}.
Formal definitions of the homology operators are provided in \appRef{sec:appendix-ph-formalism}.

% \PR{in HOLE we only track H0...}

% \marginpar{Offset_filtration in wikipedia}
% \marginpar{Offset filtration in wikipedia, cite?}

\subsubsection{Persistent Homology} 
PH is a mathematical framework that allows tracking the evolution of topological features across different scales (i.e., different values of $\epsilon$)~\cite{carlsson2009topology}.
%
% \textbf{Filtration.} 
The key concept of PH is a \textbf{filtration}, which is
% ÷Filtration or offset filtration is
a growing sequence of metric balls, used to detect the size and \textbf{scale} of topological features of a dataset~\cite{edelsbrunner2010computational}.
Mathematically, a filtration can be defined as a nested sequence of simplicial complexes,
% \[
$\emptyset = K_0 \subseteq K_1 \subseteq K_2 \subseteq \cdots \subseteq K_m$,
% \]
parameterized by increasing scale parameter $0 = \epsilon_0 < \epsilon_1 < \epsilon_2 < \cdots < \epsilon_m$.
%The Vietoris--Rips filtration is obtained by setting $K_i = \mathrm{VR}_{\epsilon_i}(X)$.
%
As the filtration parameter $\epsilon$ increases, topological features appear, known as \textbf{birth}, and disappear, known as \textbf{death}.
PH tracks these changes by computing homology groups across the filtration and recording \textbf{birth--death pairs}.
% , helping us analyze topological features of the point cloud $X$ through a sequence of \textit{simplicial complexes} built over increasing distance scales. 
% For each topological feature $\alpha$, we define:
% - Birth $b(\alpha)$: the smallest $\epsilon$ where $\alpha$ appears
% - Death $d(\alpha)$: the smallest $\epsilon > b(\alpha)$ where $\alpha$ disappears  
% - Persistence $\text{pers}(\alpha) = d(\alpha) - b(\alpha)$: the lifespan of the feature
Further, a measure known as \textbf{persistence} is the difference between the birth and the death values of a feature.
Features with high persistence are considered significant topological structures, while short-lived features are typically attributed to noise. 
% \PR{you might want to highlight/bold any terminology that you use in the rest of the paper. In particular, filtration, scales, ...?}
% In our context, we perform filtration on the activation data from the neural network as per the user's choice of distance metric. 

% \marginpar{distance metric}

% \subsubsection{Filtration and Persistent Homology}

% Given a set of points and a distance metric, the VR filtration generates a nested sequence of simplicial complexes parameterized by a non-negative scale parameter $\epsilon$.
% At each scale $\epsilon$, the VR complex includes a $k$-simplex (i.e., a set of $k+1$ points) if every pair of its vertices is at most $\epsilon$ apart according to the chosen distance metric. 
% \marginpar{cite the Gunnar Carlsson paper}

% As $\epsilon$ increases, the VR complex grows: isolated points become connected, edges fill in, triangles and higher-dimensional simplices appear, and topological features such as connected components, loops, and voids emerge and eventually disappear. 

\begin{figure}[!t]
    \centering
    \includegraphics[width=\columnwidth]{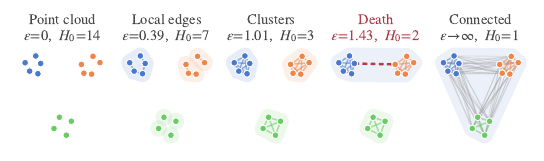}
    \caption{Example of $H_0$ PH on a three-class point cloud. As $\epsilon$ increases, the VR complex grows: nearby points form edges within each class, consolidating into three distinct groups. At $\epsilon=1.43$ two clusters merge (a \textbf{death} event, dashed edge).
    %, reducing the component count to $H_0=2$. 
    At $\epsilon\to\infty$ all points belong to a single connected component.}
    % \PR{text in the figure is a bit small}
    \label{fig:ph_illustration}
\end{figure}

% This filtration produces a sequence of nested complexes parameterized by increasing $\epsilon$, and persistent homology tracks the birth and death of topological features (connected components, loops, voids, etc.) across scales.
To calculate PH, we first compute a pairwise distance matrix for the activation point cloud using a chosen distance metric (discussed in \cref{subsec:distance-metrics}).
% \PR{distance conversation starts here and continues in the next subsection. that is confusing organization.}
We then perform the filtration process on this distance matrix.
Since HOLE focuses exclusively on $H_0$ (connected components), we only need to compute the filtration up to simplicial dimension~1 (i.e., edges), tracking how clusters form and merge across scales.
% Additionally, restricting to $H_0$ admits an efficient near-linear union-find computation.
This can be accomplished with an efficient near-linear union-find computation.
% \PR{1? 0 is points, 1 is edges}
% \PR{i guess you have the only track H0 here. It is confusing to say dimension one and that is H0.} 
% The VR-complex is constructed using multiple distance metrics. 
% The resulting filtration data is then used to draw different interpretations of the network.
\Cref{fig:ph_illustration} illustrates this process on a synthetic three-class point cloud.

% \PR{as we discussed, you may want to move the more technical description to the appendix and focus on a higher-level intuitive description here.}
% \marginpar{add a figure here -- VR filtration}
% \PR{I recommend adding an illustrative figure for calculating persistent homology. you can label it with the kep concepts that you bolded.}

% Starting from $14$ isolated points ($H_0 = 14$), the VR complex progressively connects nearby points as $\epsilon$ grows.
% Within-class edges appear first, forming simplicial complexes that consolidate into three clusters corresponding to the three classes ($H_0 = 3$).
% As $\epsilon$ increases further, inter-class edges emerge and clusters merge---each such merge constitutes a \textbf{death} event in the filtration.
% Features that persist across a wide range of $\epsilon$ values reflect genuine topological structure (i.e., well-separated classes), while short-lived features correspond to noise.

\paragraph{How HOLE Adapts PH to Activation Analysis}

 The \emph{scale at which structure is read} is selected in a label-aware way: instead of reporting the raw diagram, HOLE scores the $H_0$ clustering against the ground-truth classes with purity and homogeneity to choose the filtration scales it foregrounds (\cref{sec:cluster_quality_main}), a customization to the supervised activation-analysis setting that generic, label-free persistence summaries do not make. Together with the $H_0$ restriction motivated above, these adaptations tie the topological machinery to the specific interpretability question rather than applying it off the shelf.

\subsubsection{Distance Metrics to Highlight Features}
\label{subsec:distance-metrics}

Although HOLE does not introduce a new persistence algorithm, it does not treat PH as a black box either: the pipeline is customized to activation-space analysis.
% at the two points a generic application would leave unspecified. First, 
In particular, the \emph{filtration function} is chosen to match the geometry of learned representations.
%rather than defaulting to Euclidean distance, using cosine distance by default (motivated by the directional nature of embeddings) with Mahalanobis, geodesic, and density-normalized alternatives for correlated, manifold-structured, or unevenly sampled activations (\cref{subsec:distance-metrics}); 
Since the filtration function determines the entire persistence output, this choice is itself an expressive, problem-specific degree of freedom~\cite{ballester2025expressivityph}.
%
% The distance metric is a crucial component of the filtration process.
For neural network activation spaces, cosine distance is well-motivated as the primary metric: foundational work on distributed representations has established that learned embeddings encode semantic content primarily through the \emph{direction} of the activation vector rather than its norm~\cite{mikolov2013efficient}. 
Self-supervised and multimodal learning methods make this assumption explicit by operating entirely in cosine similarity space~\cite{chen2020simple, radford2021learning}. 
Furthermore, in high-dimensional spaces, the \emph{curse of dimensionality} causes Euclidean pairwise distances to concentrate around a narrow range, making them nearly indistinguishable and reducing discriminative power~\cite{aggarwal2001surprising}, whereas cosine similarity measures only angular separation and remains stable regardless of dimensionality. 
We therefore adopt cosine distance as the default for HOLE.
For use cases requiring alternative geometric perspectives, HOLE also supports the following metrics, each motivated by prior work.
\textit{Euclidean} distance is the standard $\ell_2$ baseline adopted by foundational TDA studies of neural network activations and weights~\cite{purvine2023experimental, rieck2019neural, watanabe2021topologicalmeasurement}, and is appropriate when activation magnitudes are meaningfully bounded (e.g., after batch normalization).
\textit{Mahalanobis} distance weights pairwise distances by the inverse covariance of the activation distribution, naturally accounting for correlated features and anisotropic spread; it has been applied to class-conditional activation modeling for out-of-distribution detection in neural networks~\cite{lee2018mahalanobis}.
\textit{Geodesic} distance approximates intrinsic manifold distances between activations via shortest paths through a $k$-nearest-neighbor graph, capturing non-linear structure in curved activation spaces whose intrinsic geometry has been shown to govern generalization~\cite{birdal2021intrinsic}.
% \textit{Manhattan} distance ($\ell_1$) sums absolute coordinate-wise differences, providing greater robustness to individual large-magnitude outlier dimensions than $\ell_2$.
% \textit{Chebyshev} distance ($\ell_\infty$) is governed entirely by the single largest coordinate difference, making it useful for detecting activations that diverge sharply in even one feature dimension.
\textit{Density-normalized} variants scale each pairwise distance by the geometric mean of the local neighborhood densities of both points~\cite{silverman2018density}, making the filtration robust to heterogeneous activation densities and outliers.
The extensible metric interface is described in \appRef{sec:custom_metrics}.
% \PR{you will need some justification for the inclusion for each of the other metrics. this mostly requires a citation to another work that uses them.}

\subsubsection{Relationship to Clustering}
At the level of $H_0$, PH of a VR filtration is closely related to single-linkage hierarchical clustering: a connected component's death coincides with the distance at which single-linkage merges two clusters, so the $H_0$ merge tree and the single-linkage dendrogram encode essentially the same information.
% HOLE, therefore, does not claim a new clustering algorithm; its contribution is threefold.
However, the PH strategy HOLE uses provides some conceptual differences.
First, treating the \emph{entire} filtration as the object of study (rather than committing to a fixed number of clusters or a single cut height) equips every merge with a \emph{persistence} that quantifies its significance and is stable under small perturbations of the activation point cloud, giving a principled basis for separating signal from noise and selecting interpretable thresholds.
Second, HOLE varies the \emph{filtration function} (Euclidean, cosine, Mahalanobis, geodesic, and density-normalized distances), examining the same activations through several complementary geometric lenses; the choice of filtration is itself expressive~\cite{ballester2025expressivityph}.
Third, our contribution is not the topological summary in isolation but the coordinated, label-aware visual analytics workflow built upon it, designed to compare how class-discriminative structure emerges, degrades, and reorganizes across layers, architectures, input noise, and compression, a comparison that a single dendrogram or a fixed $k$-partition does not support.

\section{Homological Observation of Latent Embeddings}

% \subsection{Visualization and Analysis}

% \PR{rename this subsection and only focus on the task analysis. the visualization design portion should become a subsection (instead of a subsubsection)}

While persistence diagrams and barcodes (\cref{fig:persistence_viz}) are standard tools for visual analysis when using PH, they have fundamental limitations for neural network interpretability tasks.
Persistence diagrams encode topological features as points, abstracting away the identity and composition of individual clusters, precisely the information critical for assessing whether learned representations align with semantic class structure.
Barcodes similarly represent feature lifespans without revealing cluster details.
% which data points belong to which persistent components or how cluster membership relates to ground truth labels.
% These representations excel at summarizing global topological properties but cannot 
These representations do not support tasks requiring inspection of cluster composition, identification of specific outliers, or assessment of class separability.
%at the level of individual data points.
For neural network analysis, we require visualizations that preserve the connection between topological structure and semantic content, enabling users to evaluate whether topologically-defined clusters correspond to meaningful learned representations.
% We therefore identified a series of tasks and developed visualization techniques that maintain this critical linkage between topology and semantics.

\subsection{Task Analysis}
\label{subsec:visualization}
% \subsubsection{Task Identification}

To determine the design needs of our interface, we evaluated prior work on explainability and topological analysis of deep neural networks to identify recurring analytical questions for inspecting learned representations.

% \PR{shouldn't they be [T1] - [T4] (brackets instead of parentheses).}
% \PR{this should be the first paragraph of the section. It should start with something more akin to "In order to determine the design needs of our interface, we evaluated prior work on explainability in DNNs..." You may also borrow some text from the above paragraph. The description of the low level tasks can come in the paragraph following.}

\vspace{0.2em}
\noindent
\hangindent=0.4cm
\textbf{[T1]} \textbf{Hierarchy}: Understand the evolution of class clusters in activation space during the filtration process by tracking how $H_0$ components form, persist, merge, and disappear across scales.
% This task composes \textit{Cluster} (identify connected components at each filtration level), \textit{Characterize Distribution} (of component birth/death times), \textit{Sort} (by persistence values), and \textit{Determine Range} (of filtration thresholds where meaningful structure emerges).
% The \vhmd{dendrogram} reveals multi-resolution organization of learned representations. \PR{premature to start discussing visualizations}
The importance of tracking hierarchical cluster evolution is established by prior topological analyses of neural network activations, which show that how class clusters form and merge across filtration scales directly reveals the multi-scale representational structure learned by the network~\cite{watanabe2021topologicalmeasurement, purvine2023experimental, wheeler2021activationlandscapes}.

\vspace{0.2em}
\noindent
\hangindent=0.4cm
\textbf{[T2]} \textbf{Separability}: Assess how discriminatively the network has organized different classes in activation space by measuring the degree of separation between class-specific clusters.
% This task composes \textit{Cluster} (identify class groupings), \textit{Determine Range} (measure inter-cluster distances), \textit{Characterize Distribution} (of within- vs.\ between-class distances), and \textit{Compute Derived Value} (separation metrics at optimal filtration scales).
Well-separated clusters indicate effective feature learning, as class separability in learned representations is fundamental to classification capacity and generalization~\cite{cohen2020separability}, with better linear separability in hidden layers demonstrating improved robustness and performance~\cite{verma2019manifold}.

\vspace{0.2em}
\noindent
\hangindent=0.4cm
\textbf{[T3]} \textbf{Homogeneity}: Determine whether topologically coherent clusters correspond to semantically meaningful (single-class) groupings, or whether they incorrectly merge multiple classes.
% This task composes \textit{Cluster} (identify persistent components), \textit{Retrieve Value} (obtain ground-truth class labels for points within each cluster), \textit{Compute Derived Value} (cluster purity/homogeneity metrics), and \textit{Characterize Distribution} (of class composition within clusters).
Non-homogeneous clusters suggest the network has learned representations that conflate distinct concepts, as compact intra-class clusters are predictive of better generalization~\cite{carbonnelle2021intraclass}, reduce classification capacity requirements~\cite{cohen2020separability}, and improve robustness~\cite{verma2019manifold}.
The need to assess class separability (\textbf{[T2]}) and cluster homogeneity (\textbf{[T3]}) comes from studies showing that well-separated, compact intra-class clusters in activation space are predictive of better generalization and robustness~\cite{cohen2020separability, verma2019manifold, carbonnelle2021intraclass}.

\begin{table}[!b]
  % \centering
  \caption{Each analysis task (\textbf{[T1]}--\textbf{[T4]}) decomposed into the low-level analytic primitives of Amar et al.~\cite{amar2005lowlevel}, and the HOLE views that support it, indicated by bullets~($\bullet$). The \vcflow{cluster flow} view comprises the stacked-area, compact stacked-bar, and Sankey renderings (\cref{subsec:vis-design}).}
  \label{tab:task-primitive-vis}

    \resizebox{\linewidth}{!}{%
  \footnotesize
  \setlength{\tabcolsep}{4.5pt}
  \renewcommand{\arraystretch}{1.15}
  \begin{tabular}{@{}l l *{9}{p{4pt}} p{10pt} @{\hspace{1.5em}} *{2}{p{4pt}} p{30pt}}
    \toprule
    & & \multicolumn{10}{c}{\textbf{Amar et al.\ analytic primitives}} & \multicolumn{3}{c}{\textbf{Views}} \\
    \cmidrule(lr){3-12}\cmidrule(l){13-15}
    \textbf{Task} & &
      \rotatebox{70}{Retrieve Value~} &
      \rotatebox{70}{Filter~} &
      \rotatebox{70}{Compute Derived Value~} &
      \rotatebox{70}{Find Extremum~} &
      \rotatebox{70}{Sort~} &
      \rotatebox{70}{Determine Range~} &
      \rotatebox{70}{Characterize Distribution~} &
      \rotatebox{70}{Find Anomalies~} &
      \rotatebox{70}{Cluster~} &
      \rotatebox{70}{Correlate~} &
      \rotatebox{70}{\vcflow{Cluster Flow}~} &
      \rotatebox{70}{\vblob{Blob Graph}~} &
      \rotatebox{70}{\vhmd{Heatmap Dendrogram}~} \\
    \midrule
    \textbf{[T1]} & Hierarchy    &  &  &  &  & $\bullet$ & $\bullet$ & $\bullet$ &  & $\bullet$ &  & $\bullet$ &  & $\bullet$ \\
    \textbf{[T2]} & Separability &  &  & $\bullet$ &  &  & $\bullet$ & $\bullet$ &  & $\bullet$ &  & $\bullet$ & $\bullet$ & $\bullet$ \\
    \textbf{[T3]} & Homogeneity  & $\bullet$ &  & $\bullet$ &  &  &  & $\bullet$ &  & $\bullet$ &  &  & $\bullet$ &  \\
    \textbf{[T4]} & Outliers     & $\bullet$ & $\bullet$ &  &  &  &  & $\bullet$ & $\bullet$ &  &  &  & $\bullet$ &  \\
    \bottomrule
  \end{tabular}%
  }
\end{table}

\vspace{0.2em}
\noindent
\hangindent=0.4cm
\textbf{[T4]} \textbf{Outliers}: Identify data points with anomalous topological behavior---those that remain isolated or merge very late in the filtration---which may indicate poorly learned representations for specific inputs.
% This task composes \textit{Find Anomalies} (identify outlying points), \textit{Filter} (select late-merging or low-persistence components), \textit{Retrieve Value} (obtain persistence/death times), and \textit{Characterize Distribution} (of merge times to establish what constitutes ``late'').
Prevalent outliers suggest suboptimal model training or dataset issues, and detecting such out-of-distribution examples is critical for deployment safety and reliability, as models frequently fail silently on outliers~\cite{hendrycks2017baseline}.
The need to surface outliers is motivated by empirical observations that deployed models frequently fail silently on atypical or poorly-represented inputs~\cite{hendrycks2017baseline}.
\vspace{0.2em}

We ground these tasks in the low-level analytic task taxonomy of Amar et al.~\cite{amar2005lowlevel}.
%, which identified ten primitive operations through empirical analysis of nearly 200 data analysis questions: \textit{Retrieve Value, Filter, Compute Derived Value, Find Extremum, Sort, Determine Range, Characterize Distribution, Find Anomalies, Cluster}, and \textit{Correlate}.
% These primitives form a foundational vocabulary for describing analytic activities in information visualization systems.
%Building upon this foundation, w
Each of our four higher-level tasks
%specifically designed for topological analysis of neural network activations.
% Each task 
represents a composition of multiple Amar et al.\ primitives, applied in the context of PH filtration to address the unique analytical needs of neural network interpretability.
Unlike traditional static data analysis, our tasks operate on \textit{dynamic topological features} that evolve across filtration scales, requiring coordinated application of multiple primitive operations to track hierarchical structure, assess representation quality, and identify anomalous patterns in learned representations.
\Cref{tab:task-primitive-vis} summarizes, for each task, the Amar et al.\ primitives it composes and the HOLE views that support it.
% \PR{we still need some description of how you decided these were the important tasks. this could be interviews with experts, review of literature, etc. the easiest thing would be to find a state of the art report that lists out tasks people want to perform on CNNs. right now you have some citations sprinkled in. you could also use those.}

    % \item \textbf{[T5]} \textbf{Comparative Analysis}: Certain visualizations are more suited for comparitive analysis. It's much harder to compare and interpret two persistence diagrams or barcodes.

\subsection{Cluster Quality Metrics}
\label{sec:cluster_quality_main}
Underlying all of HOLE's visualizations is a pair of cluster-quality metrics that measure how well the $H_0$ clusters at a given filtration scale align with the ground-truth classes: they are the basis on which every view selects the scale it foregrounds. For cluster assignments $\mathcal{C}$ and true labels $\mathcal{T}$ over $N$ points, \emph{purity} is the fraction of points lying in a cluster whose majority label is their own,
\[
\mathbb{P}(\mathcal{C}, \mathcal{T}) = \frac{1}{N} \sum_{k} \max_j |C_k \cap T_j|,
\]
and \emph{homogeneity} is the fraction of points whose class is captured by a single cluster,
\[
\mathbb{H}(\mathcal{C}, \mathcal{T}) = \frac{1}{N} \sum_{j} \max_k |T_j \cap C_k|,
\]
where $C_k$ denotes cluster $k$ and $T_j$ true class $j$. The two are complementary: purity is low when a cluster mixes classes, whereas homogeneity is low when a single class is split across clusters, so high values of both indicate a clean correspondence between clusters and classes. HOLE combines them into a single quality score $\mathbb{Q}(\mathcal{C}, \mathcal{T}) = \alpha\,\mathbb{P}(\mathcal{C}, \mathcal{T}) + \beta\,\mathbb{H}(\mathcal{C}, \mathcal{T})$, with weights $\alpha, \beta \ge 0$ and $\alpha + \beta = 1$; across our experiments we found $\alpha = 0.7$ and $\beta = 0.3$ to give the best balance between the two. HOLE selects the scale that maximizes $\mathbb{Q}$ over the filtration: the \vblob{blob} and contour graphs are drawn there, and the intermediate stages of the \vcflow{cluster flow} views are chosen around it. We formally define these metrics and prove the bounds, monotonicity, and complementary extremes that make this selection well-posed in \appRef{sec:cluster_quality}.

\subsection{Visualization Design}
\label{subsec:vis-design}
% \todo[inline]{Describe the design and rationale for the Sankey flow diagram visualization, including how it summarizes cluster evolution across filtration stages and supports interpretability tasks.}
% \PR{make this a subsection instead of a subsubsection}

% Based upon the analysis tools available (PH), 
Using PH as a basis for analysis,
we have developed three visualization strategies, customized from common visualization types, to address the analysis tasks we have identified.
All visualizations operate on a common input: a set of activation vectors extracted from a chosen network layer for a labeled probe set (e.g., 200 CIFAR-10 test images or 150 CoNLL-2003 test sentences), together with the ground-truth class labels.
A pairwise distance matrix is computed over these activations using the selected metric (cosine by default; see \cref{subsec:distance-metrics}), and a VR filtration is built from this matrix.
Each view presents a different facet of the resulting PH.
% \PR{you should describe here what data the visualizations and PH are using (test set w/ ground truth labels)}

% \subsubsection{Visualization Design}
% \marginpar{Why Sankey diagrams?}
\subsubsection{\vcflow{Cluster Flow Visualizations}}
% \PR{inconsistent capitalization. make sure that capitalization is the same in all per usage context.}
Understanding how class-relevant clusters emerge, persist, and merge across the filtration is central to tasks \textbf{[T1]}~(Hierarchy) and \textbf{[T2]}~(Separability).
Intuitively, as the distance threshold grows, nearby points are progressively connected: first, within-class neighbors join, then increasingly distant points, until eventually all data collapse into a single group.
\vcflow{Cluster flow visualizations} address this by showing, at each threshold, which connected components (clusters) exist, how many probe-set samples (colored by their ground-truth labels) belong to each, and how samples transition between clusters as the threshold increases.
We provide three complementary subtypes.
The \vcflow{\textit{stacked area chart}} (\cref{fig:teaser} top-left) gives a continuous overview of the entire filtration, with each colored region representing a cluster and the horizontal axis spanning all death thresholds; it serves as the natural entry point for exploration, and clicking any point selects that threshold for the other views.
The \vcflow{\textit{compact stacked bar chart}} summarizes the same cluster-composition information in less vertical space.
The \vcflow{\textit{Sankey diagram}} (\cref{fig:teaser} top-right) provides a detailed five-stage view: (1)~ground-truth labels, (2)~clusters at the first filtration threshold, (3--4)~the two thresholds whose cluster assignments best overlap with ground truth (selected by $\mathbb{Q}$ score of \cref{sec:cluster_quality_main}), and (5)~the fully merged single component.
This ground-truth-guided selection is an analytical convenience for evaluating known class structure; for exploratory use, users can manually select thresholds via the stacked area chart without reference to labels.
B\'{e}zier curves connect corresponding clusters across stages, with flow width proportional to the number of shared data points, making it immediately visible which classes remain coherent and where merging occurs (see \appRef{sec:sankey_math}).
An optional minimum-size filter suppresses transient singleton components so that dominant class-level flows remain legible (see \appRef{sec:sankey_filtering}). In the interactive dashboard, hovering over a flow reveals its cluster-composition statistics, and selecting a cluster highlights it simultaneously in the \vblob{blob graph} and \vhmd{heatmap dendrogram}, since all views are linked.
% \PR{there is some inconsistency in how you color the text here vs other places in the paper. i prefer coloring the words the way you have here, but the most important thing is that you do it consistently.}
% \PR{I have a bad feeling that a non-TDA person is going to have no idea what this means... not sure how to address that besides adding some more accessible details.}
% \PR{the source of "clusters" is ambiguous here. you should be more clear about what the data are and where their labels come from. this is maybe a more general comment (thus needs a more general introduction) that you're using the testing set (?) with their ground truth labels (?).}
% \PR{you don't discuss the filtering at all here. it is discussed in the dashboard, but that might be too disconnected.}

\subsubsection{\vblob{Blob Graphs}}
% \PR{where is the icon and color?}
Once the \vcflow{cluster flow} view has identified an informative filtration threshold, the user needs to assess whether the topological clusters correspond to semantically meaningful classes (\textbf{[T2]}~Separability, \textbf{[T3]}~Homogeneity) and to surface any poorly represented inputs (\textbf{[T4]}~Outliers).
\vblob{Blob graphs} (\cref{fig:teaser} bottom-left; see \cref{sec:app1} for examples on real data) address this by projecting the same activation vectors used for the filtration into a 2D spatial layout at the user-selected threshold.
Each point represents one probe-set sample, positioned according to its first two principal components via PCA.
PCA is chosen because it is deterministic, linear, and preserves global variance structure---properties that make the layout consistent and comparable across layers, models, and conditions, unlike stochastic methods such as t-SNE or UMAP whose embeddings differ across runs.
PCA preserves only linear variance structure, which may distort non-linear cluster relationships; however, the blob graph's cluster boundaries are derived from the full-dimensional PH, not from the projection, so the PCA layout serves as a spatial scaffold rather than the source of topological truth.
Other dimension reduction methods (e.g., LDA, UMAP) are also supported through the extensible interface (\cref{fig:teaser}).
% \PR{what are the data points?}

Cluster boundaries correspond to the $H_0$ components identified by PH at the selected death threshold $\epsilon^*$.
HOLE provides two variants of the blob graph, each suited to different analysis scenarios.
The \textit{scatter blob graph} plots individual data points colored by their ground-truth class label, with cluster boundaries drawn as alpha shapes or convex hulls around each component (see \appRef{sec:blob_math}).
This dual encoding of points by class and boundaries by topological cluster enables direct visual assessment of cluster--class alignment: a blob dominated by a single color indicates a well-learned, class-coherent representation (\textbf{[T2]}, \textbf{[T3]}), while points outside any blob or absorbed into a differently-colored cluster signal poorly learned representations (\textbf{[T4]}).
This variant is preferred for smaller point clouds or when individual-point inspection is needed.
The \textit{contour blob graph} (\cref{fig:teaser} bottom-left) renders the spatial density of points within each cluster as filled contour regions, with each region colored by its cluster index.
Points that do not belong to any cluster (outliers) are rendered individually, keeping them explicitly visible for outlier detection (\textbf{[T4]}).
This variant is well-suited for dense point clouds, where individual markers would occlude one another, and makes the dominant cluster structure immediately legible.
% \PR{just reiterating our last discussion, if contours become the primary method shown in the paper, you'll need to rearrange this section.}
% \PR{why PCA? more generally when you make decisions like this you need a reason. in this case, you can also provide context by stating the other methods are available.}
% \PR{missing a description of the contour version.}
% \end{itemize}

% https://developer.ibm.com/tutorials/awb-cluster-analysis-in-r/
\subsubsection{\vhmd{Heatmap Dendrograms}}
% Dendrograms is a visualization that helps us better understand hierarchical clustering. In \cref{fig:dendrogram_syn} we visualize the dendrogram alongside the RCM heatmap (heatmap of the pairwise distance of datapoints with Reverse-Cuthill McKee ordering).
% These dendrograms reveal the hierarchical clustering structure at different filtration scales, and enables us to identify critical thresholds where the clusters are formed.
% It also helps us better understand how individual data points or class clusters merge into larger clusters as the filtration parameter increases.
% For the heatmap, we first get our pairwise distance matrix, and then we reorder the matrix based using Reverse-Cuthill McKee ordering.
% Then we build a hierarchical clustering linkage matrix from the persistence death thresholds, which is then reordered based on the RCM data point order to maintain the same data ordering.
% Each major cluster is identified with a color. This enables us to identify critical thresholds where clusters emerge or disappear, and to trace how individual data points merge into larger clusters as the filtration parameter increases.
% Thus, this visualization solves task \textbf{[T1]}, \textbf{[T2]}, and \textbf{[T5]}.
While \vcflow{cluster flow} and \vblob{blob} views focus on individual thresholds or stage-to-stage transitions, users also need a global summary of how the entire hierarchical clustering structure relates to inter-class distances (\textbf{[T1]}, \textbf{[T2]}).
\vhmd{Heatmap dendrograms} (\cref{fig:teaser} bottom-right) address this by pairing a hierarchical dendrogram with a reordered pairwise distance heatmap, both derived from the same distance matrix used for the filtration.
The heatmap displays the full $n \times n$ pairwise distance matrix, reordered with the Reverse Cuthill--McKee (RCM) algorithm~\cite{cuthill1969reducing} to group similar data points adjacently, transforming scattered cluster patterns into visually coherent block-diagonal structures.
This reordering is crucial for interpretability, as it reveals cluster boundaries that would otherwise be obscured by arbitrary data ordering.
We then build a hierarchical clustering linkage matrix from the persistence death thresholds (see \appRef{sec:dendrogram_math}), reordered to maintain consistency with the RCM data point order (see \appRef{sec:matrix_reordering}).
The dendrogram captures how individual data merge into larger class clusters as the distance threshold increases during the filtration process, with each major cluster identified by a color.
Together, the dendrogram and heatmap make it possible to identify the critical thresholds where clusters emerge or disappear, to assess whether block-diagonal structure corresponds to class boundaries (\textbf{[T2]}), and to trace the full hierarchy of merges from individual points to the single connected component (\textbf{[T1]}).
\subsubsection{Interpretability Workflow}
\label{subsec:dashboard}

% \PR{i think this should be a subsubsection of the visualization design.}
% \PR{I don't think this really captures the workflow.}
% \PR{moved this to the end of the section.}
% \PR{maybe a remove this and integrate the the important text into the dashboard section.}
% \PR{you don't describe this visualization at all when you're introducing the visualizations.}
% \PR{a lot of this information repeats from above. i recommend merging most of the above content with 3.4. you can still have a dashboard section, but it can just focus on overall structure and interaction.}

The visualizations are combined in an interactive web-based dashboard (\cref{fig:teaser}) that supports the full HOLE interpretability workflow, from data ingestion to coordinated visual exploration.
The dashboard is implemented as a web application, allowing HOLE to be integrated into existing ML workflows and shared with collaborators without specialized software.
%
%
%
%\paragraph{Exploration}
The dashboard provides a coordinated multi-view interface with four linked panels.
Users select models, layers, and distance metrics through dropdown menus, and the views update together.

% \section{Implementation}

% \subsection{Activations and hook functions}
% \subsection{Persistent Homology}
% \externaldocument{methods}
% \externaldocument{appendix}

\section{Evaluations}

\subsection{Implementation}
HOLE is implemented in Python. 
The user loads a PyTorch model and a dataset, then selects the activation layers to probe.
% HOLE places forward hooks at those layers and extracts activations during a forward pass, forming a high-dimensional point cloud at each layer (see \cref{subsec:dnn-structure}).
The hook functions are implemented using PyTorch's module-level forward hooks, which allow capturing outputs during a forward pass without altering the network architecture.
The input data is fed into the neural network, and the activations are recorded at each layer. 
For a given batch of inputs and a specified network layer, we collect the layer activations and flatten them to form vectors in $\mathbb{R}^d$, where $d$ is the number of output units in that layer. 
The resulting $N \times d$ matrix defines a point cloud of $N$ points in $d$-dimensional space. 
These probed intermediate activations then act as point clouds for the PH pipeline (see \cref{subsec:dnn-structure}).
%, which is then used to study the layer topology. 
We use GUDHI~\cite{maria2014gudhi} to compute the PH
on each point cloud using the user's choice of distance metric (see \cref{subsec:distance-metrics}), tracking $H_0$ features across filtration scales.
% In our experiments, we conducted tests on ResNet-18, ResNet-50, and ViT-based architectures. The activation functions were ReLU and GELU. 
Source code is included for review only and will be open-sourced upon publication (see \appRef{sec:code_availability}).
%, respectively.

% \paragraph{Setup}

\subsection{Datasets and Network Architectures}
We apply HOLE to three architectures across two domains to demonstrate its model- and domain-agnostic capabilities.

\textit{Vision Transformer (ViT) on CIFAR-10:}
All experiments in the main text use \texttt{vit-base-patch16-224-in21k}~\cite{dosovitskiy2020image}, a ViT-B/16 model pre-trained on ImageNet-21k and fine-tuned on CIFAR-10 (test accuracy 96.0\%).
CIFAR-10~\cite{Krizhevsky2009LearningML} consists of 60,000 32$\times$32 color images evenly distributed across 10 classes (airplane, automobile, bird, cat, deer, dog, frog, horse, ship, truck), with a 50,000/10,000 train/test split.
Fine-tuning uses 5,000 training images; all HOLE analyses use a balanced probe set of 200 test images (20 per class).

\textit{ResNet-50 on CIFAR-10:}
Additional experiments with ResNet-50~\cite{he2016deep} fine-tuned on CIFAR-10 (test accuracy 89.5\%) are reported in \appRef{sec:rn50_analysis}, using the same dataset and probe set. Fine-tuning uses 15,000 training images.

\begin{figure}[!b]
    \centering

    \begin{subfigure}{0.48\linewidth}
         \centering
        \includegraphics[width=\linewidth]{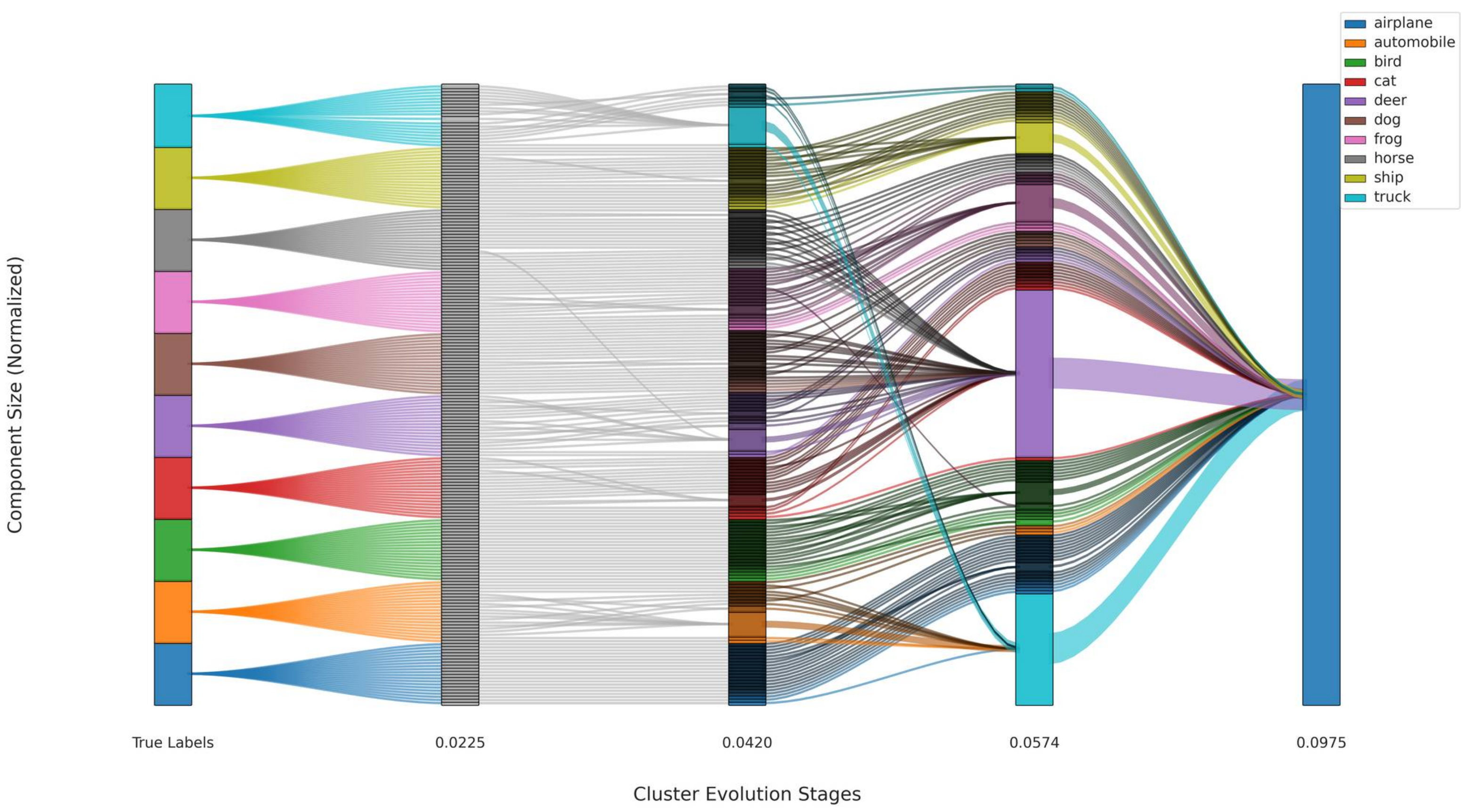}
        \caption{\vcflow{Cluster flow}, \textit{layer~9}. Classes converge rapidly into a single component, indicating weak class separation.}
        \label{fig:sankey_layer_9}
    \end{subfigure}\hfill
    \begin{subfigure}{0.48\linewidth}
        \centering
        \includegraphics[width=\linewidth]{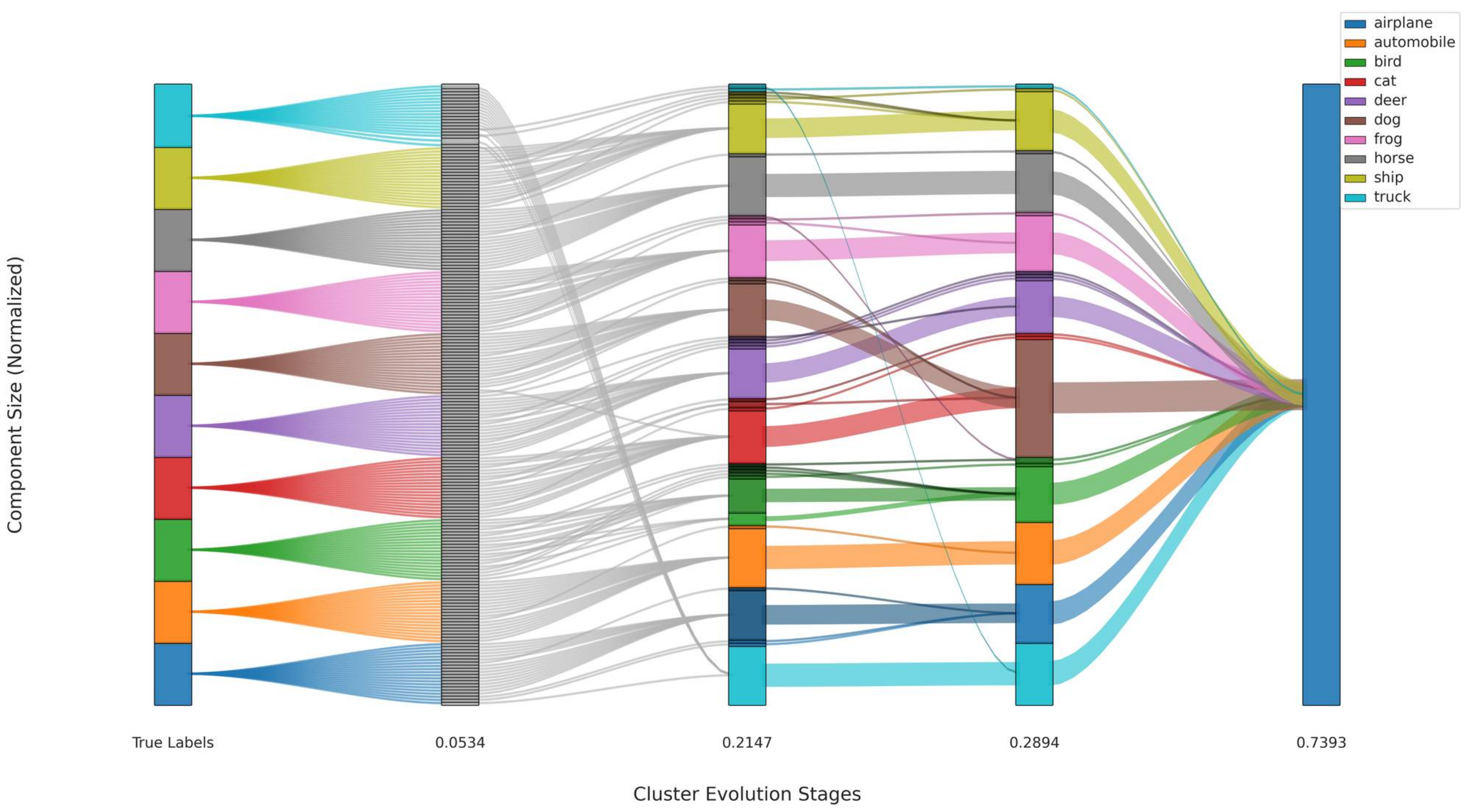}
        \caption{\vcflow{Cluster flow}, \textit{layer~11}. Class flows persist across filtration stages, revealing stronger discriminative structure.}
        \label{fig:sankey_layer_11}
    \end{subfigure}

    \caption{Learned representation analysis for ViT-B/16 on a class-balanced CIFAR-10 sample. Comparing encoder layers~9 (a) and 11 (b) via \vcflow{cluster flow diagrams} reveals the progressive emergence of class-discriminative structure in deeper layers; the corresponding layer-11 \vblob{blob graph} appears as the clean/FP32 baseline of \cref{fig:perturbation_blobs}. Enlarged version in \cref{fig:enlarged_learned_rep}.}
    \label{fig:learned_rep_comparison}
\end{figure}

\textit{BERT-base NER on CoNLL-2003:}
To demonstrate that HOLE generalizes beyond computer vision, we use \texttt{dslim/bert-base-NER}~\cite{devlin2018bert}, a BERT-base model fine-tuned for named entity recognition (collapsed entity-type F1\textsubscript{macro}~91.7\%, 110M parameters).
The CoNLL-2003 English NER corpus~\cite{tjong2003introduction} contains newswire text annotated with four entity types (PER, ORG, LOC, MISC) plus the non-entity label O.
We sample 150 sentences from the test split; token-level embeddings are extracted from each of the 12 encoder layers, with BIO tags collapsed into these five entity types.
Full results are provided in \appRef{sec:bert_ner_analysis}.

% \PR{structure of these paragraphs was a bit confusing. you say 2 discriminative architectures, then talk ViT and Resnet. The following paragraph then talks about BERT. you then jump back to datasets, switching between the vision and nlp tasks.}

% \PR{you need to discuss how much of this data was used in the experiments. if it varies by experiment put it with the experiment.}

% \PR{primary dataset or only dataset? -- RESOLVED: CIFAR-10 is the only CV dataset used; CoNLL-2003 is used for NLP. Current phrasing "The dataset used in our computer vision experiments" already scopes this correctly.}

% \PR{i'd swap the order o paragraphs, architecture first, then dataset. -- RESOLVED: architectures now appear first, datasets second.}

% \end{itemize}

% \subsection{Evaluations}

\subsection{Application 1: Learned Representation Analysis}\label{sec:app1}

% \PR{this comment extends to all of this section, but when you reference a certain visualization type, you should use the color and icon.}
Exploring the model's learned representations is a key aspect of understanding its behavior.
Performing comparative analysis across layers helps understand what layers induce disentanglement of class representations.

\subsubsection{ViT}
Using a class-balanced subset of 200 CIFAR-10 test images (20 per class, drawn uniformly at random with a fixed seed for reproducibility), we compare ViT-B/16 encoder layers~9 and~11 to illustrate how representation quality evolves with depth (\cref{fig:learned_rep_comparison}).
% \PR{which samples? how did you choose them? how many? -- RESOLVED: 200 test images, 20 per class, uniform random class-balanced sampling, np.random.seed(42). Full details in released code (vit_inference_unified.py).}

In an early layer (layer~9) of ViT, the \vcflow{cluster flow diagram} (\cref{fig:sankey_layer_9}) shows no clustering after filtration, indicating that the activation space provides little to no cluster separation at this layer.
In contrast, layer~11 exhibits markedly clearer organization. 
The \vcflow{cluster flow diagram} (\cref{fig:sankey_layer_11}) reveals several cluster formations with class separability, and individual class flows remain coherent across multiple filtration stages, demonstrating that the final encoder layers produce separable activation spaces where inter-class separation substantially exceeds intra-class variation.
The \vblob{blob graph} for this layer (\cref{fig:blob_clean}, shown as the clean baseline of the perturbation comparison) confirms that the classes form compact clusters with visible separation.
This emergence is captured quantitatively by the cluster-quality metrics (\cref{sec:cluster_quality_main}): averaged over three probe-set seeds, the homogeneity of the $H_0$ clustering against the ground-truth classes stays near its floor ($k/N \approx 0.05$) through the early and middle layers, where each class is scattered across many small components, then rises sharply to $0.77$ at layer~10 and $0.87$ at layer~11 as the number of components collapses from roughly $199$ to $33$. Purity remains above $0.95$ throughout, so it is the homogeneity, not the purity, that pinpoints where class-discriminative structure consolidates; full per-layer values appear in \appRef{sec:quantitative_cluster_quality}.
% \PR{!!}
% Additional layer-wise results are provided in the Appendix.

\subsubsection{BERT NER}

To demonstrate that HOLE generalizes beyond computer vision, we apply the same learned-representation analysis to a token-level NLP task: named entity recognition with BERT-base~\cite{devlin2018bert}.
We use \texttt{dslim/bert-base-NER} (110M parameters, 12 encoder layers) fine-tuned on CoNLL-2003~\cite{tjong2003introduction}, collapsing the BIO label set into five entity types (O, PER, ORG, LOC, MISC; macro F1~91.7\%).
We randomly sample 150 sentences from the test split, extract token-level hidden states from all encoder layers, and sub-sample up to 75 tokens per layer with balanced label representation, using cosine distance throughout.

\begin{figure}[!b]
    \centering

    \begin{subfigure}{0.48\linewidth}
        \centering
        \includegraphics[width=\linewidth]{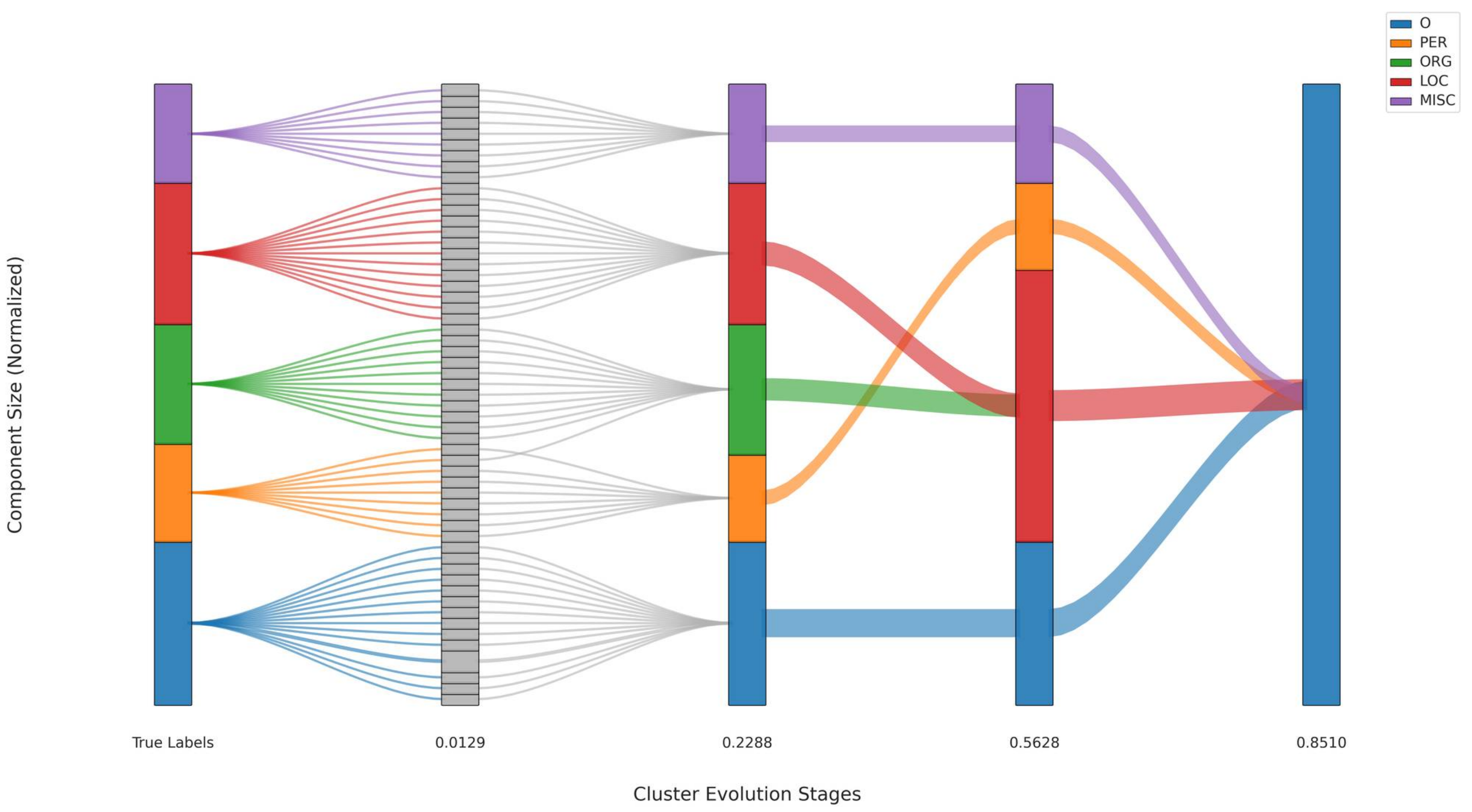}
        \caption{\vcflow{Cluster flow}. Entity-type flows persist coherently across filtration stages.}
        \label{fig:bert_ner_sankey_11}
    \end{subfigure}\hfill
    \begin{subfigure}{0.48\linewidth}
        \centering
        \includegraphics[width=\linewidth]{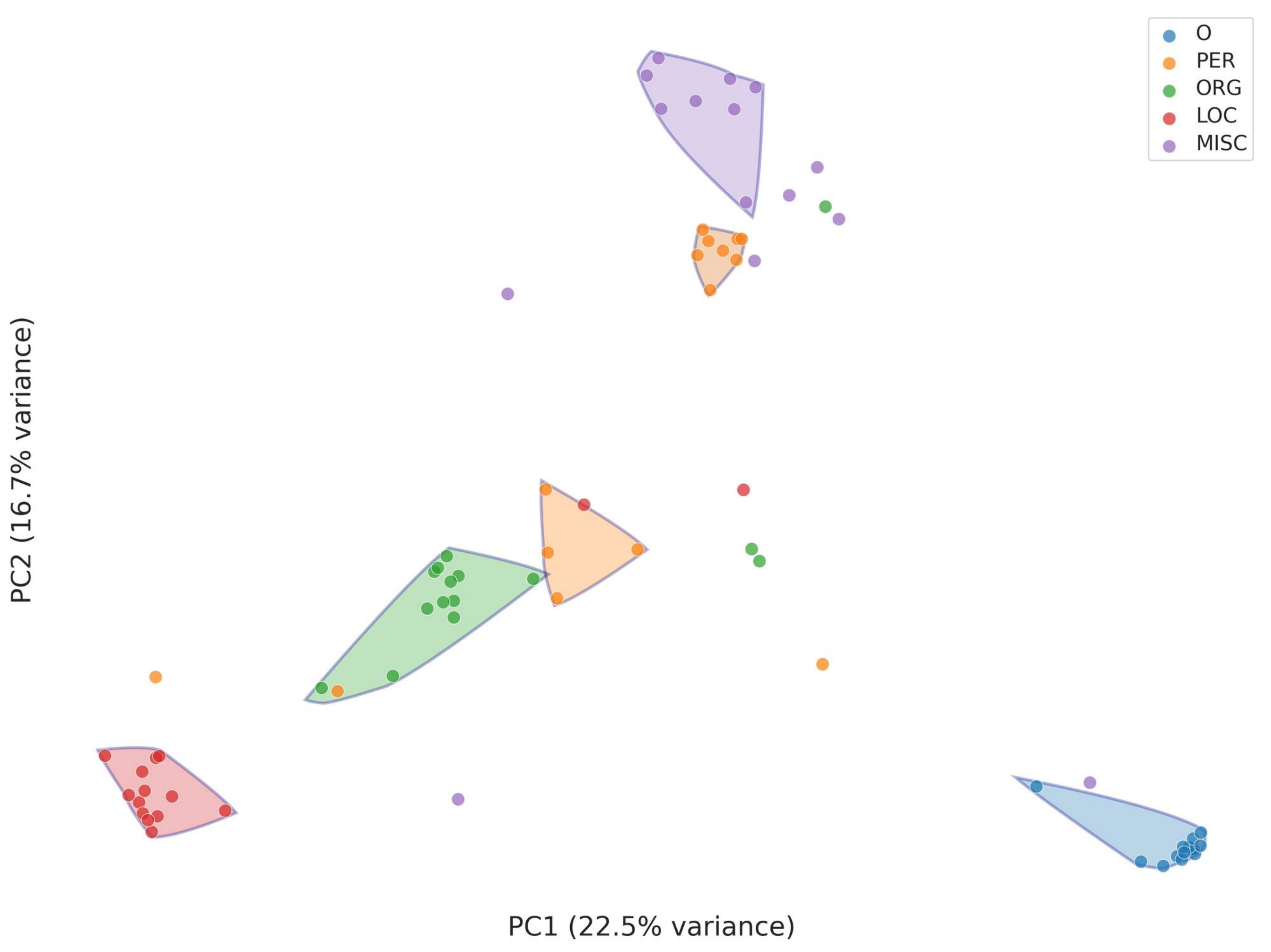}
        \caption{PCA \vblob{blob}. Entity types form compact clusters with separation from the dominant O~class.}
        \label{fig:bert_ner_blob_11}
    \end{subfigure}

    \vspace{0.6em}

    \begin{subfigure}{0.98\linewidth}
        \centering
        \includegraphics[width=\linewidth]{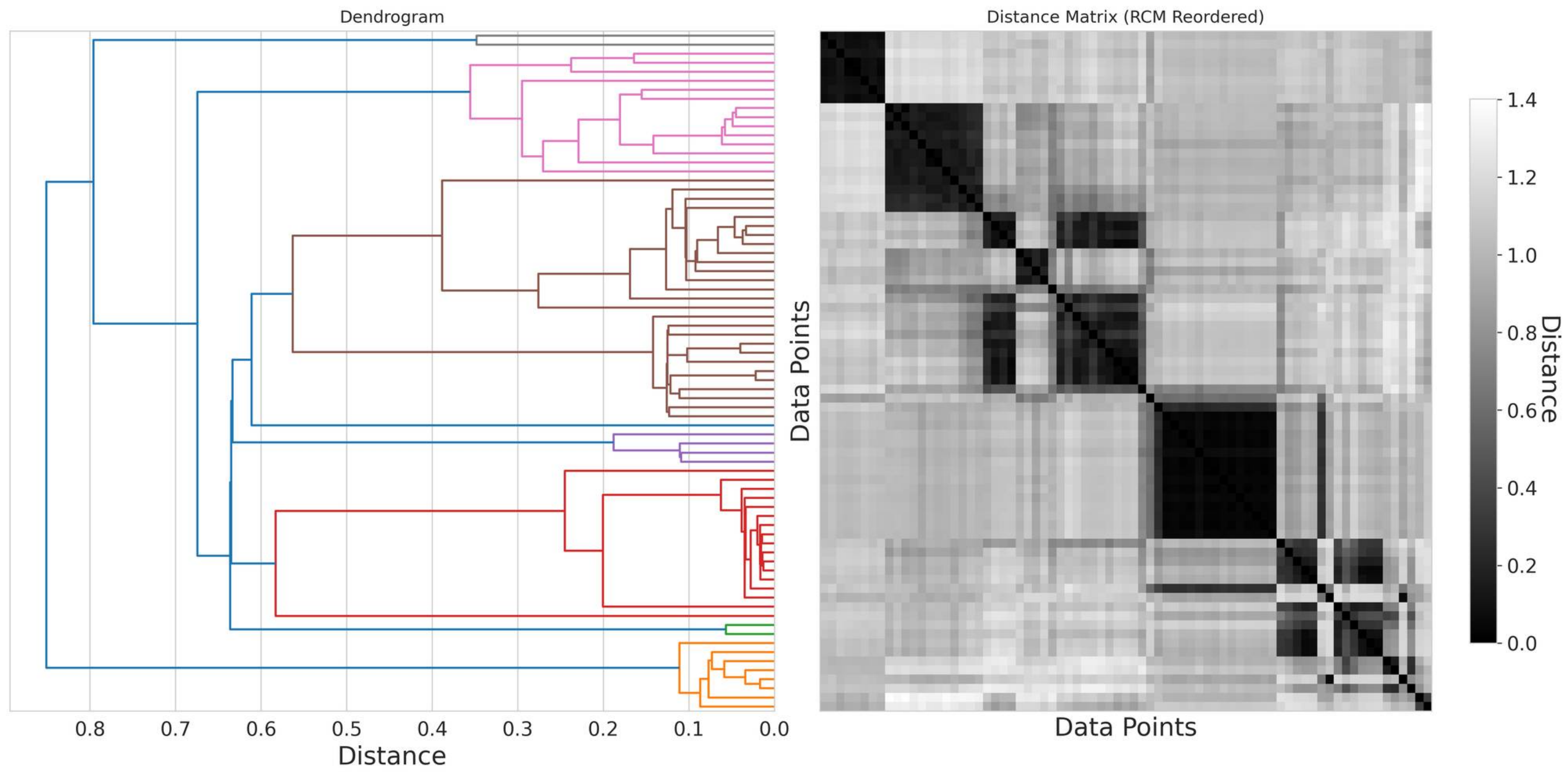}
        \caption{\vhmd{Heatmap dendrogram}. The distance matrix reveals clear block structure among entity types, and the dendrogram shows well-separated branches confirming inter-class separation.}
        \label{fig:bert_ner_hmd_11}
    \end{subfigure}
    \caption{BERT-base NER encoder layer~11 on CoNLL-2003 (cosine distance). The same class-discriminative topology observed in vision models emerges in the final encoder layer for token-level NER, as shown by \vcflow{cluster flow}, \vblob{blob graph}, and \vhmd{heatmap dendrogram}. Enlarged versions in \cref{fig:enlarged_bert_ner}.}
    \label{fig:bert_ner_learned_rep}
\end{figure}

\Cref{fig:bert_ner_learned_rep} shows the final encoder layer (layer~11).
The \vcflow{cluster flow diagram} (\cref{fig:bert_ner_sankey_11}) produces coherent per-type flows that persist across filtration stages, the \vblob{blob graph} (\cref{fig:bert_ner_blob_11}) reveals compact entity-type clusters with visible separation from the dominant O~class, and the \vhmd{heatmap dendrogram} (\cref{fig:bert_ner_hmd_11}) confirms this with clear block-diagonal structure and well-separated dendrogram branches.
Despite the architectural differences, the same class-discriminative topology observed in vision models emerges in the final encoder layer for token-level NER, demonstrating that HOLE is broadly applicable across domains and tasks.
A layer-wise comparison (layers~4 and~11) is provided in \appRef{sec:bert_ner_layers}.

\subsubsection{Implications}
Layer-level topological analysis has direct practical utility.
When selecting which layer to use as a feature extractor for downstream tasks or transfer learning, HOLE can identify the shallowest layer that achieves strong class separation, avoiding unnecessary computation.
It can also serve as a diagnostic when comparing architectures: corresponding results for ResNet-50 (test accuracy 89.5\%, compared with ViT-B/16's 96.0\%) are provided in \appRef{sec:rn50_app1}, where the weaker model exhibits measurably less topological separation, confirming that HOLE captures differences in representation quality across architectures.
The cross-domain BERT NER results further demonstrate that this analysis generalizes beyond vision: the same topological signatures of class-discriminative structure appear in token-level language model activations, confirming that HOLE provides a model- and domain-agnostic diagnostic for representation quality.
% \PR{speculation like this belongs in the discussion section, not in the evaluation. moving this leaves you with the same problem, what do i do with this information?}
% \PR{need a little reflection on why this matters, what it is useful for. might be interesting to contrast this with a model lower quality model or you could show additional levels and discuss how the discriminability improves as you get deeper into the model but at some point plateaus. basically, you could show when the model starts to discriminate.}
% \PR{this could also be a place to put in the experiment about training. here you could show the network at different epochs to show it learns the representation over training. }
% \PR{fwiw, while this is a demonstration of what HOLE can do, it doesn't have any specific utility as defined. \textbf{why would I use it for this?}}

\begin{figure}[!htb]
    \centering
    \begin{subfigure}{0.485\linewidth}
        \centering
        \includegraphics[width=\linewidth]{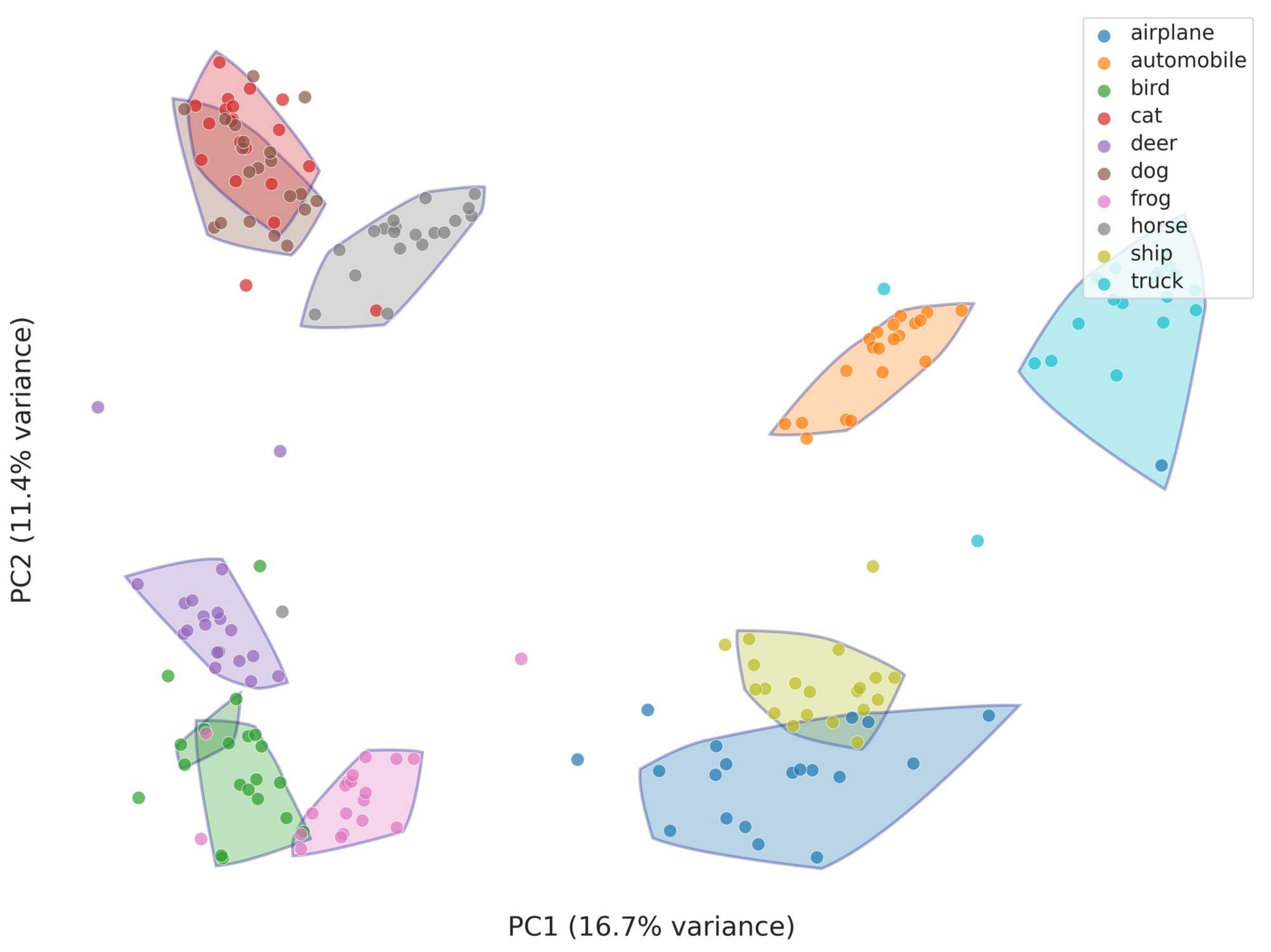}
        \caption{Clean / FP32 baseline. Classes form compact, well-separated clusters.}
        \label{fig:blob_clean}
    \end{subfigure}\hfill
    \begin{subfigure}{0.485\linewidth}
        \centering
        \includegraphics[width=\linewidth]{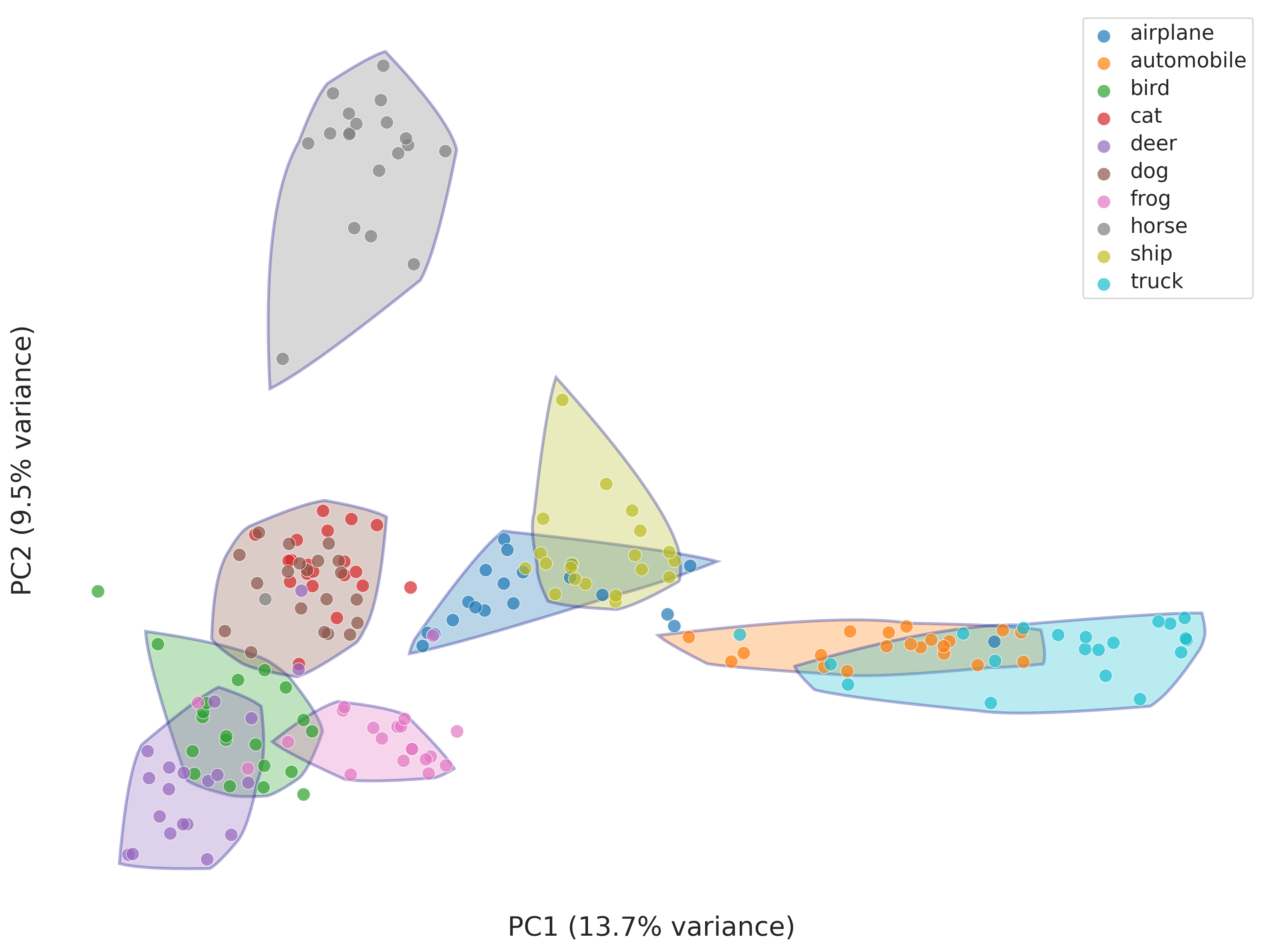}
        \caption{Application~2: Gaussian noise. Cluster boundaries blur; coarse structure is preserved.}
        \label{fig:blob_gaussian}
    \end{subfigure}

    \vspace{0.6em}

    \begin{subfigure}{0.485\linewidth}
        \centering
        \includegraphics[width=\linewidth]{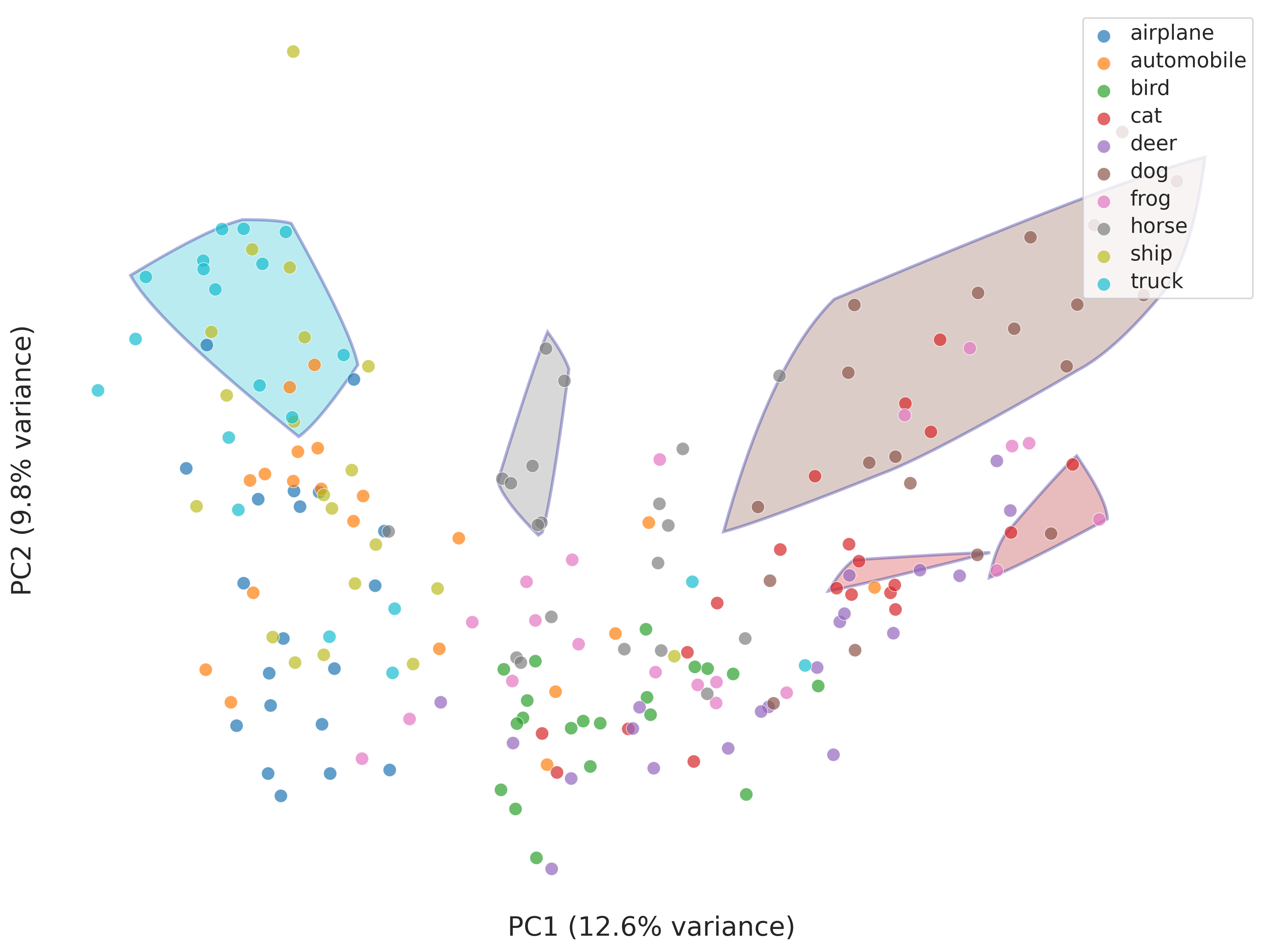}
        \caption{Application~2: Salt \& Pepper noise. Clusters expand and overlap substantially across classes.}
        \label{fig:blob_sp}
    \end{subfigure}\hfill
    \begin{subfigure}{0.485\linewidth}
        \centering
        \includegraphics[width=\linewidth]{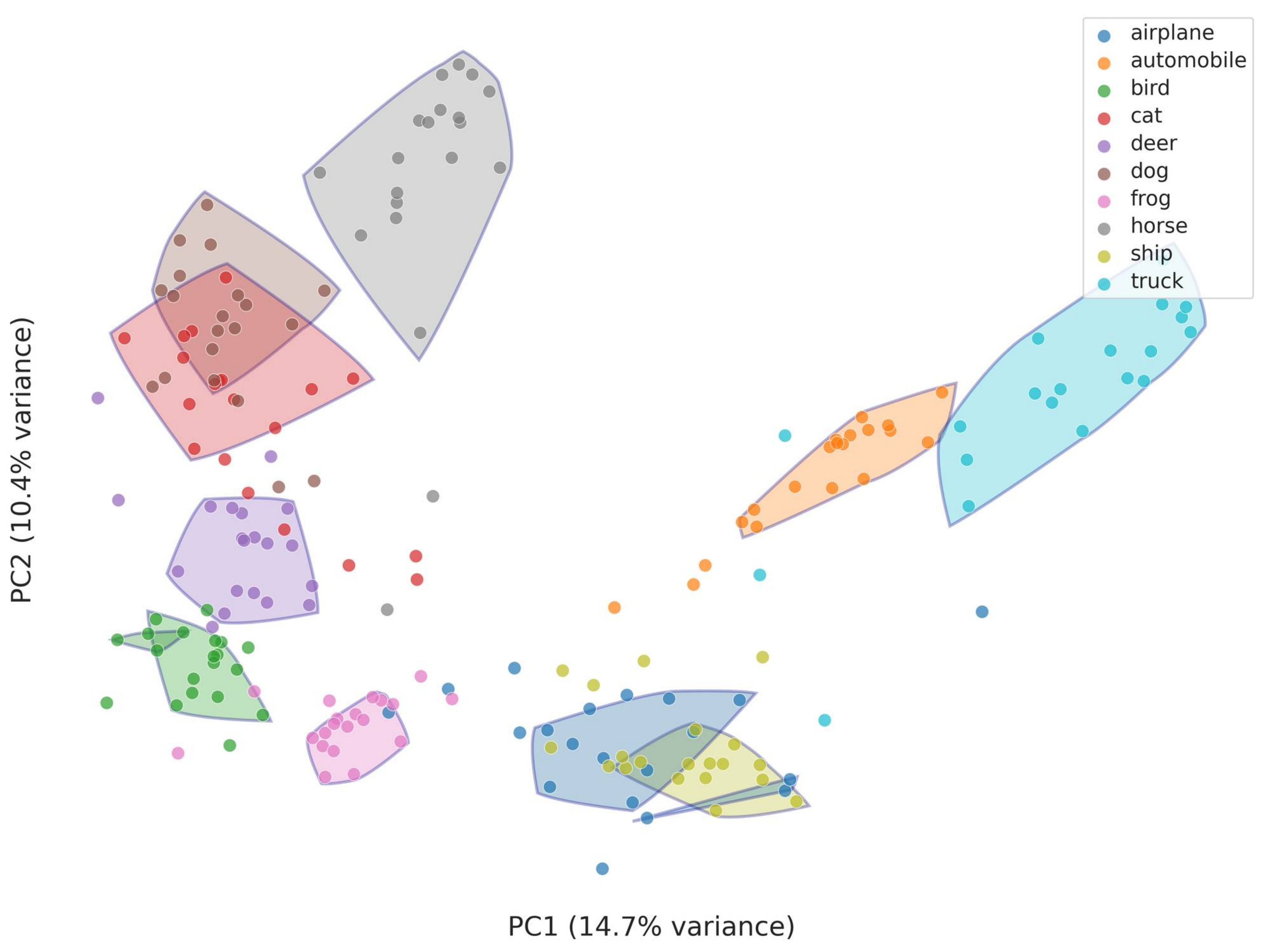}
        \caption{Application~3: INT8 quantized. Clusters fragment and overlap; inter-class margins shrink.}
        \label{fig:blob_int8}
    \end{subfigure}
    \caption{PCA \vblob{blob visualizations} of ViT-B/16 encoder layer~11 activations (class-balanced CIFAR-10, cosine distance). A shared clean/FP32 baseline (a) is compared against two input-noise perturbations (b, c; Application~2) and INT8 weight quantization (d; Application~3). Cluster compactness and class separation deteriorate progressively under each perturbation. Enlarged versions in \cref{fig:enlarged_noise_blobs,fig:enlarged_quant_blobs}.}
    \label{fig:perturbation_blobs}
\end{figure}

\subsection{Application 2: Robustness Analysis Under Noise}\label{sec:app2}

Understanding how noise corrupts a model's internal representations, rather than only its final outputs, is essential for real-world deployment.
We can use HOLE to understand how these model perturbations affect the stability and degradation of topological features.
In this example, ViT-B/16 encoder layer~11 is evaluated under two noise variations: Gaussian (additive, zero-mean) and Salt \& Pepper (impulse noise, 10\% pixel corruption).
Examples of all noise types applied to CIFAR-10 are shown in \cref{fig:all_noise_types}, along with technical specifications in \appRef{sec:appendix_robustness}.
The clean baseline representations are characterized in \cref{fig:learned_rep_comparison}.

\subsubsection{Filtration Flow Under Noise}
The \vcflow{cluster flow diagrams} in \cref{fig:perturbation_sankeys} show how topological features persist across filtration scales under each noise condition.
For the clean baseline, \cref{fig:sankey_layer_11} (Application~1) shows each class maintaining a coherent, color-coded flow that persists across several filtration stages.
Under Gaussian noise (\cref{fig:sankey_gaussian}), the class flows still separate from one another at early thresholds but merge sooner than in the clean case, indicating that the noise shrinks the effective inter-class distances in activation space.
Under Salt \& Pepper noise (\cref{fig:sankey_sp}), the flows degrade: classes collapse into a small number of large components much earlier in the filtration and individual class trajectories become difficult to isolate, signaling severe disruption of the inter-class geometry.

\subsubsection{Cluster Structure Under Noise}
The \vblob{blob visualizations} in \cref{fig:perturbation_blobs} corroborate the filtration analysis by showing the spatial layout of the layer~11 activation space.
Under clean inputs (\cref{fig:blob_clean}) the ten CIFAR-10 classes form compact, well-separated clusters in the projection.
Gaussian noise (\cref{fig:blob_gaussian}) introduces moderate blurring of cluster boundaries; however, most classes remain individually identifiable, though several visually similar classes (e.g., cat/dog, automobile/truck) begin to overlap, consistent with the earlier merging observed in the \vcflow{cluster flow diagram}.
Salt \& Pepper noise (\cref{fig:blob_sp}) causes substantially more damage: clusters expand, boundaries blur across many class pairs, and the overall layout becomes less organized, matching the rapid flow collapse seen in \cref{fig:sankey_sp}.

% \PR{the prior 2 paragraphs seem out of order. aren't we supposed to use the sankey to then investigate the blobs?}

\subsubsection{Implications}
These results show that topological analysis captures representational degradation at a finer granularity than aggregate performance metrics.
Clean inputs achieve 96.0\% accuracy (macro F1 0.960).
Under Gaussian noise, accuracy drops only moderately to 91.5\% (macro F1 0.914), and inter-class separation erodes only slightly while the global cluster topology remains largely intact.
By contrast, Salt \& Pepper noise---despite affecting only 10\% of pixels---causes accuracy to plummet to 62.0\% (macro F1 0.632), a drop of 34.0 percentage points.
This disproportionate degradation suggests that impulsive, high-contrast perturbations are particularly disruptive to the feature geometry learned by the transformer.
Such layer-level, representation-space diagnosis can guide targeted defenses such as noise-specific augmentation or robustness-aware fine-tuning.
% \PR{it would be really great if you showed the results after fine tuning. we may not have time for that, but it would demonstrate the utility more.}
Analogous results for ResNet-50, which exhibits substantially more severe topological degradation under both noise types, are provided in \appRef{sec:rn50_app2}.

\subsection{Application 3: Model Compression}\label{sec:app3}
% TODO: Add contour version here.
Deploying neural networks in resource-constrained settings requires compression techniques such as quantization and pruning, which reduce memory and compute at the cost of altering model parameters.
A critical but often overlooked question is whether these modifications preserve the \emph{topology} of learned representations---structural changes that may go undetected by standard accuracy benchmarks.
We evaluate INT8 dynamic quantization on ViT-B/16, which reduces linear-layer weight precision to 8-bit integers at inference time while retaining full-precision activations.
Details of the compression methods and accuracy benchmarks are in \appRef{sec:compression_experiments}.

\subsubsection{Filtration Flow Under Quantization}
\begin{figure}[!t]
    \centering
    \begin{subfigure}{0.48\linewidth}
        \centering
        \includegraphics[width=\linewidth]{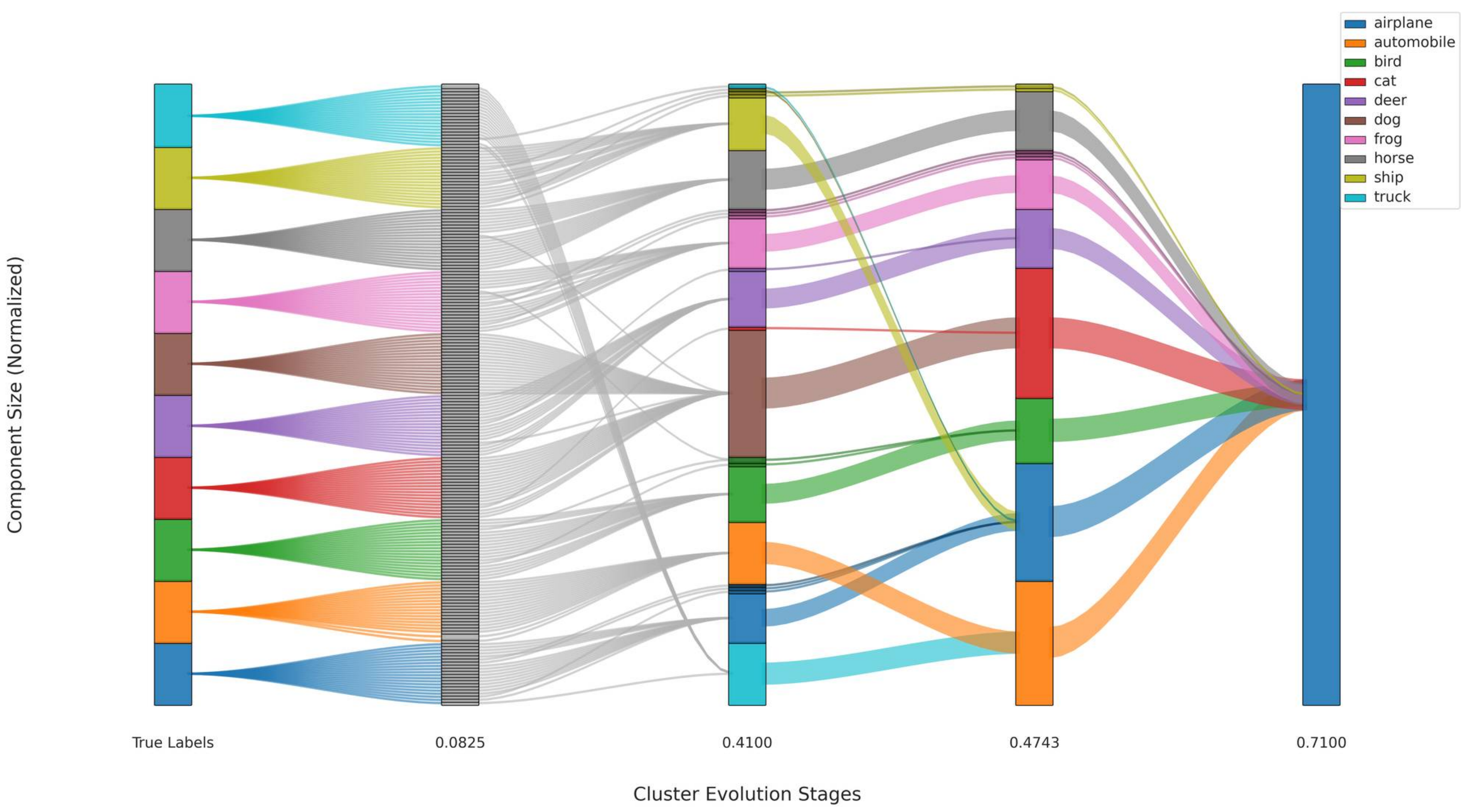}
        \caption{Application~2: Gaussian noise. Class flows remain coherent but cross-class merging occurs earlier than the clean baseline (\cref{fig:sankey_layer_11}).}
        \label{fig:sankey_gaussian}
    \end{subfigure}\hfill
    \begin{subfigure}{0.48\linewidth}
        \centering
        \includegraphics[width=\linewidth]{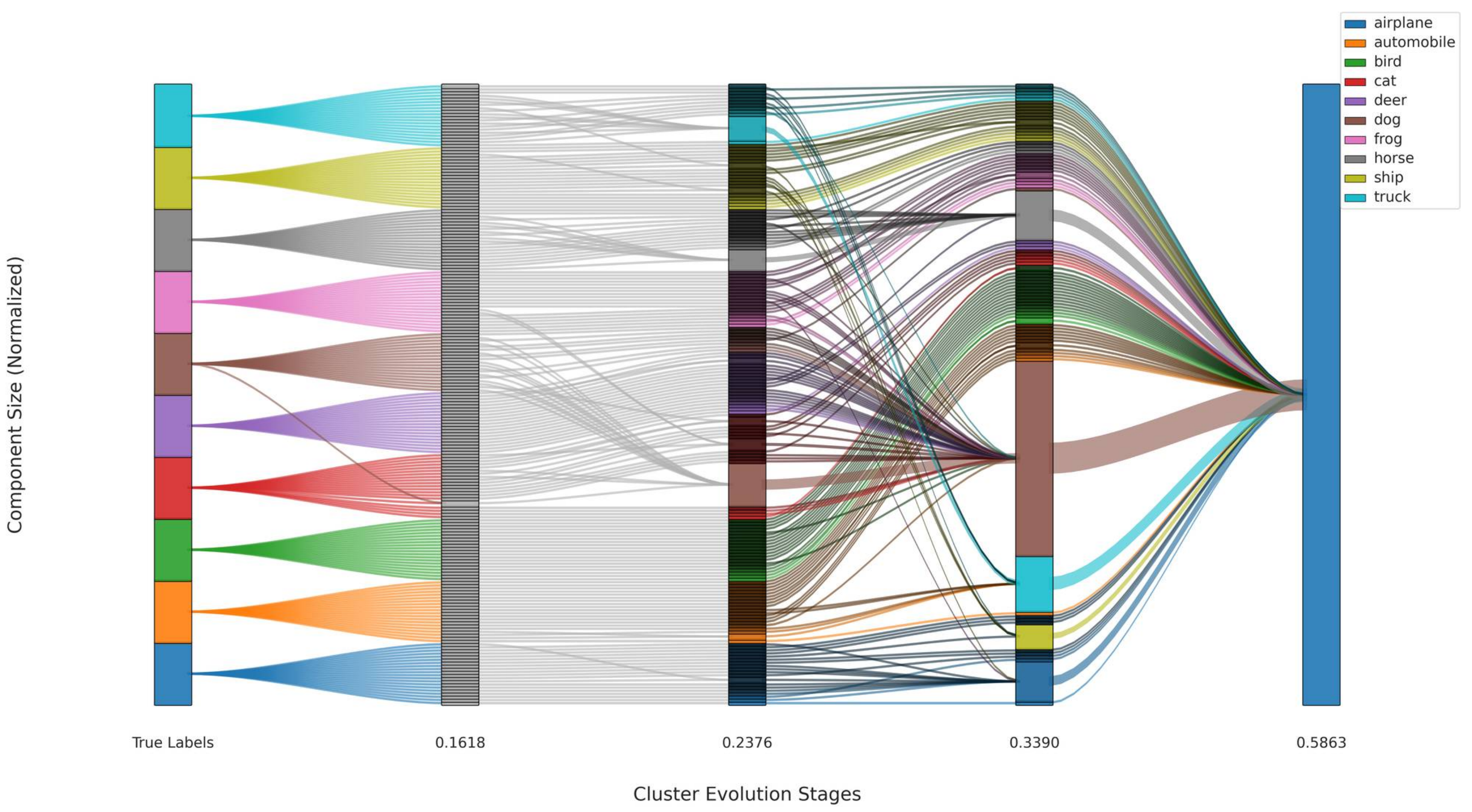}
        \caption{Application~2: Salt \& Pepper noise. Class flows collapse rapidly; most classes merge into a single component by the second filtration stage.}
        \label{fig:sankey_sp}
    \end{subfigure}

    \vspace{0.5em}

    \begin{subfigure}{0.48\linewidth}
        \centering
        \includegraphics[width=\linewidth]{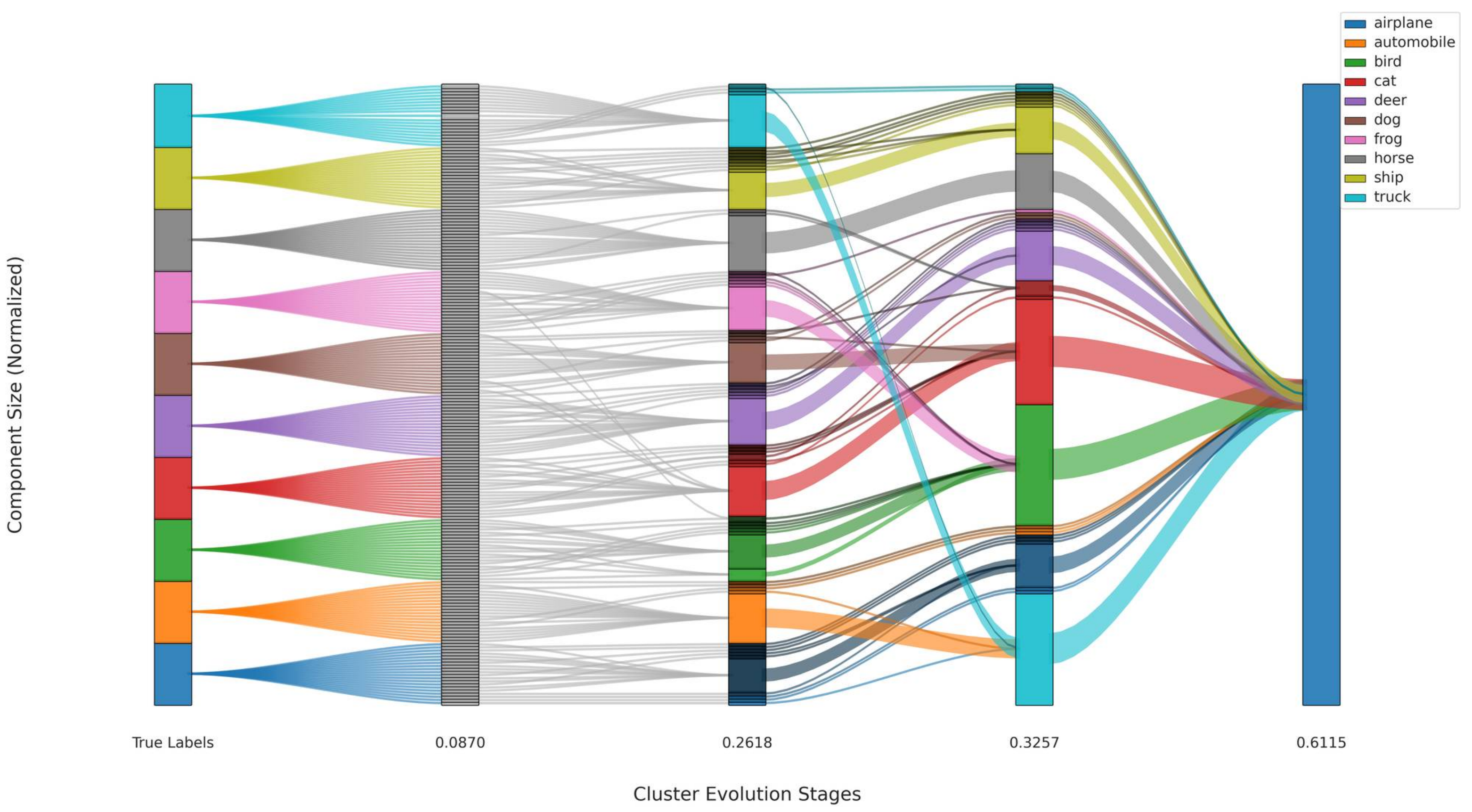}
        \caption{Application~3: INT8 quantization. Class flows fragment and merge prematurely relative to the FP32 baseline (\cref{fig:sankey_layer_11}).}
        \label{fig:sankey_int8}
    \end{subfigure}
    \caption{\vcflow{Cluster flow diagrams} for ViT-B/16 encoder layer~11 under perturbation (class-balanced CIFAR-10, cosine distance): (a-b) input noise from Application~2 and (c) INT8 weight quantization from Application~3. All are compared against the clean/FP32 baseline, which is the layer-11 cluster flow of \cref{fig:sankey_layer_11}. Enlarged versions in \cref{fig:enlarged_noise_sankeys,fig:enlarged_quant_sankeys}.}
    \label{fig:perturbation_sankeys}
\end{figure}

% The INT8 \vcflow{cluster flow diagram} (\cref{fig:perturbation_sankeys}); its uncompressed FP32 baseline is the layer-11 cluster flow already established in Application~1 (\cref{fig:sankey_layer_11}).
In FP32 (\cref{fig:sankey_layer_11}), well-separated per-class flows persist coherently across several filtration stages.
After INT8 quantization (\cref{fig:sankey_int8}), this coherence breaks down: some class flows collapse into shared components earlier in the filtration, while others develop thin, fragmented ribbons---a topological signature of the local neighborhood disruptions introduced by weight snapping and the suppression of low-variance activation directions.

\subsubsection{\vblob{Blob Structure} Before and After Quantization}
The PCA \vblob{blob visualizations} in \cref{fig:perturbation_blobs} corroborate the filtration analysis.
In FP32 (\cref{fig:blob_clean}), class clusters are compact and largely non-overlapping, consistent with the strong inter-class geometry observed in Application~1.
After INT8 quantization (\cref{fig:blob_int8}), several clusters fragment into sub-groups, visually similar classes overlap more extensively, and inter-cluster margins shrink, matching the premature merging seen in the \vcflow{cluster flow diagram}---indicating that weight discretisation disrupts the inter-class geometry at layer~11 (the final encoder layer, and therefore the most discriminative) even though the accuracy drop is only 2.0 percentage points ($96.0\%\to94.0\%$).
This damage is also quantitative: at the class-optimal filtration scale, INT8 lowers the homogeneity of the late layers relative to FP32, e.g., from $0.71$ to $0.61$ at layer~10 and from $0.85$ to $0.80$ at layer~11 (mean over three seeds), while purity stays high and, as noted, accuracy barely changes. The homogeneity therefore exposes a fragmentation of the class clusters that accuracy does not report (\appRef{sec:quantitative_cluster_quality}).
% \PR{is layer~11 the right one to evaluate? perhaps the network becomes more discriminative in later layers?}

% \PR{same comment, shouldn't the prior paragraphs be in the opposite order?}

\subsubsection{Implications}
These results demonstrate that HOLE surfaces representational degradation from compression that would otherwise remain hidden behind a small accuracy change.
The 2.0 percentage point accuracy drop ($96.0\%\to94.0\%$) from INT8, combined with the substantially altered representation topology, suggests that accuracy is an insufficient proxy for compression quality.
Topological diagnostics of this kind can inform deployment decisions---for instance, whether calibration data, quantization-aware fine-tuning, or a less aggressive bit-width is warranted before a model is shipped.
We deliberately chose dynamic INT8 quantization as it is a widely used compression technique in practice.
A complementary result for ResNet-50, where the same INT8 scheme preserves accuracy entirely (89.5\%~$\to$~89.5\%) yet still induces detectable topological shifts, is provided in \appRef{sec:rn50_app3}.
Extending this analysis to other compression techniques such as intermediate bit-widths (e.g., FP16, INT4) or structured pruning is a natural direction for future work.
% \PR{do you have an example with less compression?}
\section{Discussion}

This work presents HOLE (Homological Observation of Latent Embeddings), an approach that applies PH to intermediate activation spaces for interpretable analysis of discriminative neural networks.
By treating layer activations as point clouds and using PH, HOLE reveals structural patterns in learned representations---class separation dynamics, layer-wise feature evolution, and robustness characteristics---that are invisible to traditional attribution-based or neuron-level interpretability methods.
Its model-agnostic design allows consistent application across discriminative architectures, from convolutional networks and vision transformers to encoder-based language models, spanning both vision and natural language processing domains.
% \PR{this description ignores the visualizations that you specifically designed for the tasks.}

To make these topological structures interpretable, HOLE provides a coordinated suite of visualizations, each designed around the analytical tasks identified in \cref{subsec:visualization}.
\vcflow{Cluster flow diagrams}, which consist of Sankey and stacked-area-chart variants, trace how class clusters form, merge, and split across the filtration (\textbf{[T1]}~Hierarchy, \textbf{[T2]}~Separability).
\vblob{Blob graphs} show the spatial layout of clusters via projection, making inter-class overlap (\textbf{[T2]}~Separability), cluster--class alignment (\textbf{[T3]}~Homogeneity), and outliers (\textbf{[T4]}) immediately visible.
\vhmd{Heatmap dendrograms} reveal hierarchical inter-class distance structure, supporting both hierarchy (\textbf{[T1]}) and separability (\textbf{[T2]}) assessment.
Underpinning these views, HOLE adopts cosine distance as its default metric, motivated by the directional nature of learned embeddings~\cite{mikolov2013efficient} and the concentration-of-distances problem in high-dimensional Euclidean spaces~\cite{aggarwal2001surprising}, while also supporting other metric alternatives for different analytical perspectives.
An interactive dashboard links all views to support rapid, coordinated exploration across models, layers, and metrics.
% \PR{ah, here are the visualizaitons. maybe switch around the order so that this comes as the second paragraph or maybe integrates with the first.}

Our three evaluation applications demonstrate the potential of this combined topological and visual approach. 
% \PR{this statement is too strong}
The learned representation analysis (\cref{sec:app1}) shows that HOLE can indicate the layers where class-discriminative topology first emerges, informing feature extractor selection for transfer learning and architectural comparison across models of differing quality.
The robustness analysis (\cref{sec:app2}) reveals that topological signatures capture representational degradation from input noise at a granularity that accuracy alone cannot provide---e.g., distinguishing the moderate erosion caused by Gaussian noise from the severe geometric disruption inflicted by Salt~\&~Pepper perturbations.
The compression analysis (\cref{sec:app3}) shows that even when INT8 quantization preserves most classification accuracy, HOLE exposes substantial topological damage to inter-class geometry, providing a more sensitive diagnostic for deployment decisions.

% \paragraph{Clustering and Dimensionality Reduction.}
% A common way to 

\paragraph{What PH Contributes Beyond Clustering}
As noted in \cref{sec:topological-analysis}, restricting to $H_0$ makes our filtration formally equivalent to single-linkage hierarchical clustering, so the value of the topological framing lies in how that structure is used rather than in the merge tree alone. 
Working with the full filtration lets HOLE treat the birth, death, and persistence of components as first-class quantities, so model quality is read from \emph{how long} class-level components persist and \emph{at which scales} they merge rather than from a single fixed partition; this is precisely what allows the robustness and compression analyses to separate moderate from severe representational damage when accuracy is unchanged. 
% Because the filtration function is a free parameter, 
The same activations can also be examined through several geometric lenses (cosine, Euclidean, Mahalanobis, geodesic, density-normalized), each surfacing different structure, so the choice of filtration is itself expressive~\cite{ballester2025expressivityph}. 
Finally, HOLE's contribution is not the merge tree in isolation but the coordinated, label-aware visual-analytics workflow built upon it, which supports comparison across layers, architectures, noise, and compression in a way that a single dendrogram or a fixed $k$-partition does not.

\paragraph{Relationship to Dimensionality Reduction}
A common way to inspect activation spaces is with low-dimensional projection methods, while HOLE differs by computing the topological summaries on the \emph{full-dimensional} activation point cloud. 
% A common way to inspect activation spaces is to project them to two dimensions with PCA, t-SNE, or UMAP and read off apparent clusters by eye. 
% HOLE differs in where the analysis happens: its topological summaries are computed on the \emph{full-dimensional} activation point cloud, so cluster formation and merging are measured in the space the network actually uses rather than in a projection that discards most of it. 
This distinction matters because low-dimensional embeddings can mislead, e.g., t-SNE and UMAP are stochastic and highly sensitive to hyperparameters such as perplexity and neighborhood size, making their apparent clusters and densities artifacts of projection. 
% PCA retains only linear variance, while t-SNE and UMAP are stochastic and highly sensitive to hyperparameters such as perplexity and neighborhood size, so their apparent clusters and densities can be artifacts of the projection and can change from one run to the next. 
Because HOLE's clusters are defined by the filtration in the original space, they do not inherit these distortions, and its deterministic summaries remain directly comparable across layers, models, noise, and compression.
HOLE does use PCA to position the points within the \vblob{blob graph}, so the topological conclusions are independent of the projection.
% HOLE does use PCA within the \vblob{blob graph}, but only as a spatial scaffold for display: the cluster boundaries themselves are derived from the full-dimensional PH, so the topological conclusions are independent of the projection.

\subsection{Limitations and Future Directions}

While HOLE provides a principled topological lens for neural network interpretability, the current work has several limitations that suggest directions for future investigation.

\paragraph{Scope of Topological Analysis}
We restrict HOLE to $H_0$ because class-discriminative structure in activation spaces is reflected in cluster formation and merging.
$H_1$ (loops) or higher-order features could in principle detect topological complexity within class clusters, but we have not observed cases where this provides interpretive value for discriminative models.

\paragraph{Scope of Architectures}
% Our diverse experimental setup (\cref{sec:bert_ner_analysis}) demonstrates that HOLE generalizes beyond vision models to token-level NLP tasks.
Our experimental setup (\appRef{sec:bert_ner_analysis}) demonstrates that HOLE generalizes across different discriminative architectures.
The current evaluation does not cover generative architectures where sequential dependencies introduce additional complexity.
% Extending the similar topological analysis methodologies to such other models, latent spaces of variational autoencoders~\cite{kingma2013auto} and generative adversarial networks~\cite{goodfellow2014generative} could provide insights into mode collapse, sample diversity, and generation quality. 
Extending similar topological analysis methodologies to such models could provide insights into mode collapse, sample diversity, and generation quality.
% \PR{maybe be more explicit--is this analysis methodology even relevant to non-discriminative architectures?}

\paragraph{Training Dynamics}
Extending the learned representation analysis to track how representations evolve during training, like identifying at which epoch class-separating topology first emerges, could provide a topological lens on training dynamics and inform early stopping or design decisions.

% \paragraph{Distance metric assumptions.}
% Several of the available distance formulations assume Euclidean geometry in activation space, which may not faithfully reflect the intrinsic structure of the underlying space, particularly with respect to learned decision boundaries.
% Further investigation into what these metrics capture---and how they relate to the decisions the model is making---could yield more meaningful topological signatures. Geodesic distances on learned manifolds and density-adaptive metrics that account for local neighborhood structure~\cite{lou2020neural} are promising directions.

\paragraph{Filtration Threshold Sensitivity}
The scales at which HOLE renders its views are not hand-tuned but selected by the purity--homogeneity score (\cref{sec:cluster_quality_main}), and the persistence stability theorem~\cite{cohen2007stability} bounds how much the diagram can move under small perturbations of the probe set.
The highlighted thresholds nonetheless depend on this selection: when two filtration scales score almost equally the chosen intermediate views can shift between them, and the label-aware criterion is unavailable in fully unsupervised settings, where a heuristic such as the largest persistence gap would be required.
A systematic sensitivity analysis over $\epsilon$, quantifying how cluster quality and the resulting visual conclusions vary along the filtration, is a natural quantitative extension.

% \paragraph{Adversarial robustness.}
% Our robustness experiments are limited to random noise perturbations.
% Systematic investigation of topological changes under adversarial perturbations~\cite{szegedy2013intriguing} may reveal fundamental principles of adversarial vulnerability and inform topology-based defense mechanisms. 
% \PR{not bad, but also could be cut for space, as it is not a direct limitation.}

\paragraph{Scalability and Probe-Set Size}
% Our experiments use class-balanced probe sets of 200 CIFAR-10 images (20 per class) and up to 75 BERT tokens.
% This size is a deliberate design choice rather than a limitation of the method: because HOLE analyses the \emph{topological} structure of activation spaces (connected components, merge thresholds, and hierarchical relationships) rather than estimating distributional statistics, it does not require the large sample counts needed for density estimation or statistical testing.
% With 20 points per class, the birth and death thresholds of class-level clusters are already stable, and the persistence stability theorem~\cite{cohen2007stability} guarantees that small perturbations in the probe set produce bounded changes in the persistence diagram.
The persistence stability theorem~\cite{cohen2007stability} guarantees that small perturbations in the probe set produce bounded changes in the persistence diagram.
We verify this empirically in \appRef{sec:stability_analysis}, where repeating the analysis across 10 different random seeds produces consistent topological conclusions, with filtered layer~11 cluster counts matching the number of ground-truth classes (9--10 out of 10) in every trial.
% We further test robustness to \emph{distribution shift} and \emph{uneven sampling} in \cref{sec:shift_imbalance}: under CIFAR-10-C corruptions the layer-11 class structure degrades by fragmenting rather than mixing, in a manner that tracks the accuracy drop, and under imbalanced probe sets well-represented classes retain their clusters while starved classes gracefully drop out.
We further test robustness to \emph{distribution shift} and \emph{uneven sampling} in \appRef{sec:shift_imbalance}.
Regarding scalability, the aggregate views (\vcflow{cluster flow} and \vblob{blob graphs}) are constrained primarily by the number of classes rather than the number of data points, and scale comfortably to larger probe sets.
% The evaluation of these bounds are shown in \cref{sec:scalability_eval}, where we profile the pipeline along probe-set size (to $2000$ samples), number of classes (to the $1000$ classes of ImageNet-1k), and network width (to the $1408$-dimensional EVA-Giant): runtime is dominated by the quadratic dependence on probe-set size, network width contributes only a linear, sub-dominant term, and the aggregate views scale with the number of classes rather than with the size of the network.
The evaluation of these bounds is shown in \appRef{sec:scalability_eval}, where we profile the pipeline along probe-set size, class count, and network width; runtime is dominated by the quadratic dependence on probe-set size, and the aggregate views scale with the number of classes rather than the size of the network.

% In contrast, the \vhmd{heatmap dendrogram} faces both computational and visual scalability limits: the pairwise distance matrix is quadratic in the number of samples, and beyond approximately 500 points the dendrogram becomes visually occluded, reducing individual merge events to indistinguishable color bands rather than traceable branches.
% Similarly, larger probe sets (beyond approximately 250 samples) introduce numerous transient singleton components in the \vcflow{cluster flow} views, requiring more aggressive noise filtering to maintain legibility.
% For production use-cases requiring larger sample counts, coreset approximations~\cite{feldman2020turning} that maintain topological fidelity while reducing the effective sample count, or hierarchical aggregation strategies, would help extend these analyses.
% \PR{this reminds me, did you talk about the number of testing inputs you used? did you test the actual scalability at all? this is a question that will come up with reviewers.}

These limitations notwithstanding, PH offers a mathematically grounded framework for understanding and validating model behavior, bridging the gap between theoretical understanding and practical model analysis.

\setstretch{1}

\bibliographystyle{IEEEtran}

\bibliography{main}

\begin{IEEEbiography}
[{\includegraphics[width=1in,height=1.25in,clip,keepaspectratio]{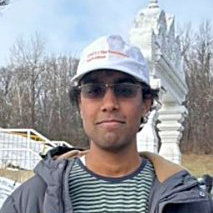}}]{Sudhanva Manjunath Athreya} is currently working toward the PhD degree with the Kahlert School of Computing, University of Utah, where he is affiliated with the Scientific Computing and Imaging (SCI) Institute and the Huntsman Mental Health Institute (HMHI). His research interests include computational topology, interpretability of machine learning models, and developing foundation models for clinical data.
\end{IEEEbiography}

\begin{IEEEbiography}[{\includegraphics[width=1in,height=1.25in,clip,keepaspectratio]{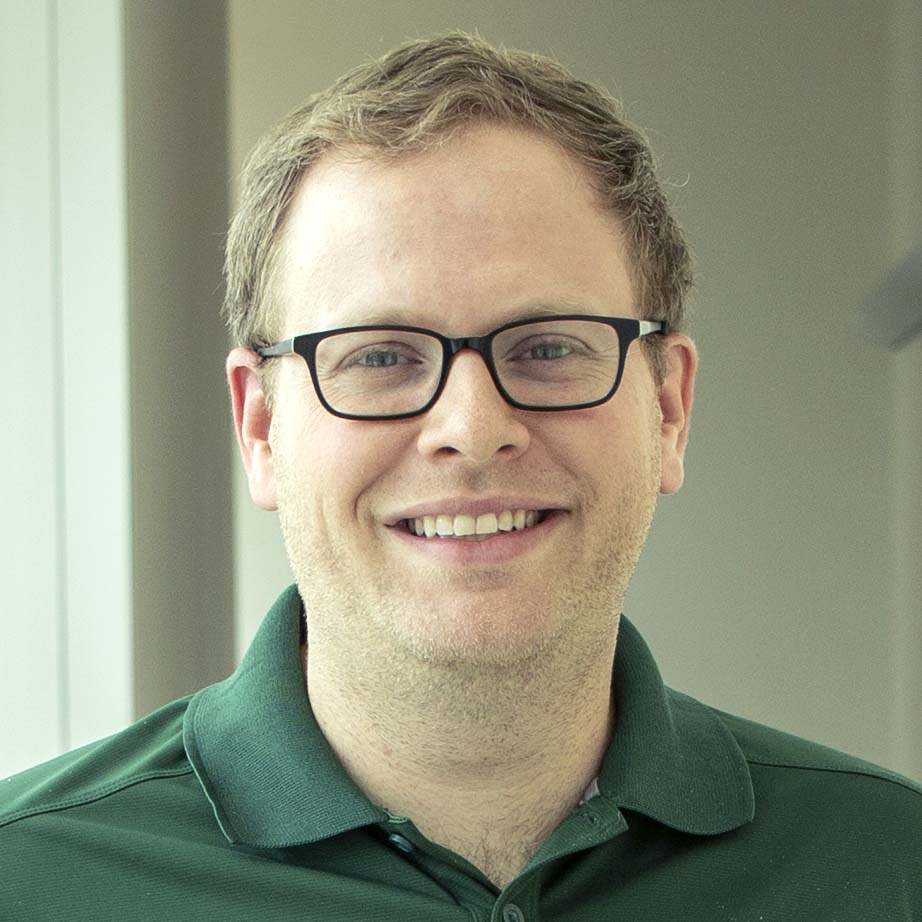}}]{Paul Rosen} received a PhD degree from Computer Science Department, Purdue University. He is an associate professor with the University of Utah. His research interests include applying geometry-- and topology-based approaches to problems in information visualization. Along with his collaborators, he has received best paper awards or honorable mentions at IEEE VIS, IEEE Pacific Vis, CG\&A, IVAPP, and SIBGRAPI. He received a National Science Foundation CAREER Award in 2019.
\end{IEEEbiography}
\vfill

%% Appendix section
\clearpage
\setcounter{figure}{0}
\setcounter{page}{1}
\renewcommand{\thefigure}{A.\arabic{figure}}
\appendices
% This appendix provides additional details, experimental results, and supplementary material
% that support the main paper but are not essential for understanding the core contributions.

\section{Persistent Homology: Formal Definitions}
\label{sec:appendix-ph-formalism}

Let $X = \{x_1, x_2, \ldots, x_n\} \subset \mathbb{R}^d$ be a finite point cloud (in our case, neural network activations).
A $k$-simplex $\sigma$ is the convex hull of $k+1$ affinely independent points:
\[
\sigma = \{t_0 v_0 + t_1 v_1 + \cdots + t_k v_k \mid t_i \geq 0,\ \sum_{i=0}^k t_i = 1\}
\]
where $v_0, v_1, \ldots, v_k$ are the vertices. A 0-simplex is a vertex, a 1-simplex an edge, a 2-simplex a triangle, and so forth.
A \textit{face} of a $k$-simplex is any simplex formed by a subset of its vertices.
A \textit{simplicial complex} $K$ is a collection of simplices where every face of a simplex in $K$ is also in $K$~\cite{carlsson2009topology}.

The $k$-th homology group $H_k(K)$ captures $k$-dimensional topological features: $H_0$ counts connected components, $H_1$ counts loops, $H_2$ counts voids, etc.
Homology is defined through the chain complex:
\[
\cdots \xrightarrow{\partial_{k+1}} C_k(K) \xrightarrow{\partial_k} C_{k-1}(K) \xrightarrow{\partial_{k-1}} \cdots \xrightarrow{\partial_1} C_0(K) \xrightarrow{\partial_0} 0
\]
where $C_k(K)$ is the vector space of $k$-chains and $\partial_k$ is the boundary operator. The $k$-th homology group is:
\[
H_k(K) = \frac{\ker(\partial_k)}{\mathrm{im}(\partial_{k+1})} = \frac{Z_k(K)}{B_k(K)}
\]
where $Z_k(K)$ are $k$-cycles (closed chains) and $B_k(K)$ are $k$-boundaries.

To construct simplicial complexes from point clouds, we use the \textit{Vietoris--Rips (VR) complex}~\cite{edelsbrunner2002persistence}.
Given a distance function $d: X \times X \to \mathbb{R}$ and radius $\epsilon \geq 0$, the VR complex is defined as:
\[
\mathrm{VR}_\epsilon(X) = \left\{ \sigma \subseteq X \mid \forall\, x_i, x_j \in \sigma,\ d(x_i, x_j) \leq \epsilon \right\}
\]
At a given radius $\epsilon$, a $k$-simplex is formed among a set of points if all pairwise distances are $\leq \epsilon$.

\section{Implementation Details}
\label{sec:implementation_details}

This section provides detailed implementation specifics that may be useful for reproducibility but are too technical for the main paper.

% \subsection{Software Dependencies}
% \label{sec:software_dependencies}

% Our implementation relies on the following software packages and their specific versions:

% \begin{itemize}
%     \item Python 3.8.5
%     \item PyTorch 1.7.0
%     \item NumPy 1.19.2
%     \item Scikit-learn 0.23.2
%     \item Matplotlib 3.3.2
%     \item Seaborn 0.11.0
% \end{itemize}

\subsection{Hardware Configuration}
\label{sec:hardware_config}

The experiments were conducted on a Lambda Vector One deep learning workstation with the following hardware configuration:
\begin{itemize}
    \item CPU: AMD Ryzen Threadripper 7960X (48) @ 7.786GHz
    \item GPU: 2 x NVIDIA GeForce RTX 4090
    \item RAM: 256GB 2400 MHz DDR5
\end{itemize}

\subsection{Training and Evaluation Setup}
\label{sec:train_eval_setup}

We fine-tune all CIFAR-10 models with weight decay of 0.01 and a linear learning rate scheduler unless otherwise specified.
For ResNet-50, we use a learning rate of $5\times10^{-5}$ and a per-device batch size of 128 for 25 epochs (best checkpoint at epoch 15).
For ViT-B/16, we use a learning rate of $2\times10^{-5}$ and a per-device batch size of 10 for 10 epochs.
For BERT-base NER, we use the pre-trained \texttt{dslim/bert-base-NER} model and evaluate directly on CoNLL-2003 test sentences with a batch size of 16 and max sequence length of 128.

\subsection{Model Sizes and CIFAR-10 Performance}
\label{sec:model_sizes_perf}

\begin{table}[htb]
    \centering
    \caption{Parameter counts and performance for models used in our study.}
    \label{tab:model_params_acc}
    \resizebox{\linewidth}{!}{%
    \small
    \begin{tabular}{lccc}
        \toprule
        \textbf{Model} & \textbf{Task} & \textbf{Parameters (M)} & \textbf{Performance} \\
        \midrule
        ResNet-50 & CIFAR-10 & 25.6 & 89.5\% Acc. \\
        ViT-B/16  & CIFAR-10 & 85.8 & 96.0\% Acc. \\
        BERT-base NER & CoNLL-2003 & 110.0 & 91.7\% F1\textsubscript{macro} \\
        \bottomrule
    \end{tabular}}
\end{table}

\subsection{Compression Experiments}
\label{sec:compression_experiments}

\paragraph{Quantization}
Quantization reduces the precision of numerical representations in neural networks, mapping continuous floating-point values to a discrete set via uniform quantization~\cite{gholami2021survey,krishnamoorthi2018quantizing}. 
For a value $x$ with scale factor $s$ and zero-point $z$, the quantized representation is $q = \mathrm{round}\!\left(\tfrac{x-z}{s}\right)$, where $s$ controls level spacing and $z$ ensures zero is exactly representable~\cite{jacob2018quantization}. 
Quantization strategies include: \emph{post-training quantization}~\cite{banner2019post,nagel2019data} (no retraining required); \emph{quantization-aware training}~\cite{jacob2018quantization,esser2019learned} (incorporates rounding during training); \emph{dynamic quantization}~\cite{wu2020integer} (scale/zero-point computed at runtime); and \emph{static quantization}~\cite{migacz2017nvidia} (precomputed from calibration data). 
The step-wise nature of quantized values perturbs local neighborhoods in activation space, making topological analysis particularly informative~\cite{yao2021hawq}.

% \paragraph{Pruning}
% Pruning sparsifies networks by zeroing low-magnitude weights~\cite{lecun1989optimal,hassibi1993second,han2015deep}, exploiting the observation that many networks are over-parameterized~\cite{frankle2018lottery}. 
% We evaluate two strategies: \emph{global unstructured pruning} (L1-magnitude ranking across all layers) and \emph{layer-wise unstructured pruning} (L1-magnitude ranking independently per layer)~\cite{molchanov2016pruning}. 
% Global pruning preserves the most important weights network-wide; layer-wise pruning maintains the relative importance distribution within each layer but may remove globally significant weights~\cite{blalock2020state}.

\paragraph{Quantization Methods}
\begin{itemize}
    \item \textbf{Static Quantization (Custom Implementation)}
    \begin{itemize}
        \item INT8: Uniform quantization with per-tensor scale/zero-point
        \item INT4: 4-bit quantization ($\sim$4$\times$ compression ratio)
        \item INT2: 2-bit quantization ($\sim$8$\times$ compression ratio)
    \end{itemize}
    \item \textbf{Dynamic Quantization}
    \begin{itemize}
        \item PyTorch INT8 dynamic quantization (CPU only)
        \item Applied to Linear layers only
    \end{itemize}
\end{itemize}

% \paragraph{Pruning Methods}
% \begin{itemize}
%     \item \textbf{Global Unstructured Pruning}: 30\% magnitude-based pruning across all Linear/Conv2d layers
%     \item \textbf{Layer-wise Unstructured Pruning}: 20\% magnitude-based pruning per layer
% \end{itemize}

\paragraph{Compression Results}

\textbf{ResNet-50} (Original: 89.5\% accuracy, 90MB)
\begin{itemize}
    \item Dynamic INT8: 89.5\% accuracy (no degradation)
\end{itemize}

\textbf{ViT-B/16} (Original: 96.0\% accuracy, 328MB)
\begin{itemize}
    \item Dynamic INT8: 94.0\% accuracy (\textminus2.0\% degradation)
\end{itemize}

\paragraph{Key Compression Insights}
\begin{itemize}
    \item \textbf{Dynamic INT8} preserves accuracy for ResNet-50 but introduces a 2\% drop for ViT-B/16, suggesting that transformer architectures are more sensitive to linear-layer quantization than convolutional ones
    \item \textbf{Topological disruption is detectable even when accuracy is preserved}, as shown in Application~3 and \appRef{sec:rn50_app3}
\end{itemize}

\section{Advanced Mathematical Derivations}
\label{sec:mathematical_derivations}

This section contains detailed mathematical derivations and theoretical foundations that support our work.

\subsection{Distance Metric Properties and Derivations}
\label{sec:distance_metric_properties}

\subsubsection{Euclidean Distance Properties}
\label{sec:euclidean_properties}

The Euclidean distance $d_E(x, y) = \|x - y\|_2$ satisfies the metric axioms and has the following key properties:

\begin{lemma}[Euclidean Distance Invariance]
For any orthogonal transformation $Q \in \mathbb{R}^{d \times d}$ with $Q^T Q = I$, the Euclidean distance is invariant: $d_E(Qx, Qy) = d_E(x, y)$.
\end{lemma}

\begin{proof}
$d_E(Qx, Qy) = \|Qx - Qy\|_2 = \|Q(x - y)\|_2 = \sqrt{(x-y)^T Q^T Q (x-y)} = \sqrt{(x-y)^T(x-y)} = d_E(x, y)$.
\end{proof}

\subsubsection{Mahalanobis Distance Analysis}
\label{sec:mahalanobis_analysis}

The Mahalanobis distance incorporates the covariance structure $\Sigma$ of the data. For a given covariance matrix $\Sigma \in \mathbb{R}^{d \times d}$, the Mahalanobis distance is:
\[
d_M(x, y) = \sqrt{(x - y)^T \Sigma^{-1} (x - y)}
\]

\begin{theorem}[Mahalanobis Distance Whitening Property]
The Mahalanobis distance with covariance matrix $\Sigma$ is equivalent to the Euclidean distance in the whitened space defined by the transformation $W = \Sigma^{-1/2}$.
\end{theorem}

\begin{proof}
Let $\tilde{x} = \Sigma^{-1/2}x$ and $\tilde{y} = \Sigma^{-1/2}y$. Then:
\begin{align}
d_M(x, y) &= \sqrt{(x - y)^T \Sigma^{-1} (x - y)} \\
&= \sqrt{(x - y)^T (\Sigma^{-1/2})^T \Sigma^{-1/2} (x - y)} \\
&= \|\Sigma^{-1/2}(x - y)\|_2 \\
&= \|\tilde{x} - \tilde{y}\|_2 = d_E(\tilde{x}, \tilde{y})
\end{align}
\end{proof}

\subsubsection{Cosine Distance Geometric Interpretation}
\label{sec:cosine_geometry}

The cosine distance $d_C(x, y) = 1 - \frac{x \cdot y}{\|x\| \|y\|}$ measures angular separation.

\begin{proposition}[Cosine Distance Angular Relationship]
For vectors $x, y \in \mathbb{R}^d$, the cosine distance satisfies:
\[
d_C(x, y) = 1 - \cos(\theta)
\]
where $\theta$ is the angle between $x$ and $y$.
\end{proposition}

This property makes cosine distance particularly suitable for analyzing activation patterns where magnitude scaling is less important than directional alignment.

\subsubsection{Manhattan Distance Properties}
\label{sec:manhattan_properties}

The Manhattan distance ($\ell_1$) sums the absolute coordinate-wise differences:
\[
d_1(x, y) = \sum_{i=1}^{d} |x_i - y_i|.
\]
It satisfies the metric axioms and weights all coordinate deviations equally in an additive sense, making it more robust to individual large-magnitude outlier dimensions than $\ell_2$.

\subsubsection{Chebyshev Distance Properties}
\label{sec:chebyshev_properties}

The Chebyshev distance ($\ell_\infty$) is governed entirely by the single largest coordinate difference:
\[
d_\infty(x, y) = \max_{i=1,\dots,d} |x_i - y_i|.
\]
It satisfies the metric axioms and can be useful for detecting activations that diverge sharply in even one feature dimension.

\subsubsection{Geodesic Distance}
\label{sec:geodesic_distance}

When activation vectors lie on or near a curved manifold embedded in $\mathbb{R}^d$, straight-line distances can underestimate the intrinsic separation between points. Geodesic distance approximates the manifold distance via a $k$-nearest-neighbor graph.

Given a point cloud $X = \{x_1, \dots, x_n\}$, we construct a weighted graph $G = (V, E)$ where each point $x_i$ is connected to its $k$ nearest neighbors under the Euclidean metric, with edge weight $w_{ij} = \|x_i - x_j\|_2$. The geodesic distance is then defined as the shortest-path distance on $G$:
\[
d_G(x_i, x_j) = \min_{\pi \in \mathcal{P}(i,j)} \sum_{(u,v) \in \pi} w_{uv},
\]
where $\mathcal{P}(i,j)$ is the set of all paths from $x_i$ to $x_j$ in $G$. This is computed via Dijkstra's algorithm for sparse graphs or Floyd--Warshall for dense graphs. For disconnected components, HOLE falls back to the Euclidean distance between the components.

\subsection{Extensible Metric Interface and Custom Distance Metrics}
\label{sec:custom_metrics}

Although cosine distance is recommended as the default metric for analyzing transformer-based neural network activations (see \cref{subsec:distance-metrics}), HOLE exposes a flexible metric interface that allows practitioners to substitute any pairwise distance function for the VR-filtration step. This is useful when domain knowledge suggests that a different geometric perspective is more appropriate. The currently supported built-in alternatives are:

\begin{itemize}
    \item \textbf{Euclidean distance} ($\ell_2$): The standard additive distance metric. Appropriate when activations are expected to reside in isotropic, well-separated clusters and magnitude differences are meaningful.
    \item \textbf{Manhattan distance} ($\ell_1$): Sums absolute coordinate-wise differences, providing robustness to large-magnitude outlier dimensions.
    \item \textbf{Chebyshev distance} ($\ell_\infty$): Determined by the single largest coordinate difference, useful for detecting sharp divergence in individual feature dimensions.
    \item \textbf{Mahalanobis distance}: Accounts for feature covariance and anisotropic scaling, making it suitable when learned representations exhibit strong inter-feature correlations or when the covariance structure of the activation distribution is of interest.
    \item \textbf{Geodesic distance}: Approximates intrinsic manifold distance via a $k$-nearest-neighbor graph, capturing non-linear relationships in curved activation spaces.
    \item \textbf{Density-normalized variants}: Any of the above base metrics can be normalized by local neighborhood density, providing robustness to heterogeneous sampling densities and outliers.
\end{itemize}

These alternatives are exposed through a common API in HOLE, meaning that practitioners can also pass any user-defined callable that computes a valid pairwise distance matrix. This design follows the principle of metric extensibility: because the downstream persistent homology computation only requires a symmetric, non-negative pairwise distance matrix satisfying the triangle inequality, any proper metric (or even a valid dissimilarity measure) can be substituted without modifying the rest of the pipeline.

\subsection{Computational Complexity Analysis}
\label{sec:complexity_analysis}

We provide detailed complexity analysis for the core algorithmic components of HOLE.

\subsubsection{VR Complex Construction Complexity}
\label{sec:vr_complexity}

\begin{theorem}[VR Complex Construction Complexity]
For a point cloud of size $n$ in $\mathbb{R}^d$, constructing the VR complex up to dimension $k$ (fixed) has:
\begin{itemize}
    \item Time complexity: $O(n^2 d + n^{k+1})$
    \item Space complexity: $O(n^{k+1})$
\end{itemize}
\end{theorem}

\begin{proof}
The pairwise distance matrix requires $O(n^2 d)$ time. The number of potential $k$-simplices is $\binom{n}{k+1}$, and each requires checking $\binom{k+1}{2}$ pairwise distances against the threshold $\epsilon$, giving $O(\binom{n}{k+1} \cdot \binom{k+1}{2}) = O(n^{k+1} \cdot k^2)$ time. For fixed $k$, the simplex enumeration cost is $O(n^{k+1})$, yielding total time $O(n^2 d + n^{k+1})$. For $k \geq 1$ the latter term dominates. Storage of all simplices requires $O(n^{k+1})$ space.
\end{proof}

\subsubsection{Persistent Homology Computation}
\label{sec:ph_complexity}

\begin{theorem}[Persistent Homology Complexity]
Computing persistent homology for a VR complex with $m$ simplices has complexity $O(m^3)$ in the worst case using standard matrix reduction algorithms.
\end{theorem}

For our application focusing on 0-dimensional persistence (connected components), the complexity reduces significantly:

\begin{corollary}[$H_0$ Persistence Complexity]
Computing 0-dimensional persistent homology can be achieved in $O(m \alpha(m))$ time using Union-Find data structures, where $\alpha$ is the inverse Ackermann function.
\end{corollary}

\subsection{Stability Theory for Persistent Homology}
\label{sec:stability_theory}

Stability results ensure that small perturbations in the input data lead to small changes in the persistent homology.

\begin{theorem}[Stability of Persistence Diagrams]
Let $f, g: X \to \mathbb{R}$ be two functions on a metric space $(X, d_X)$. Then the bottleneck distance between their persistence diagrams satisfies:
\[
d_B(\text{Dgm}(f), \text{Dgm}(g)) \leq \|f - g\|_\infty
\]
where $\|f - g\|_\infty = \sup_{x \in X} |f(x) - g(x)|$.
\end{theorem}

For our application with VR complexes:

\begin{corollary}[VR Complex Stability]
Let $X, Y$ be finite point clouds with Hausdorff distance $d_H(X, Y) \leq \delta$. Then:
\[
d_B(\text{Dgm}(\text{VR}_\bullet(X)), \text{Dgm}(\text{VR}_\bullet(Y))) \leq 2\delta
\]
\end{corollary}

This stability result is crucial for understanding robustness of our topological analysis to noise in neural network activations.

% \subsection{Noise Model Mathematics}
% \label{sec:noise_models}

% We model various noise types affecting neural network activations and their impact on topological features.

% \subsubsection{Gaussian Noise Model}
% \label{sec:gaussian_noise}

% Let $X = \{x_1, \ldots, x_n\}$ be clean activation vectors and $\tilde{X} = \{x_1 + \epsilon_1, \ldots, x_n + \epsilon_n\}$ be noisy activations where $\epsilon_i \sim \mathcal{N}(0, \sigma^2 I)$.

% \begin{theorem}[Gaussian Noise Impact on Persistence]
% Under Gaussian noise with variance $\sigma^2$, the expected change in birth times of topological features is bounded by:
% \[
% \mathbb{E}[|\tilde{b}_i - b_i|] \leq C\sigma\sqrt{\log n}
% \]
% for some constant $C$ depending on the geometry of the point cloud.
% \end{theorem}

% \subsubsection{Quantization Noise Model}
% \label{sec:quantization_noise}

% For $k$-bit quantization, we model the quantization error as uniform noise:
% \[
% q(x) = \text{round}\left(\frac{x - x_{\min}}{x_{\max} - x_{\min}} \cdot (2^k - 1)\right) \cdot \frac{x_{\max} - x_{\min}}{2^k - 1} + x_{\min}
% \]

% The quantization error satisfies $|x - q(x)| \leq \frac{x_{\max} - x_{\min}}{2^{k+1}}$.

% \begin{proposition}[Quantization Impact on Topology]
% $k$-bit quantization introduces distortions in persistence diagrams bounded by:
% \[
% d_B(\text{Dgm}(X), \text{Dgm}(q(X))) \leq \frac{\|X\|_{\max}}{2^k}
% \]
% where $\|X\|_{\max} = \max_{i,j} \|x_i - x_j\|$.
% \end{proposition}

\subsection{Extended Filtration Theory}
\label{sec:filtration_theory}

We provide theoretical foundations for the offset filtrations used in our analysis.

\subsubsection{Offset Filtration Properties}
\label{sec:offset_filtration}

For a point cloud $X \subset \mathbb{R}^d$ and offset function $f_r(x) = \min_{p \in X} \|x - p\| - r$, the offset filtration is:
\[
X_r = \{x \in \mathbb{R}^d : f_r(x) \leq 0\} = \bigcup_{p \in X} B(p, r)
\]

\begin{theorem}[Offset Filtration and the VR--\v{C}ech Interleaving]
The \v{C}ech complex at radius $r$ satisfies $\check{C}_r(X) \simeq \bigcup_{p \in X} B(p, r)$ by the Nerve theorem. The VR and \v{C}ech complexes are related by the inclusion chain $\check{C}_r(X) \subseteq \mathrm{VR}_r(X) \subseteq \check{C}_{r\sqrt{2}}(X)$ in Euclidean space, so the VR complex at radius $r$ carries the same homotopy type as $\bigcup_{p \in X} B(p, r')$ for some $r \leq r' \leq r\sqrt{2}$.
\end{theorem}

\subsubsection{Multiscale Analysis}
\label{sec:multiscale}

We extend our analysis to multiple scales simultaneously:

\textbf{Definition (Multiscale Persistence):}
For scales $0 < r_1 < r_2 < \cdots < r_k$, the multiscale persistence module is:
\[
\mathcal{M} = H_*(\text{VR}_{r_1}(X)) \to H_*(\text{VR}_{r_2}(X)) \to \cdots \to H_*(\text{VR}_{r_k}(X))
\]

This enables hierarchical analysis of topological features at different granularities.

\subsection{Visualization Algorithm Mathematics}
\label{sec:visualization_math}

We provide comprehensive mathematical foundations for our novel visualization approaches implemented in the HOLE framework.

\subsubsection{\vcflow{Cluster Flow} Diagram Computation}
\label{sec:sankey_math}

For cluster evolution across filtration stages, we define the flow matrix $F_{i,j}^{(t,t+1)}$ representing the fraction of cluster $i$ at stage $t$ that flows to cluster $j$ at stage $t+1$:

\[
F_{i,j}^{(t,t+1)} = \frac{|C_i^{(t)} \cap C_j^{(t+1)}|}{|C_i^{(t)}|}
\]

where $C_i^{(t)}$ denotes cluster $i$ at stage $t$.

\begin{theorem}[Flow Conservation]
The flow matrix satisfies: $\sum_j F_{i,j}^{(t,t+1)} = 1$ for all clusters $i$ at stage $t$.
\end{theorem}

For our flow visualization, we carefully select the most relevant stages to visualize the evolution of the clusters.

Flow thickness in \vcflow{cluster flow} diagrams is computed with logarithmic scaling for better visibility:

\textbf{Flow Thickness Formula:}
For flow with count $c$ and maximum flow $c_{\max}$:
\[
\text{thickness}(c) = \max\left(0.003, \min\left(0.04, \frac{c}{c_{\max}} \cdot 0.06\right)\right)
\]

\textbf{Bézier Flow Curves:}
Flow paths use cubic Bézier curves with control points at distance $0.3 \cdot |x_2 - x_1|$ from endpoints:
\[
B(t) = (1-t)^3 P_0 + 3(1-t)^2 t P_1 + 3(1-t)t^2 P_2 + t^3 P_3
\]
where $P_0, P_3$ are endpoints and $P_1, P_2$ are control points.

\subsubsection{Cluster Quality Metrics}
\label{sec:cluster_quality}

We define the two cluster-quality measures that underlie every HOLE visualization, then prove the properties that justify using them to select filtration scales. A clustering $\mathcal{C} = \{C_1, C_2, \dots\}$ partitions the $N$ points into non-empty parts, and $\mathcal{T} = \{T_1, \dots, T_K\}$ are the $K$ ground-truth classes.

\textbf{Definition (Clustering Purity):}
For cluster labels $\mathcal{C}$ and true labels $\mathcal{T}$, the purity is:
\[
\mathbb{P}(\mathcal{C}, \mathcal{T}) = \frac{1}{N} \sum_{k} \max_j |C_k \cap T_j|
\]
where $C_k$ is cluster $k$ and $T_j$ is true class $j$.

\textbf{Definition (Clustering Homogeneity):}
The homogeneity measures how well each true class is contained in a single cluster:
\[
\mathbb{H}(\mathcal{C}, \mathcal{T}) = \frac{1}{N} \sum_{j} \max_k |T_j \cap C_k|
\]

\textbf{Combined Quality Score:}
We combine them into $\mathbb{Q}(\mathcal{C}, \mathcal{T}) = \alpha \cdot \mathbb{P}(\mathcal{C}, \mathcal{T}) + \beta \cdot \mathbb{H}(\mathcal{C}, \mathcal{T})$, with weights $\alpha, \beta \ge 0$ and $\alpha + \beta = 1$, to select the threshold where clusters best match true labels. Empirically we found $\alpha = 0.7$ and $\beta = 0.3$ to give the best balance between the two.

Purity can be read as a size-weighted \emph{precision} (label each cluster by its majority class and count the agreeing points) and homogeneity as the dual size-weighted \emph{recall} (the largest fraction of each class captured by a single cluster). The following two theorems establish why both measures are needed and why maximizing $\mathbb{Q}$ selects a meaningful scale.

\begin{theorem}[Bounds and monotonicity]
\label{thm:cluster_bounds}
Let $\mathcal{C}$ be any clustering of the $N$ points.
\emph{(a)} $\max_j |T_j|/N \le \mathbb{P}(\mathcal{C}, \mathcal{T}) \le 1$ and $K/N \le \mathbb{H}(\mathcal{C}, \mathcal{T}) \le 1$; purity equals $1$ if and only if every cluster lies within one class, homogeneity equals $1$ if and only if every class lies within one cluster, and both equal $1$ if and only if clusters and classes are in one-to-one correspondence.
\emph{(b)} If $\mathcal{C}'$ refines $\mathcal{C}$ (every part of $\mathcal{C}'$ is contained in a part of $\mathcal{C}$), then $\mathbb{P}(\mathcal{C}') \ge \mathbb{P}(\mathcal{C})$ and $\mathbb{H}(\mathcal{C}') \le \mathbb{H}(\mathcal{C})$.
\end{theorem}
\begin{proof}
\emph{(a, upper bounds.)} The parts $C_k$ partition the points, so $\sum_k \max_j |C_k \cap T_j| \le \sum_k |C_k| = N$, giving $\mathbb{P} \le 1$, with equality if and only if $\max_j |C_k \cap T_j| = |C_k|$ for every $k$, that is, each cluster lies within one class. Exchanging the roles of clusters and classes gives $\mathbb{H} \le 1$ with equality if and only if each class lies within one cluster; both are tight together exactly when the two partitions coincide.
\emph{(b)} It suffices to consider splitting one part $C = C' \cup C''$. For purity, let $j^\star = \arg\max_j |C \cap T_j|$; then $|C \cap T_{j^\star}| = |C' \cap T_{j^\star}| + |C'' \cap T_{j^\star}| \le \max_j |C' \cap T_j| + \max_j |C'' \cap T_j|$, so the purity sum does not decrease under refinement. For homogeneity, $|T_j \cap C| = |T_j \cap C'| + |T_j \cap C''| \ge \max\bigl(|T_j \cap C'|, |T_j \cap C''|\bigr)$ for every class $j$, so no term $\max_k |T_j \cap C_k|$ increases and homogeneity does not increase.
\emph{(a, lower bounds.)} By (b), purity is smallest at the coarsest clustering (a single cluster), where it equals $\max_j |T_j|/N$, and homogeneity is smallest at the finest clustering ($N$ singletons), where each class contributes $1$ and it equals $K/N$.
\end{proof}

\begin{theorem}[Complementary extremes]
\label{thm:cluster_extremes}
\emph{(i)} If every point forms its own cluster (maximal fragmentation), then $\mathbb{P} = 1$ while $\mathbb{H} = K/N$.
\emph{(ii)} If all points form a single cluster (no separation), then $\mathbb{H} = 1$ while $\mathbb{P} = \max_j |T_j| / N$.
Consequently neither measure alone separates a faithful clustering from a degenerate one: purity is maximal under maximal fragmentation, and homogeneity is maximal under total collapse.
\end{theorem}
\begin{proof}
(i) With singleton clusters, each $C_k$ meets exactly one class, so $\max_j |C_k \cap T_j| = |C_k| = 1$ and $\mathbb{P} = N/N = 1$; each class $T_j$ is spread over $|T_j|$ distinct singletons, so its largest intersection with any single cluster is $1$, and summing over the $K$ classes gives $\mathbb{H} = K/N$.
(ii) With one cluster $C_1 = X$, $\max_j |C_1 \cap T_j| = \max_j |T_j|$, so $\mathbb{P} = \max_j |T_j|/N$ (equal to $1/K$ for balanced classes); each class lies entirely in $C_1$, so $\max_k |T_j \cap C_k| = |T_j|$ and $\mathbb{H} = \sum_j |T_j|/N = 1$.
\end{proof}

\textbf{Scale selection.}
These two theorems are what make $\mathbb{Q}$ usable as the basis for HOLE's views. The filtration produces a nested family $\{\mathcal{C}(\epsilon)\}_{\epsilon \ge 0}$ whose $H_0$ components only merge as $\epsilon$ grows, so $\mathcal{C}(\epsilon)$ coarsens monotonically from $N$ singletons to a single component; by Theorem~\ref{thm:cluster_bounds}(b), $\mathbb{P}(\mathcal{C}(\epsilon))$ is therefore non-increasing and $\mathbb{H}(\mathcal{C}(\epsilon))$ non-decreasing in $\epsilon$, so $\mathbb{Q}$ genuinely trades the two against each other. Because the clustering changes only at the finitely many death times $0 = \epsilon_0 < \epsilon_1 < \dots < \epsilon_m$ of the $H_0$ persistence diagram, $\mathbb{Q}(\mathcal{C}(\epsilon), \mathcal{T})$ is piecewise constant, and HOLE selects
\[
\epsilon^\star = \operatorname*{arg\,max}_{\epsilon \in \{\epsilon_0, \dots, \epsilon_m\}} \mathbb{Q}\bigl(\mathcal{C}(\epsilon), \mathcal{T}\bigr).
\]
By Theorem~\ref{thm:cluster_extremes} the score is $\alpha + \beta\,(K/N)$ at the fragmented end and $\alpha\,(\max_j |T_j|/N) + \beta$ at the merged end, both strictly below the $\mathbb{Q} \to 1$ of a class-aligned clustering, so whenever the representation carries genuine class structure the maximizer $\epsilon^\star$ falls at an interior scale. The \vblob{blob} and contour graphs are drawn at $\epsilon^\star$, and the intermediate stages of the \vcflow{cluster flow} Sankey are the highest-scoring scales around it, while its first and last stages are fixed at the ground-truth labels and the single merged component. This also explains the distribution-shift signature (\appRef{sec:shift_imbalance}): corruption fragments each class into small but individually pure pieces, which by Theorem~\ref{thm:cluster_extremes} keeps the sample-weighted purity high while homogeneity, equal to the mean per-class purity for a class-balanced probe set, collapses.

\subsubsection{\vblob{Blob Visualization} Mathematics}
\label{sec:blob_math}

For cluster separation analysis, we implement a boundary computation algorithm.

\textbf{\vblob{Blob Visualization} Boundary Algorithm:}
Given cluster points $P = \{p_1, \ldots, p_k\} \subset \mathbb{R}^2$, we compute the convex hull:

\begin{algorithm}[H]
\begin{algorithmic}[1]
\State Compute convex hull $H = \text{ConvexHull}(P)$
\State Calculate centroid $c = \frac{1}{k}\sum_{i=1}^k p_i$
\State For each vertex $v \in H$:
\State \quad $v' = v + \alpha \cdot (v - c)$ where $\alpha = 0.15$
\State Return expanded vertices $\{v'\}$
\end{algorithmic}
\end{algorithm}

\textbf{Smooth Hull Generation:}
For improved visual appeal, we generate smooth boundaries using Bézier interpolation:

\textbf{Definition (Smooth Hull):}
Between consecutive hull vertices $v_i$ and $v_{i+1}$, we create a quadratic Bézier curve with control point:
\[
c_i = v_i + \beta \cdot \frac{v_{i+1} - v_i}{\|v_{i+1} - v_i\|} + \gamma \cdot \frac{v_i - c}{\|v_i - c\|}
\]
where $\beta = 0.3 \cdot \|v_{i+1} - v_i\|$ and $\gamma = 0.3\beta$.

The curve is parameterized as: $B(t) = (1-t)^2 v_i + 2(1-t)t c_i + t^2 v_{i+1}$ for $t \in [0,1]$.

\subsubsection{\vhmd{Dendrogram} Construction from Persistence}
\label{sec:dendrogram_math}

We construct dendrograms using the ultrametric induced by the persistence filtration:

\textbf{Definition (Persistence Ultrametric):}
For points $x, y \in X$, define $d_{\text{pers}}(x, y)$ as the smallest filtration value $r$ such that $x$ and $y$ belong to the same connected component in $\text{VR}_r(X)$.

\begin{theorem}[Ultrametric Property]
$d_{\text{pers}}$ satisfies the ultrametric inequality: $d_{\text{pers}}(x, z) \leq \max(d_{\text{pers}}(x, y), d_{\text{pers}}(y, z))$.
\end{theorem}

\textbf{Linkage Matrix Construction:}
We build the linkage matrix $L \in \mathbb{R}^{(n-1) \times 4}$ where each row $L_i = [c_1, c_2, d, s]$ represents the merge of clusters $c_1$ and $c_2$ at distance $d$ creating a cluster of size $s$.

\textbf{Algorithm (Persistence-Based Linkage):}
\begin{algorithm}[H]
\begin{algorithmic}[1]
\State Sort edges $(i,j)$ by distance $d_{ij}$
\State Initialize Union-Find structure $U$
\State For each edge $(i,j)$ in sorted order:
\State \quad If $U.\text{find}(i) \neq U.\text{find}(j)$:
\State \quad \quad Record merge: $[U.\text{find}(i), U.\text{find}(j), d_{ij}, \text{size}]$
\State \quad \quad $U.\text{union}(i,j)$
\end{algorithmic}
\end{algorithm}

\subsubsection{Matrix Reordering for Visualization}
\label{sec:matrix_reordering}

For distance matrix visualization, we employ the Reverse Cuthill--McKee (RCM) algorithm:

\textbf{Definition (RCM Reordering):}
Given adjacency matrix $A$ derived from distance matrix $D$ by thresholding at percentile $p$:
\[
A_{ij} = \begin{cases}
1 & \text{if } D_{ij} \leq \text{percentile}_p(D) \\
0 & \text{otherwise}
\end{cases}
\]

The RCM algorithm produces a permutation $\pi$ that minimizes the bandwidth of $A$.

\textbf{Bandwidth Minimization:}
The bandwidth of a matrix $A$ with permutation $\pi$ is:
\[
\beta(A, \pi) = \max_{A_{\pi(i),\pi(j)} \neq 0} |\pi(i) - \pi(j)|
\]

RCM seeks $\pi^* = \arg\min_\pi \beta(A, \pi)$.

% \subsubsection{Dimensionality Reduction Mathematics}
% \label{sec:dimred_math}

% Our blob visualization employs multiple dimensionality reduction techniques with optimized parameters.

% \textbf{PCA Formulation:}
% For activation matrix $X \in \mathbb{R}^{n \times d}$, PCA finds the projection $W \in \mathbb{R}^{d \times 2}$ that maximizes variance:
% \[
% W^* = \arg\max_{W^T W = I} \text{tr}(W^T \Sigma W)
% \]
% where $\Sigma = \frac{1}{n-1}(X - \bar{X})^T(X - \bar{X})$ is the covariance matrix.

% \textbf{Adaptive t-SNE Parameters:}
% For t-SNE visualization, we adapt the perplexity parameter based on sample size:
% \[
% \text{perplexity} = \min\left(30, \max\left(5, \left\lfloor\frac{n}{4}\right\rfloor\right)\right)
% \]
% where $n$ is the number of samples.

\subsubsection{Color Assignment}
\label{sec:color_mapping}

We implement a color assignment algorithm for consistent visualization across stages:

\textbf{Distinct Color Generation:}
Using the golden ratio $\phi = 0.618033988749895$, we generate colors in HSV space:
\[
\begin{aligned}
h_i &= (i \cdot \phi) \bmod 1 \\
s_i &= 0.6 + (i \bmod 4) \cdot 0.1 \\
v_i &= 0.7 + (i \bmod 3) \cdot 0.1
\end{aligned}
\]

\textbf{Color Distance Constraint:}
For color uniqueness, we enforce minimum Euclidean distance in RGB space:
\[
\min_{j < i} \|c_i - c_j\|_2 \geq 0.2
\]
where $c_i$ is the RGB representation of color $i$.

% \subsubsection{Persistence Barcode and Diagram Mathematics}
% \label{sec:persistence_viz_math}

% \textbf{Barcode Normalization:}
% For persistence barcode visualization, we handle infinite persistence with normalization:
% \[
% \text{infinity} = \max\{\text{death}(f) : \text{death}(f) < \infty\} + 0.1 \cdot \text{range}
% \]

% \textbf{Diagram Point Sizing:}
% In persistence diagrams, point sizes are scaled by persistence:
% \[
% \text{size}(b, d) = 20 + 80 \cdot \frac{d - b}{\max\{d' - b' : (b', d') \in \text{diagram}\}}
% \]

% \subsection{Topological Data Analysis for Pruning}
% \label{sec:tda_pruning}

% We provide theoretical analysis of how network pruning affects topological structure.
% % \marginpar{write about pruning}
% This result provides theoretical justification for using topological analysis to guide pruning strategies.

\section{Dataset Descriptions}
\label{sec:dataset_descriptions}

Detailed descriptions of the datasets used in our experiments:

\subsection{Robustness Experiments}
\label{sec:appendix_robustness}

\begin{figure}[!htbp]
    \centering
    \begin{subfigure}{0.30\linewidth}
        \includegraphics[width=\linewidth]{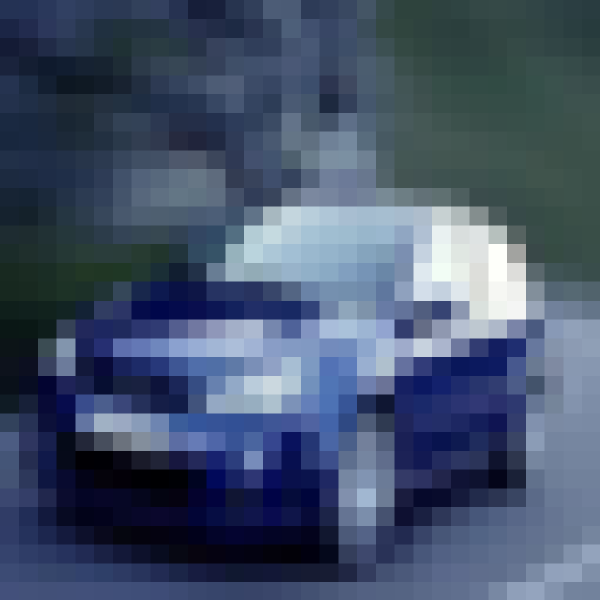}
        \caption{Original}
    \end{subfigure}\hfill
    \begin{subfigure}{0.30\linewidth}
        \includegraphics[width=\linewidth]{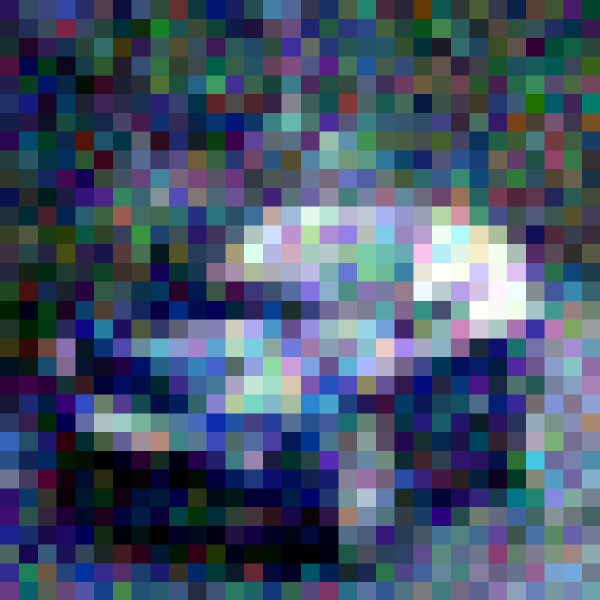}
        \caption{Gaussian}
    \end{subfigure}\hfill
    \begin{subfigure}{0.30\linewidth}
        \includegraphics[width=\linewidth]{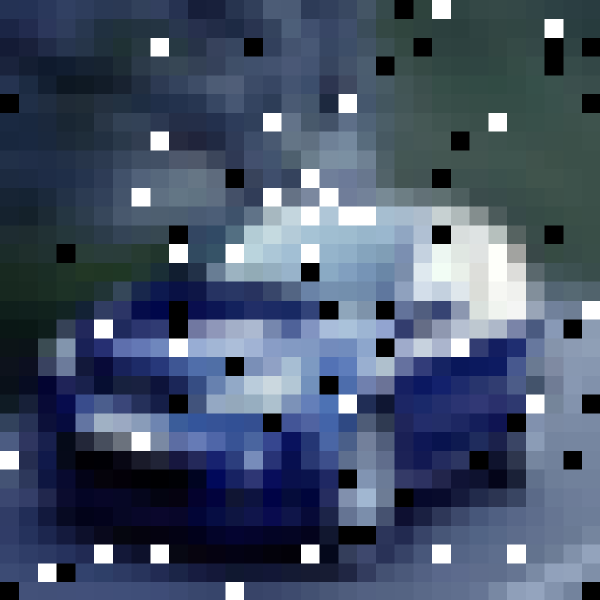}
        \caption{Salt \& Pepper}
    \end{subfigure}

    \vspace{0.5em}

    \begin{subfigure}{0.30\linewidth}
        \includegraphics[width=\linewidth]{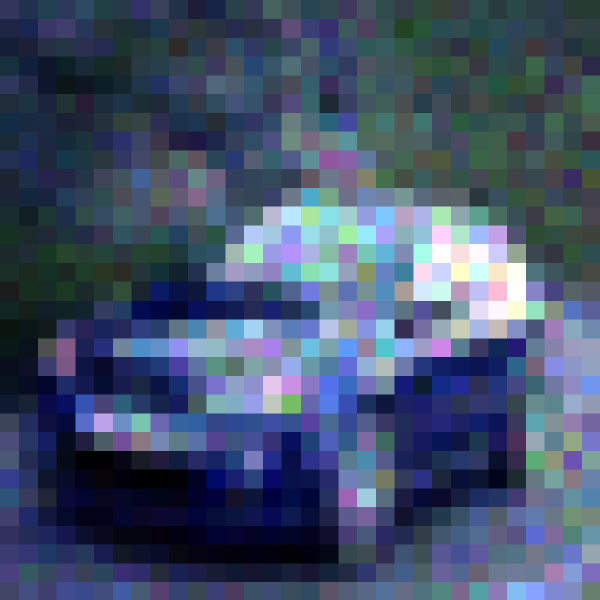}
        \caption{Speckle}
    \end{subfigure}\hfill
    \begin{subfigure}{0.30\linewidth}
        \includegraphics[width=\linewidth]{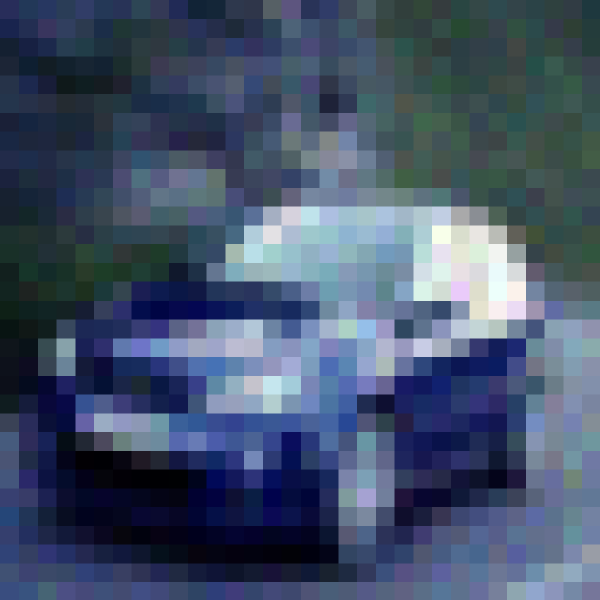}
        \caption{Poisson}
    \end{subfigure}\hfill
    \begin{subfigure}{0.30\linewidth}
        \includegraphics[width=\linewidth]{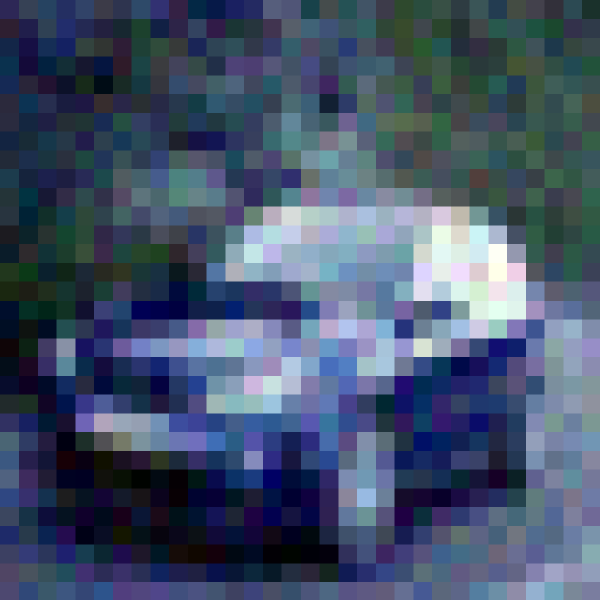}
        \caption{Uniform}
    \end{subfigure}
    \caption{All noise types applied to CIFAR-10 example images. The main text (Application~2) focuses on Gaussian and Salt \& Pepper; the remaining types are shown here for completeness.}
    \label{fig:all_noise_types}
\end{figure}

\paragraph{Noise Model Specifications}
\begin{itemize}
    \item \textbf{Gaussian}: Additive white Gaussian noise (AWGN) with zero mean and variance $\sigma^2$, applied per pixel; we use $\sigma=0.1$ in our experiments
    \item \textbf{Salt \& Pepper}: Impulse noise with probability $p$ of setting pixels to extremal values (0 or 255); we use $p=0.10$
    \item \textbf{Speckle}: Multiplicative noise $x' = x + x \cdot \mathcal{N}(0, \sigma^2)$, where variance scales with signal intensity
    \item \textbf{Poisson}: Shot noise $x' \sim \text{Poisson}(\lambda x)$, modeling photon-counting effects
    \item \textbf{Uniform}: Additive noise sampled from $\mathcal{U}(-a, a)$
\end{itemize}

\paragraph{Robustness Results}

\textbf{ResNet-50} (Original: 89.5\%)
\begin{itemize}
    \item Gaussian: 63.0\% (\textminus26.5pp)
    \item Salt \& Pepper: 10.0\% (\textminus79.5pp)
    \item Speckle: 59.0\% (\textminus30.5pp)
    \item Poisson: 10.0\% (\textminus79.5pp)
    \item Uniform: 76.0\% (\textminus13.5pp)
\end{itemize}

\textbf{ViT-B/16} (Probe-set baseline: 96.0\%)
\begin{itemize}
    \item Gaussian: 91.5\% (\textminus4.5pp)
    \item Salt \& Pepper: 62.0\% (\textminus34.0pp) \emph{Most destructive}
    \item Speckle: 94.5\% (\textminus1.5pp) \emph{Most robust}
    \item Poisson: 89.5\% (\textminus6.5pp)
    \item Uniform: 93.0\% (\textminus3.0pp)
\end{itemize}

\paragraph{Robustness Insights}
\begin{itemize}
    \item \textbf{Vision Transformer significantly more robust} to noise than ResNet-50: ViT retains $>$89\% accuracy under Gaussian/Speckle/Poisson/Uniform, while ResNet-50 drops below 63\% for all noise types except Uniform (76\%)
    \item \textbf{Salt \& Pepper and Poisson noise most destructive} for ResNet-50, both collapsing accuracy to 10.0\%; for ViT, Salt \& Pepper is most harmful (62.0\%)
    \item \textbf{Topological signatures amplify these differences}: HOLE reveals cluster breakdown even when accuracy numbers partially conceal the representational damage (see \appRef{sec:rn50_app2})
\end{itemize}

\subsection{Datasets}
\label{sec:realworld_dataset}

\paragraph{CIFAR-10}
CIFAR-10~\cite{Krizhevsky2009LearningML} consists of 60,000 32$\times$32 color images in 10 classes (airplane, automobile, bird, cat, deer, dog, frog, horse, ship, truck), split into 50,000 training and 10,000 test images.
For fine-tuning, we use a subset of the training split: 15,000 images for ResNet-50 and 5,000 for all other models (ViT-B/16, ResNet-18/34, MobileNetV2, ConvNeXt, EfficientNet-B0), with a further 90/10 train/validation split in each case.
For all HOLE analyses (robustness, quantization, and layer-level experiments), we use a balanced probe set of 200 test images (20 per class, selected with a fixed random seed of 42) to ensure equal class representation in the topological analysis.

\paragraph{CoNLL-2003}
The CoNLL-2003 English NER dataset~\cite{tjong2003introduction} provides token-level named entity annotations in four categories: PER (person), ORG (organization), LOC (location), and MISC (miscellaneous), plus a non-entity label O.
We randomly sample 150 sentences from the test split (seed 42) and tokenise them with a maximum sequence length of 128.
The original fine-grained BIO tags are collapsed to the five entity types above for HOLE analysis.
To keep the distance matrix computation tractable, token embeddings are subsampled to a maximum of 75 tokens per analysis run.

% \todo{add code link here}
\section{Code Availability}
\label{sec:code_availability}

The source code for reproducing all experiments in this paper is available at:
% \url{https://github.com/FoxHound0x00/hole}
\url{https://github.com/anonymous/hole} % anonymized for review

The repository includes:
\begin{itemize}
    \item Training scripts for all models
    \item Evaluation scripts and metrics
    \item Data preprocessing utilities
    \item Visualization tools
    \item Pre-trained model checkpoints
\end{itemize}

\section{\vcflow{Cluster Flow} Visualization: Filtering}
\label{sec:sankey_filtering}

The \vcflow{cluster flow diagrams} help us see if there is class level clustering in the activation spaces at a certain distance threshold.
In practice, during filtration there might be clusters which consist of very few points or outliers that do not contribute to the overall understanding of the data.
To make the \vcflow{cluster flow diagrams} more interpretable, we can apply a filter to remove these small clusters.
HOLE supports an optional \emph{minimum-size filter} that suppresses these ephemeral components whose normalized size falls below a configurable threshold at each filtration stage, thus leaving only the substantial persistent flows visible, and making dominant structural transitions immediately apparent.
Filtered components are not discarded from the topological computation; they are simply omitted from the \vcflow{cluster flow} rendering so that the meaningful flows remain legible.

\begin{figure}[!htb]
    \centering
    \includegraphics[width=0.90\linewidth]{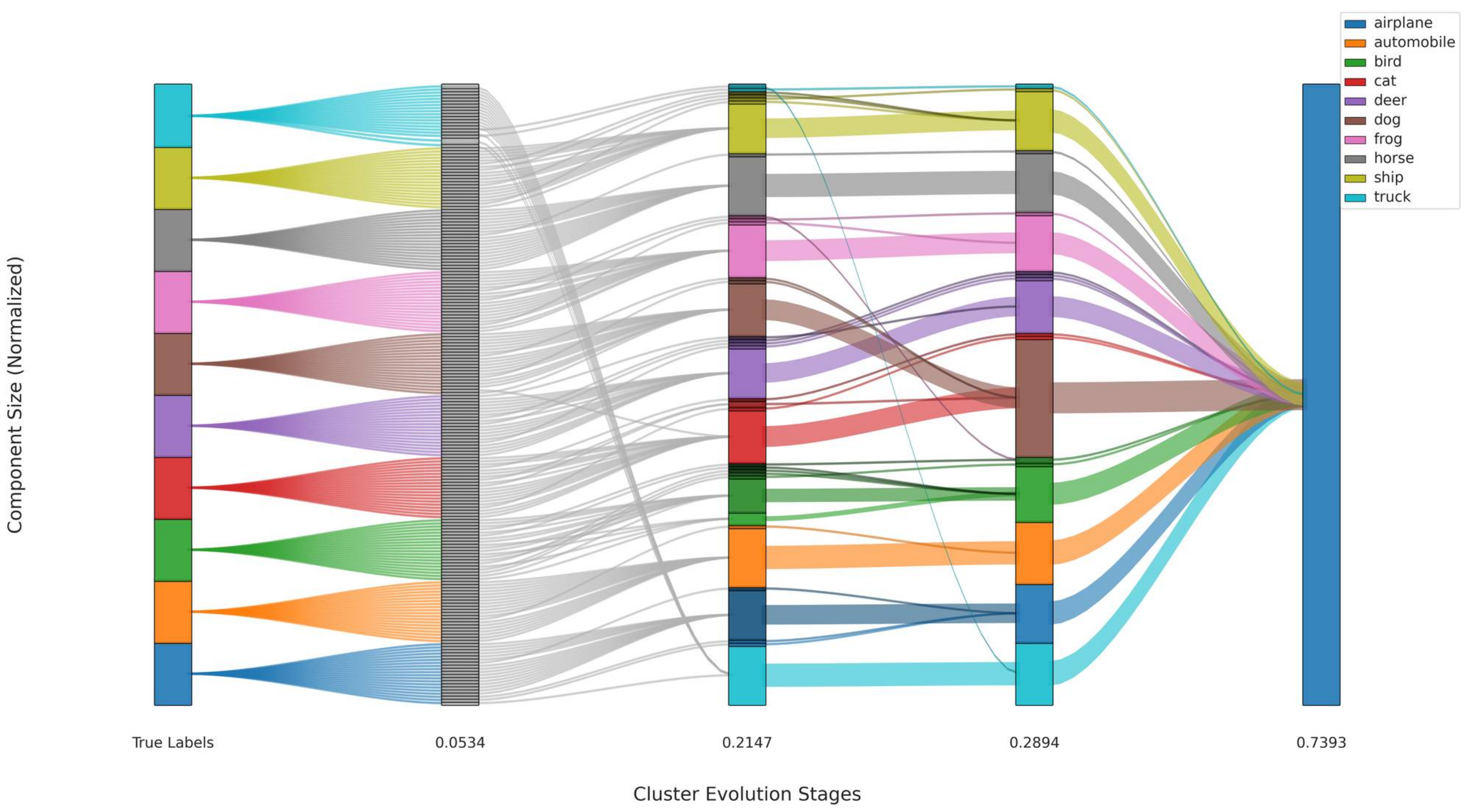}
    \caption{Unfiltered \vcflow{cluster flow} diagram for ViT-B/16 encoder layer~11, class-balanced CIFAR-10, cosine distance. Transient singleton and near-singleton components produce dense, overlapping ribbons that obscure the dominant class-level flows. Compare with the filtered version in the main text (\cref{fig:sankey_layer_11}).}
    \label{fig:sankey_nf}
\end{figure}

\Cref{fig:sankey_nf} shows the unfiltered \vcflow{cluster flow diagram} for ViT-B/16 encoder layer~11 (class-balanced CIFAR-10, cosine distance), which can be compared directly against the filtered version shown in \cref{fig:sankey_layer_11}.
% The unfiltered diagram is populated by dozens of thin gray ribbons representing transient small components; at the first filtration threshold these already outnumber the primary class flows, making it difficult to trace how individual classes evolve.
This filtering step has no impact on the underlying persistent homology computation and can be disabled to inspect fine-grained component behavior when needed.

%% ----------------------------------------------------------------
\section{ViT-B/16: Enlarged Experiment Figures}
\label{sec:vit_enlarged}
%% ----------------------------------------------------------------

This section reproduces the ViT-B/16 experiment figures from the main paper at full page width for improved legibility.

\begin{figure*}[!htb]
    \centering
    \begin{subfigure}{\linewidth}
        \centering
        \includegraphics[width=\linewidth]{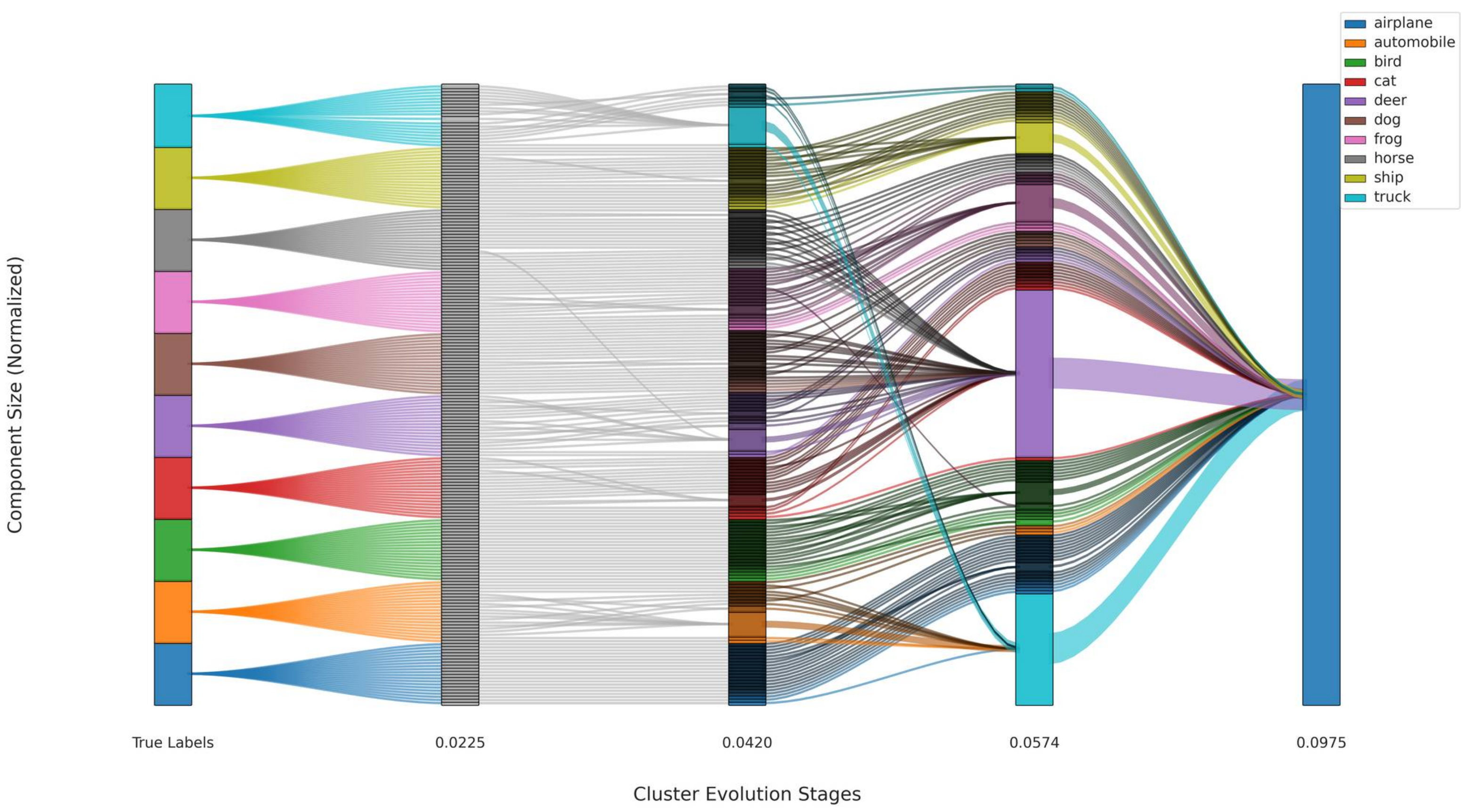}
        \caption{\vcflow{Cluster flow}, \textit{layer~9}. Classes converge rapidly into a single component, indicating weak class separation.}
        \label{fig:enlarged_sankey_layer_9}
    \end{subfigure}

    \vspace{0.5em}

    \begin{subfigure}{\linewidth}
        \centering
        \includegraphics[width=\linewidth]{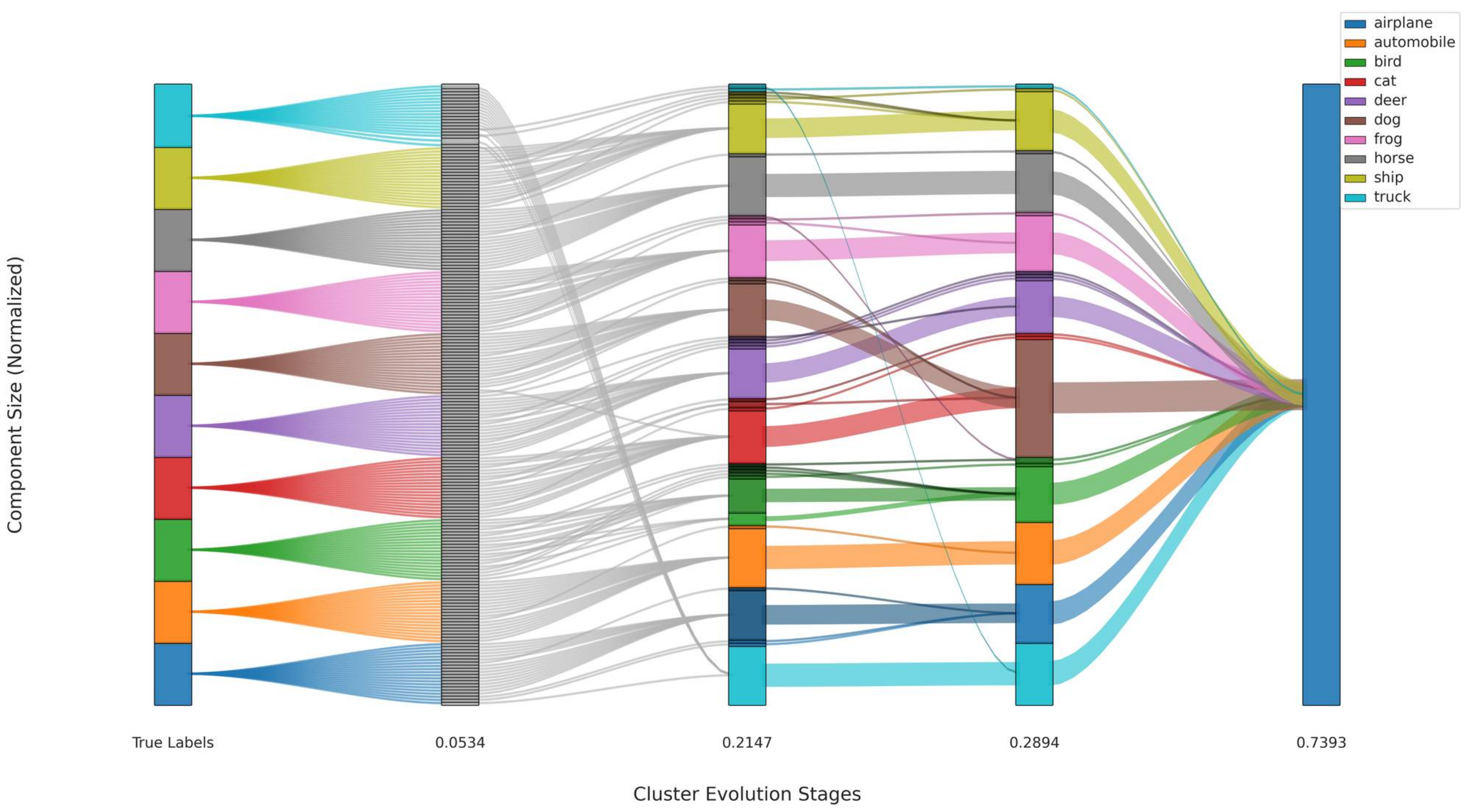}
        \caption{\vcflow{Cluster flow}, \textit{layer~11}. Class flows persist across filtration stages, showing stronger separation.}
        \label{fig:enlarged_sankey_layer_11}
    \end{subfigure}
    \caption{Enlarged version of \cref{fig:learned_rep_comparison}. \vcflow{Cluster flow} comparison for ViT-B/16 on class-balanced CIFAR-10, comparing encoder layers~9 and~11. The corresponding \vblob{blob graph} for layer~11 is shown in \cref{fig:enlarged_blob_layer_11}.}
    \label{fig:enlarged_learned_rep}
\end{figure*}

\begin{figure*}[!htb]
    \centering
    \includegraphics[width=\linewidth]{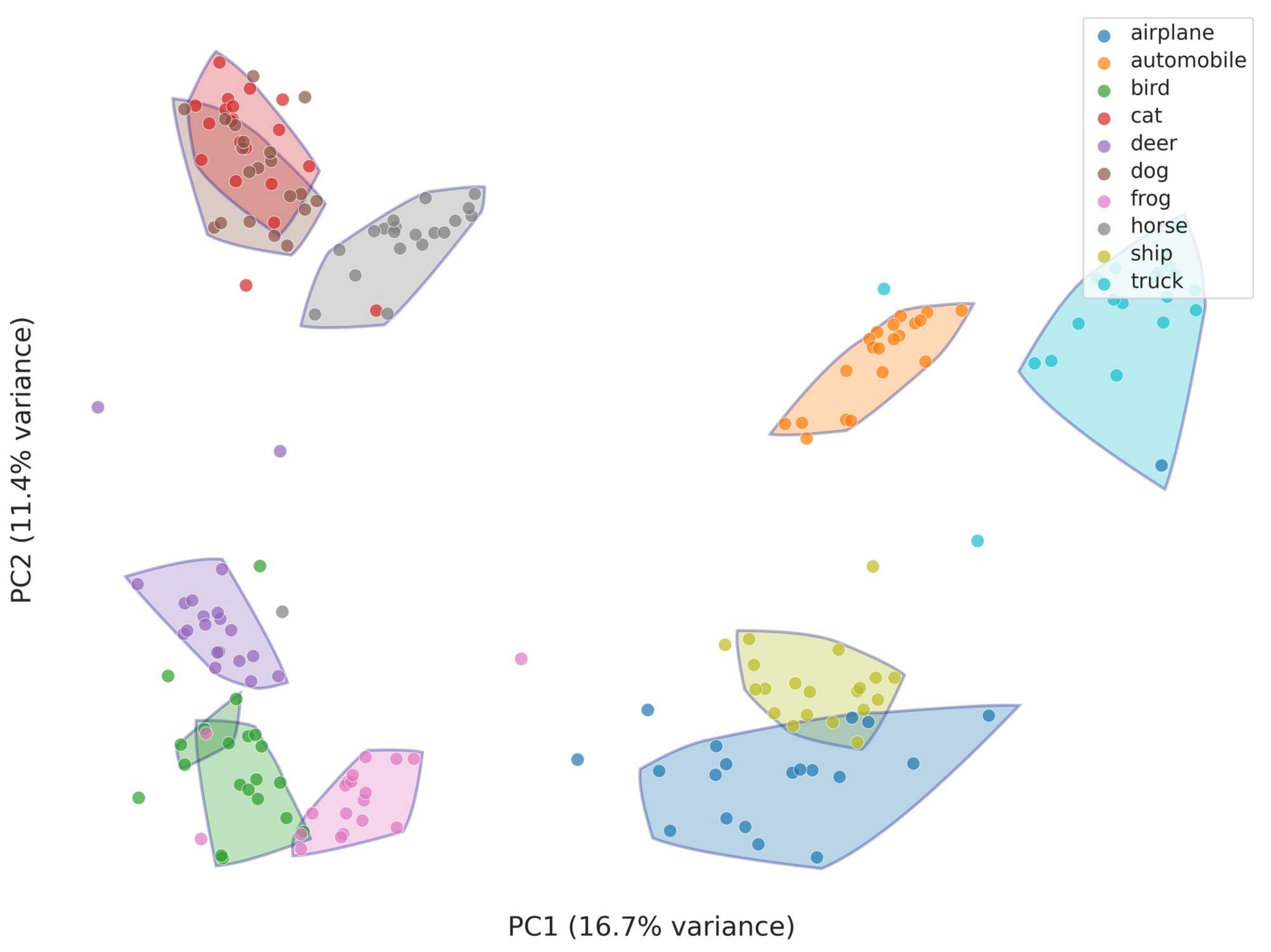}
    \caption{Enlarged version of \cref{fig:learned_rep_comparison}. \vblob{Blob graph} for ViT-B/16 encoder layer~11, class-balanced CIFAR-10. Classes form compact, well-separated clusters in the PCA projection.}
    \label{fig:enlarged_blob_layer_11}
\end{figure*}

\begin{figure*}[!htb]
    \centering
    \begin{subfigure}{\linewidth}
        \centering
        \includegraphics[width=\linewidth]{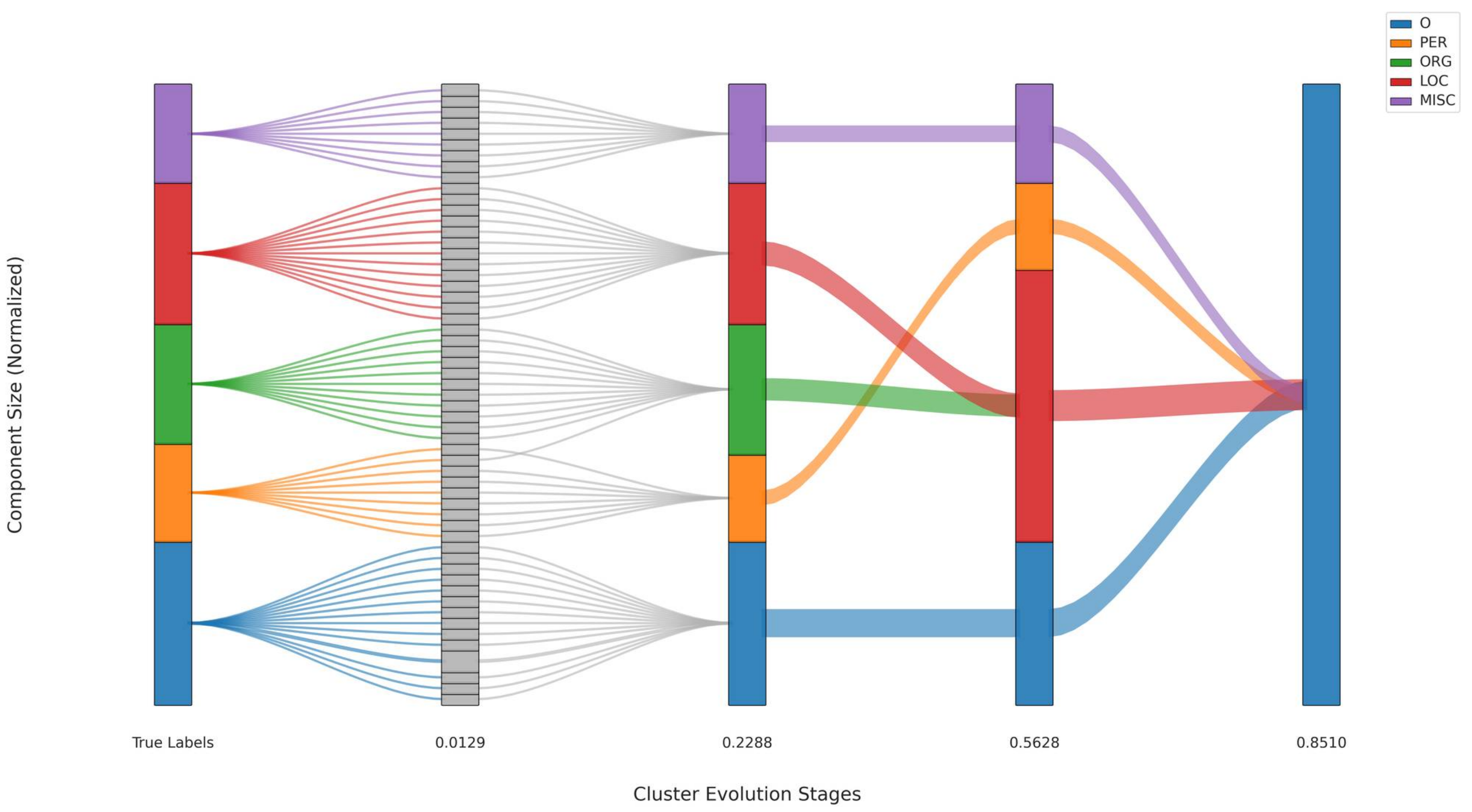}
        \caption{\vcflow{Cluster flow}. Entity-type flows persist coherently across filtration stages.}
        \label{fig:enlarged_bert_sankey}
    \end{subfigure}

    \vspace{0.8em}

    \begin{subfigure}{0.48\linewidth}
        \centering
        \includegraphics[width=\linewidth]{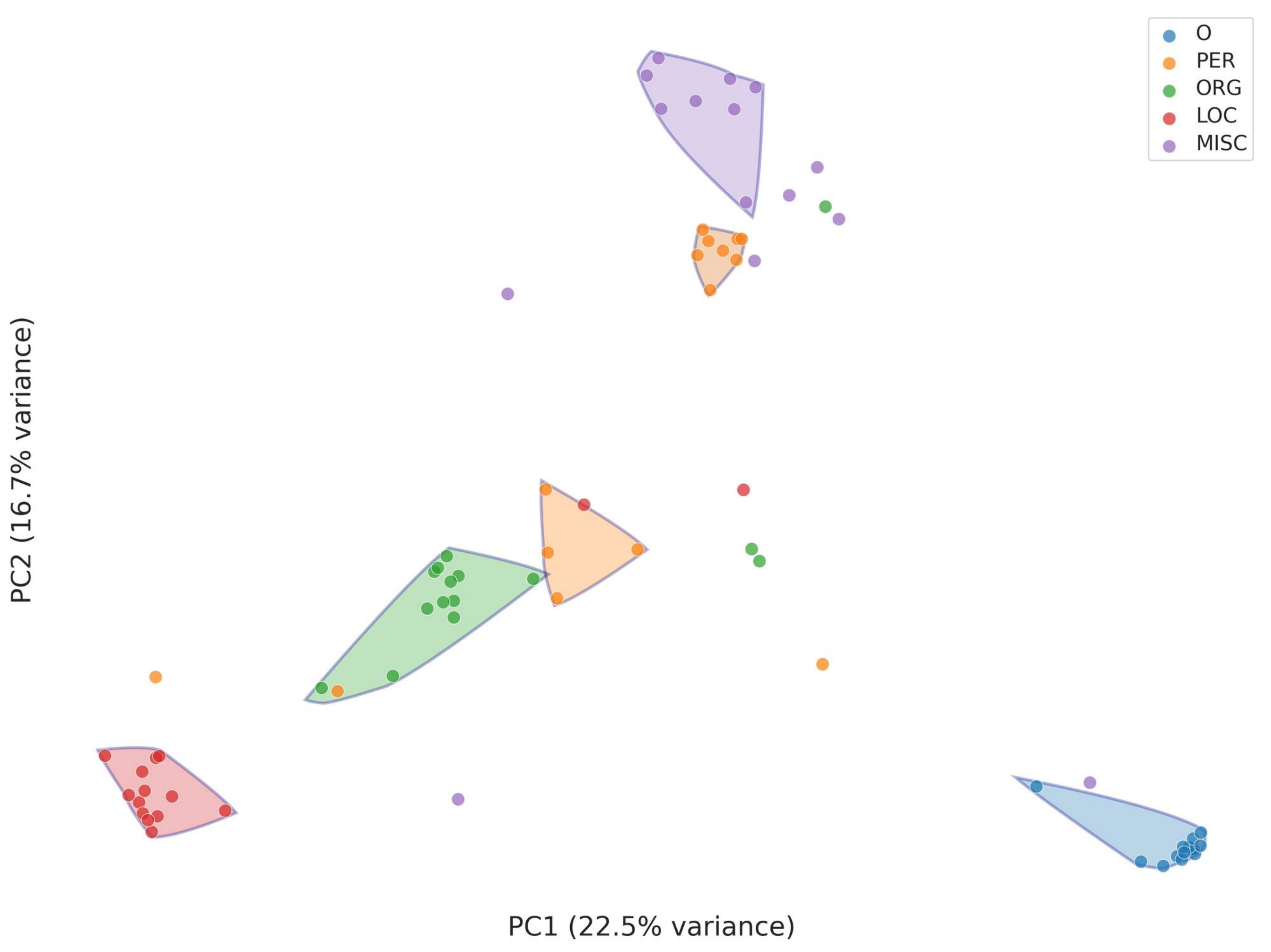}
        \caption{PCA \vblob{blob}. Entity types form compact clusters with separation from the dominant O~class.}
        \label{fig:enlarged_bert_blob}
    \end{subfigure}\hfill
    \begin{subfigure}{0.48\linewidth}
        \centering
        \includegraphics[width=\linewidth]{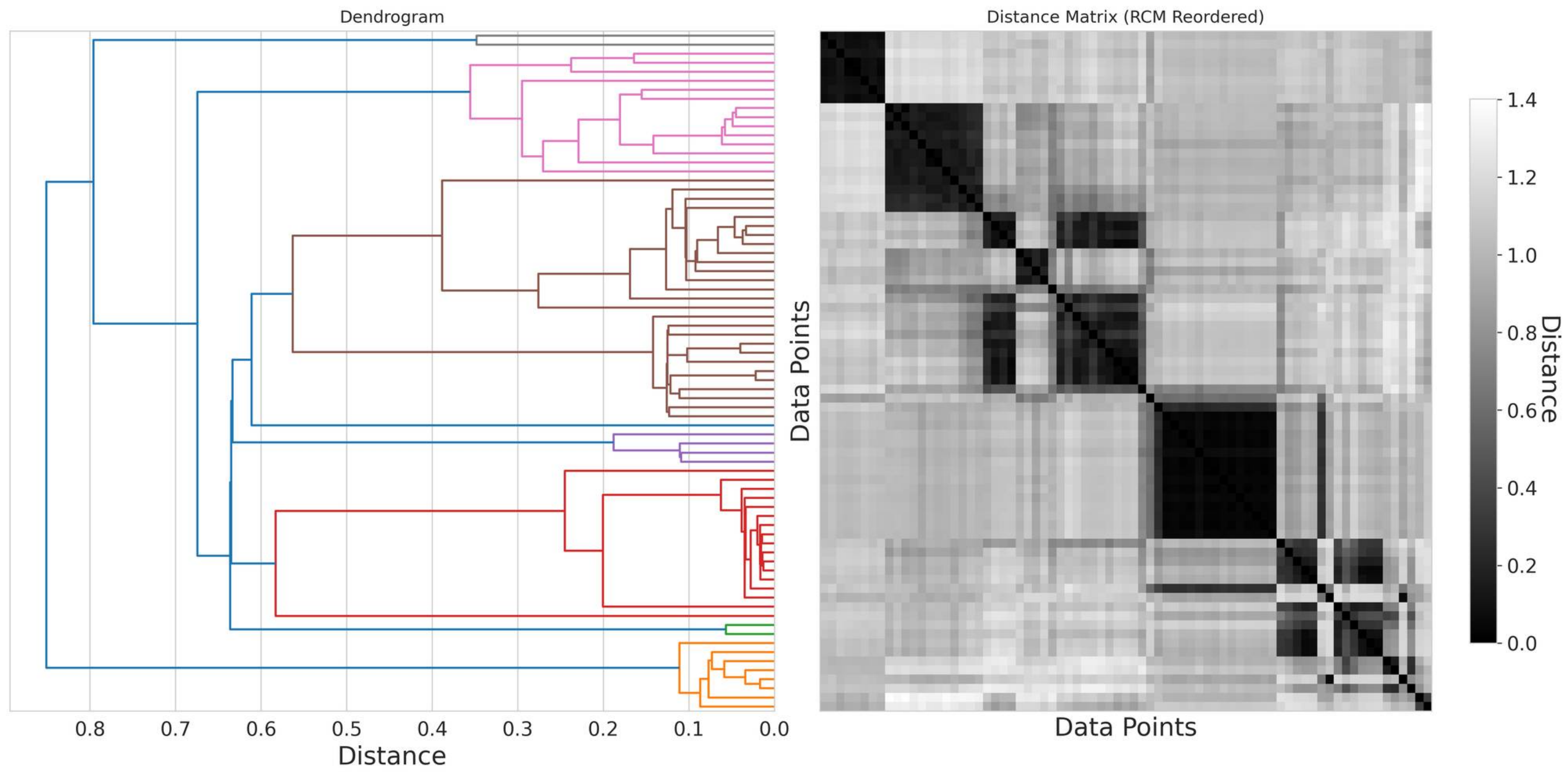}
        \caption{\vhmd{Heatmap dendrogram}. Block-diagonal structure confirms inter-class separation.}
        \label{fig:enlarged_bert_hmd_11}
    \end{subfigure}
    \caption{Enlarged version of \cref{fig:bert_ner_learned_rep}. BERT-base NER encoder layer~11 on CoNLL-2003 (cosine distance).}
    \label{fig:enlarged_bert_ner}
\end{figure*}

\begin{figure*}[!htb]
    \centering
    \begin{subfigure}{0.48\linewidth}
        \centering
        \includegraphics[width=\linewidth]{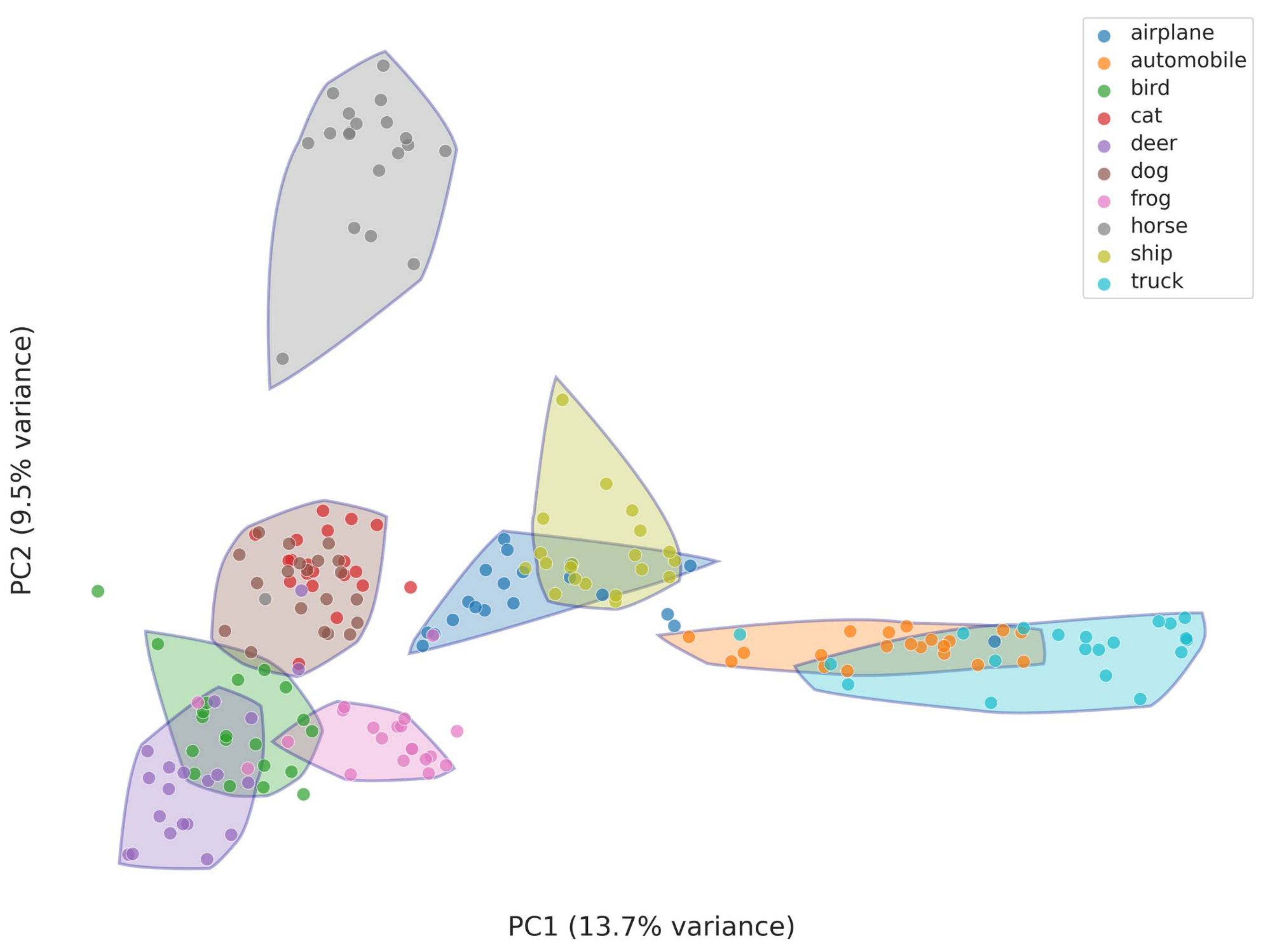}
        \caption{Gaussian noise.}
        \label{fig:enlarged_blob_gaussian}
    \end{subfigure}\hfill
    \begin{subfigure}{0.48\linewidth}
        \centering
        \includegraphics[width=\linewidth]{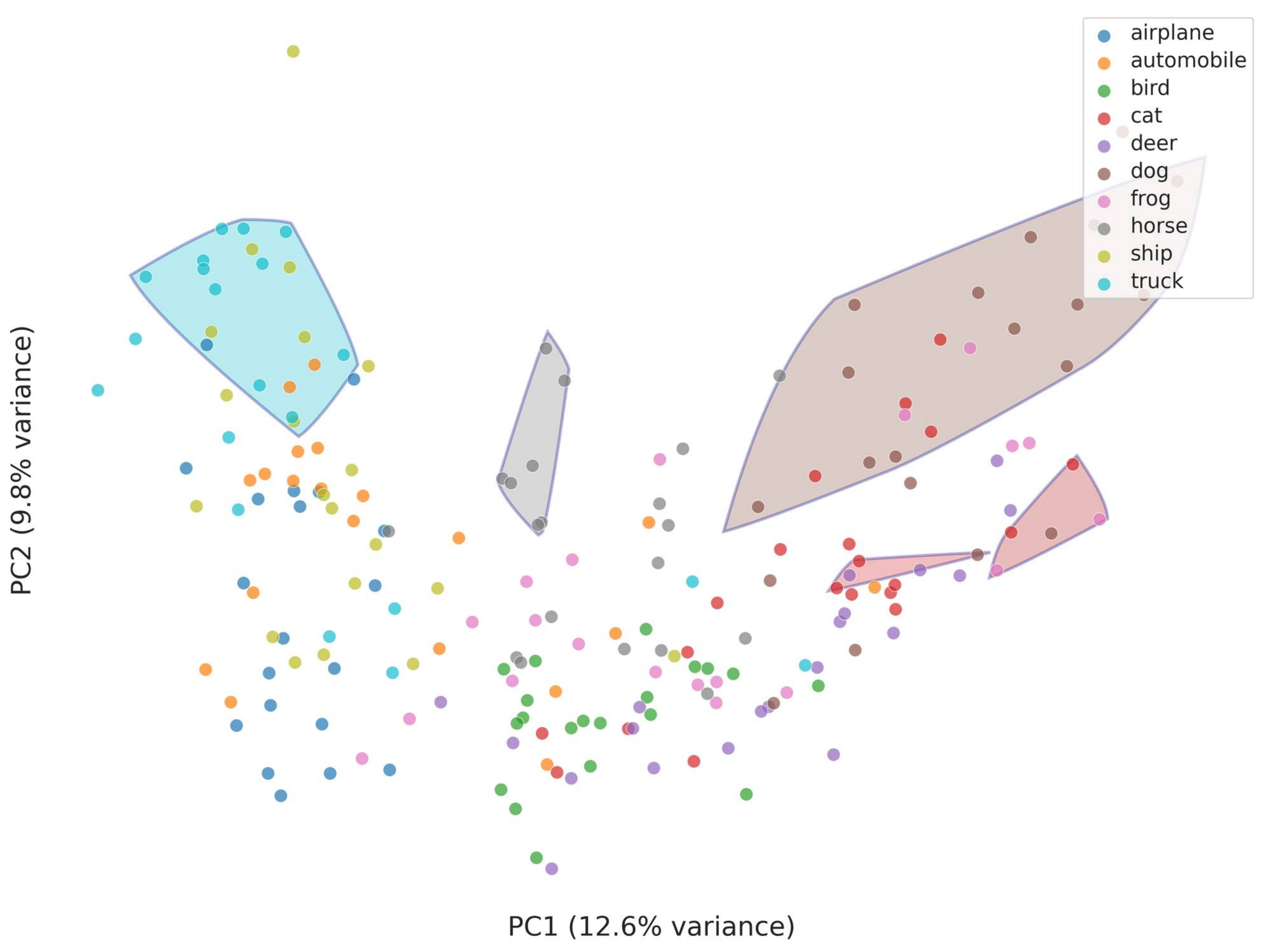}
        \caption{Salt \& Pepper noise.}
        \label{fig:enlarged_blob_sp}
    \end{subfigure}
    \caption{Enlarged version of the input-noise panels of \cref{fig:perturbation_blobs}. PCA \vblob{blob visualizations} of ViT-B/16 encoder layer~11 under noise (class-balanced CIFAR-10, cosine distance). The clean baseline is shown in \cref{fig:enlarged_blob_layer_11}.}
    \label{fig:enlarged_noise_blobs}
\end{figure*}

\begin{figure*}[!htb]
    \centering
    \includegraphics[width=\linewidth]{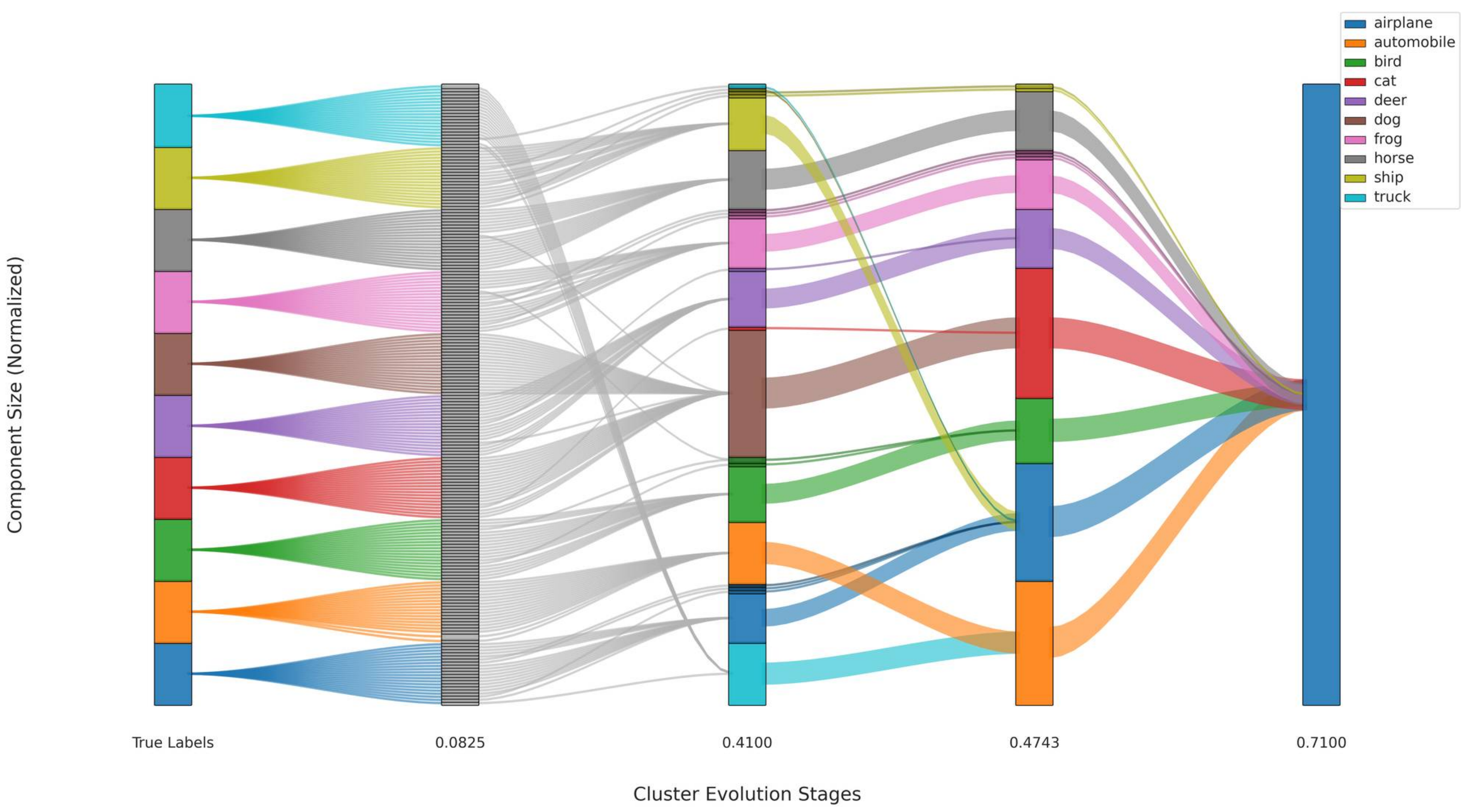}
    \caption{Enlarged version of the Gaussian-noise panel of \cref{fig:perturbation_sankeys}. Gaussian noise \vcflow{cluster flow diagram} for ViT-B/16 encoder layer~11 (class-balanced CIFAR-10, cosine distance). Class flows remain coherent but cross-class merging occurs earlier than the clean baseline.}
    \label{fig:enlarged_noise_sankeys}
\end{figure*}

\begin{figure*}[!htb]
    \centering
    \includegraphics[width=\linewidth]{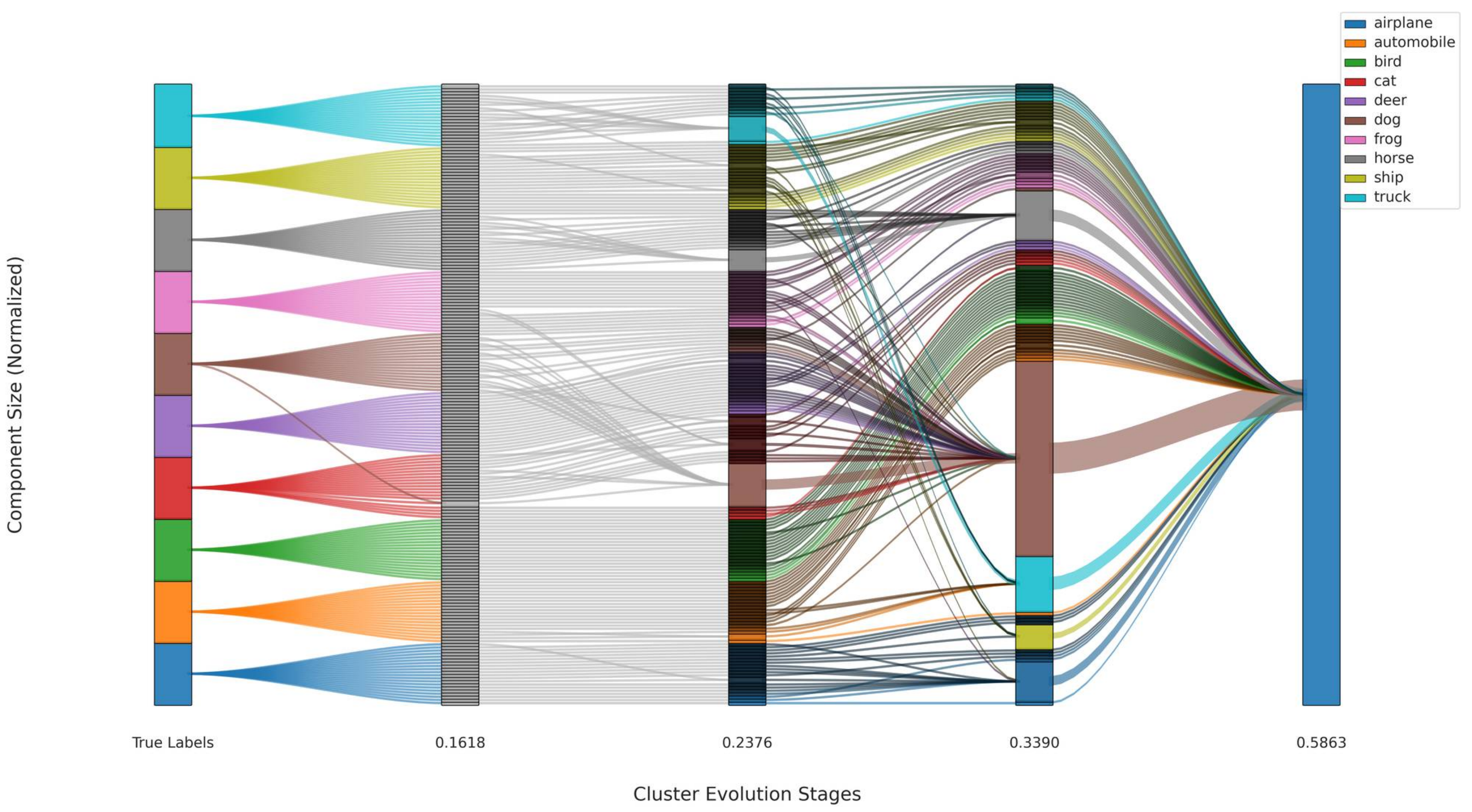}
    \caption{Enlarged version of the Salt \& Pepper panel of \cref{fig:perturbation_sankeys}. Salt \& Pepper noise \vcflow{cluster flow diagram} for ViT-B/16 encoder layer~11 (class-balanced CIFAR-10, cosine distance). Class flows collapse rapidly; most classes merge into a single component by the second filtration stage.}
    \label{fig:enlarged_sankey_sp}
\end{figure*}

\begin{figure*}[!htb]
    \centering
    \includegraphics[width=\linewidth]{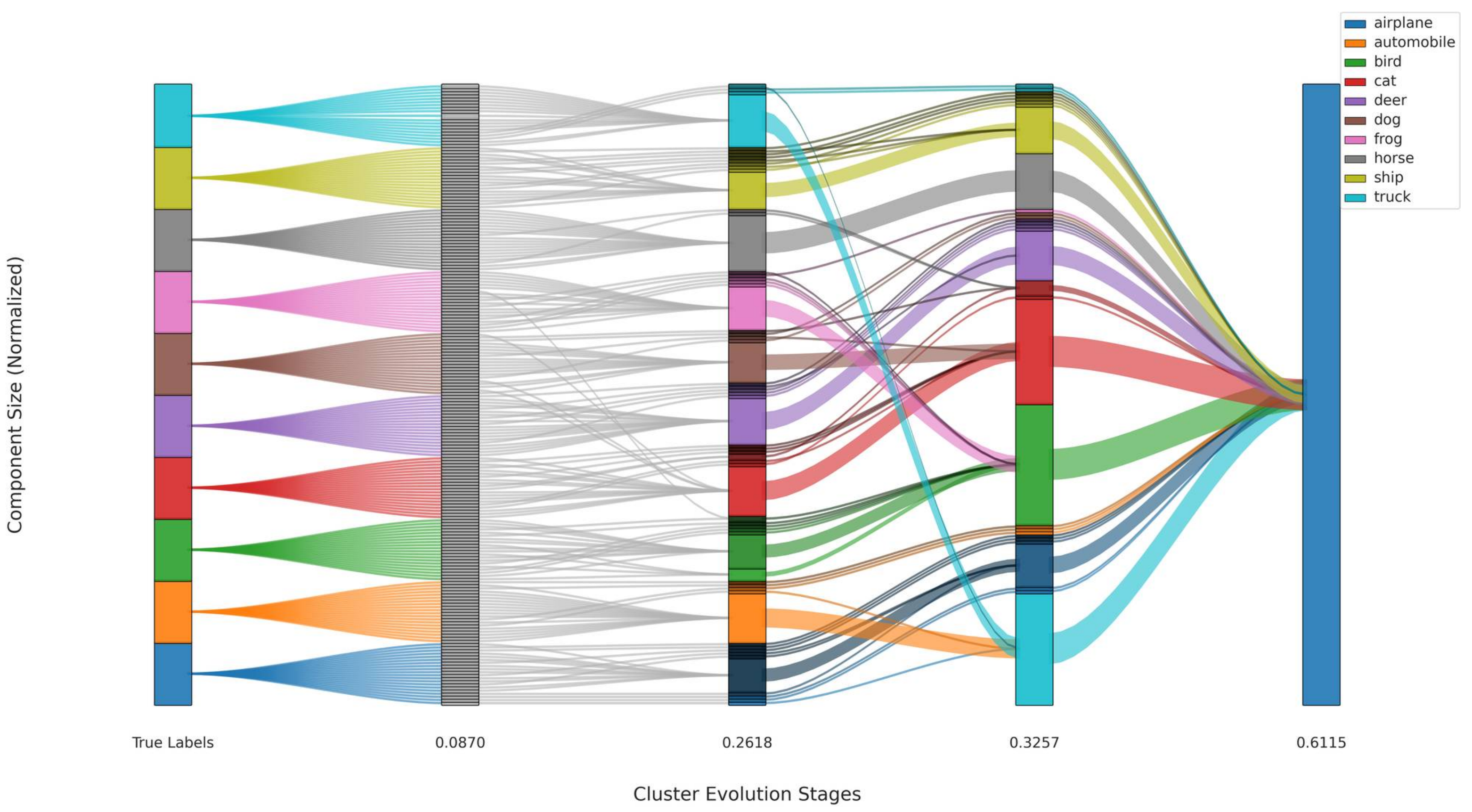}
    \caption{Enlarged version of the INT8 panel of \cref{fig:perturbation_sankeys}. INT8 quantized \vcflow{cluster flow diagram} for ViT-B/16 encoder layer~11 (class-balanced CIFAR-10, cosine distance). Class flows fragment and merge prematurely compared to the FP32 baseline (\cref{fig:enlarged_sankey_layer_11}).}
    \label{fig:enlarged_quant_sankeys}
\end{figure*}

\begin{figure*}[!htb]
    \centering
    \begin{subfigure}{0.48\linewidth}
        \centering
        \includegraphics[width=\linewidth]{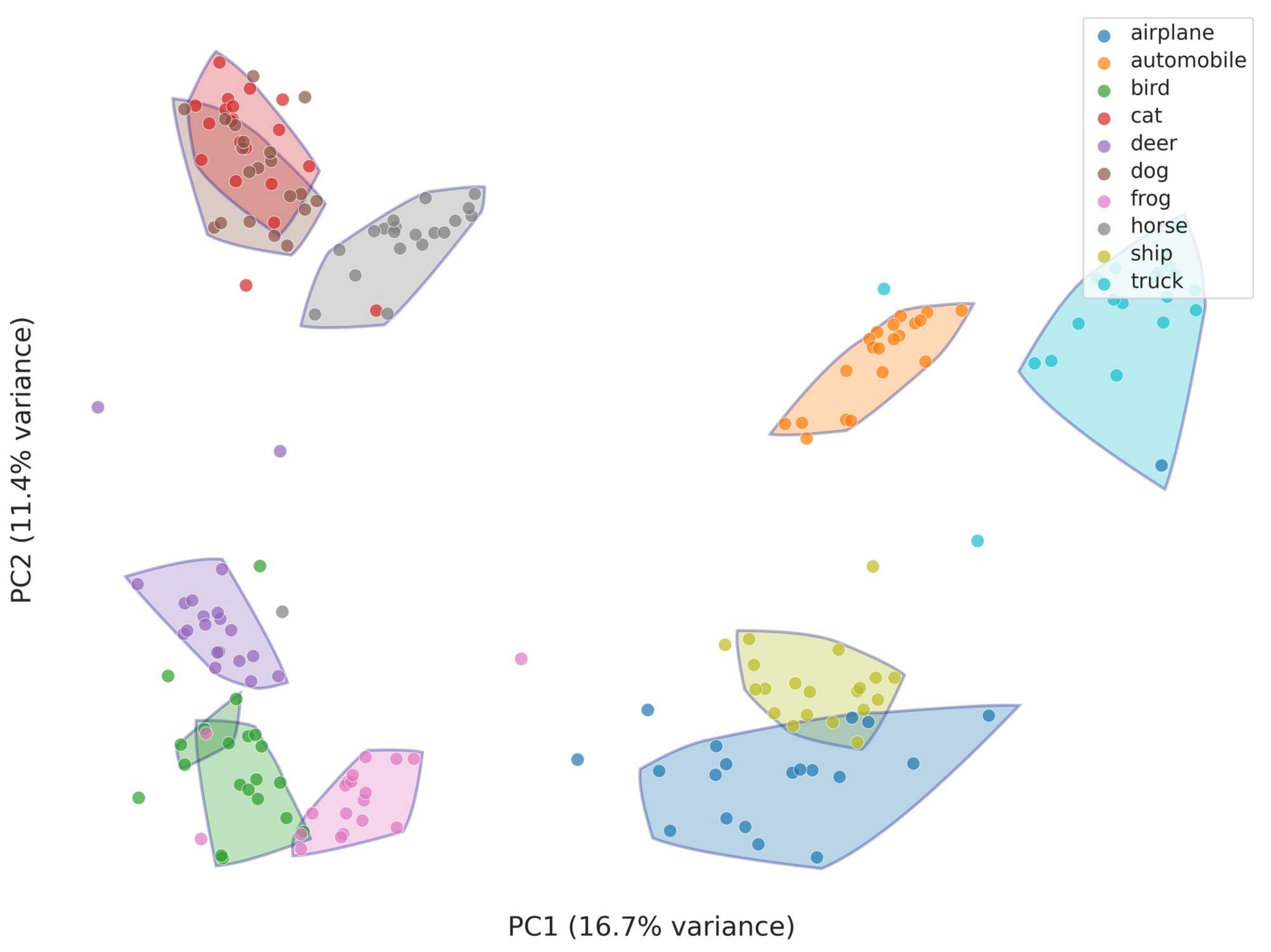}
        \caption{FP32 baseline.}
        \label{fig:enlarged_blob_fp32}
    \end{subfigure}\hfill
    \begin{subfigure}{0.48\linewidth}
        \centering
        \includegraphics[width=\linewidth]{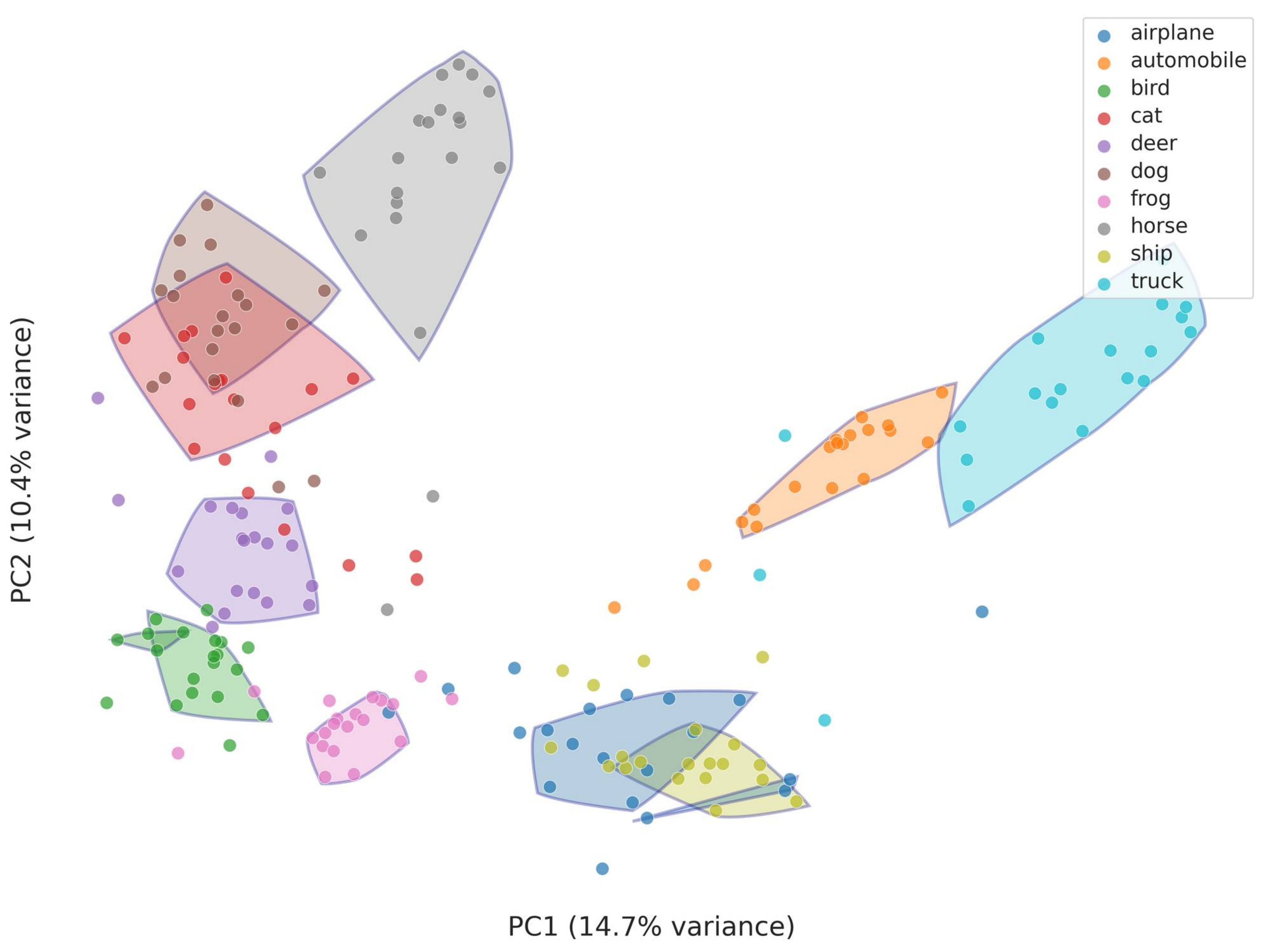}
        \caption{INT8 quantized.}
        \label{fig:enlarged_blob_int8}
    \end{subfigure}
    \caption{Enlarged version of the quantization panels of \cref{fig:perturbation_blobs}. PCA \vblob{blob visualizations} of ViT-B/16 encoder layer~11 before and after INT8 dynamic quantization (class-balanced CIFAR-10, cosine distance).}
    \label{fig:enlarged_quant_blobs}
\end{figure*}

%% ----------------------------------------------------------------
\section{ResNet-50 Analysis}
\label{sec:rn50_analysis}
%% ----------------------------------------------------------------

This section mirrors the three experiments in the main paper for a ResNet-50 backbone fine-tuned on CIFAR-10 (89.5\% test accuracy, 25.6M parameters).
The probe layer is \emph{stage~4} (the output of the fourth residual block group, i.e., index~3 when zero-indexed), the deepest convolutional feature map before the global average-pooling head.
All figures use cosine distance, which is the more discriminative metric for ResNet-50 stage~4 activations in $\mathbb{R}^{2048}$ due to the high-dimensional concentration effect discussed in \cref{subsec:distance-metrics}~\cite{aggarwal2001surprising,beyer1999nearest}.

\subsection{Application 1: Learned Representation Analysis (ResNet-50)}
\label{sec:rn50_app1}

\begin{figure*}[!htb]
    \centering
    \begin{subfigure}{0.8\linewidth}
        \centering
        \includegraphics[width=\linewidth]{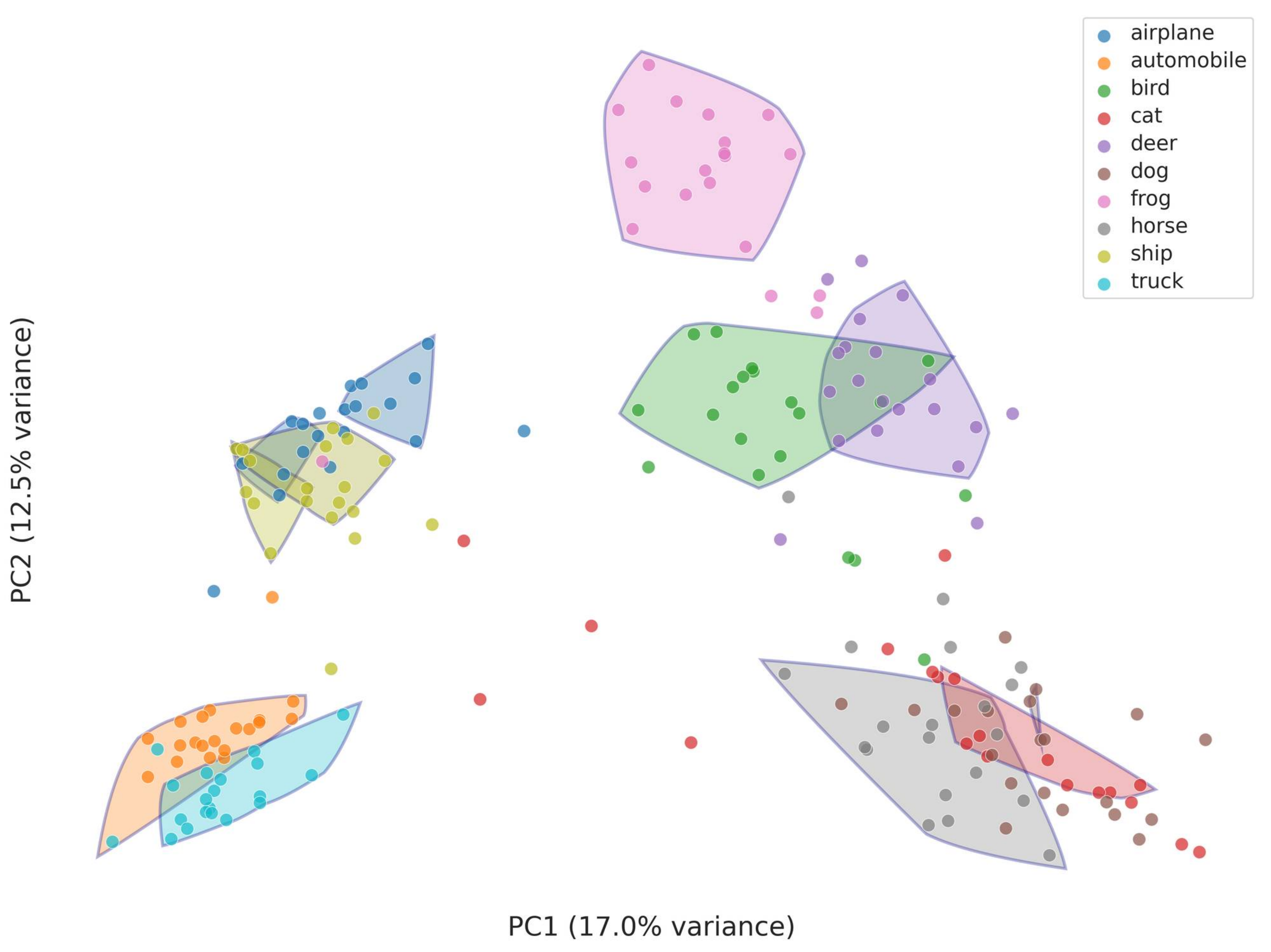}
        \caption{PCA \vblob{blob graph}, cosine distance. Classes are moderately separated with broad intra-class spread, consistent with the lower test accuracy (89.5\%) compared to ViT-B/16.}
        \label{fig:rn50_blob_cos}
    \end{subfigure}

    \vspace{0.5em}

    \begin{subfigure}{0.8\linewidth}
        \centering
        \includegraphics[width=\linewidth]{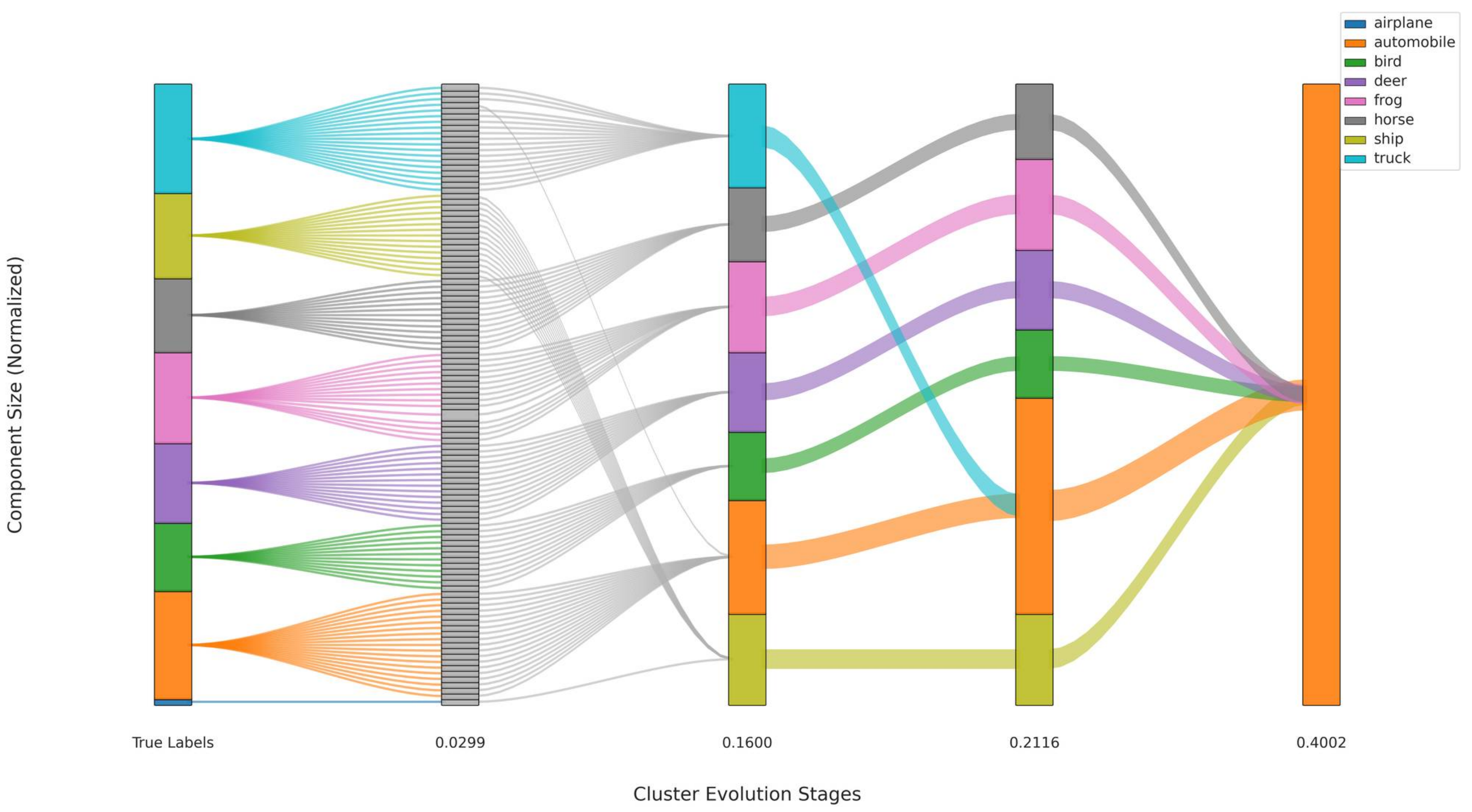}
        \caption{\vcflow{Cluster flow}, cosine distance. Per-class flows persist across filtration stages but merge earlier than in ViT-B/16 (\cref{fig:learned_rep_comparison}), reflecting weaker class separation.}
        \label{fig:rn50_sankey_cos}
    \end{subfigure}

    \caption{Learned representation analysis for ResNet-50 stage~4, class-balanced CIFAR-10, cosine distance. The \vblob{blob graph} shows moderate class-level separation, while the \vcflow{cluster flow} reveals that per-class components merge at relatively early filtration thresholds.}
    \label{fig:rn50_learned_rep}
\end{figure*}

ResNet-50 stage~4 produces moderate class-level separation (89.5\% accuracy), with broader intra-class spread than ViT-B/16 layer~11 (\cref{fig:learned_rep_comparison}).
The cosine \vcflow{cluster flow} (\cref{fig:rn50_sankey_cos}) shows that per-class flows merge earlier than in ViT-B/16, confirming that the weaker model produces less topologically distinct class representations.

\subsection{Application 2: Robustness Analysis Under Noise (ResNet-50)}
\label{sec:rn50_app2}

\begin{figure*}[!htb]
    \centering
    \begin{subfigure}{0.8\linewidth}
        \centering
        \includegraphics[width=\linewidth]{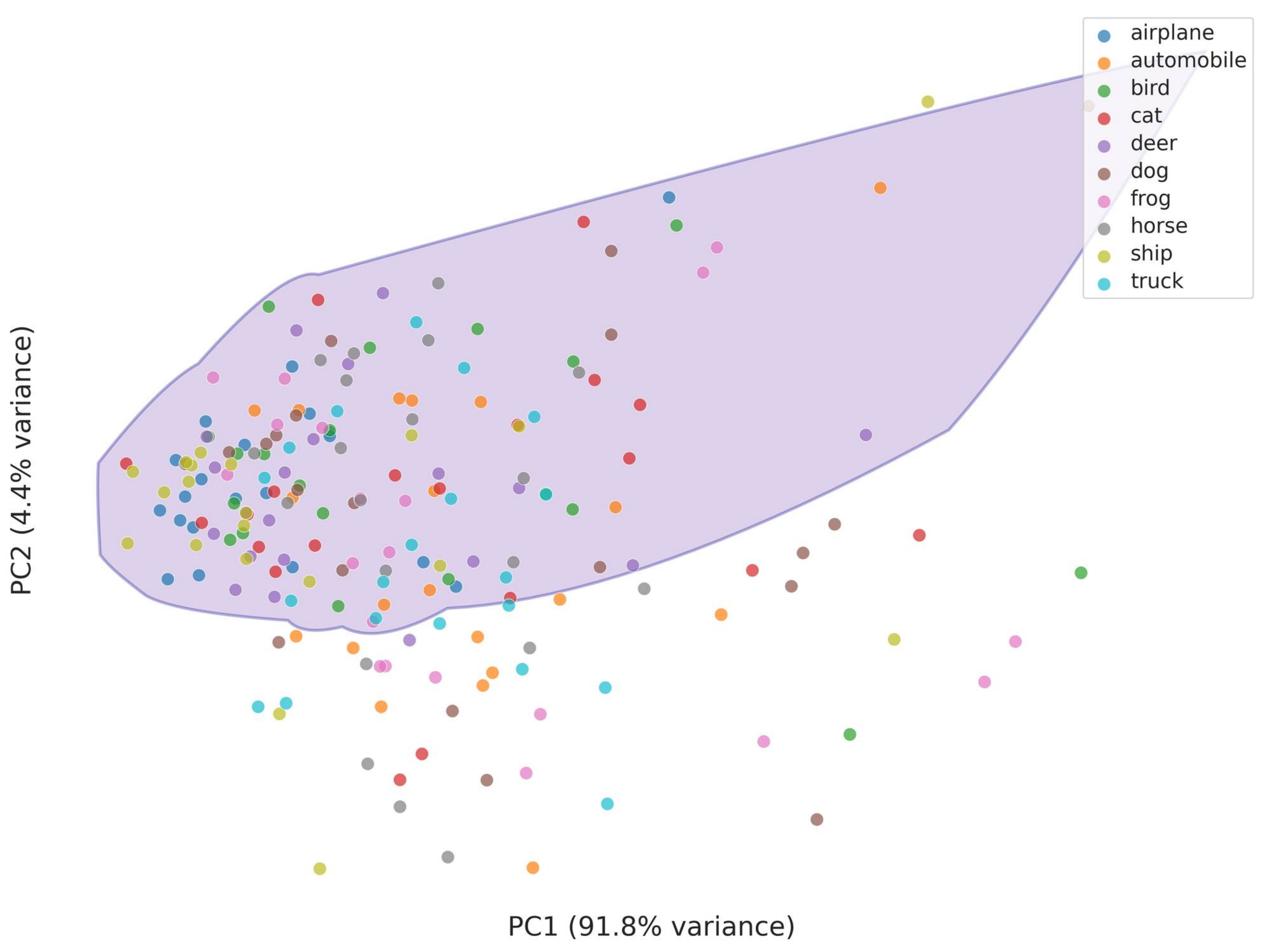}
        \caption{Salt \& Pepper \vblob{blob graph}, cosine distance. Near-complete cluster collapse; accuracy 10.0\%.}
        \label{fig:rn50_blob_sp_cos}
    \end{subfigure}

    \vspace{0.5em}

    \begin{subfigure}{0.8\linewidth}
        \centering
        \includegraphics[width=\linewidth]{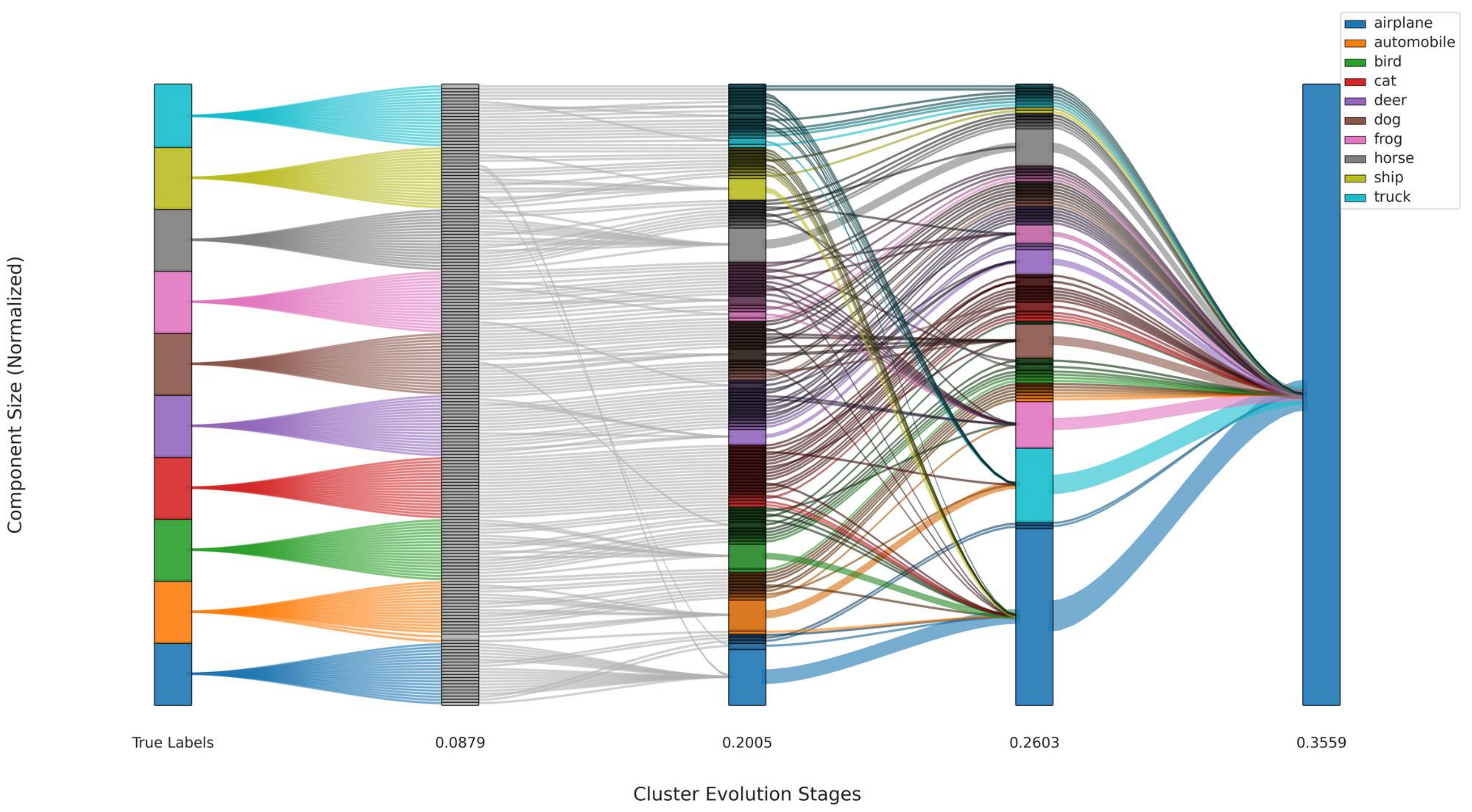}
        \caption{Gaussian noise \vcflow{cluster flow}, cosine distance (non-filtered). Class trajectories partially survive at early filtration thresholds; accuracy 63.0\%.}
        \label{fig:rn50_sankey_gauss_cos}
    \end{subfigure}

    \caption{ResNet-50 stage~4 under noise, class-balanced CIFAR-10, cosine distance. Salt \& Pepper noise (top) destroys cluster structure entirely, consistent with the 10.0\% accuracy collapse. Gaussian noise (bottom, 63.0\% accuracy) preserves partial per-class structure visible in the \vcflow{cluster flow}.}
    \label{fig:rn50_noise}
\end{figure*}

ResNet-50 is substantially more vulnerable to input noise than ViT-B/16 (\cref{fig:perturbation_blobs,fig:perturbation_sankeys}).
Under Salt \& Pepper noise (10.0\% accuracy), the cosine \vblob{blob graph} (\cref{fig:rn50_blob_sp_cos}) shows near-complete cluster collapse, confirming that extreme impulse corruption destroys the feature geometry.
For Gaussian noise (63.0\% accuracy), the cosine \vcflow{cluster flow} (\cref{fig:rn50_sankey_gauss_cos}) retains partially visible per-class trajectories, indicating that some class structure survives moderate additive noise even in the weaker ResNet-50 backbone.

\subsection{Application 3: Model Compression (ResNet-50)}
\label{sec:rn50_app3}

\begin{figure*}[!htb]
    \centering
    \includegraphics[width=0.85\linewidth]{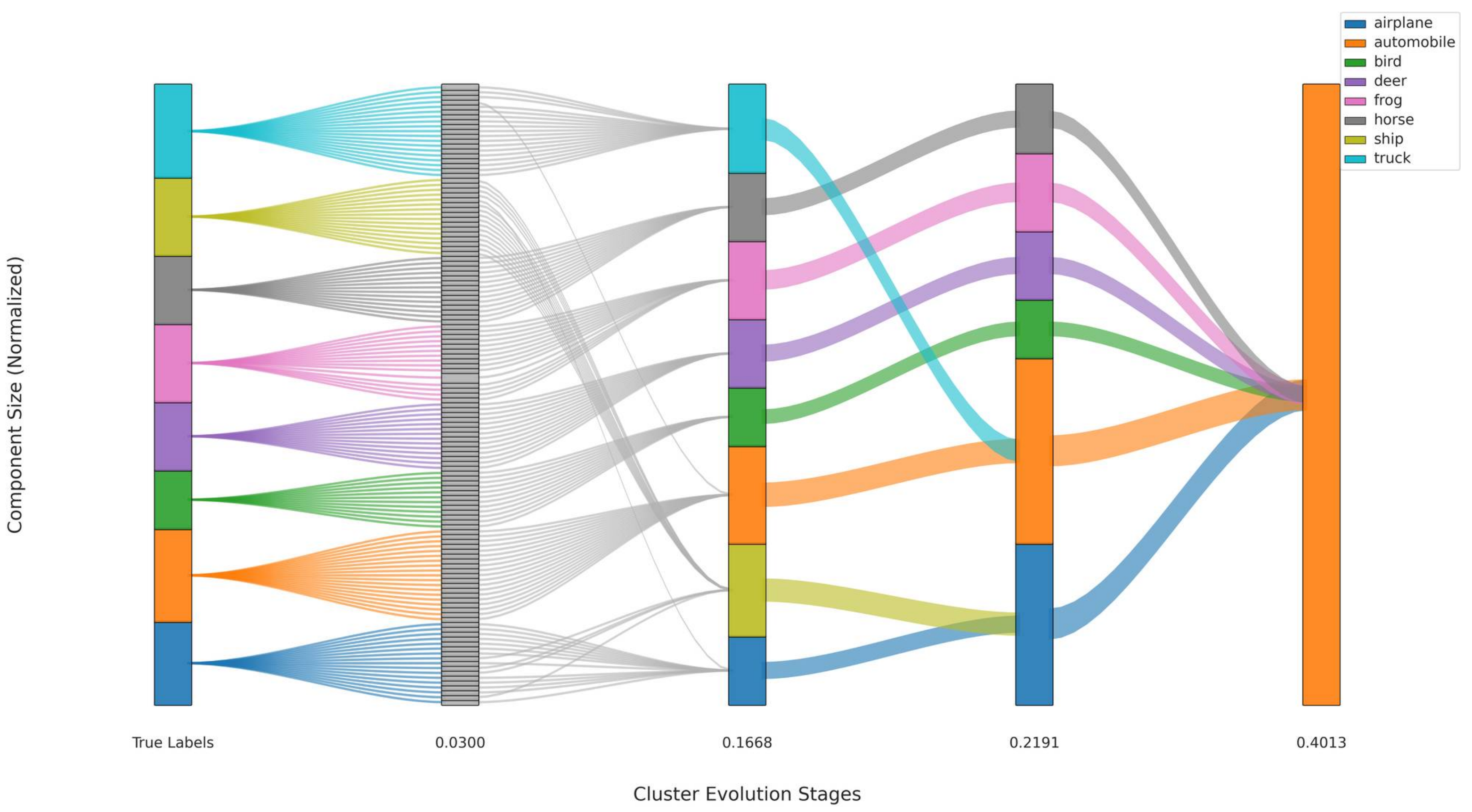}
    \caption{ResNet-50 stage~4 INT8 dynamic quantization, cosine distance, class-balanced CIFAR-10, filtered \vcflow{cluster flow} diagram. Weight snapping introduces directional noise; flows degrade relative to the FP32 baseline (\cref{fig:rn50_sankey_cos}). Accuracy is identical (89.5\%) yet the \vcflow{cluster flow} diagrams expose topological changes from weight discretisation.}
    \label{fig:rn50_quant}
\end{figure*}

Dynamic INT8 quantization preserves ResNet-50's test accuracy entirely (89.5\%~$\to$~89.5\%), yet the cosine \vcflow{cluster flow diagrams} expose topological perturbations that the accuracy metric cannot.
In the FP32 baseline (\cref{fig:rn50_sankey_cos}), class flows are coherent across filtration stages with strong per-class persistence.
After INT8 quantization (\cref{fig:rn50_quant}), flows degrade, a topological fingerprint of local neighborhood perturbations introduced by weight snapping.
As with ViT (Application~3 in the main text), HOLE reveals representational changes that accuracy benchmarks cannot.

%% ----------------------------------------------------------------
% BERT NER Analysis -- main content moved to experiments.tex (Application 1)
\label{sec:bert_ner_analysis} % keep label so existing \cref's resolve

\section{BERT NER: Layer-wise Comparison}
\label{sec:bert_ner_layers}

\begin{figure*}[!htb]
    \centering
    \begin{subfigure}{\linewidth}
        \centering
        \includegraphics[width=\linewidth]{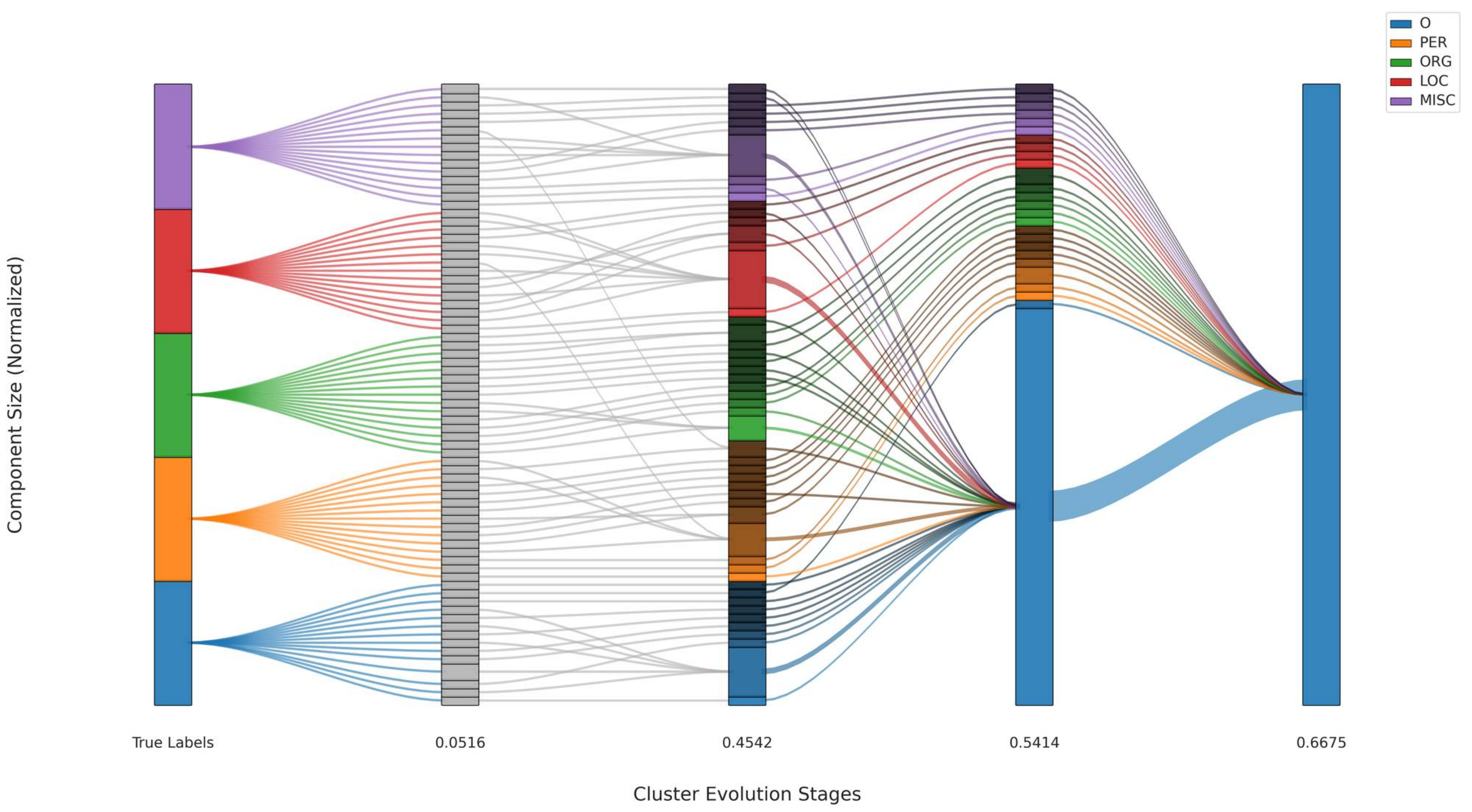}
        \caption{\vcflow{Cluster flow}, \textit{layer~4}. Entity-type flows merge rapidly; little discriminative structure.}
        \label{fig:bert_ner_sankey_4}
    \end{subfigure}

    \vspace{0.6em}

    \begin{subfigure}{\linewidth}
        \centering
        \includegraphics[width=\linewidth]{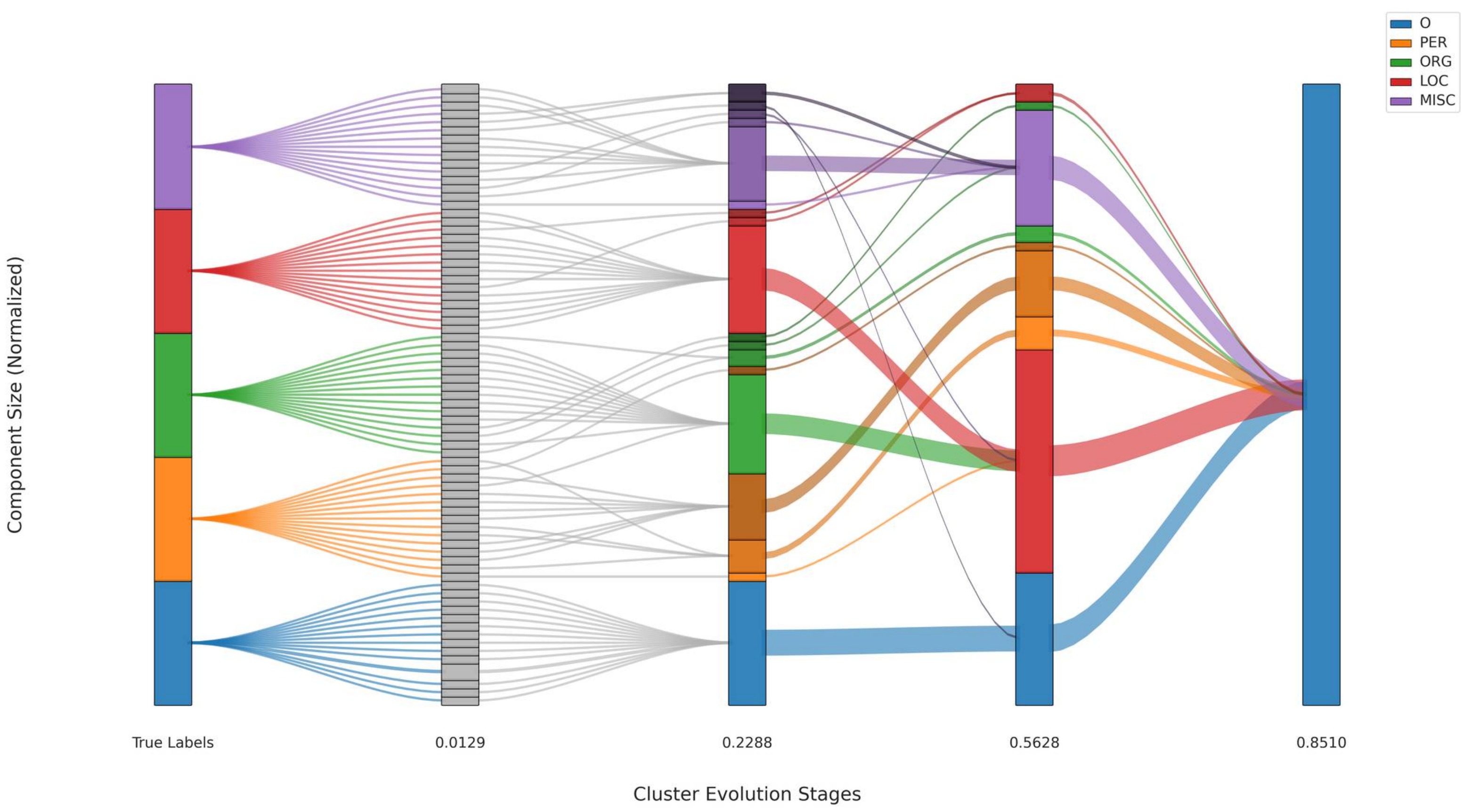}
        \caption{\vcflow{Cluster flow}, \textit{layer~11}. Coherent per-type flows persist across filtration stages.}
        \label{fig:bert_ner_sankey_11_appendix}
    \end{subfigure}
    \caption{BERT-base NER layer-wise analysis (CoNLL-2003, cosine distance). Entity-type clustering progressively emerges from layer~4 (a) to layer~11 (b). The corresponding \vhmd{heatmap dendrogram} for layer~11 is shown in \cref{fig:enlarged_bert_hmd_11}.}
    \label{fig:bert_ner_layer_comparison}
\end{figure*}

\Cref{fig:bert_ner_layer_comparison} compares BERT encoder layers~4 and~11 for NER entity-type clustering.
In the early layer (layer~4), the \vcflow{cluster flow diagram} shows minimal entity-type separation: most flows merge rapidly, indicating that the activation space has not yet developed discriminative structure for the NER task.
By layer~11, coherent per-type flows persist across multiple filtration stages, mirroring the progressive emergence of class-discriminative topology observed in the vision experiments (\cref{sec:app1}).

\section{Stability Analysis Across Probe-Set Seeds}
\label{sec:stability_analysis}
%% ----------------------------------------------------------------

To assess whether HOLE's topological conclusions depend on the particular probe-set sample, we repeat the learned representation analysis (\cref{sec:app1}) for ViT-B/16 on CIFAR-10 using 10 different random seeds (seeds: 42, 123, 256, 314, 555, 678, 821, 937, 1001, 1234), each producing a different class-balanced subset of 200 test images (20 per class).
All figures use cosine distance and the unfiltered cluster flow variant to expose the full component structure.

\paragraph{Quantitative Summary}
\Cref{tab:stability_summary} reports the probe-set accuracy and the number of $H_0$ components at the first filtration threshold for encoder layers~9 and~11 across all 10 seeds.
Layer~11 filtered cluster counts are consistently 9 or 10 (matching the 10 CIFAR-10 classes), confirming that the class-discriminative topology is a property of the model, not an artifact of any single probe sample.
Early-layer cluster counts (layer~9) are higher and more variable, reflecting the weaker and noisier representational structure at that depth.

\begin{table}[!htb]
    \centering
    \caption{Stability of HOLE across 10 random probe-set seeds for ViT-B/16 on CIFAR-10 (cosine distance, class-balanced, 200 images per seed). Cluster counts are reported at the first filtration threshold (unfiltered). Filtered layer~11 counts are consistently 9--10, matching the number of ground-truth classes.}
    \label{tab:stability_summary}
    \small
    \begin{tabular}{lcrrr}
        \toprule
        \textbf{Seed} & \textbf{Accuracy} & \textbf{L9 (nf)} & \textbf{L11 (nf)} & \textbf{L11 (f)} \\
        \midrule
        42   & 96.0\% & 164 & 39 & 10 \\
        123  & 96.0\% & 180 & 18 & 10 \\
        256  & 98.0\% & 126 & 40 &  9 \\
        314  & 96.5\% & 135 & 38 & 10 \\
        555  & 97.0\% & 105 & 35 & 10 \\
        678  & 95.0\% & 150 & 31 & 10 \\
        821  & 95.0\% & 149 & 34 &  9 \\
        937  & 97.5\% & 165 & 52 &  9 \\
        1001 & 96.0\% & 182 & 26 & 10 \\
        1234 & 97.5\% & 156 & 31 &  9 \\
        \midrule
        \textbf{Mean$\pm$Std} & 96.5$\pm$1.0\% & 151$\pm$23 & 34$\pm$9 & 9.6$\pm$0.5 \\
        \bottomrule
    \end{tabular}
\end{table}

\begin{figure*}[!htb]
    \centering
    \begin{subfigure}{0.48\linewidth}
        \includegraphics[width=\linewidth]{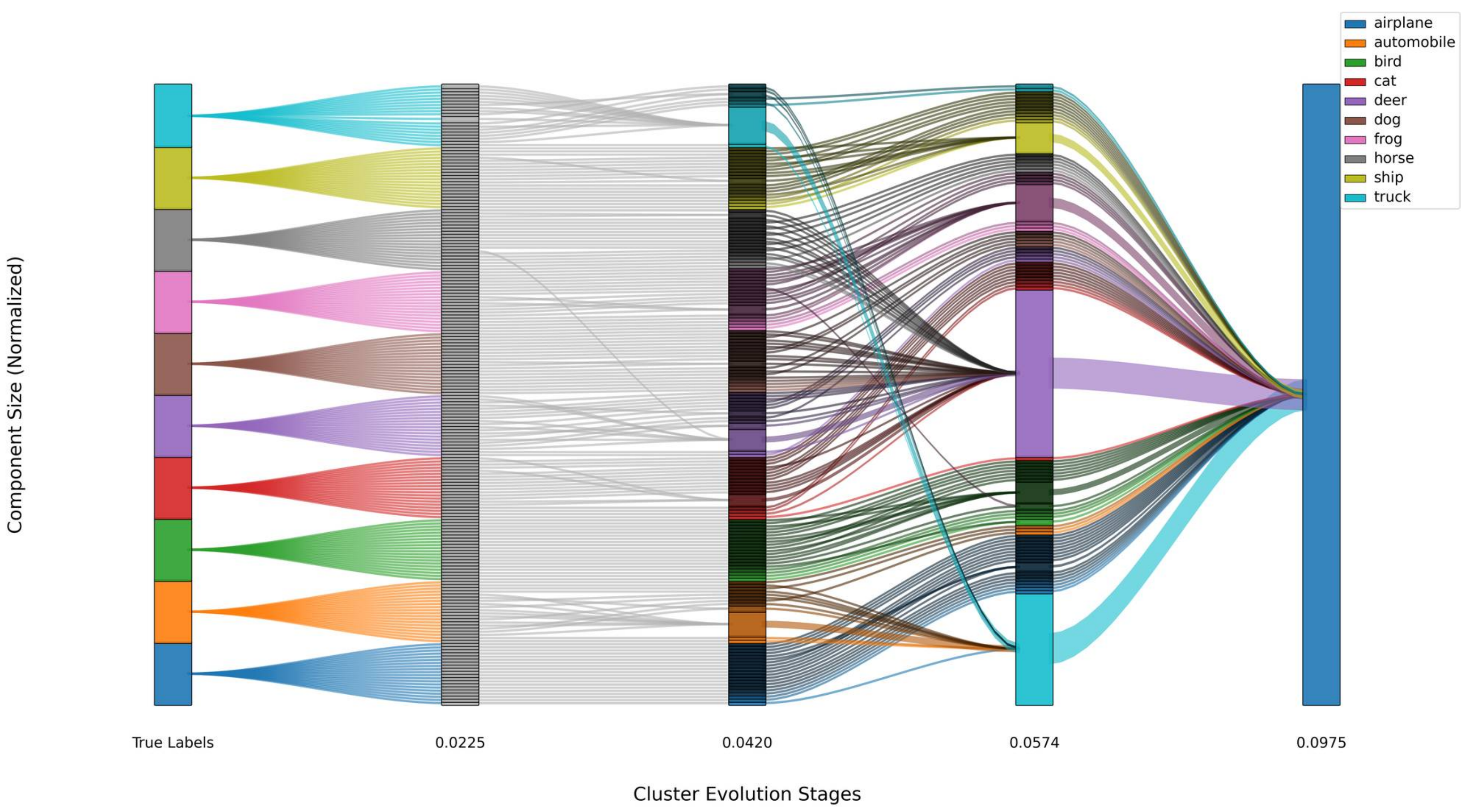}
        \caption{Seed 42 -- \vcflow{cluster flow}}
    \end{subfigure}\hfill
    \begin{subfigure}{0.48\linewidth}
        \includegraphics[width=\linewidth]{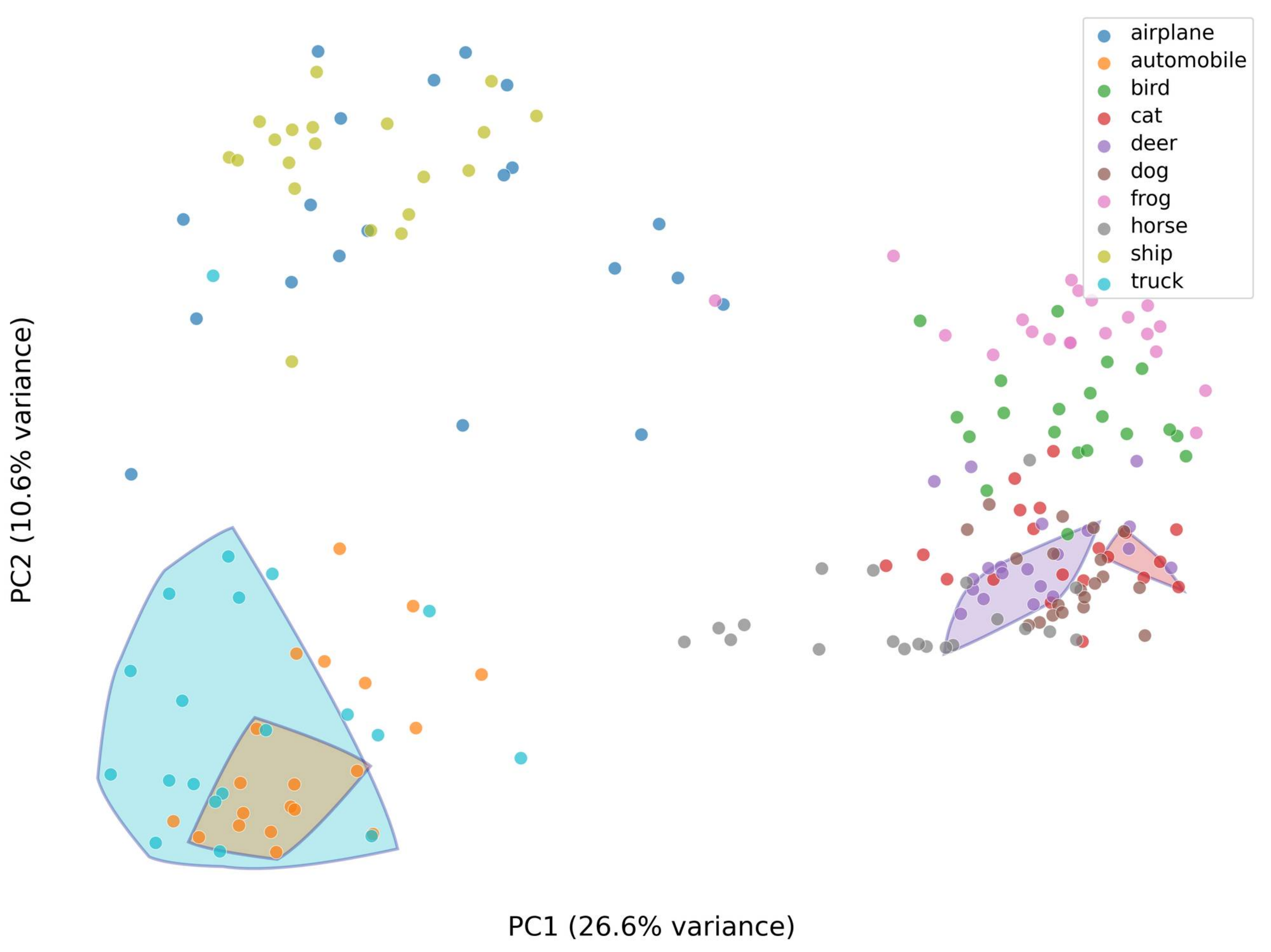}
        \caption{Seed 42 -- \vblob{blob}}
    \end{subfigure}

    \vspace{0.4em}

    \begin{subfigure}{0.48\linewidth}
        \includegraphics[width=\linewidth]{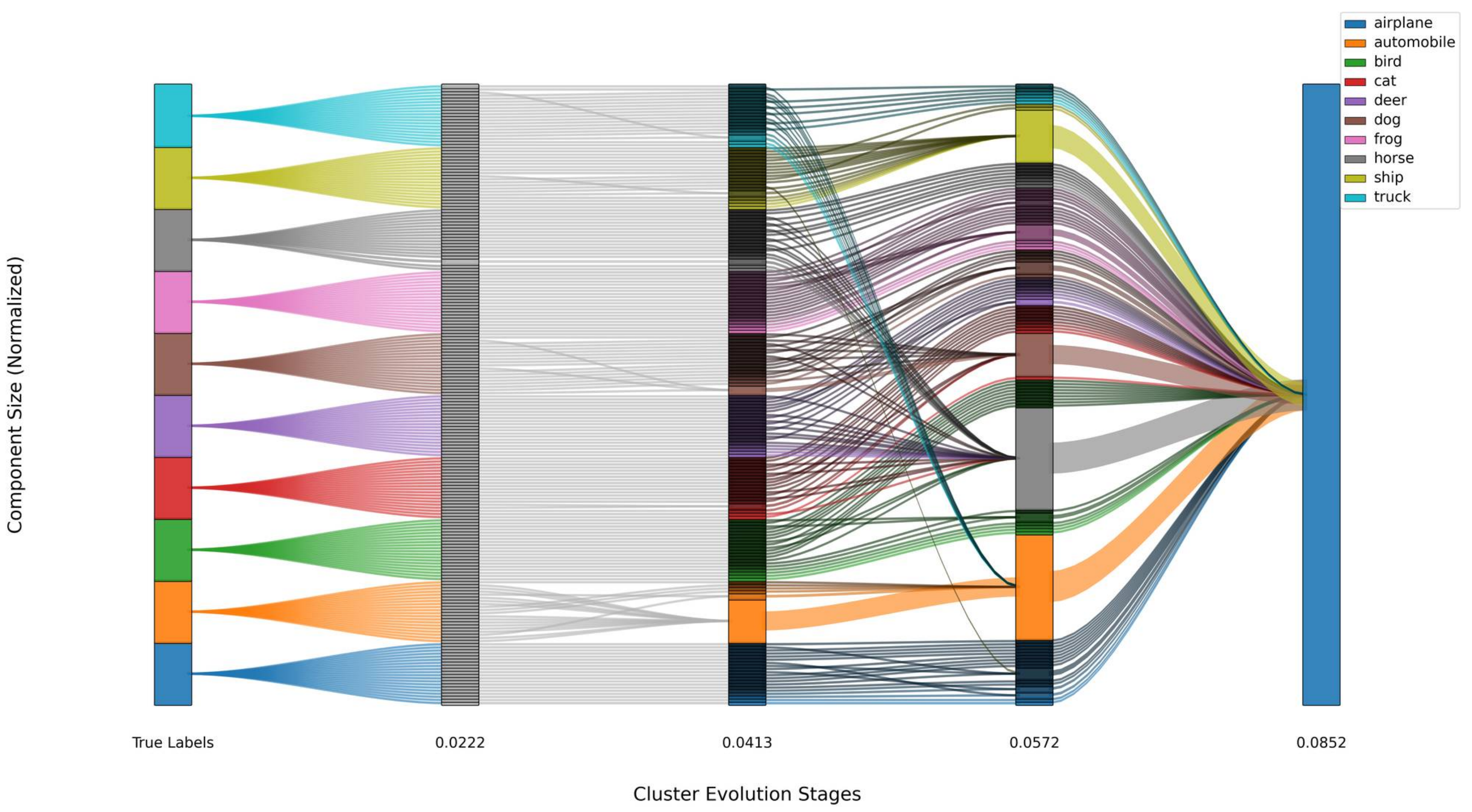}
        \caption{Seed 123 -- \vcflow{cluster flow}}
    \end{subfigure}\hfill
    \begin{subfigure}{0.48\linewidth}
        \includegraphics[width=\linewidth]{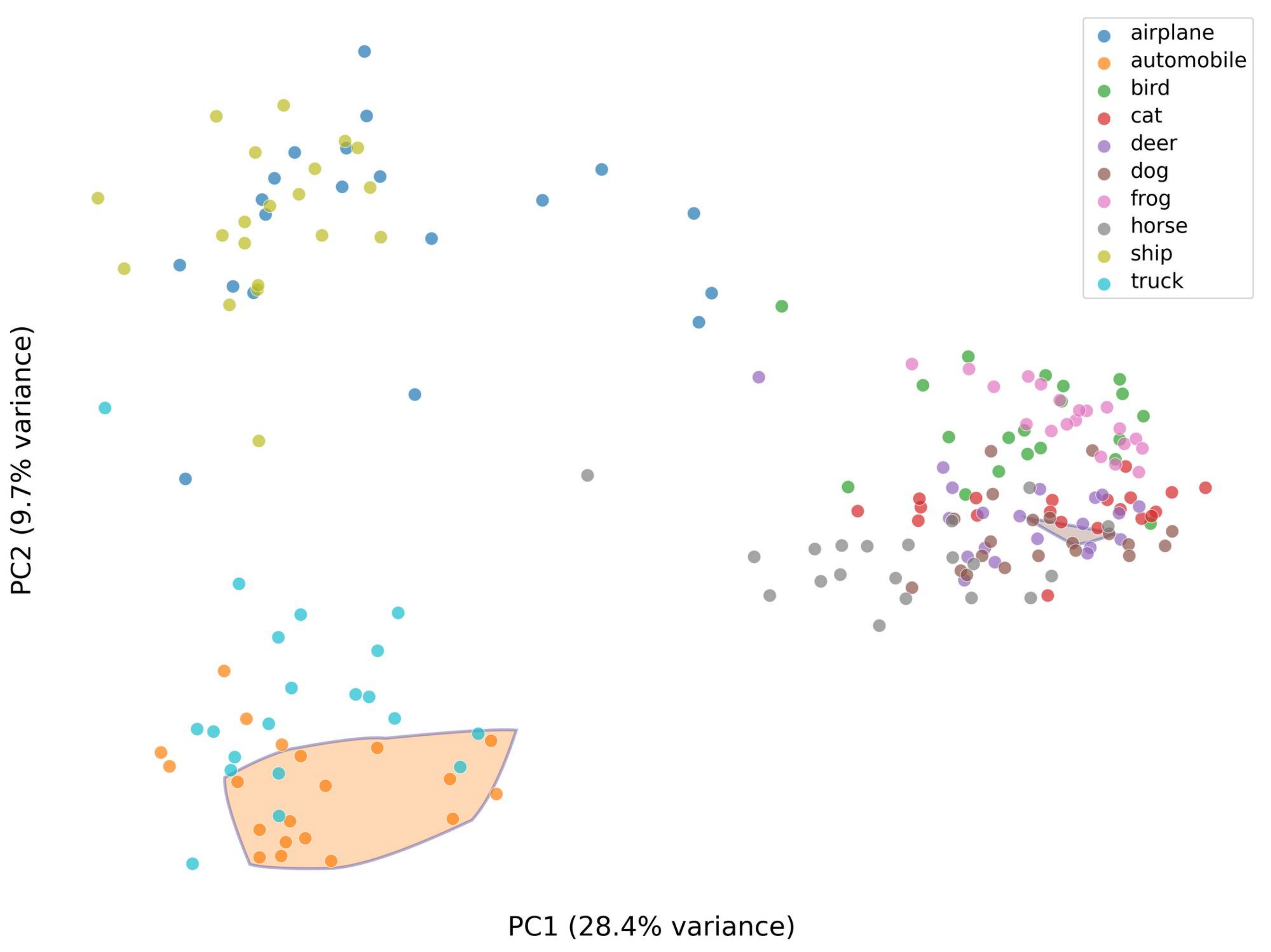}
        \caption{Seed 123 -- \vblob{blob}}
    \end{subfigure}

    \vspace{0.4em}

    \begin{subfigure}{0.48\linewidth}
        \includegraphics[width=\linewidth]{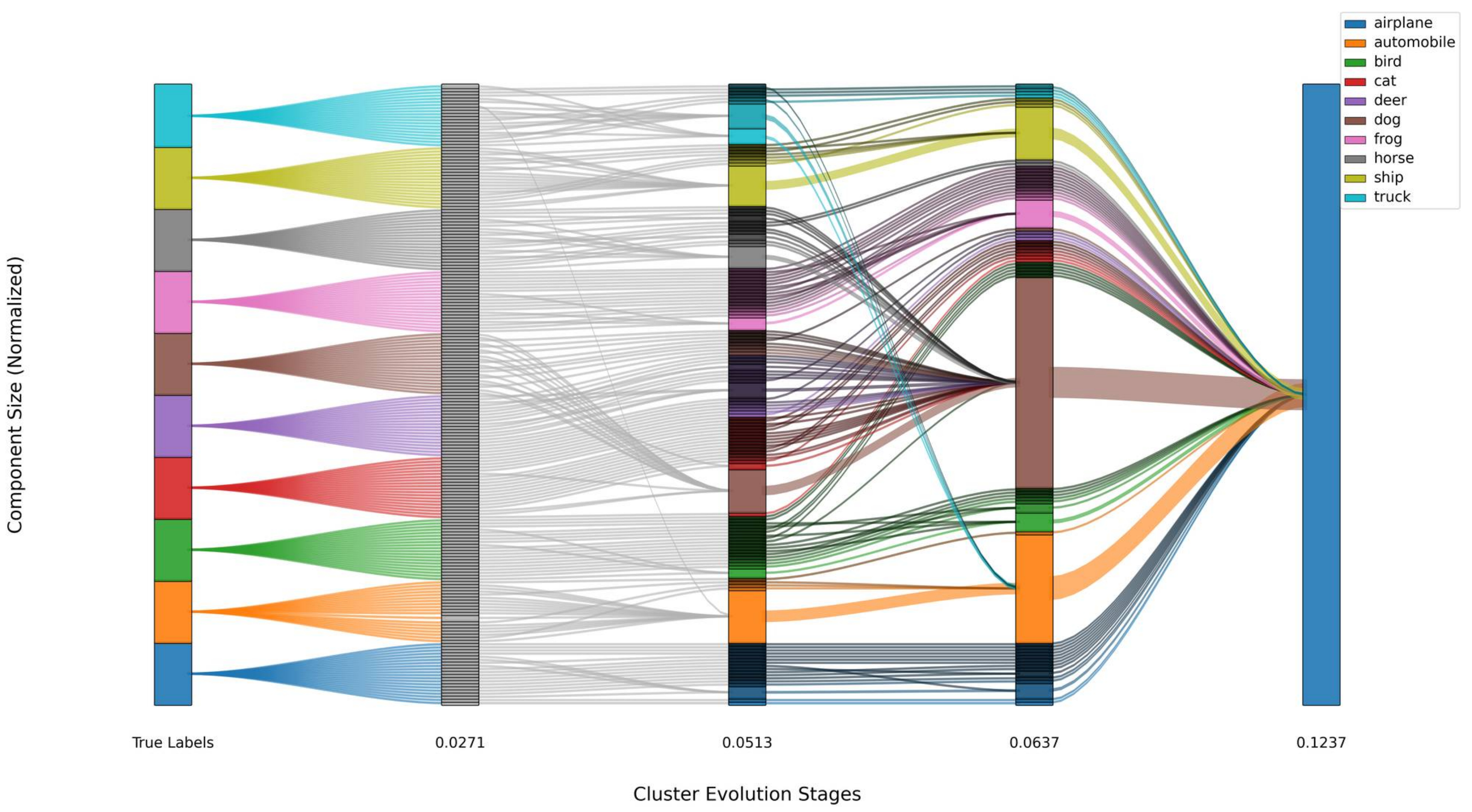}
        \caption{Seed 256 -- \vcflow{cluster flow}}
    \end{subfigure}\hfill
    \begin{subfigure}{0.48\linewidth}
        \includegraphics[width=\linewidth]{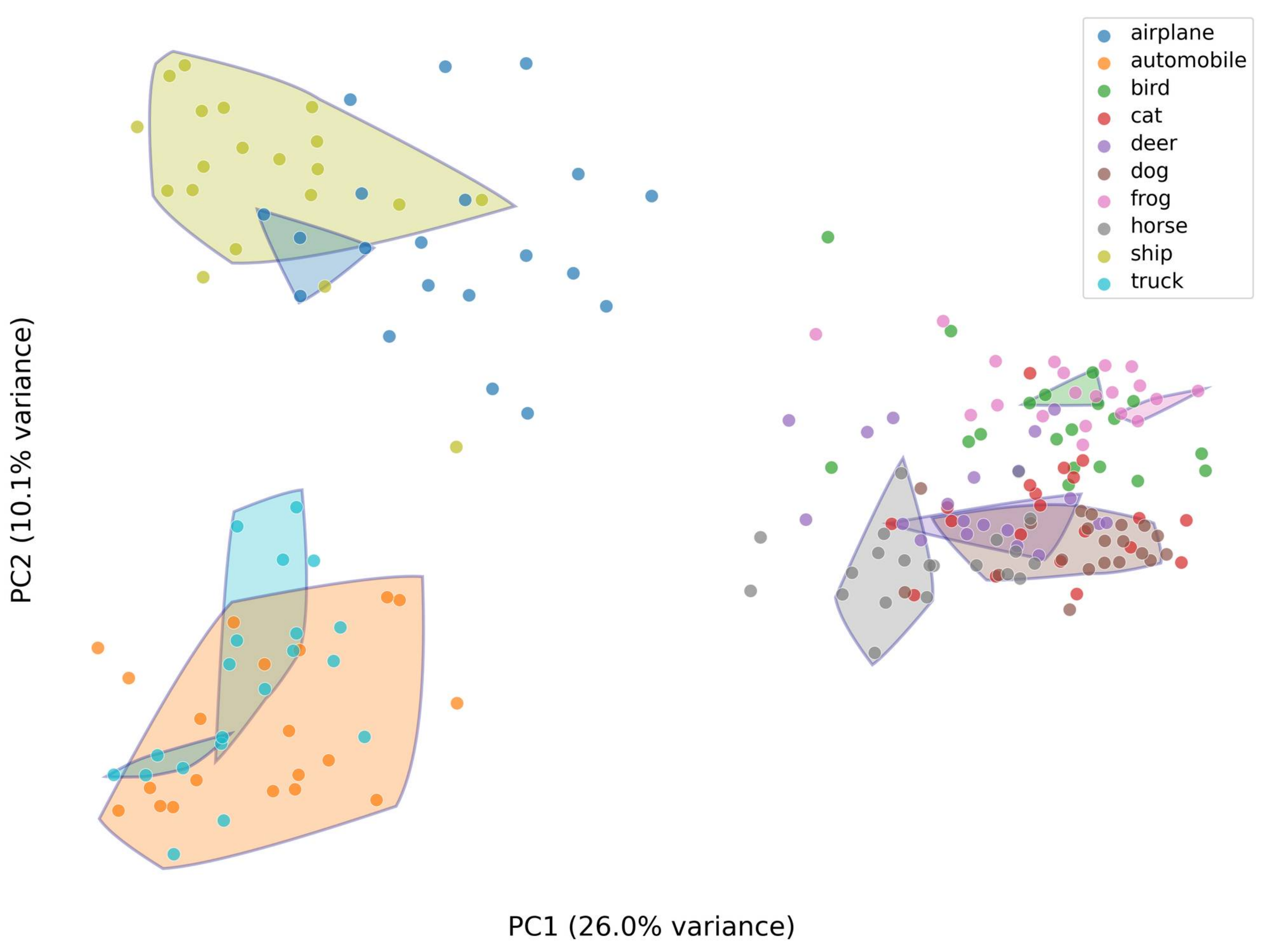}
        \caption{Seed 256 -- \vblob{blob}}
    \end{subfigure}
    \caption{Stability analysis for ViT-B/16 encoder layer~9 (seeds 42--256). Each row: \vcflow{cluster flow} (left) and \vblob{blob graph} (right). All seeds show rapid merging and weak class separation. Continued in \cref{fig:stability_l9_p2}.}
    \label{fig:stability_l9}
    \label{fig:stability_l9_sankeys}
\end{figure*}

\begin{figure*}[!htb]
    \centering
    \begin{subfigure}{0.48\linewidth}
        \includegraphics[width=\linewidth]{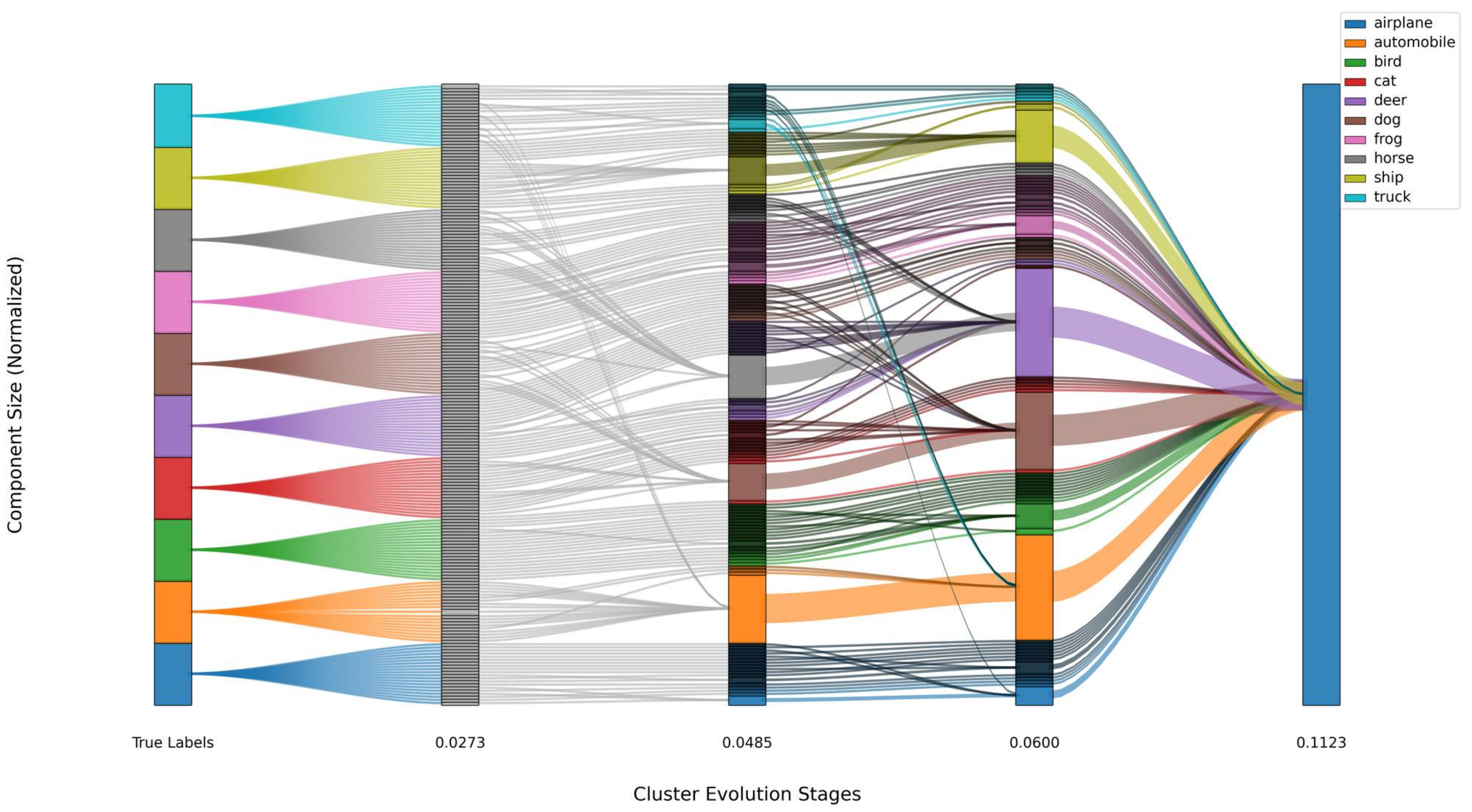}
        \caption{Seed 314 -- \vcflow{cluster flow}}
    \end{subfigure}\hfill
    \begin{subfigure}{0.48\linewidth}
        \includegraphics[width=\linewidth]{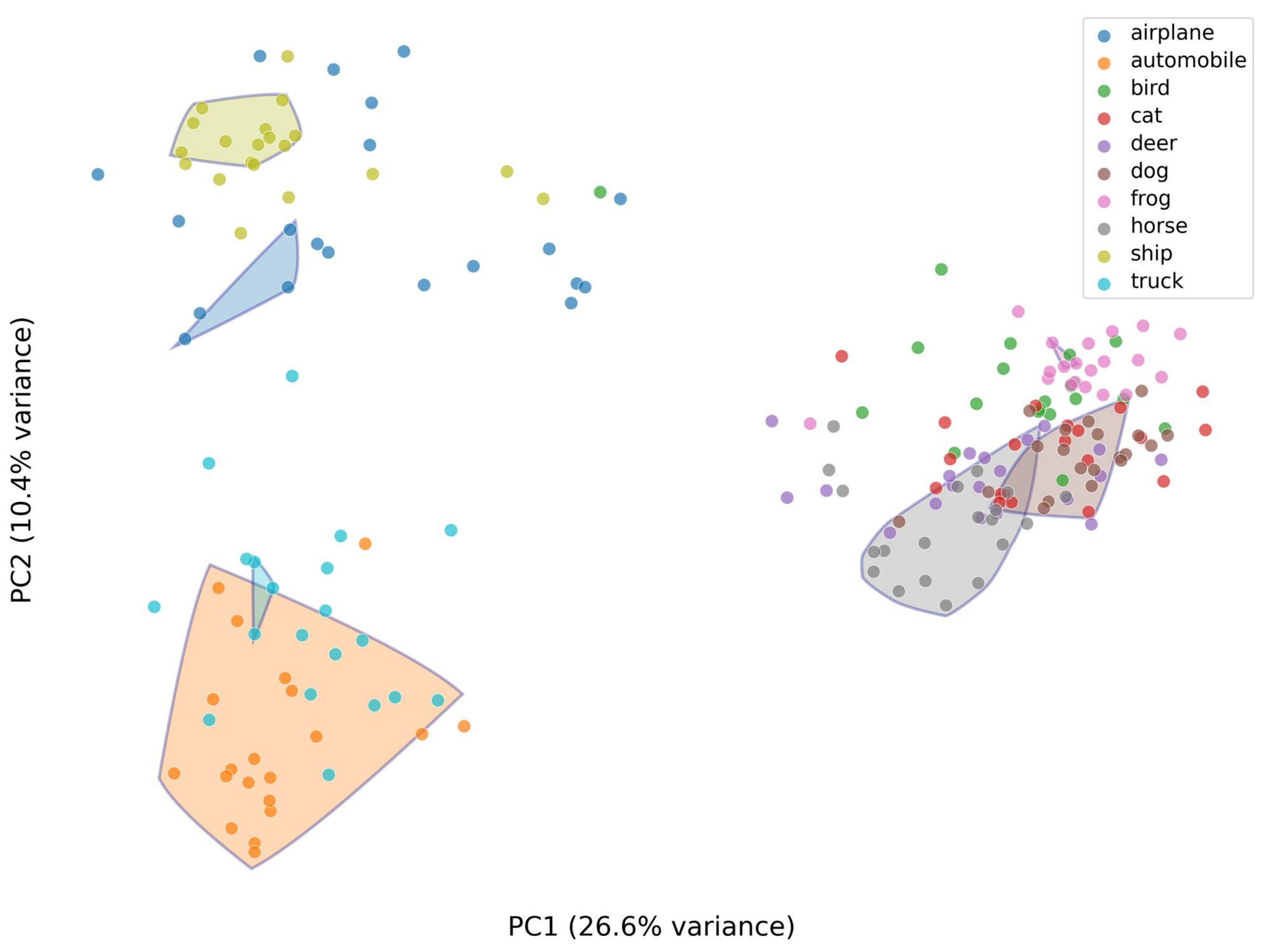}
        \caption{Seed 314 -- \vblob{blob}}
    \end{subfigure}

    \vspace{0.4em}

    \begin{subfigure}{0.48\linewidth}
        \includegraphics[width=\linewidth]{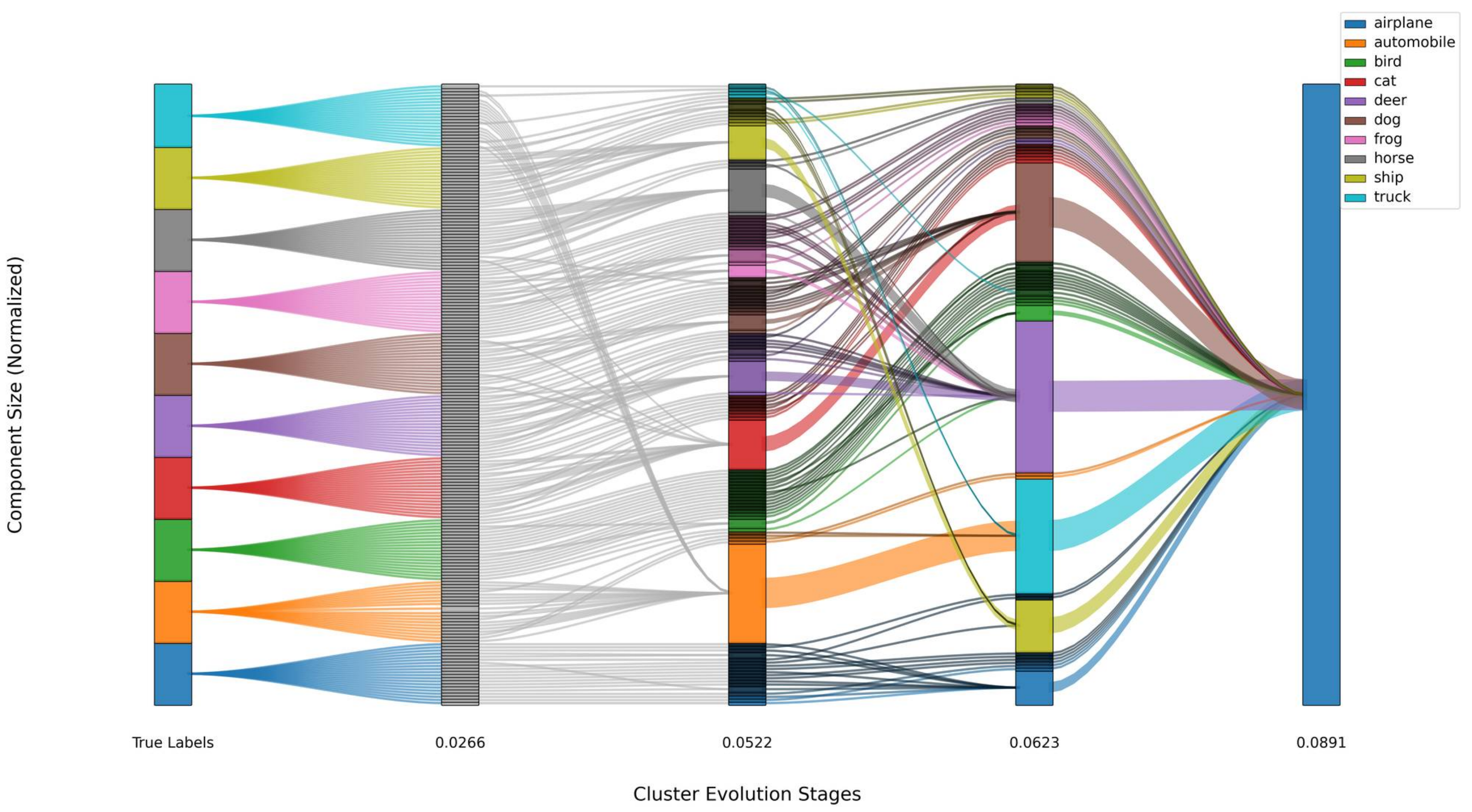}
        \caption{Seed 555 -- \vcflow{cluster flow}}
    \end{subfigure}\hfill
    \begin{subfigure}{0.48\linewidth}
        \includegraphics[width=\linewidth]{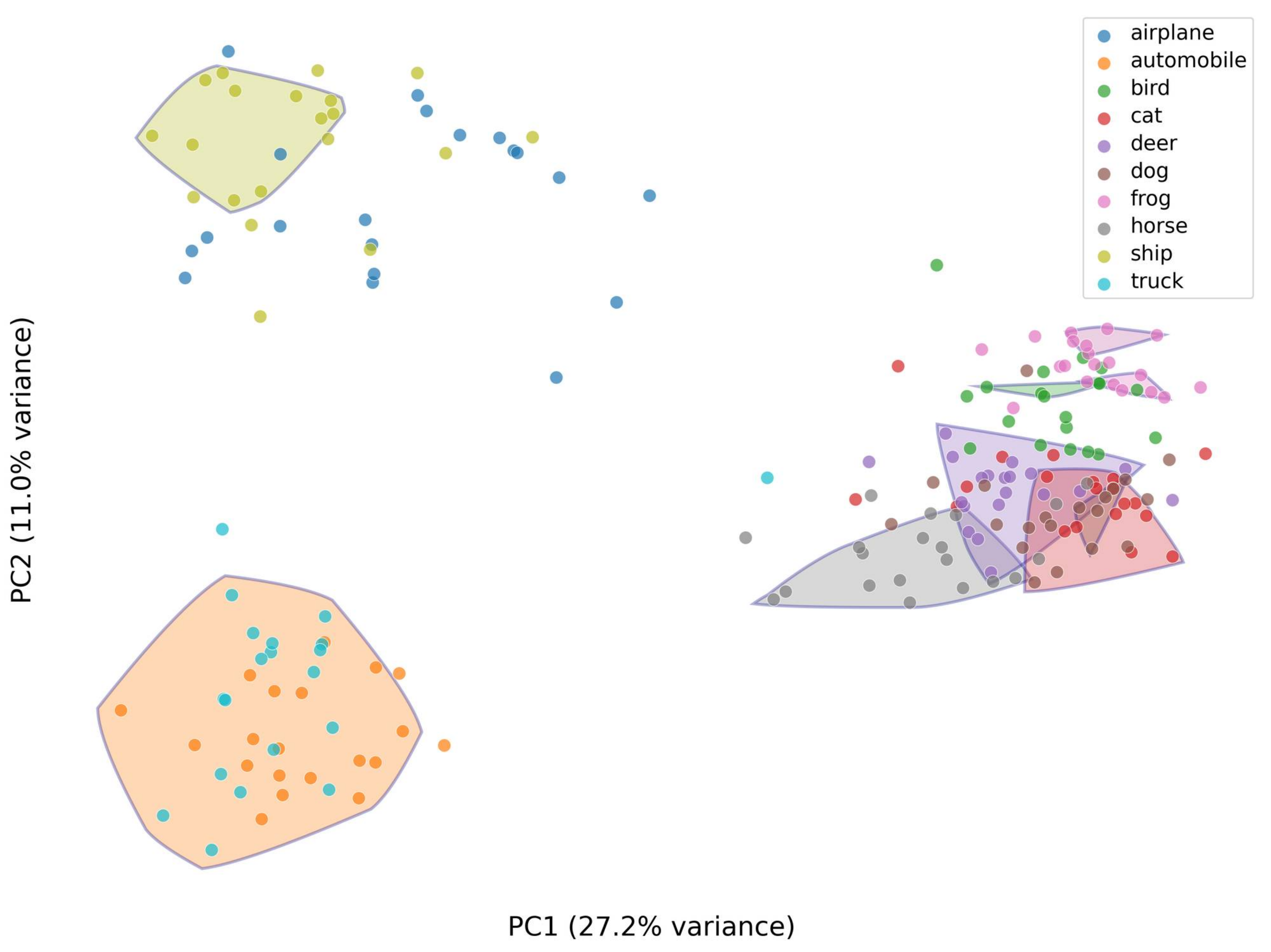}
        \caption{Seed 555 -- \vblob{blob}}
    \end{subfigure}

    \vspace{0.4em}

    \begin{subfigure}{0.48\linewidth}
        \includegraphics[width=\linewidth]{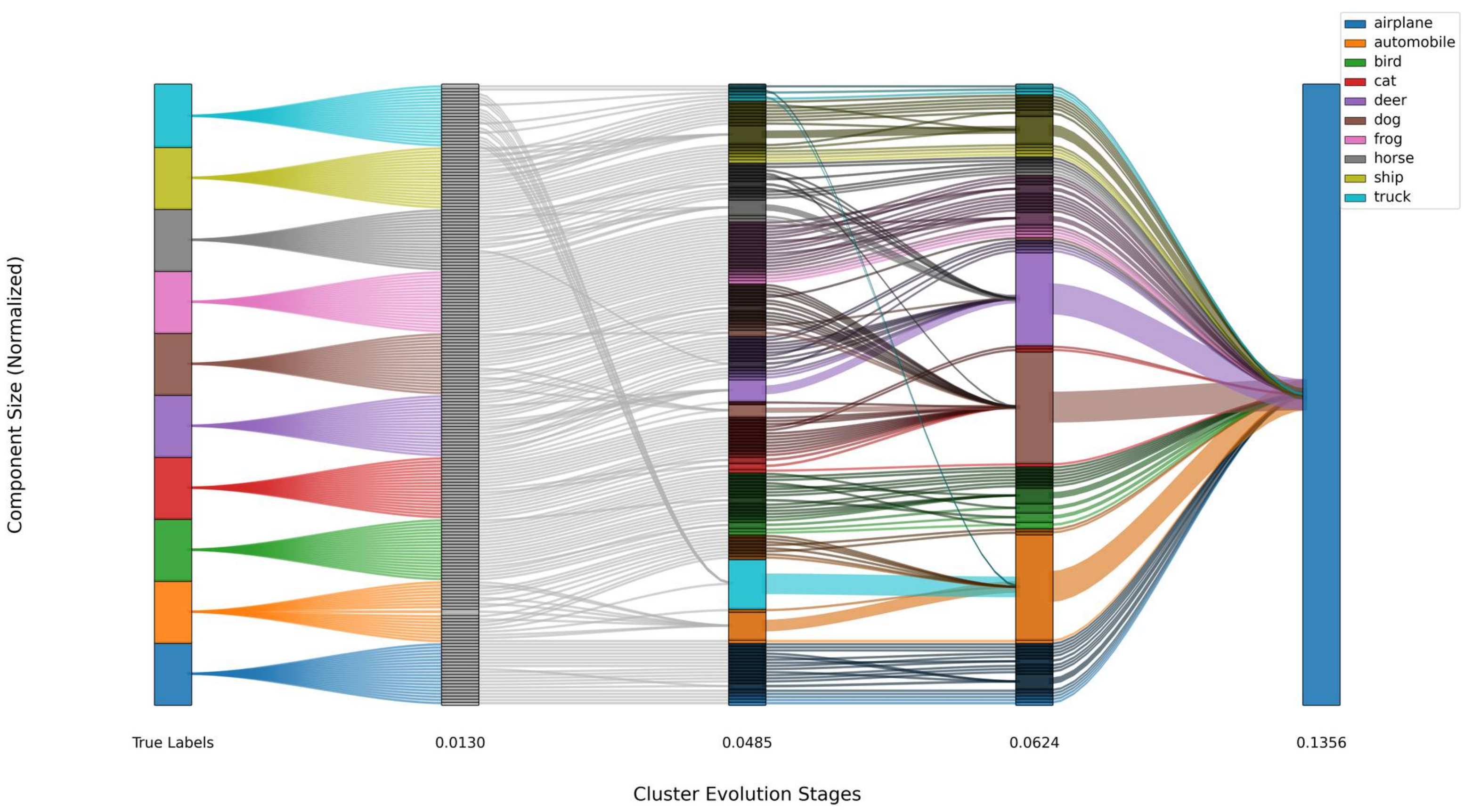}
        \caption{Seed 678 -- \vcflow{cluster flow}}
    \end{subfigure}\hfill
    \begin{subfigure}{0.48\linewidth}
        \includegraphics[width=\linewidth]{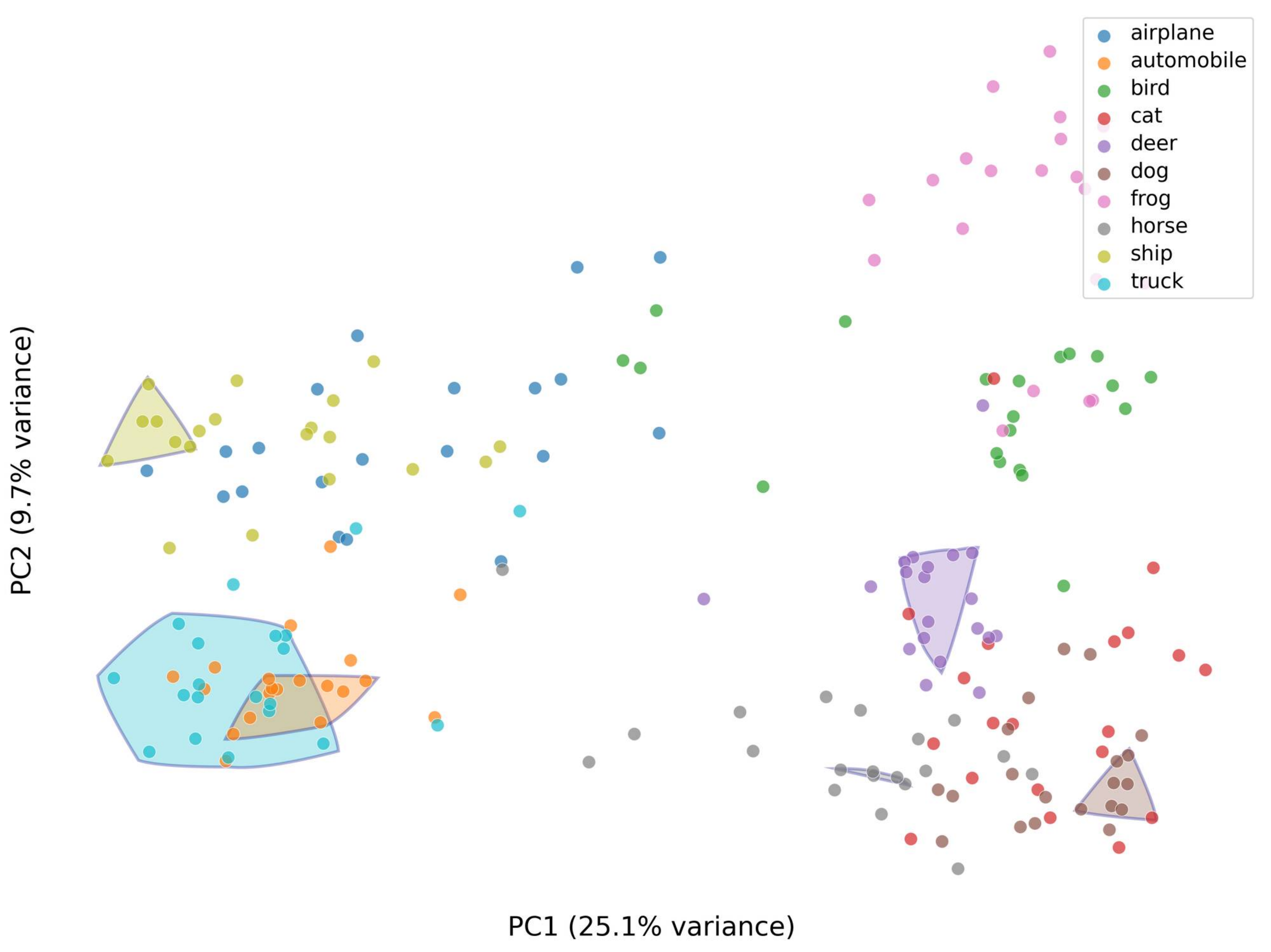}
        \caption{Seed 678 -- \vblob{blob}}
    \end{subfigure}
    \caption{Stability analysis for ViT-B/16 encoder layer~9 (seeds 314--678), continued from \cref{fig:stability_l9}. Continued in \cref{fig:stability_l9_p3}.}
    \label{fig:stability_l9_p2}
\end{figure*}

\begin{figure*}[!htb]
    \centering
    \begin{subfigure}{0.48\linewidth}
        \includegraphics[width=\linewidth]{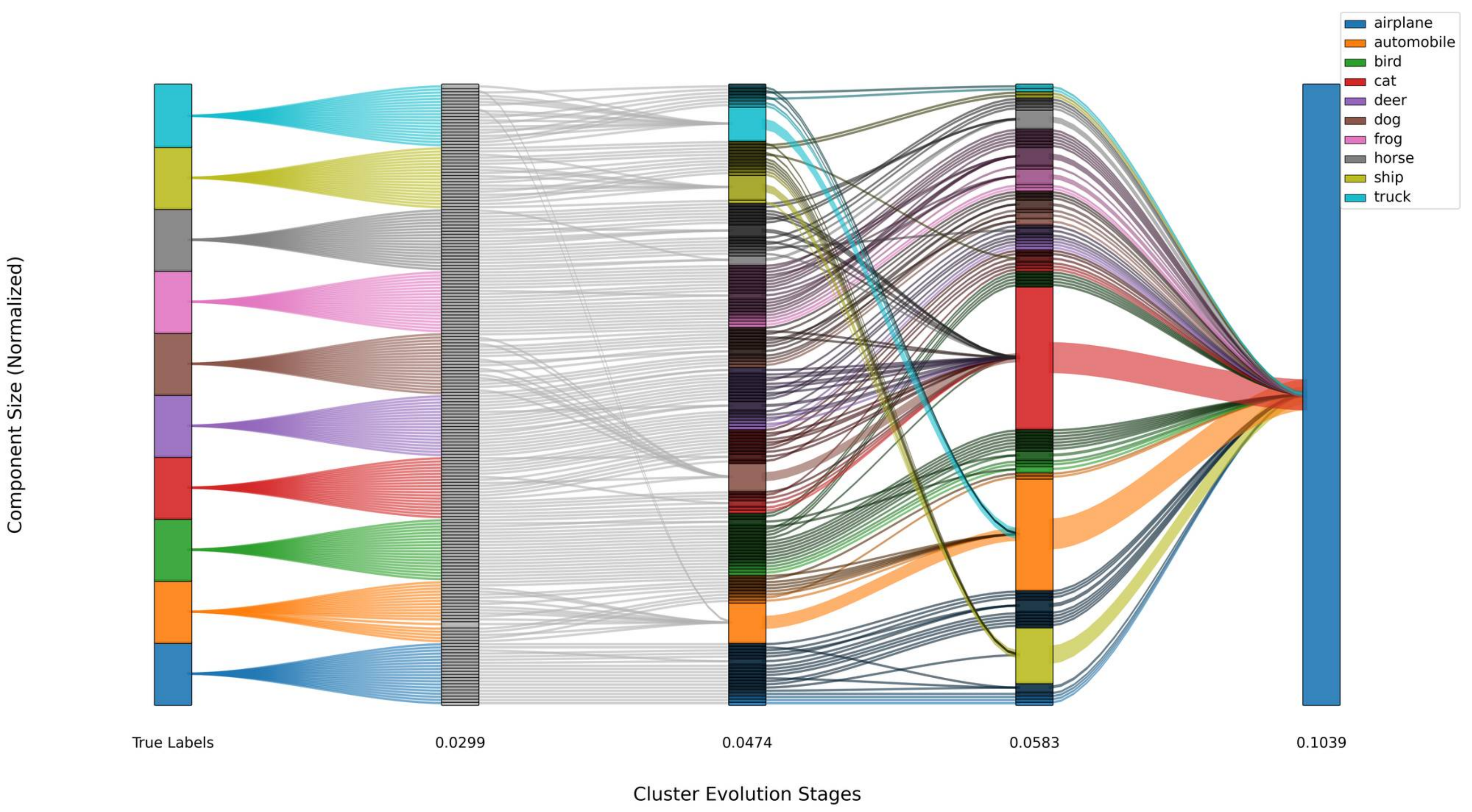}
        \caption{Seed 821 -- \vcflow{cluster flow}}
    \end{subfigure}\hfill
    \begin{subfigure}{0.48\linewidth}
        \includegraphics[width=\linewidth]{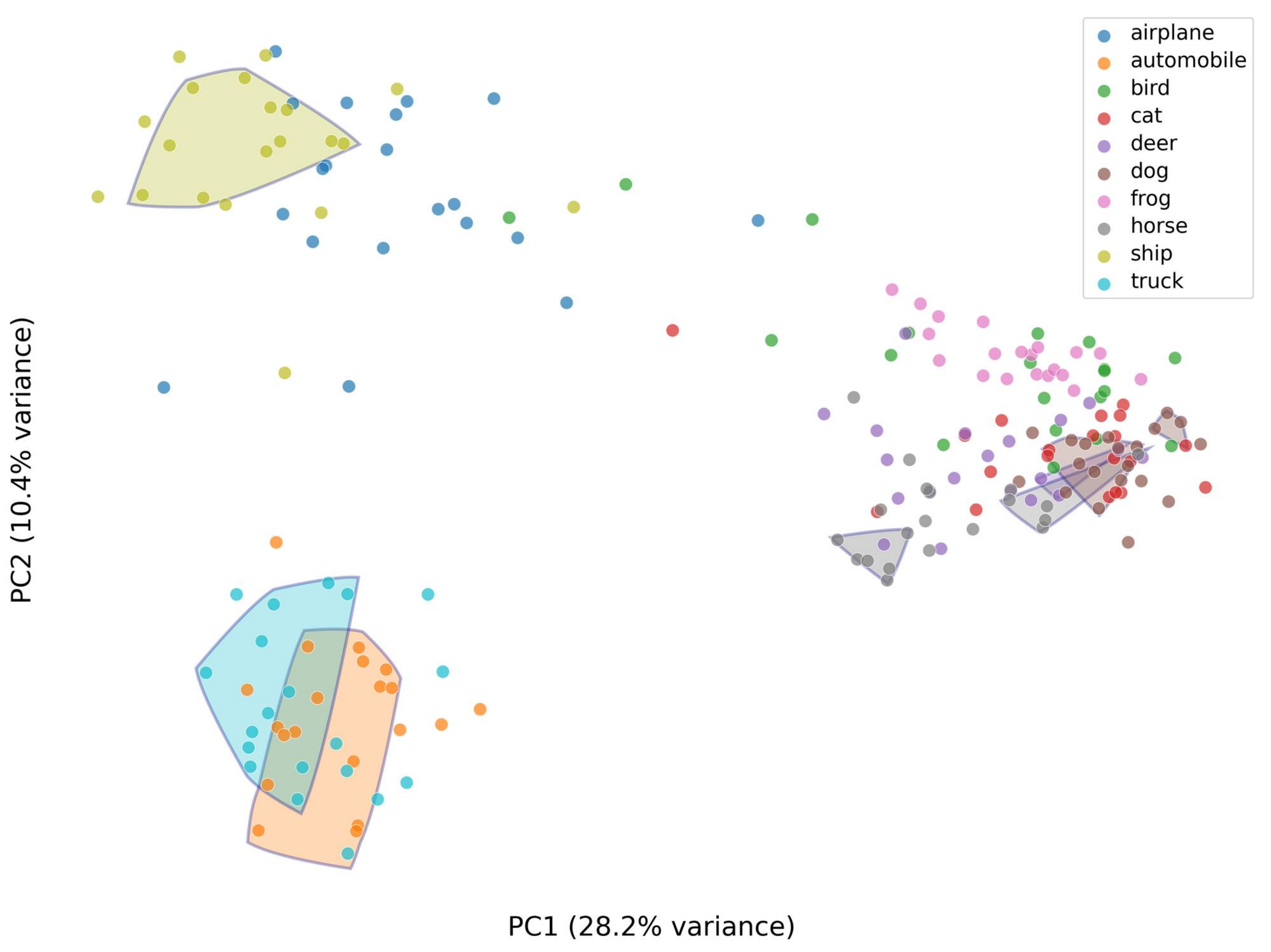}
        \caption{Seed 821 -- \vblob{blob}}
    \end{subfigure}

    \vspace{0.4em}

    \begin{subfigure}{0.48\linewidth}
        \includegraphics[width=\linewidth]{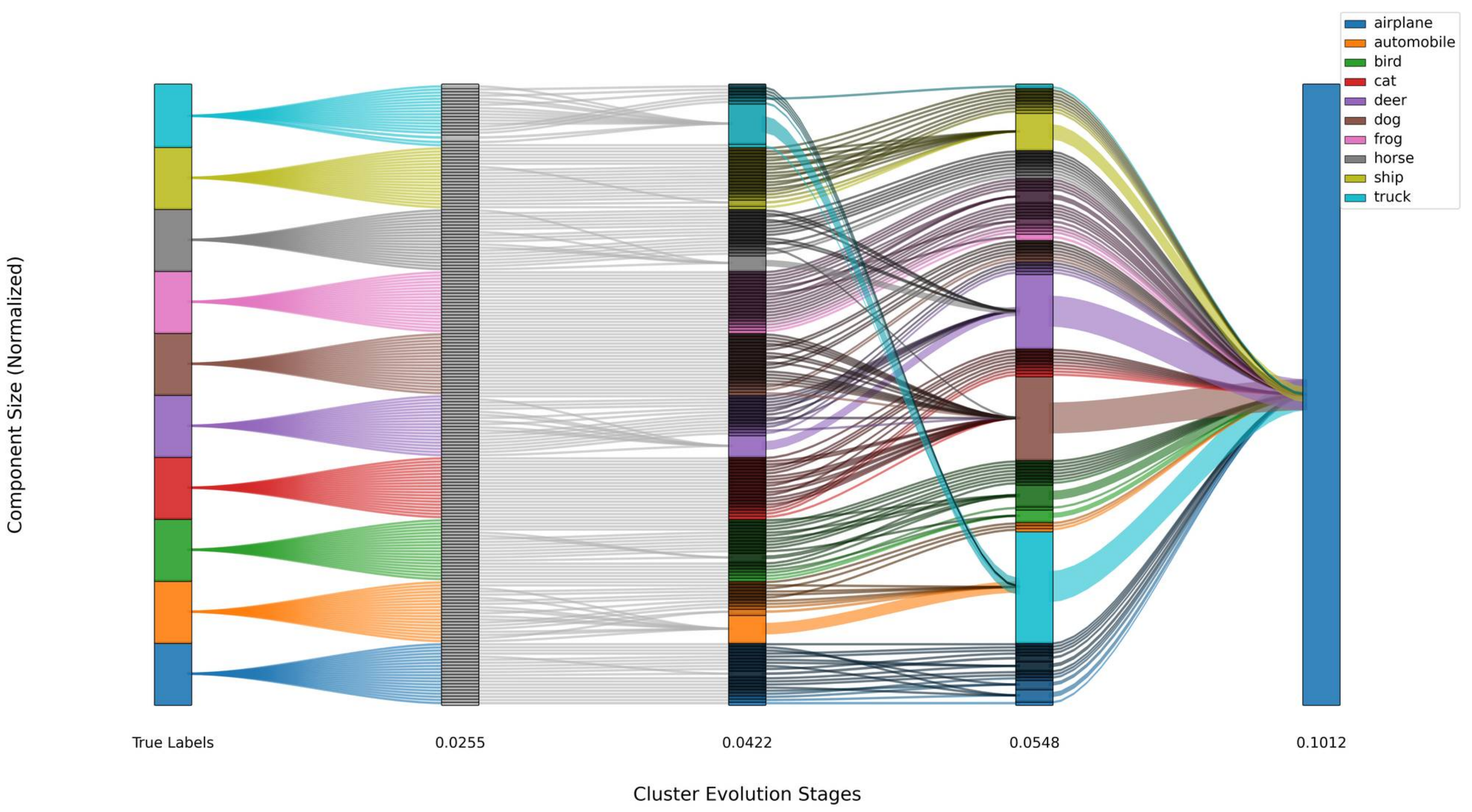}
        \caption{Seed 937 -- \vcflow{cluster flow}}
    \end{subfigure}\hfill
    \begin{subfigure}{0.48\linewidth}
        \includegraphics[width=\linewidth]{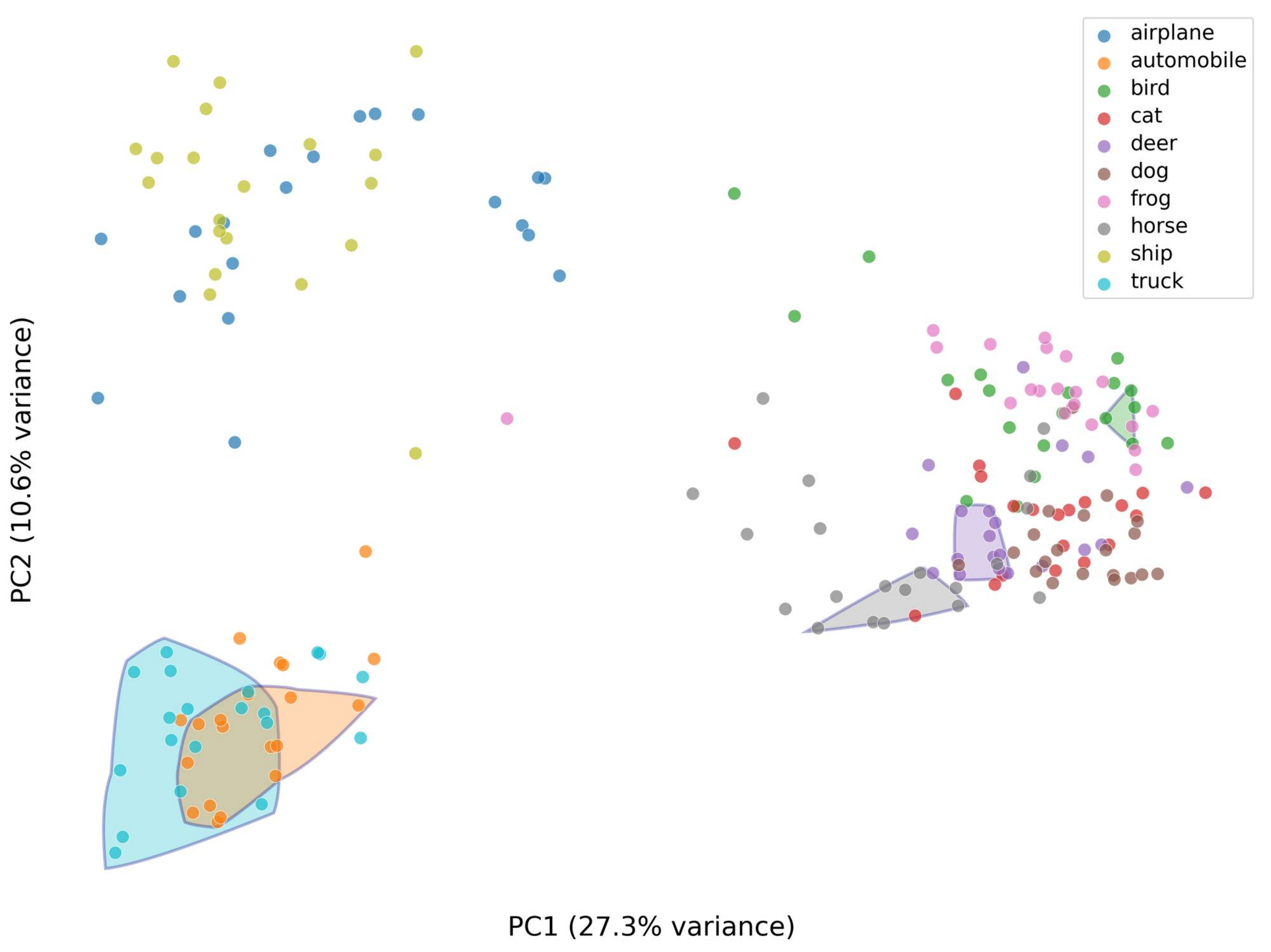}
        \caption{Seed 937 -- \vblob{blob}}
    \end{subfigure}

    \vspace{0.4em}

    \begin{subfigure}{0.48\linewidth}
        \includegraphics[width=\linewidth]{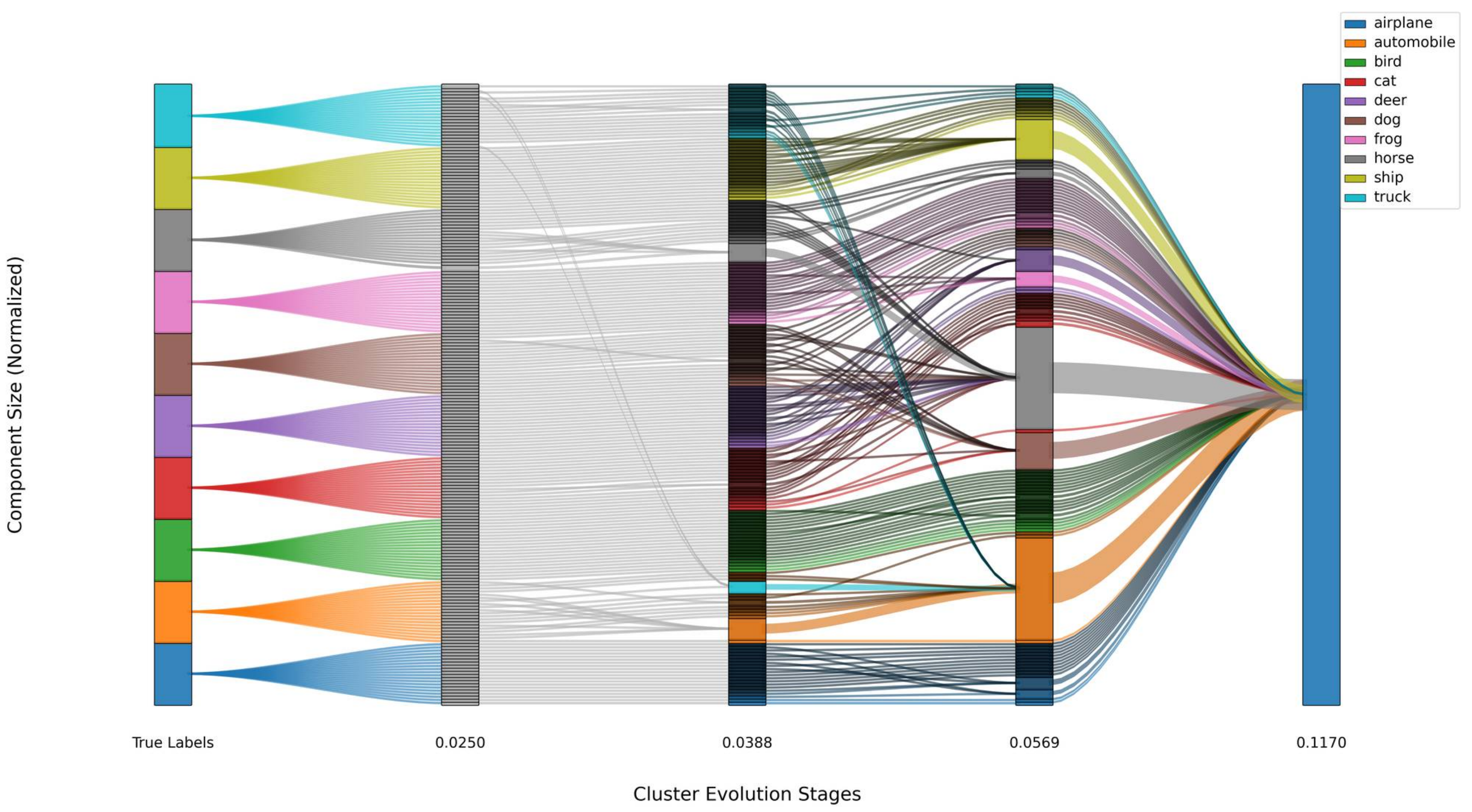}
        \caption{Seed 1001 -- \vcflow{cluster flow}}
    \end{subfigure}\hfill
    \begin{subfigure}{0.48\linewidth}
        \includegraphics[width=\linewidth]{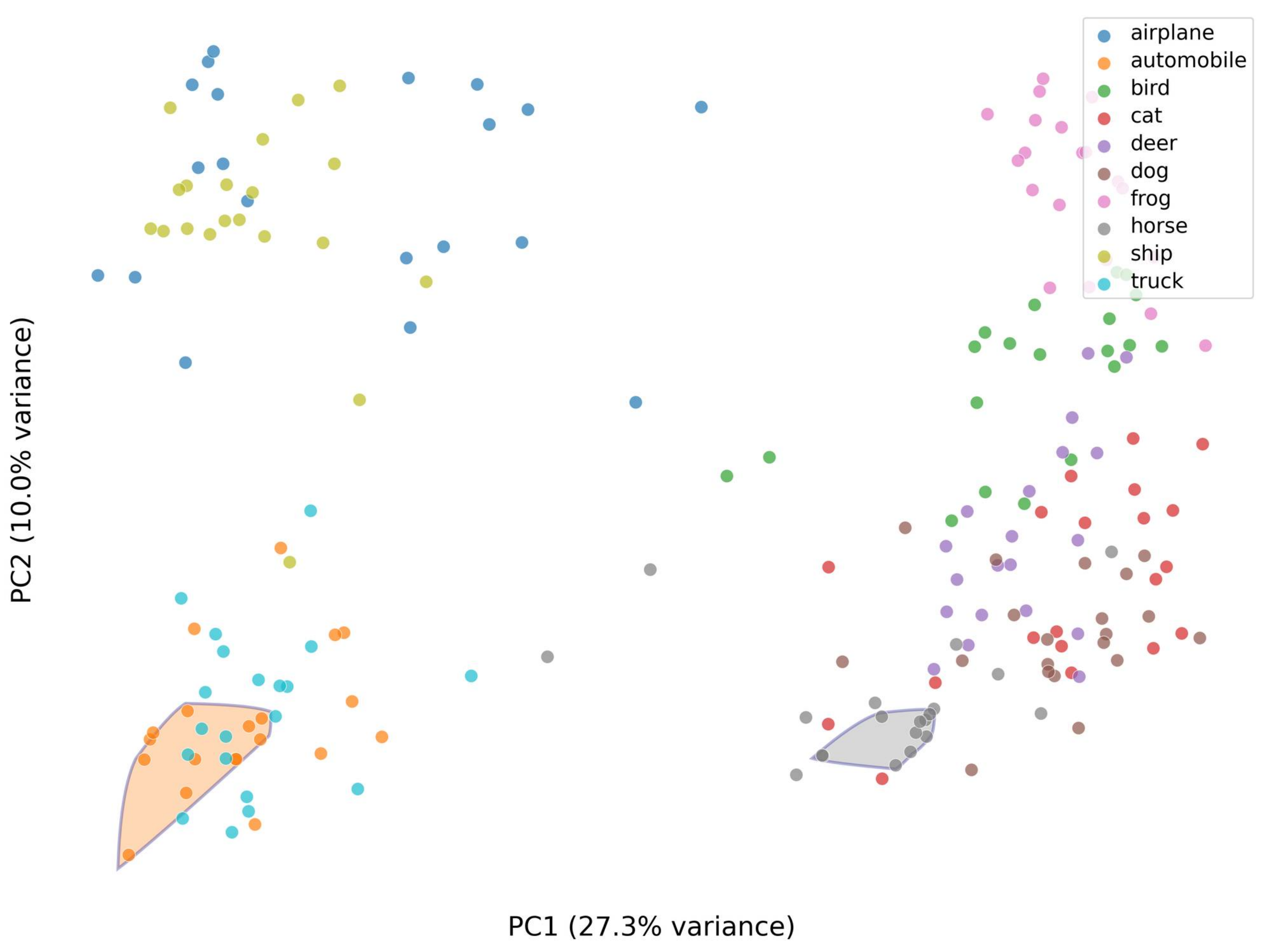}
        \caption{Seed 1001 -- \vblob{blob}}
    \end{subfigure}
    \caption{Stability analysis for ViT-B/16 encoder layer~9 (seeds 821--1001), continued from \cref{fig:stability_l9_p2}. Continued in \cref{fig:stability_l9_p4}.}
    \label{fig:stability_l9_p3}
\end{figure*}

\begin{figure*}[!htb]
    \centering
    \begin{subfigure}{0.48\linewidth}
        \includegraphics[width=\linewidth]{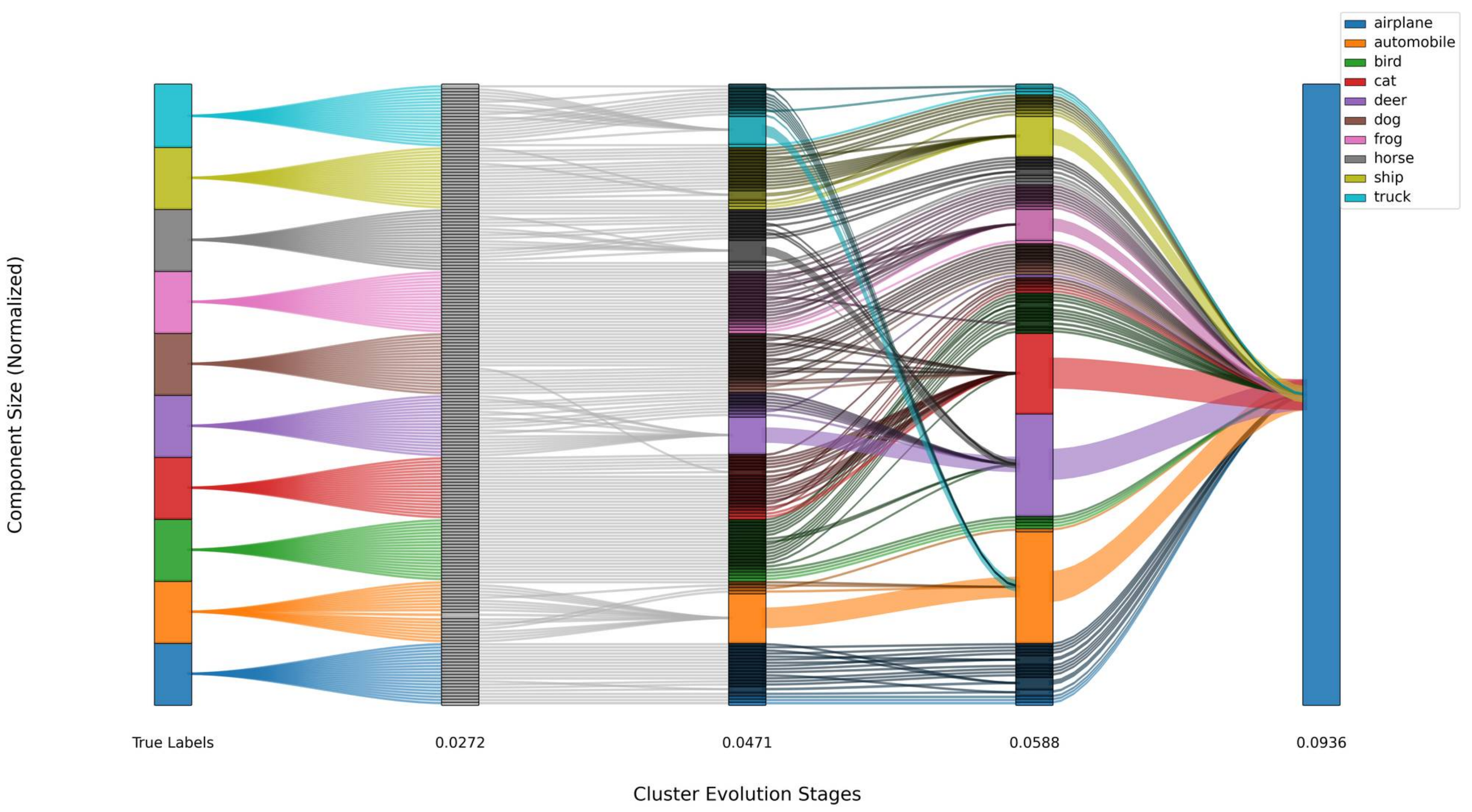}
        \caption{Seed 1234 -- \vcflow{cluster flow}}
    \end{subfigure}\hfill
    \begin{subfigure}{0.48\linewidth}
        \includegraphics[width=\linewidth]{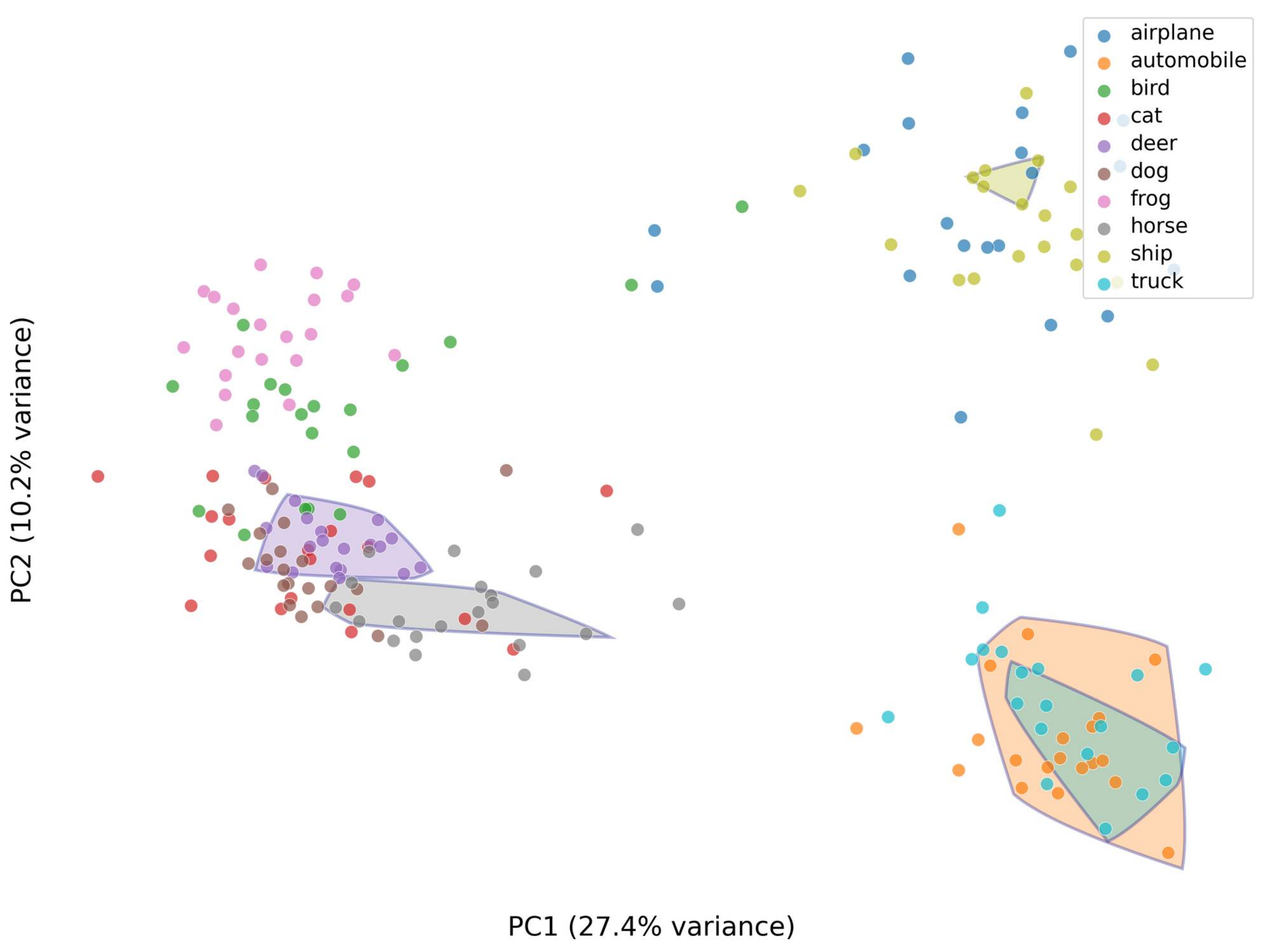}
        \caption{Seed 1234 -- \vblob{blob}}
    \end{subfigure}
    \caption{Stability analysis for ViT-B/16 encoder layer~9 (seed 1234), continued from \cref{fig:stability_l9_p3}. Weak class separation persists across all seeds, confirming this is a model property not a sampling artifact.}
    \label{fig:stability_l9_p4}
    \label{fig:stability_l9_part2}
\end{figure*}

\begin{figure*}[!htb]
    \centering
    \begin{subfigure}{0.48\linewidth}
        \includegraphics[width=\linewidth]{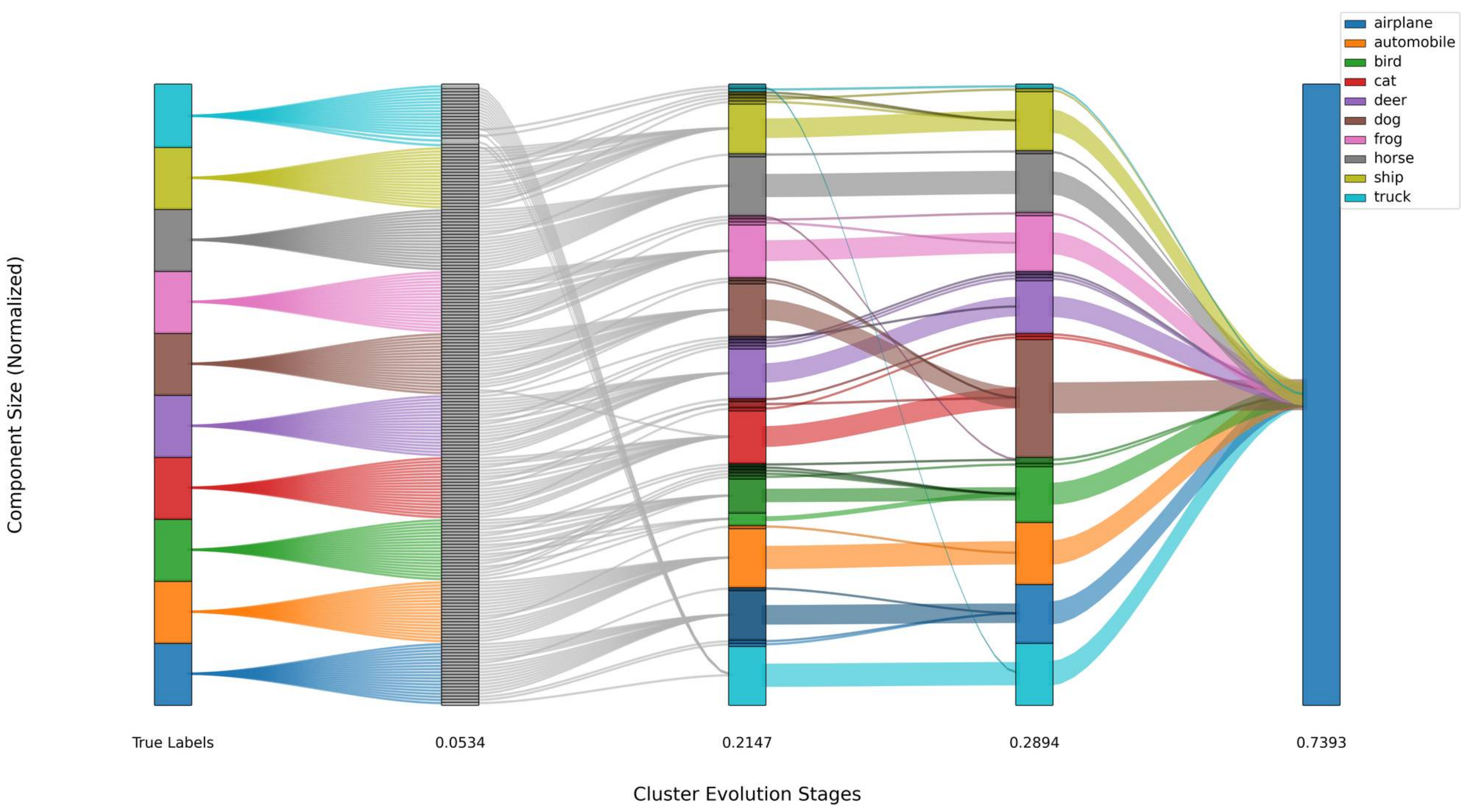}
        \caption{Seed 42 -- \vcflow{cluster flow}}
    \end{subfigure}\hfill
    \begin{subfigure}{0.48\linewidth}
        \includegraphics[width=\linewidth]{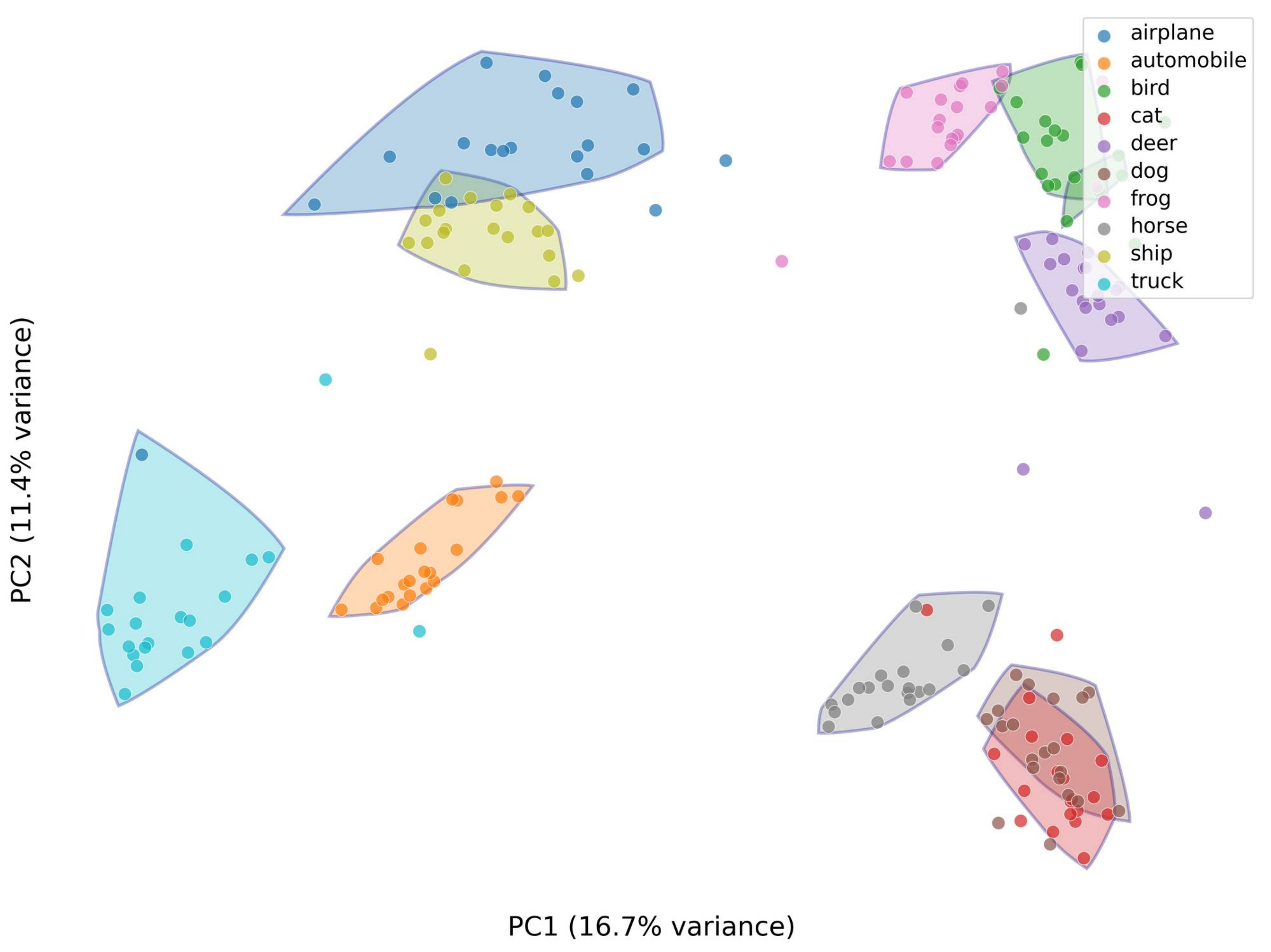}
        \caption{Seed 42 -- \vblob{blob}}
    \end{subfigure}

    \vspace{0.4em}

    \begin{subfigure}{0.48\linewidth}
        \includegraphics[width=\linewidth]{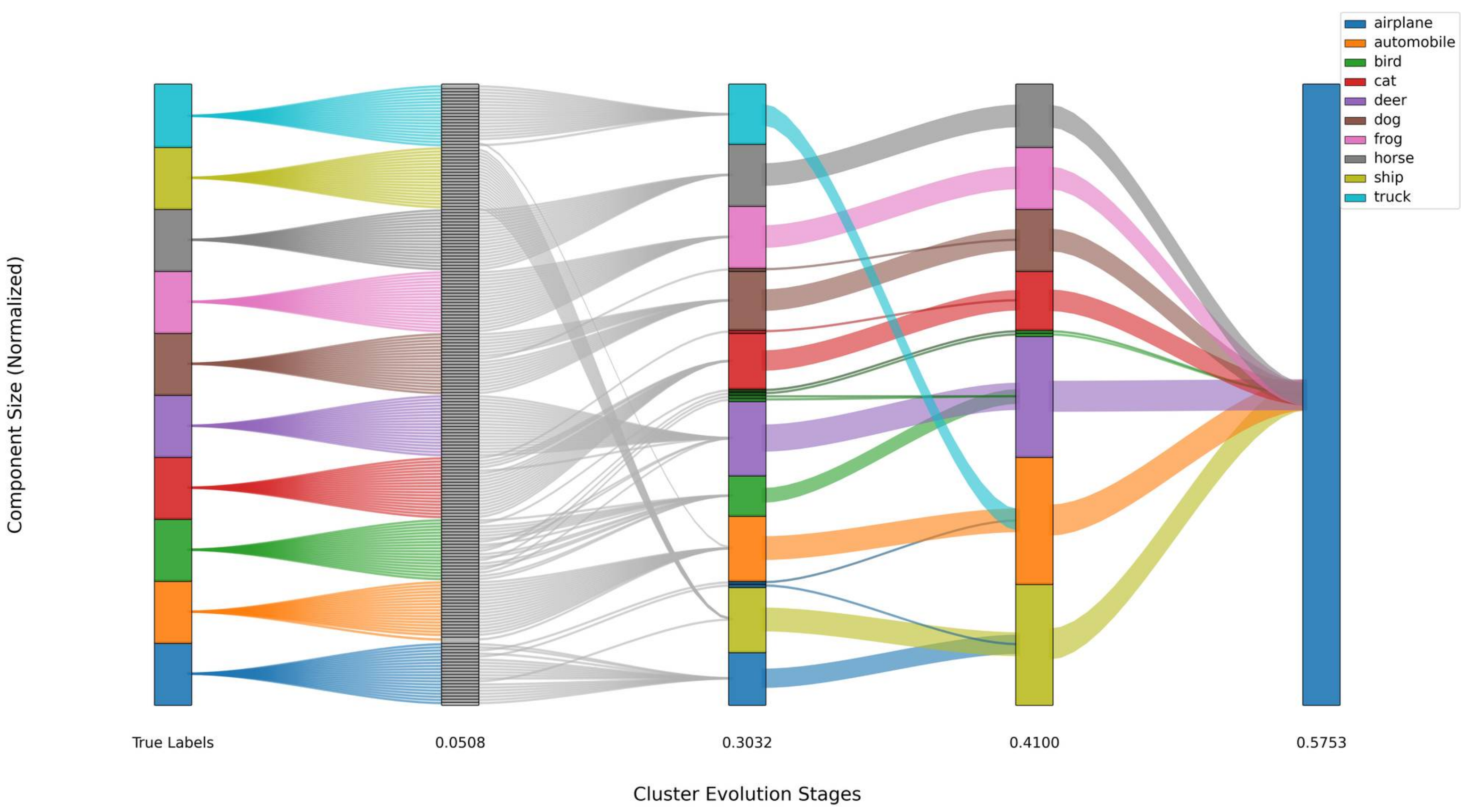}
        \caption{Seed 123 -- \vcflow{cluster flow}}
    \end{subfigure}\hfill
    \begin{subfigure}{0.48\linewidth}
        \includegraphics[width=\linewidth]{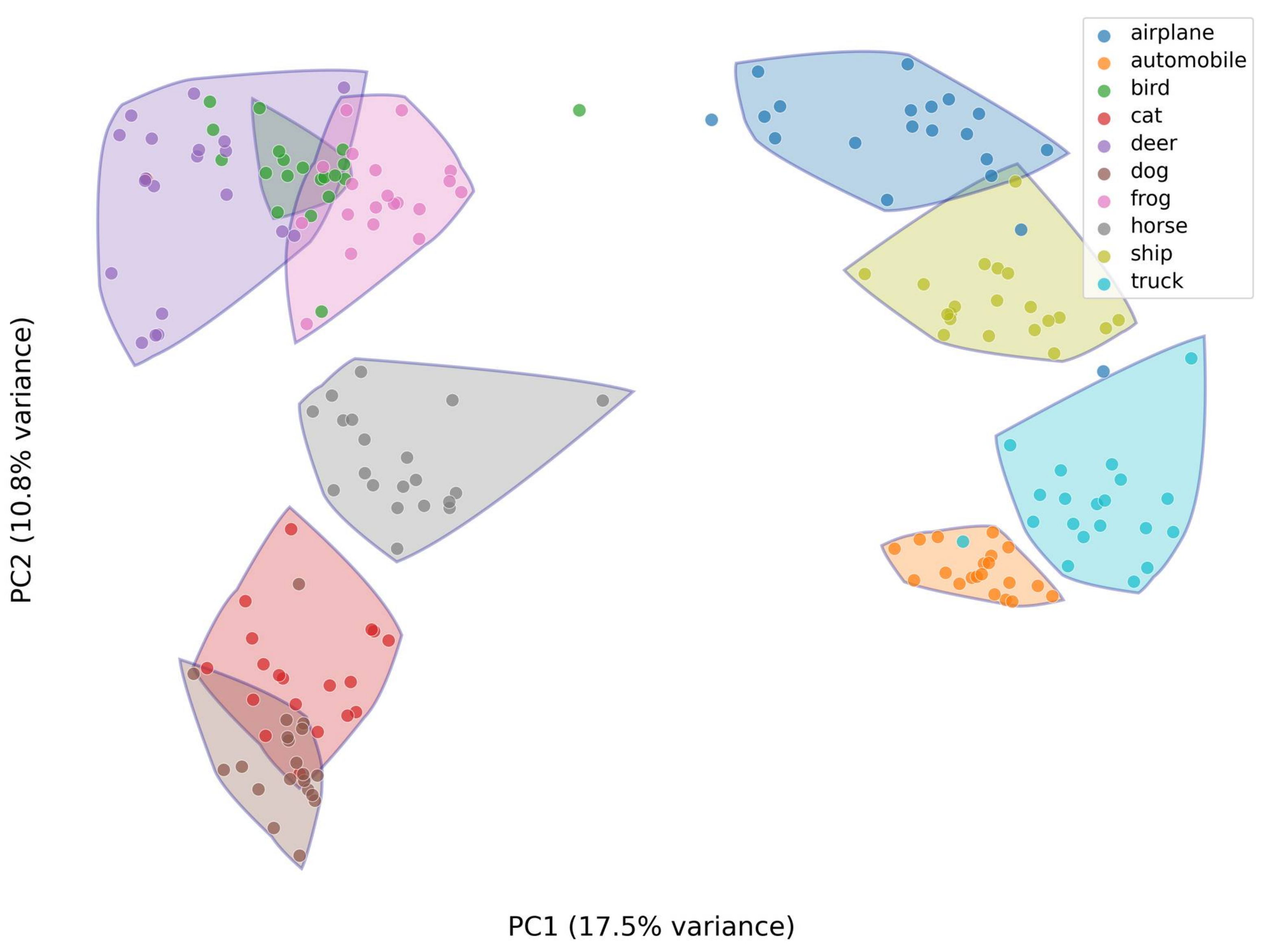}
        \caption{Seed 123 -- \vblob{blob}}
    \end{subfigure}

    \vspace{0.4em}

    \begin{subfigure}{0.48\linewidth}
        \includegraphics[width=\linewidth]{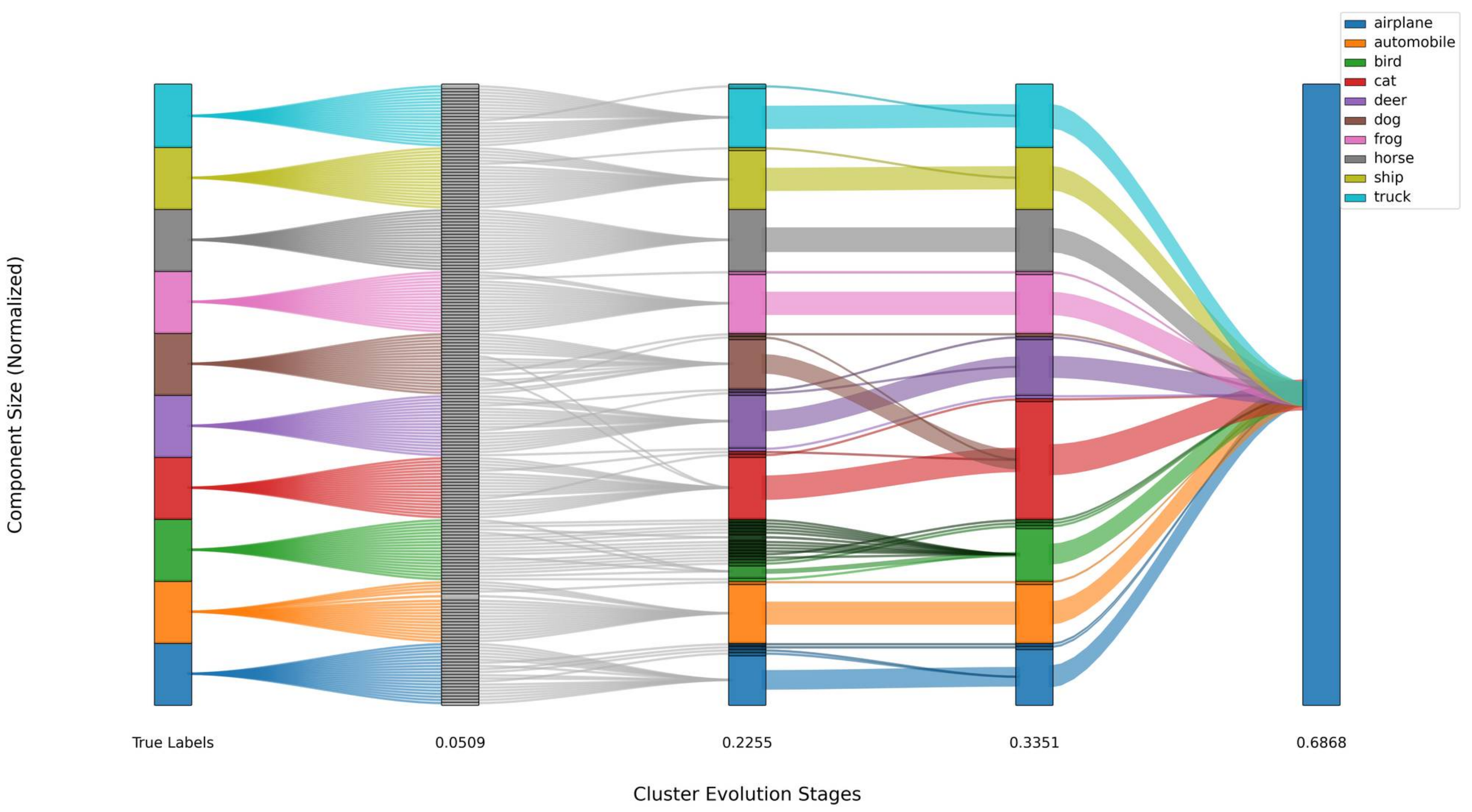}
        \caption{Seed 256 -- \vcflow{cluster flow}}
    \end{subfigure}\hfill
    \begin{subfigure}{0.48\linewidth}
        \includegraphics[width=\linewidth]{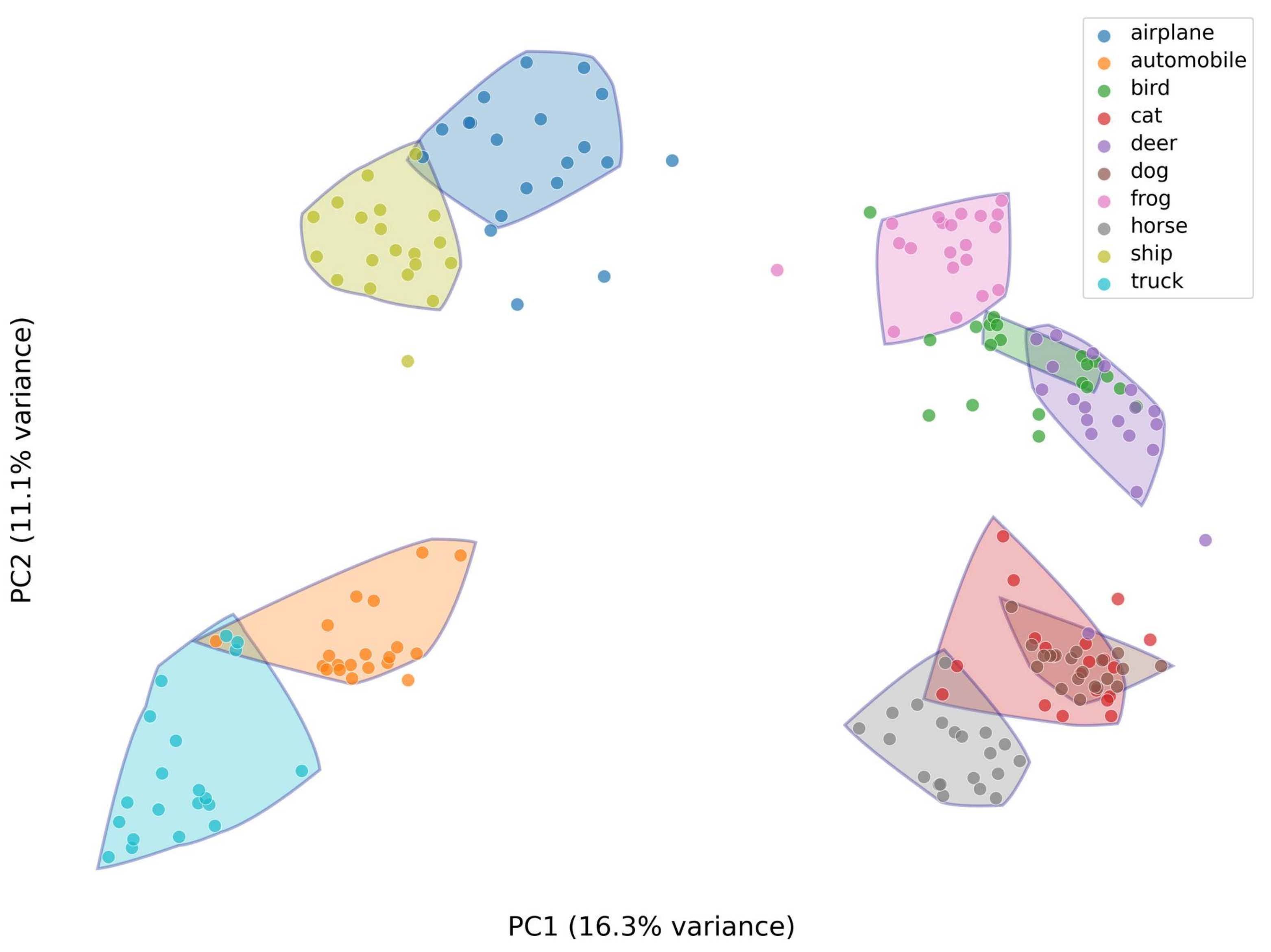}
        \caption{Seed 256 -- \vblob{blob}}
    \end{subfigure}
    \caption{Stability analysis for ViT-B/16 encoder layer~11 (seeds 42--256). Each row: \vcflow{cluster flow} (left) and \vblob{blob graph} (right). Coherent per-class flows and compact clusters persist across all seeds. Continued in \cref{fig:stability_l11_p2}.}
    \label{fig:stability_l11}
    \label{fig:stability_l11_sankeys}
\end{figure*}

\begin{figure*}[!htb]
    \centering
    \begin{subfigure}{0.48\linewidth}
        \includegraphics[width=\linewidth]{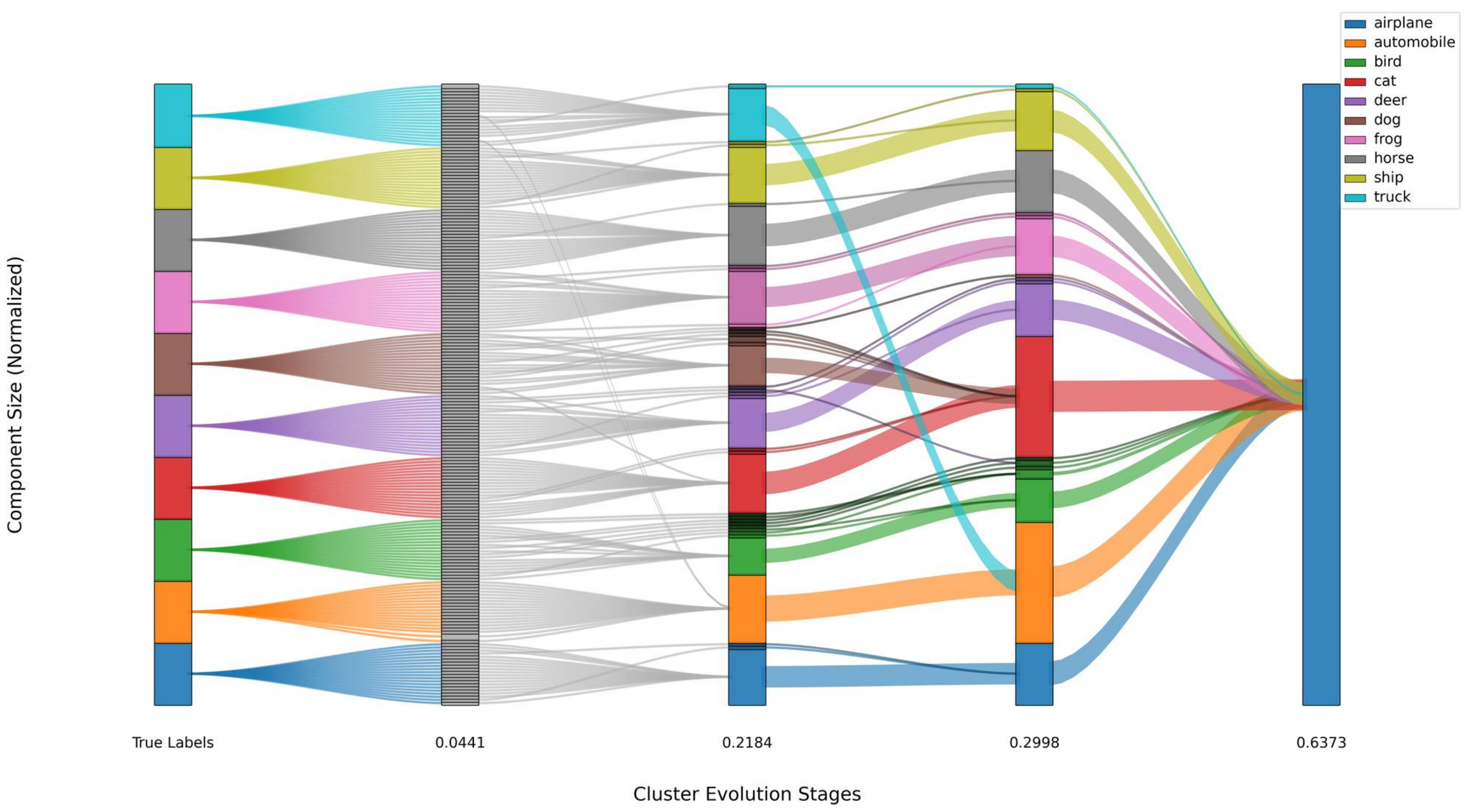}
        \caption{Seed 314 -- \vcflow{cluster flow}}
    \end{subfigure}\hfill
    \begin{subfigure}{0.48\linewidth}
        \includegraphics[width=\linewidth]{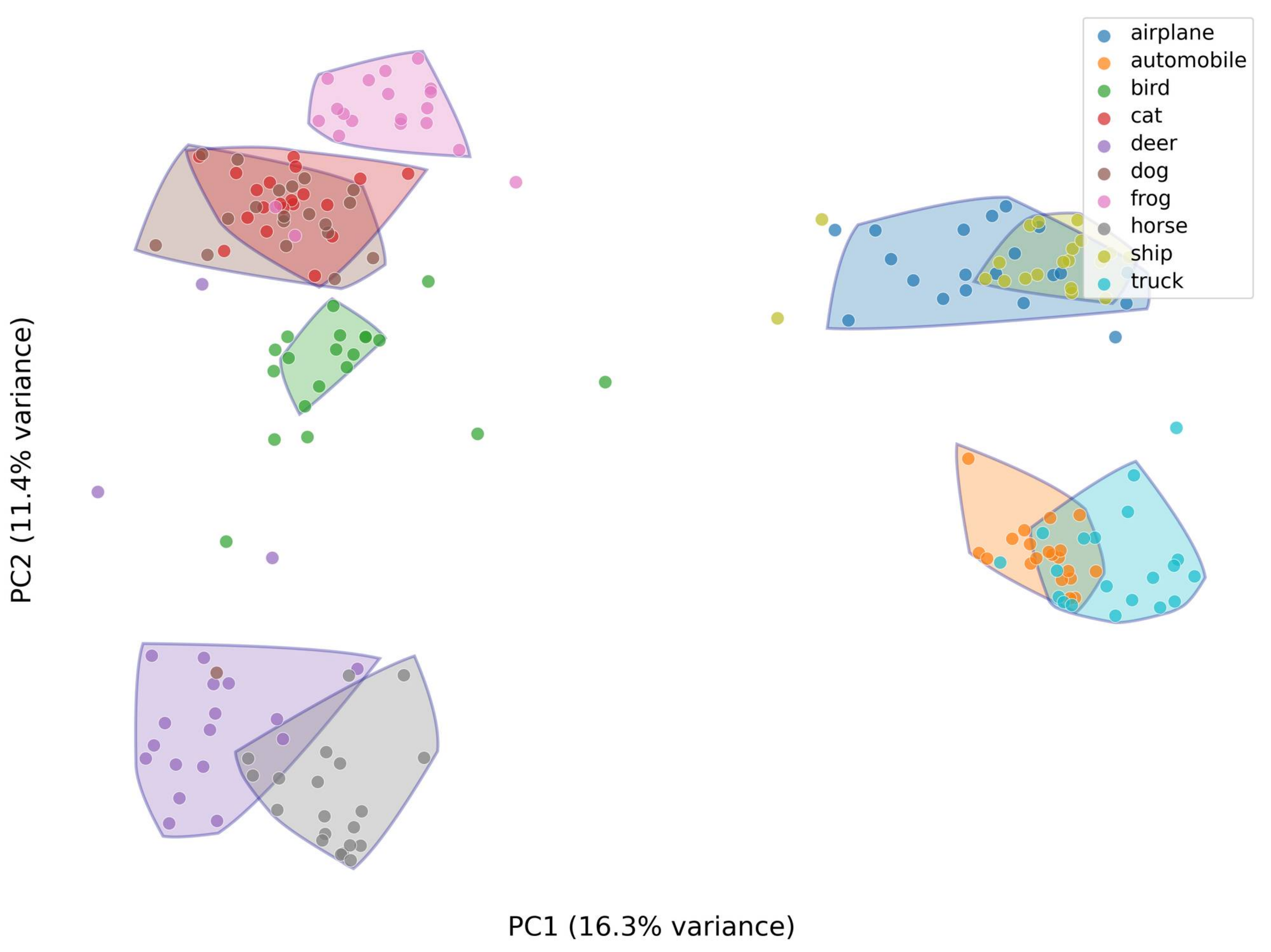}
        \caption{Seed 314 -- \vblob{blob}}
    \end{subfigure}

    \vspace{0.4em}

    \begin{subfigure}{0.48\linewidth}
        \includegraphics[width=\linewidth]{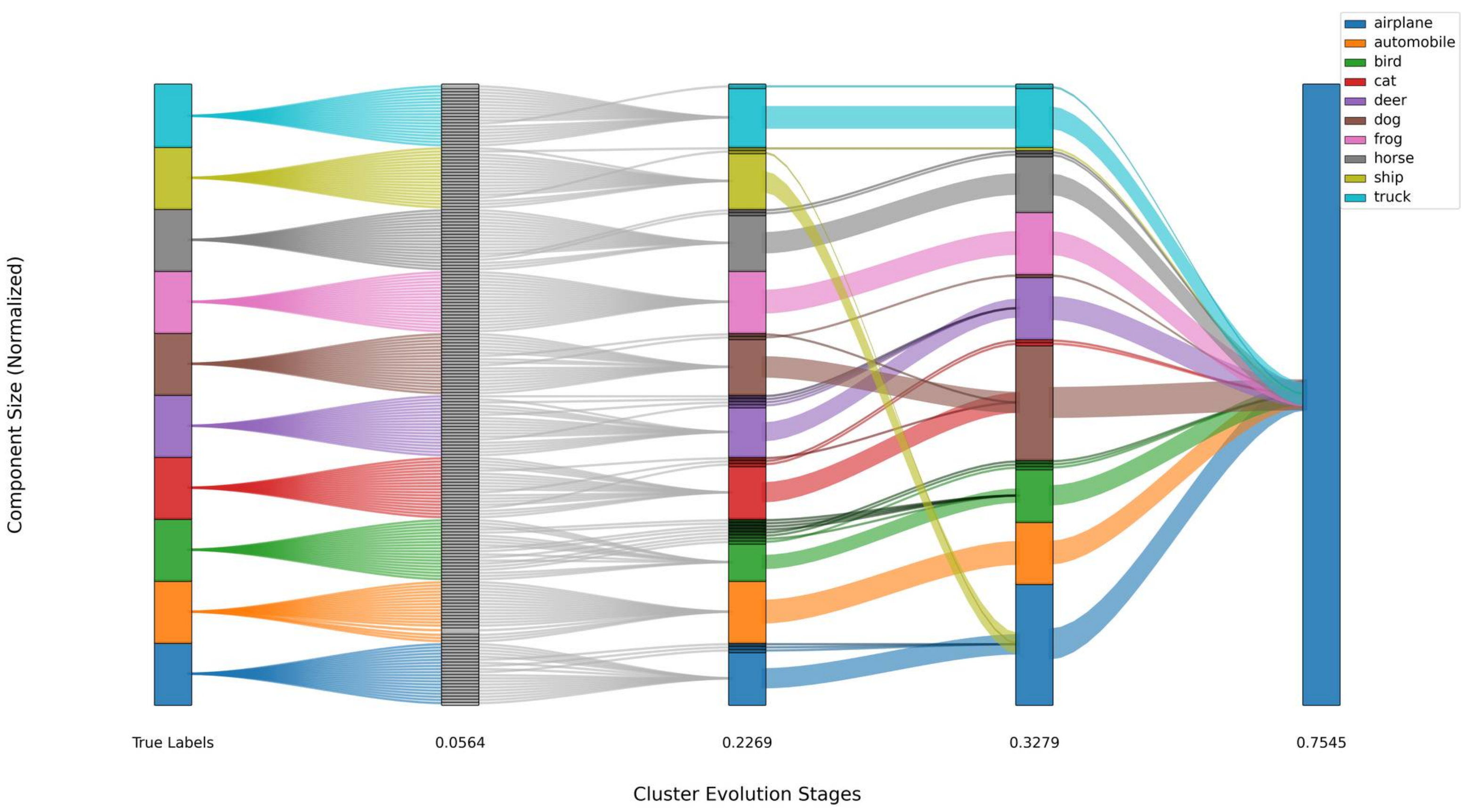}
        \caption{Seed 555 -- \vcflow{cluster flow}}
    \end{subfigure}\hfill
    \begin{subfigure}{0.48\linewidth}
        \includegraphics[width=\linewidth]{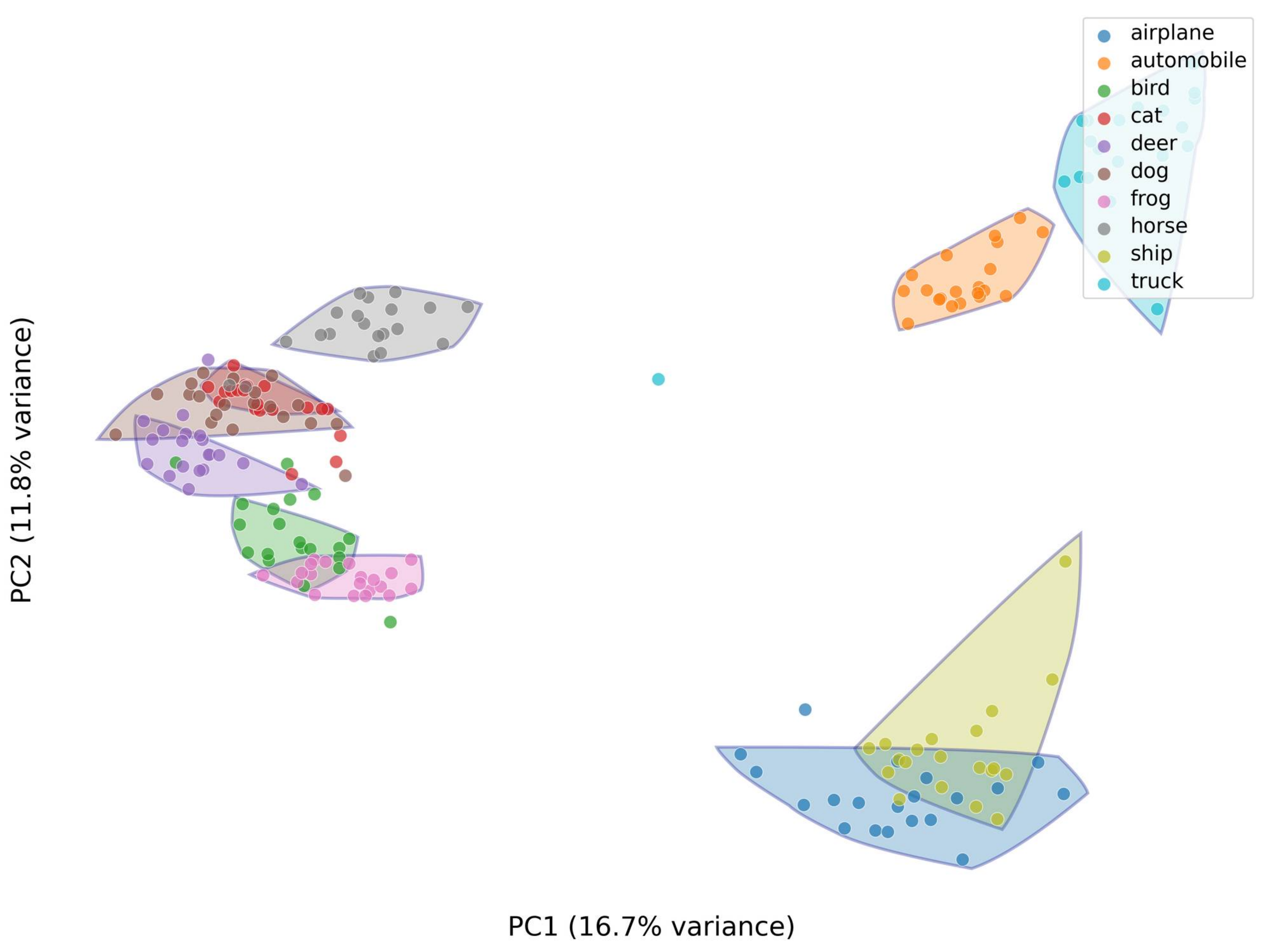}
        \caption{Seed 555 -- \vblob{blob}}
    \end{subfigure}

    \vspace{0.4em}

    \begin{subfigure}{0.48\linewidth}
        \includegraphics[width=\linewidth]{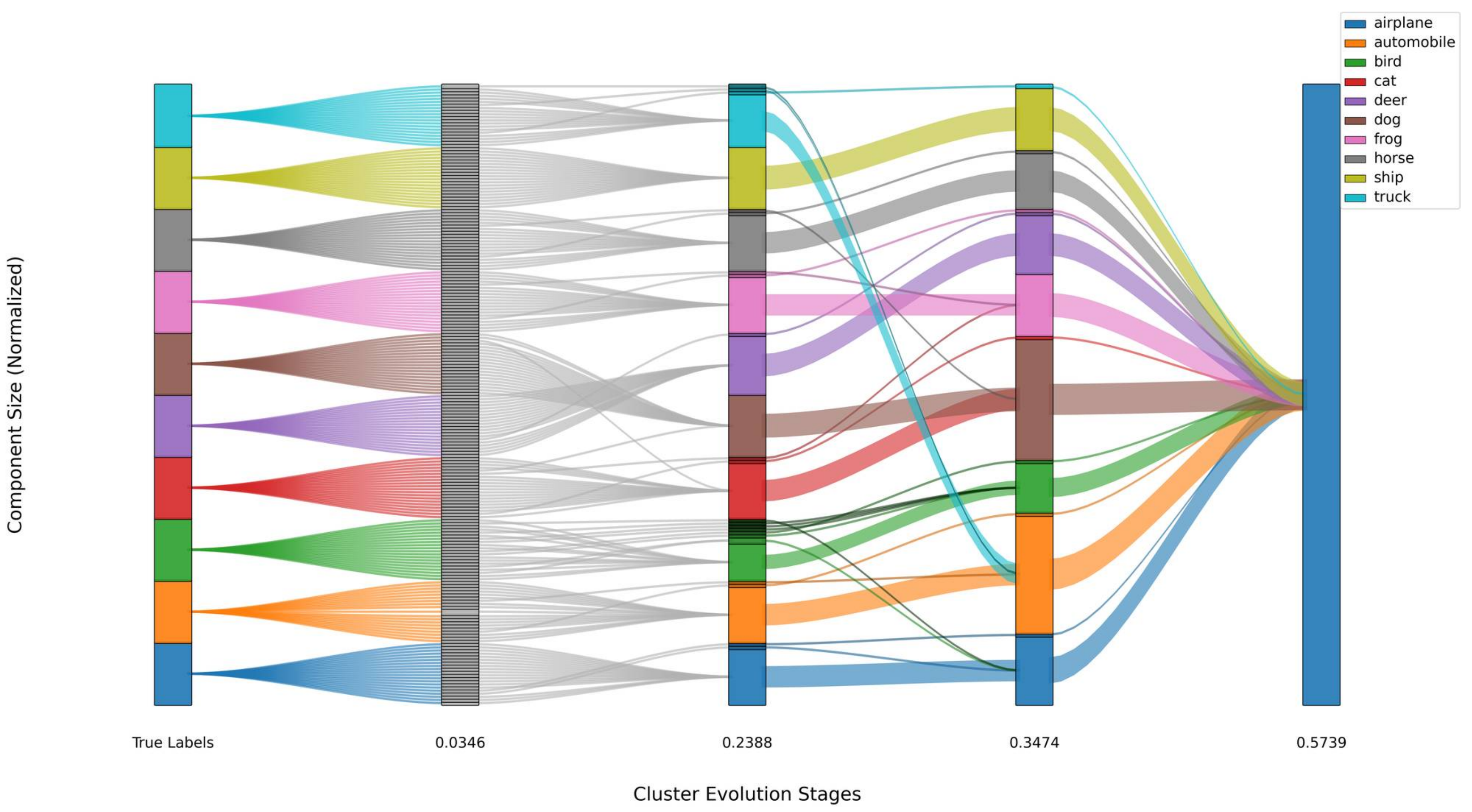}
        \caption{Seed 678 -- \vcflow{cluster flow}}
    \end{subfigure}\hfill
    \begin{subfigure}{0.48\linewidth}
        \includegraphics[width=\linewidth]{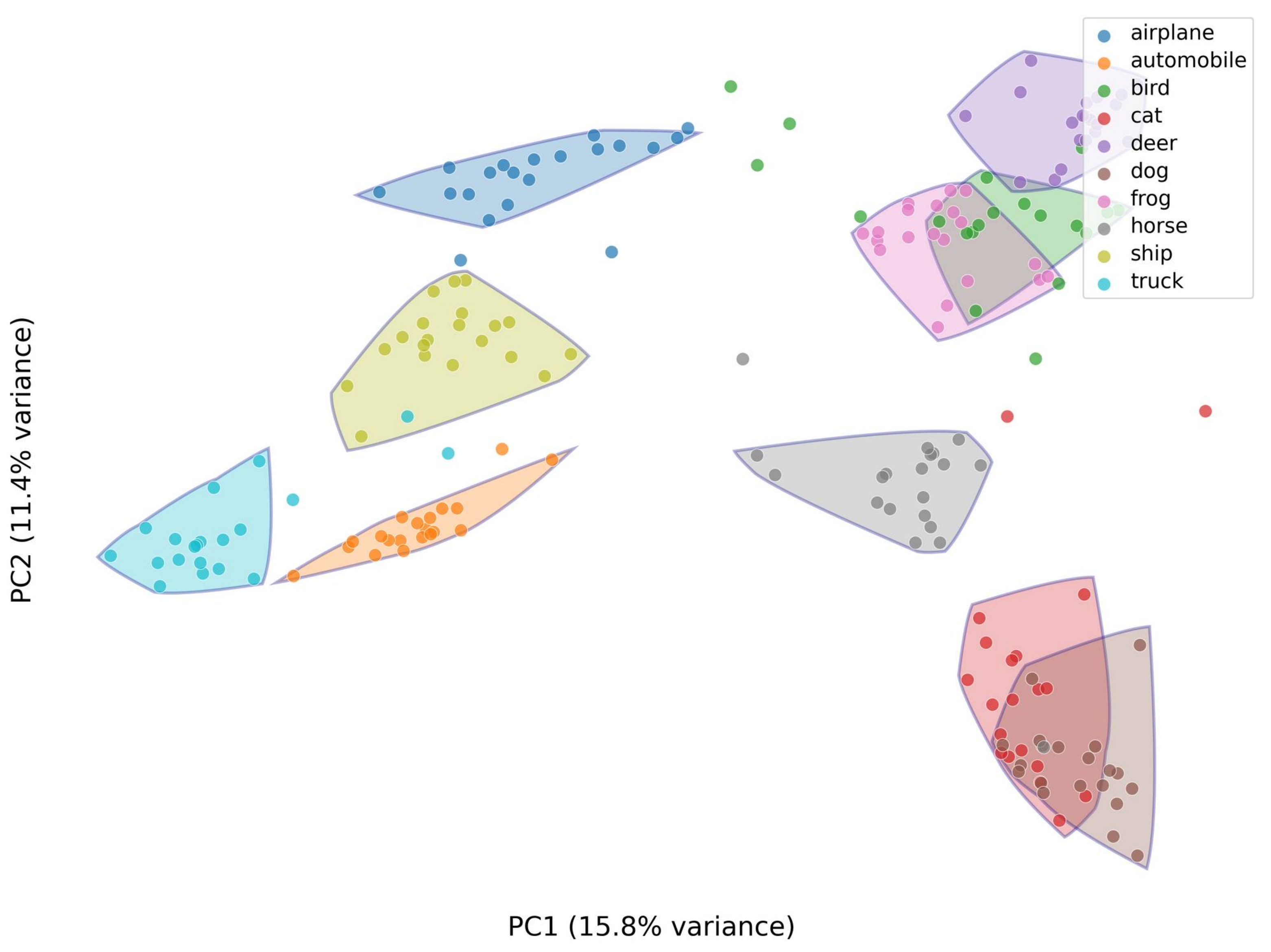}
        \caption{Seed 678 -- \vblob{blob}}
    \end{subfigure}
    \caption{Stability analysis for ViT-B/16 encoder layer~11 (seeds 314--678), continued from \cref{fig:stability_l11}. Continued in \cref{fig:stability_l11_p3}.}
    \label{fig:stability_l11_p2}
\end{figure*}

\begin{figure*}[!htb]
    \centering
    \begin{subfigure}{0.48\linewidth}
        \includegraphics[width=\linewidth]{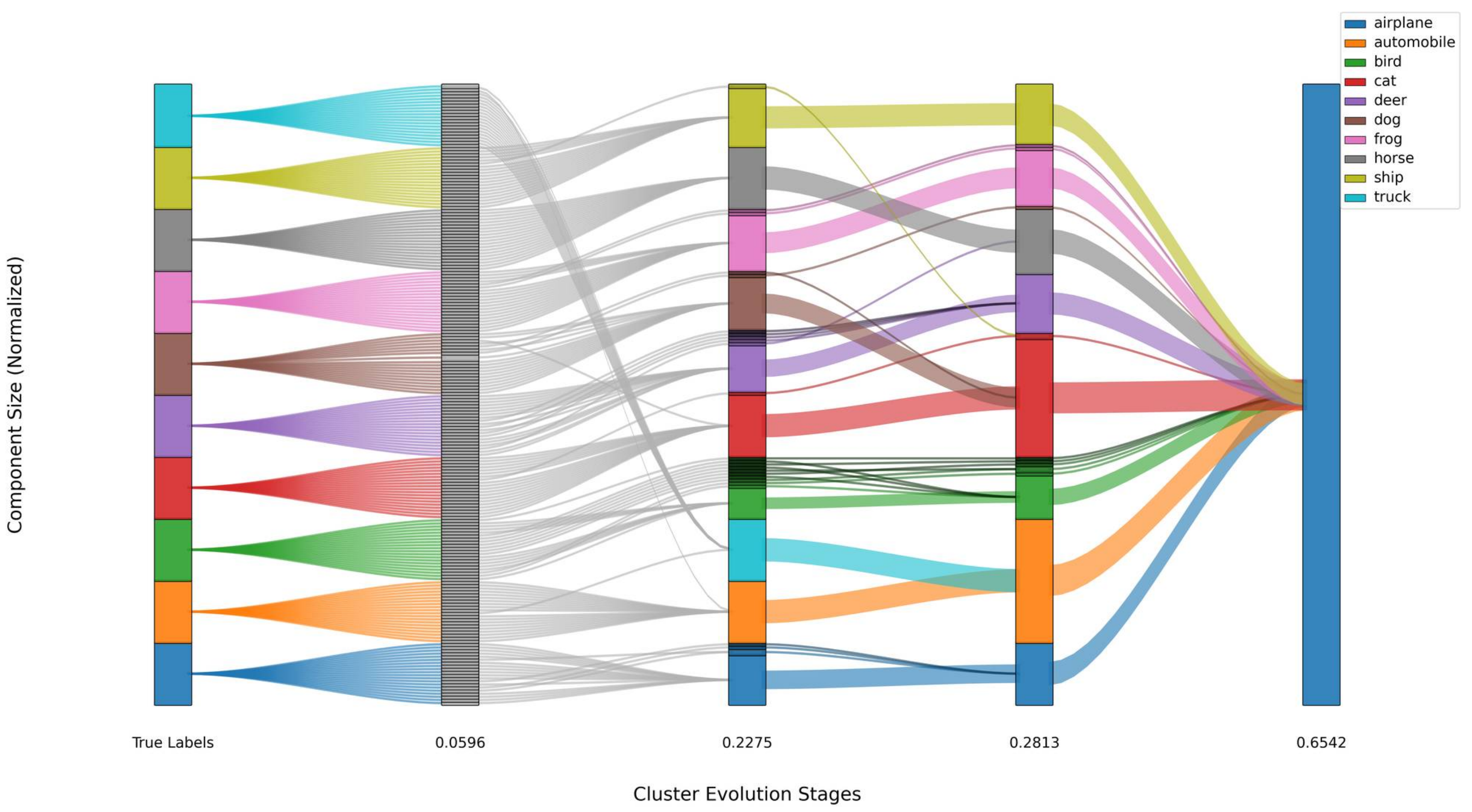}
        \caption{Seed 821 -- \vcflow{cluster flow}}
    \end{subfigure}\hfill
    \begin{subfigure}{0.48\linewidth}
        \includegraphics[width=\linewidth]{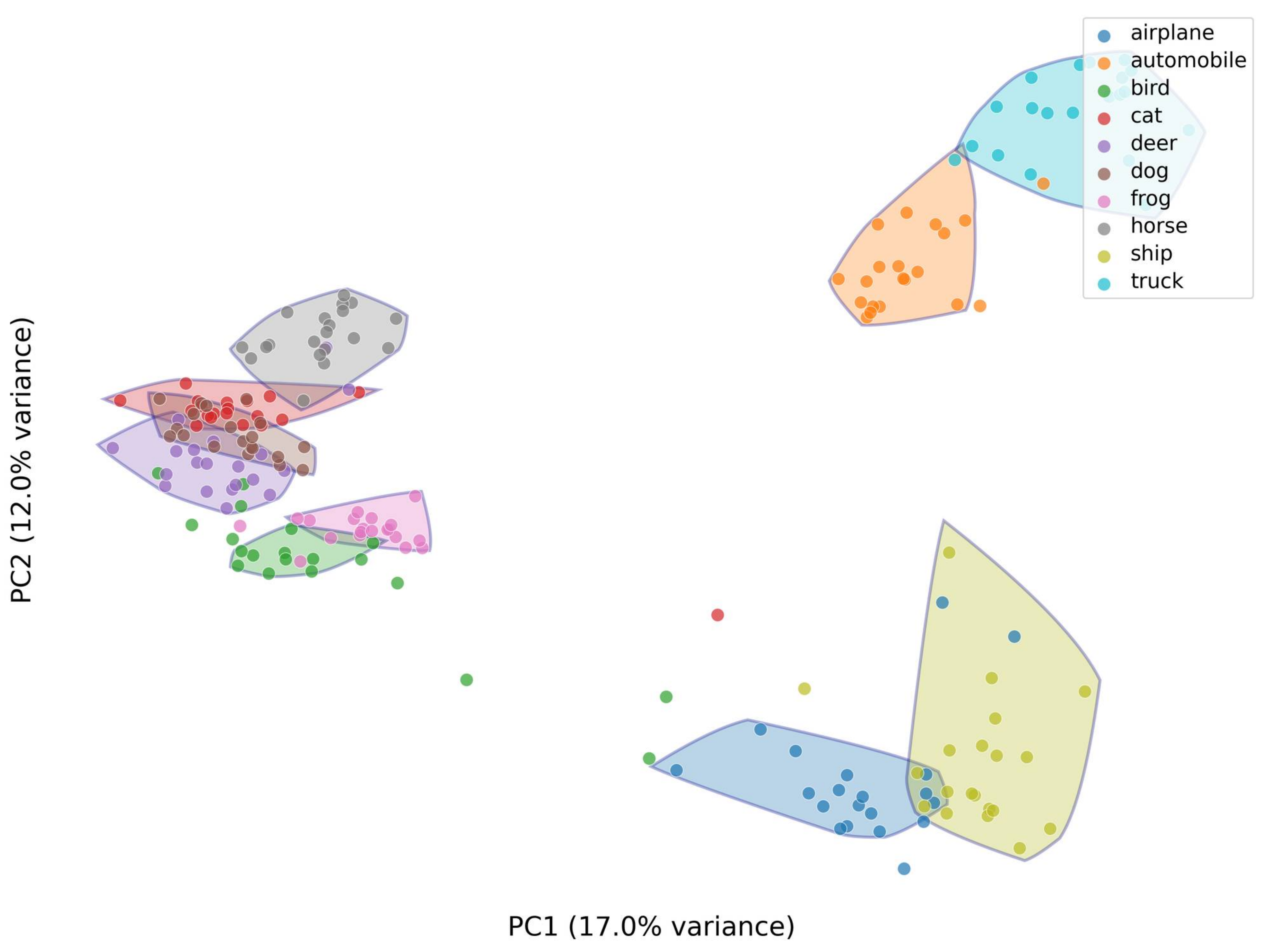}
        \caption{Seed 821 -- \vblob{blob}}
    \end{subfigure}

    \vspace{0.4em}

    \begin{subfigure}{0.48\linewidth}
        \includegraphics[width=\linewidth]{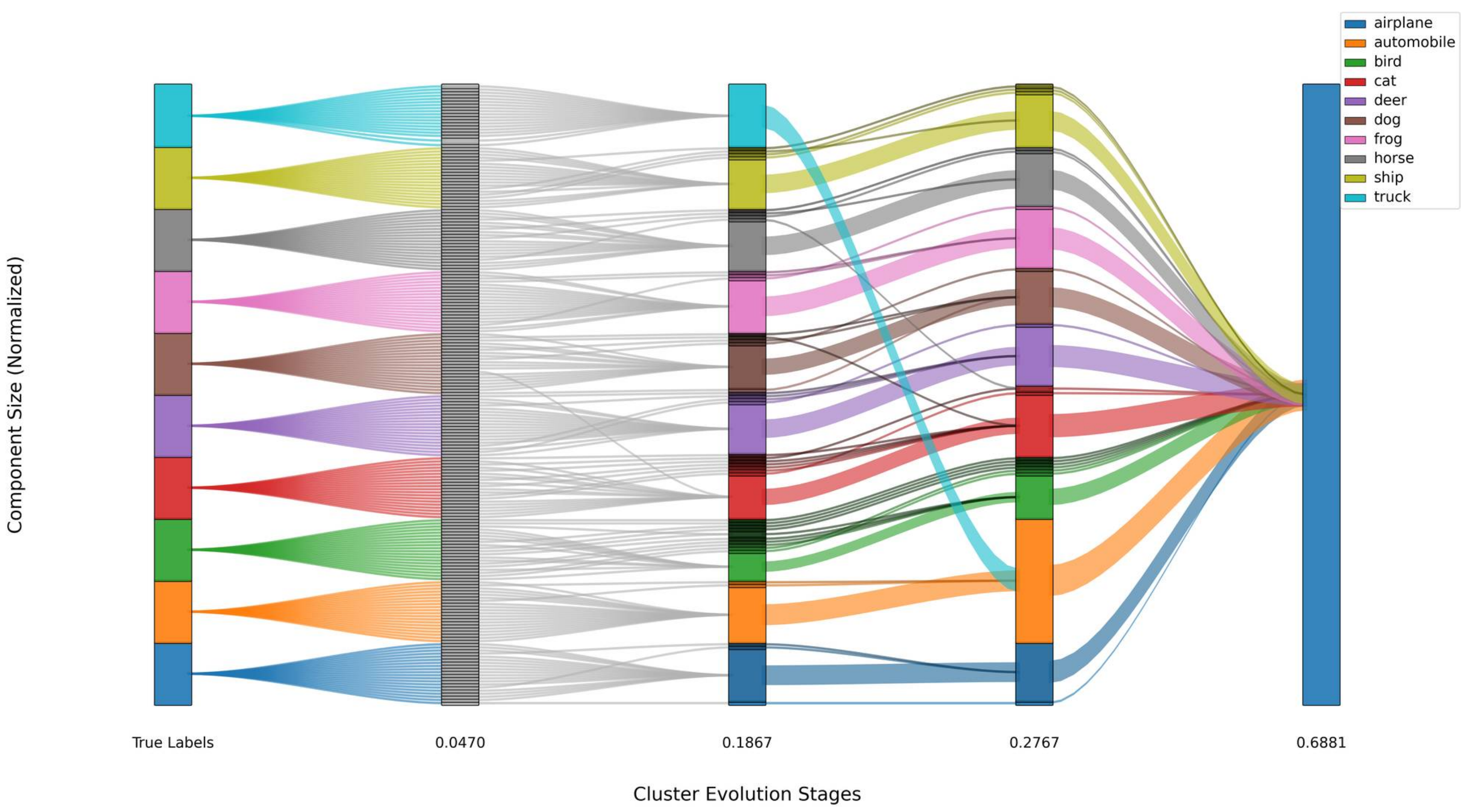}
        \caption{Seed 937 -- \vcflow{cluster flow}}
    \end{subfigure}\hfill
    \begin{subfigure}{0.48\linewidth}
        \includegraphics[width=\linewidth]{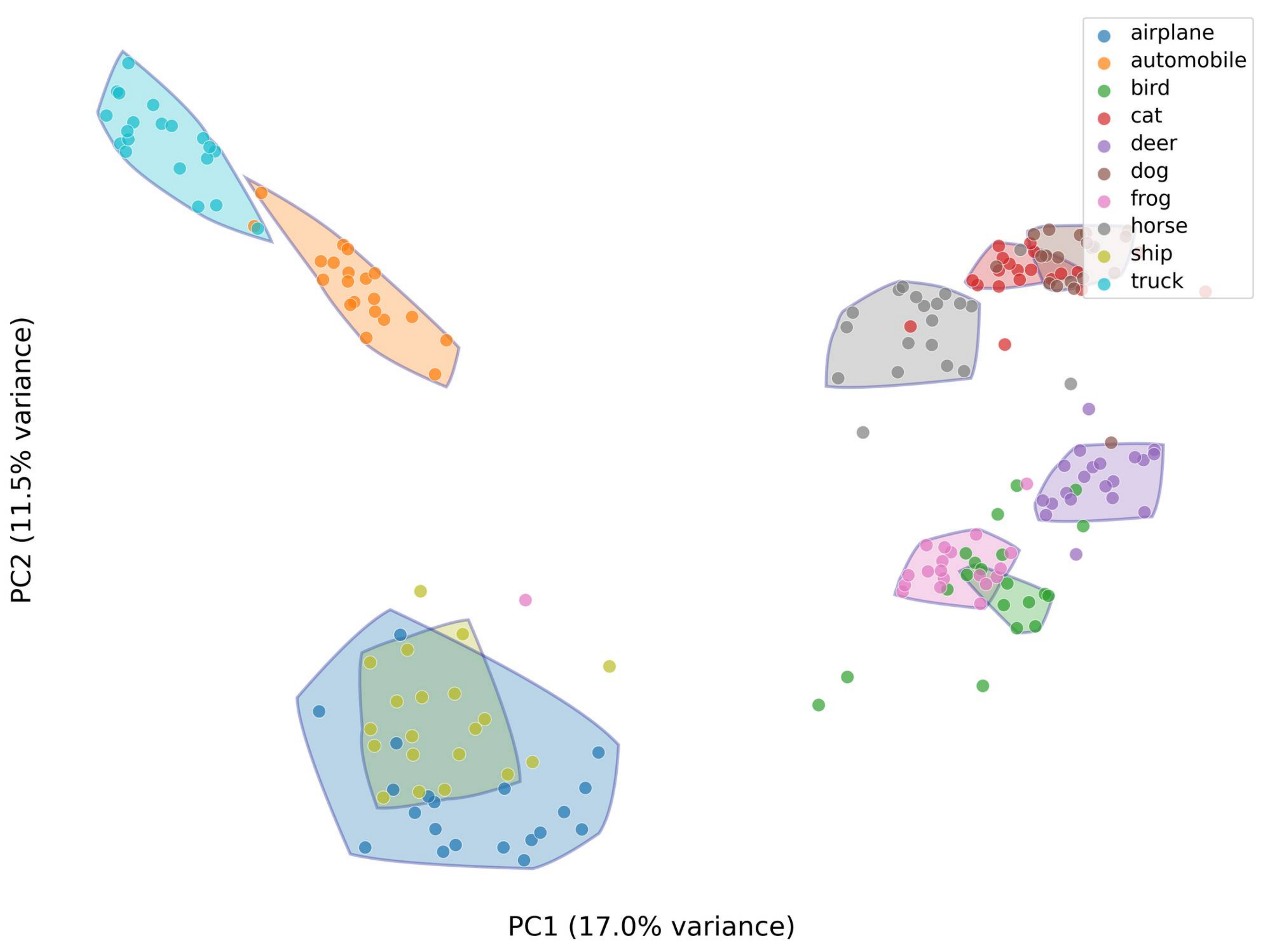}
        \caption{Seed 937 -- \vblob{blob}}
    \end{subfigure}

    \vspace{0.4em}

    \begin{subfigure}{0.48\linewidth}
        \includegraphics[width=\linewidth]{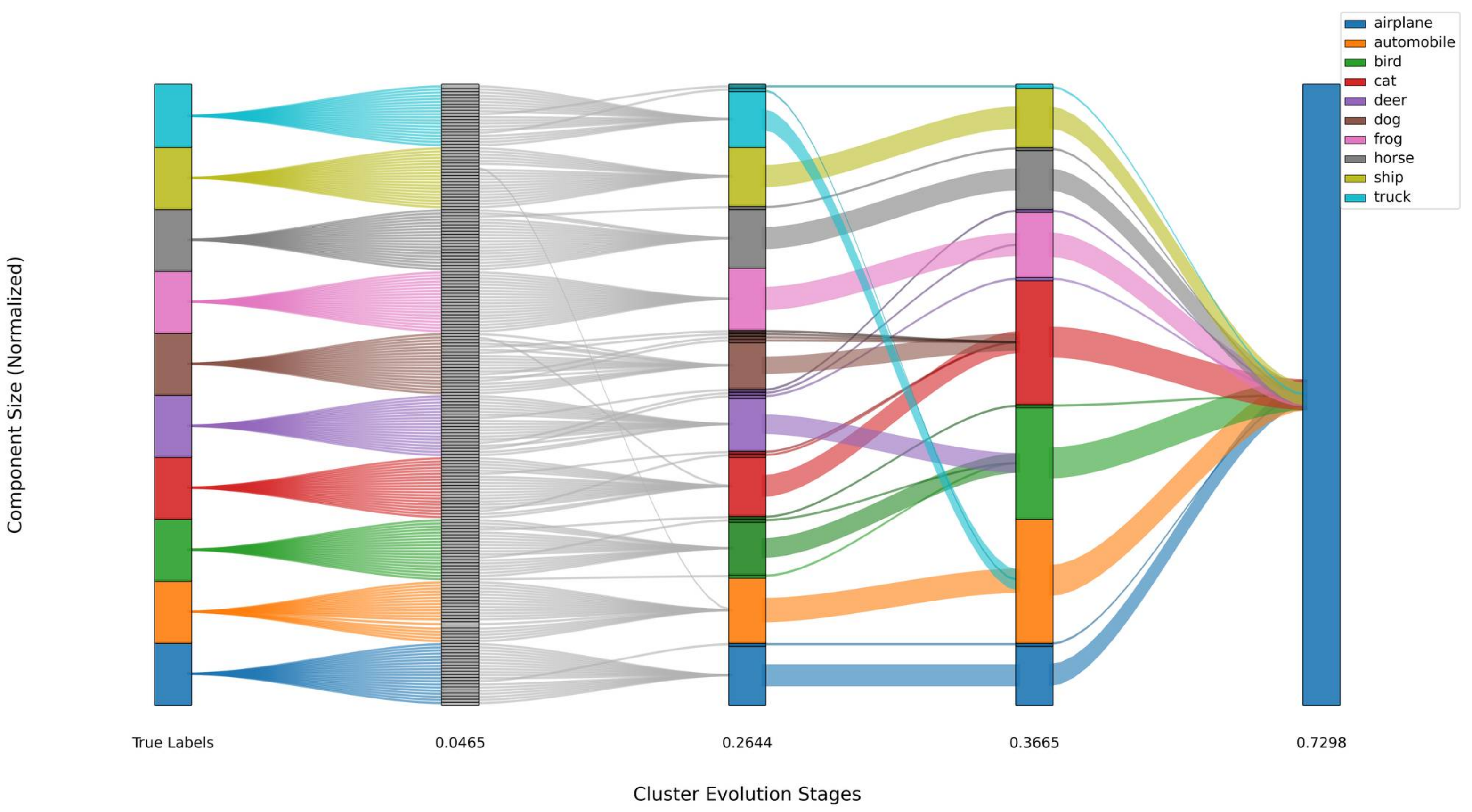}
        \caption{Seed 1001 -- \vcflow{cluster flow}}
    \end{subfigure}\hfill
    \begin{subfigure}{0.48\linewidth}
        \includegraphics[width=\linewidth]{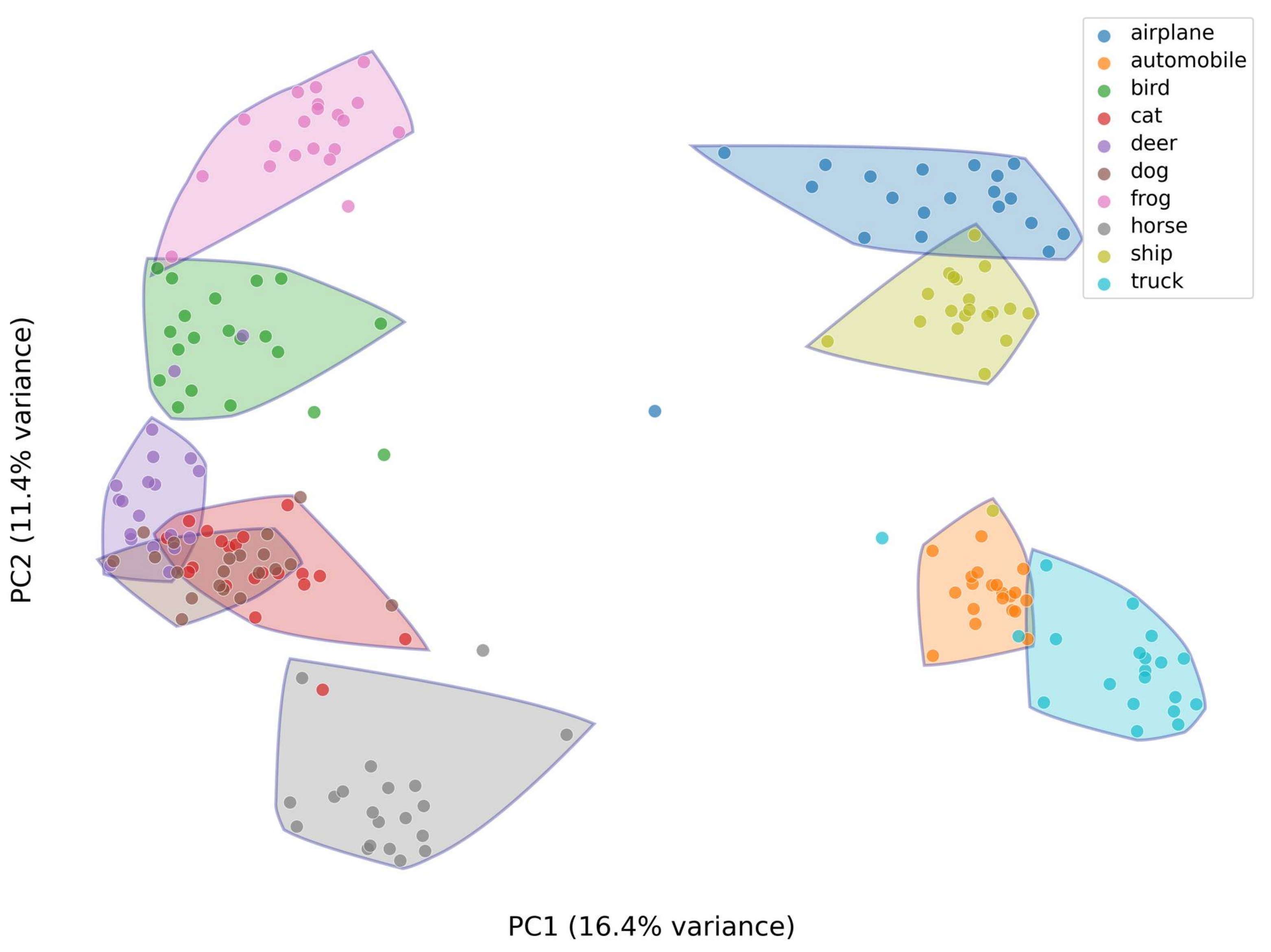}
        \caption{Seed 1001 -- \vblob{blob}}
    \end{subfigure}
    \caption{Stability analysis for ViT-B/16 encoder layer~11 (seeds 821--1001), continued from \cref{fig:stability_l11_p2}. Continued in \cref{fig:stability_l11_p4}.}
    \label{fig:stability_l11_p3}
\end{figure*}

\begin{figure*}[!htb]
    \centering
    \begin{subfigure}{0.48\linewidth}
        \includegraphics[width=\linewidth]{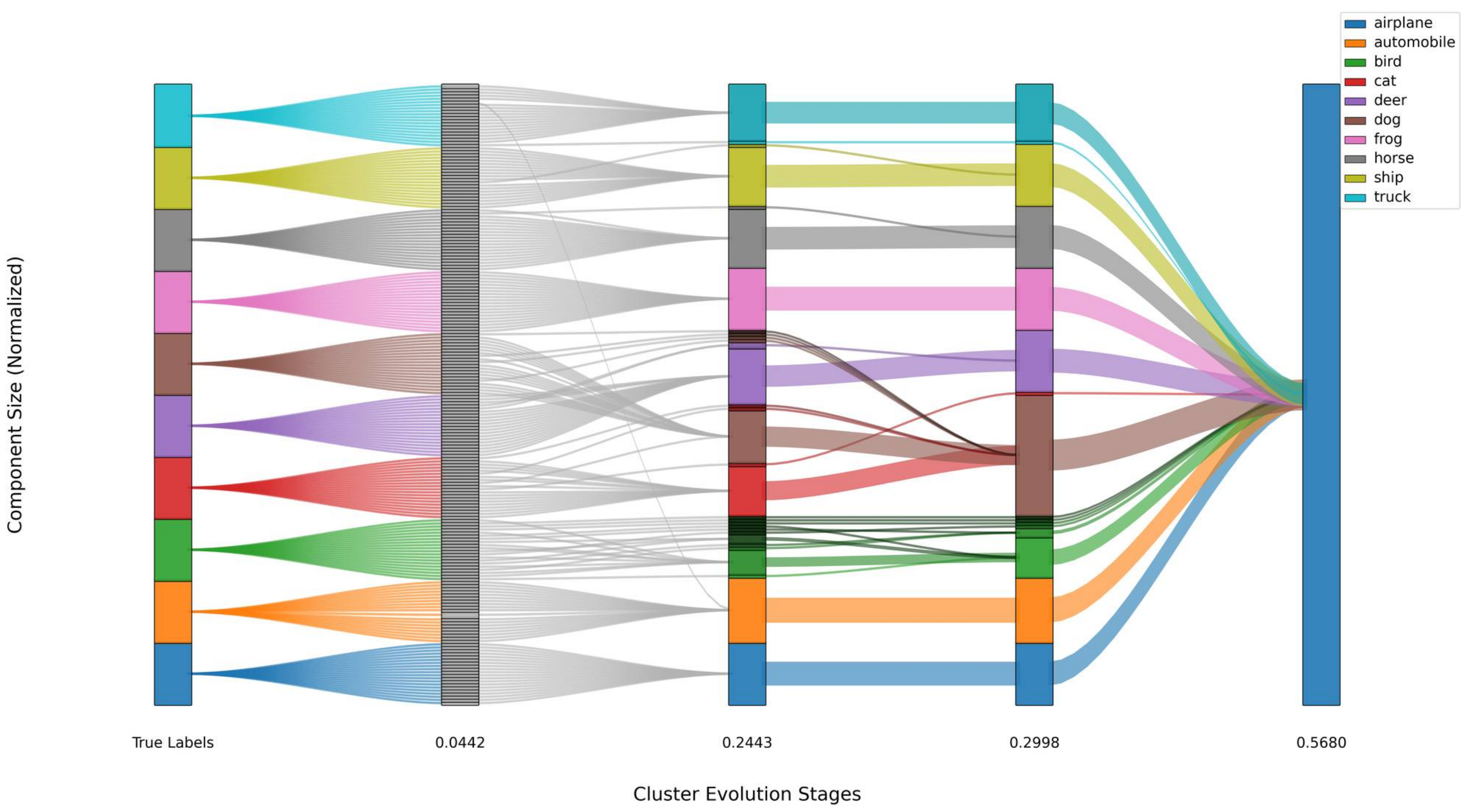}
        \caption{Seed 1234 -- \vcflow{cluster flow}}
    \end{subfigure}\hfill
    \begin{subfigure}{0.48\linewidth}
        \includegraphics[width=\linewidth]{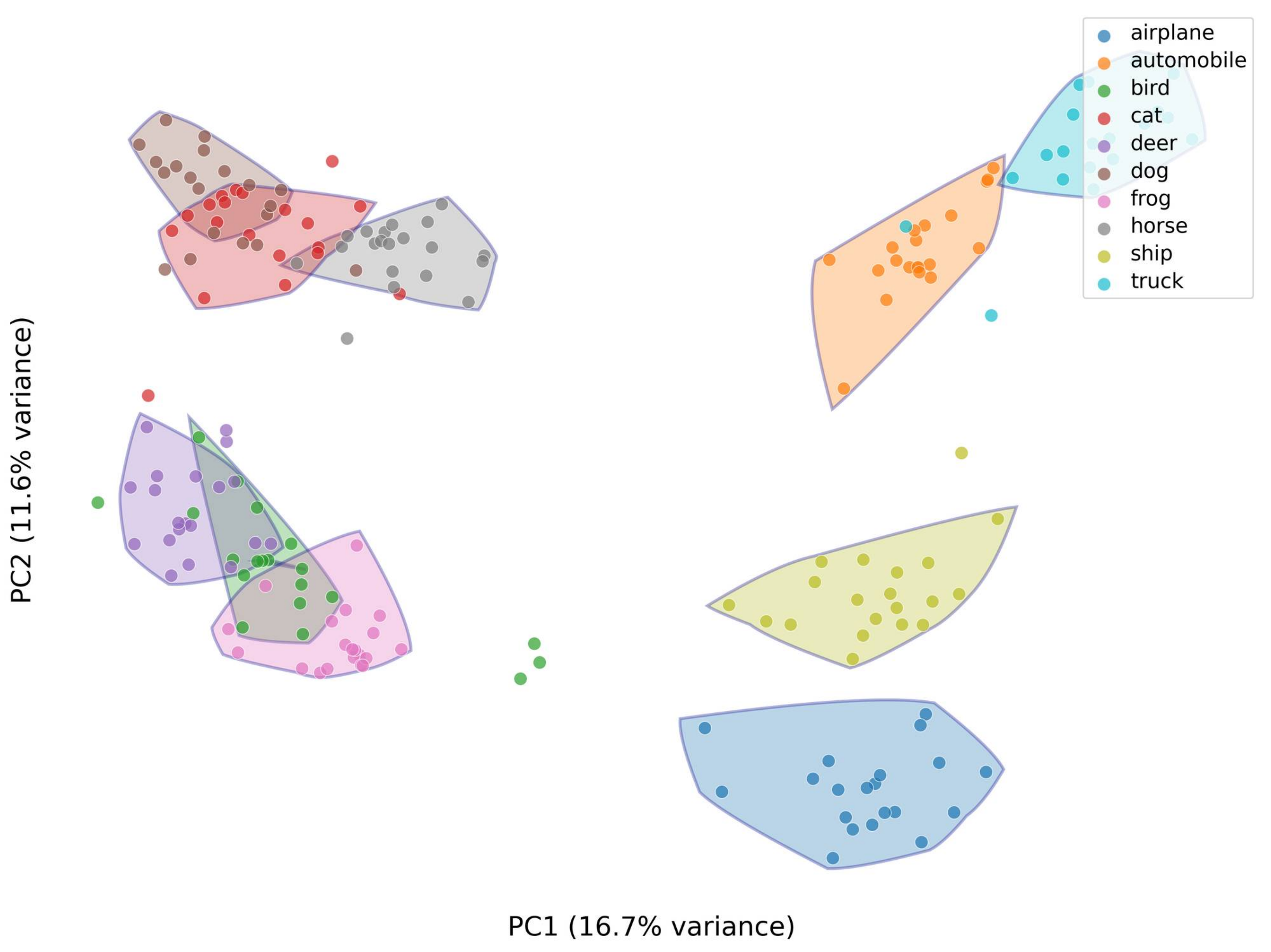}
        \caption{Seed 1234 -- \vblob{blob}}
    \end{subfigure}
    \caption{Stability analysis for ViT-B/16 encoder layer~11 (seed 1234), continued from \cref{fig:stability_l11_p3}. Class clusters remain compact and well-separated regardless of the specific probe-set sample.}
    \label{fig:stability_l11_p4}
    \label{fig:stability_l11_part2}
    \label{fig:stability_l11_blobs}
\end{figure*}

\paragraph{Visual Comparison}
\Cref{fig:stability_l9,fig:stability_l9_p2,fig:stability_l9_p3,fig:stability_l9_p4} show the \vcflow{cluster flow diagrams} (left) and \vblob{blob graphs} (right) for layer~9 across all 10 seeds; rapid merging and weak class separation are consistent regardless of seed.
\Cref{fig:stability_l11,fig:stability_l11_p2,fig:stability_l11_p3,fig:stability_l11_p4} show the same layout for layer~11, where class flows remain coherent across filtration stages and blob graphs show compact, well-separated class clusters in every seed.

%% --- Layer~9: 3 rows per figure (seeds 0-2, 3-5, 6-8, 9) ---

%% --- Layer~11: 3 rows per figure (seeds 0-2, 3-5, 6-8, 9) ---

\section{Stability Under Distribution Shift and Uneven Sampling}
\label{sec:shift_imbalance}

The probe-set-seed analysis of \appRef{sec:stability_analysis} varies the sample while holding its distribution fixed. To assess whether HOLE's topological observations also hold under \emph{distribution shift} and \emph{uneven sampling}, we run two companion experiments that reuse the same ViT-B/16 model, cosine metric, embedding extraction, and cluster-flow pipeline, changing only the data source or the sampler. All metrics are reported at encoder layer~11, where class-discriminative structure is strongest, using two quantities at the middle filtration threshold: \emph{global purity}, the sample-weighted fraction of points whose cluster majority label matches their own, and \emph{per-class purity}, the fraction of a class's points contained in its single largest cluster, averaged over classes.

\subsection{Distribution Shift (CIFAR-10-C)}

We replace the clean CIFAR-10 probe set with corrupted images from CIFAR-10-C~\cite{hendrycks2019benchmarking}, taking one corruption per category (Gaussian noise, motion blur, fog, and JPEG compression) at severities $1$, $3$, and $5$, with three seeds each. This generalizes the additive-noise experiment of \cref{sec:app2} to the standard corruption benchmark.

\begin{figure*}[!htb]
  \centering
  \begin{subfigure}{0.32\linewidth}
    \includegraphics[width=\linewidth]{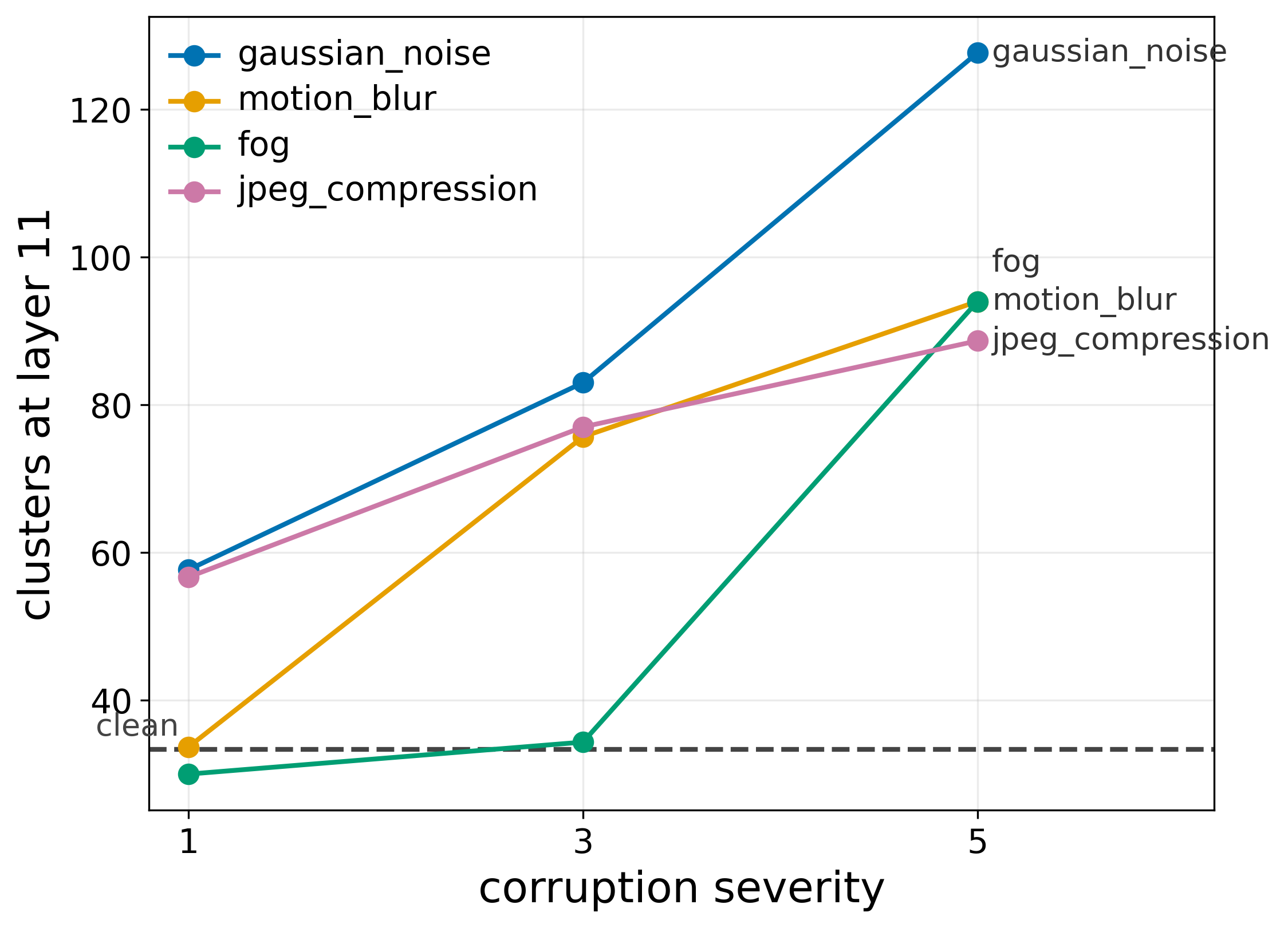}
    \caption{Component count vs.\ severity}
  \end{subfigure}\hfill
  \begin{subfigure}{0.32\linewidth}
    \includegraphics[width=\linewidth]{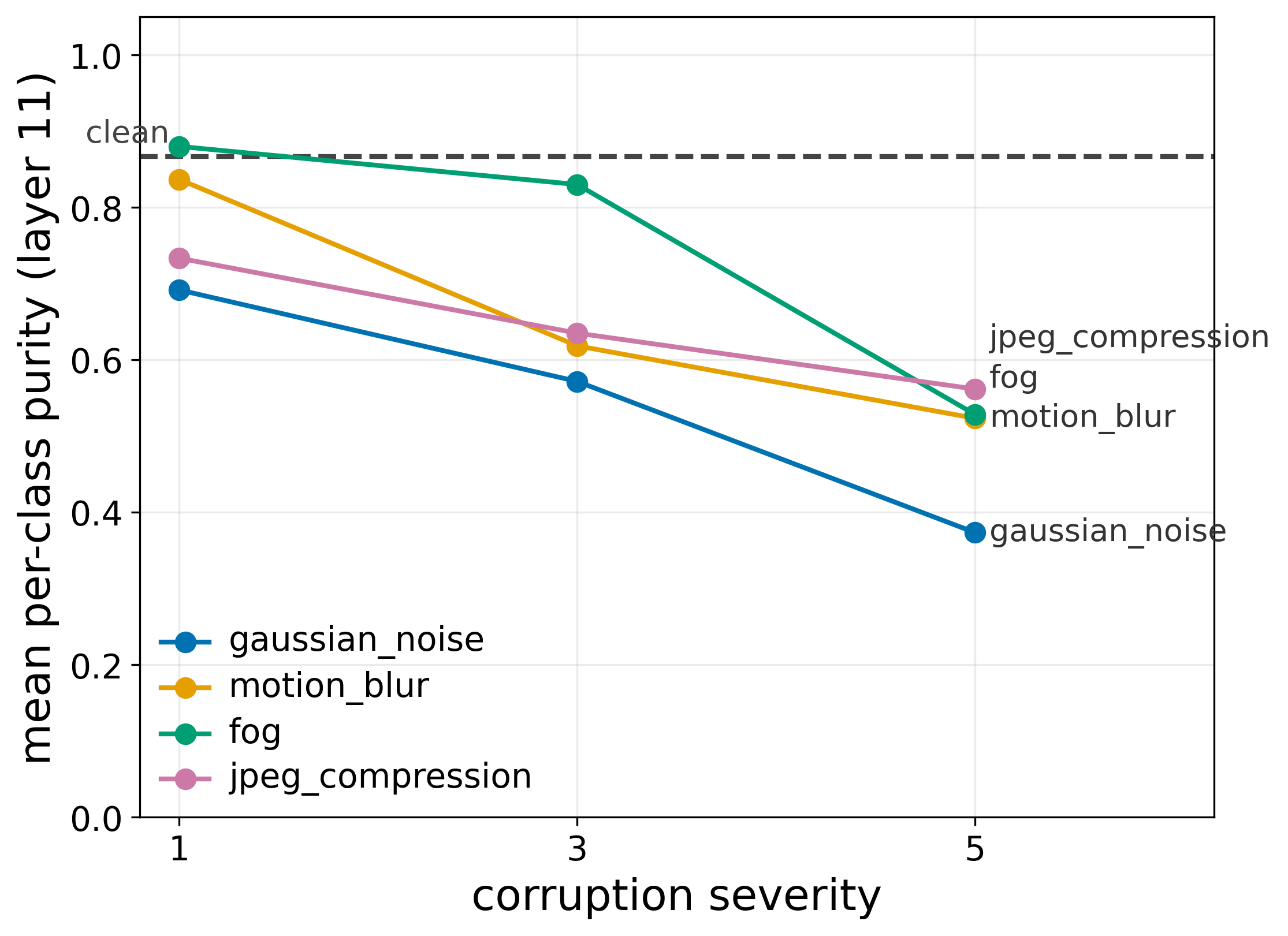}
    \caption{Per-class purity vs.\ severity}
  \end{subfigure}\hfill
  \begin{subfigure}{0.32\linewidth}
    \includegraphics[width=\linewidth]{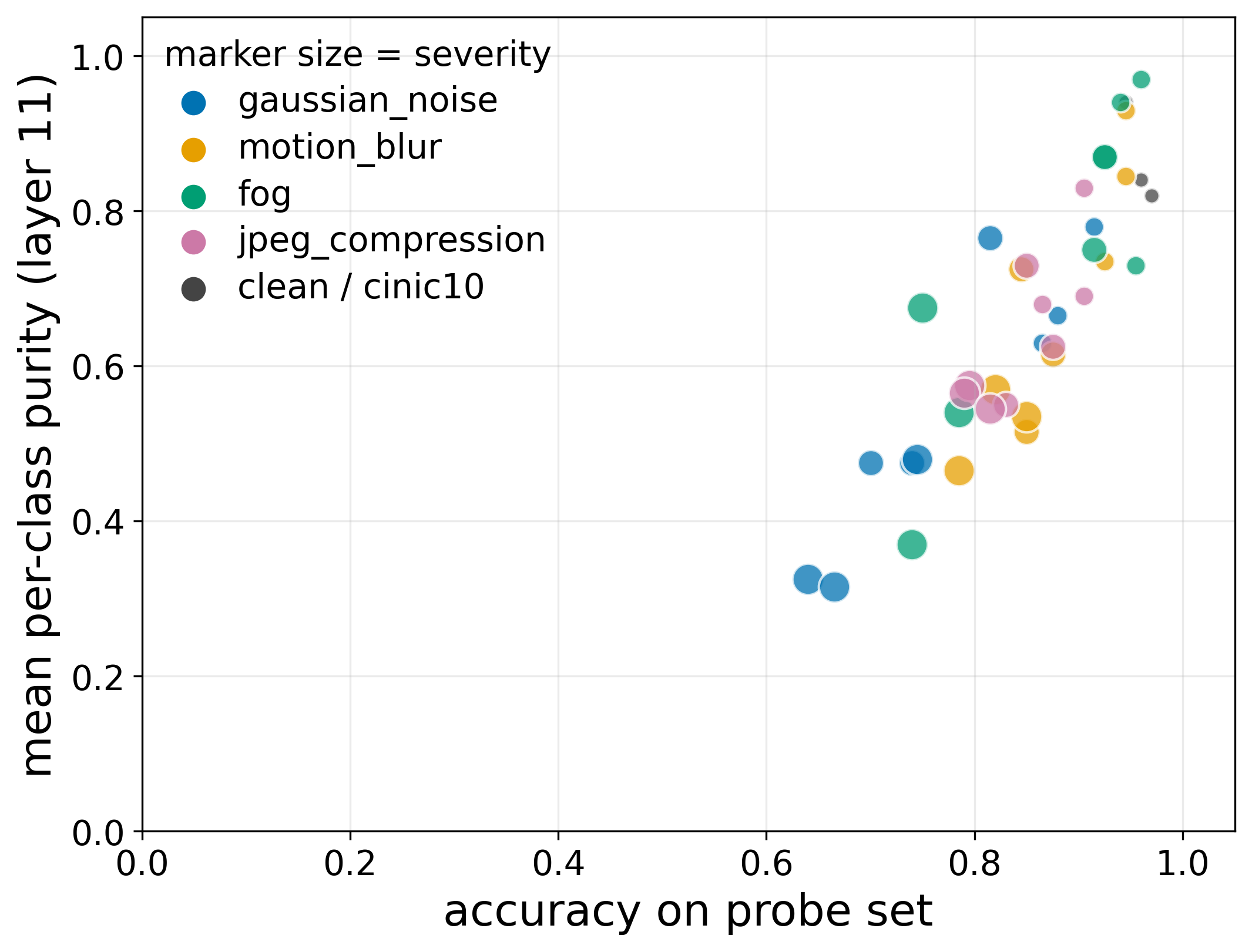}
    \caption{Per-class purity vs.\ accuracy}
  \end{subfigure}
  \caption{Distribution-shift analysis at ViT-B/16 layer~11 (CIFAR-10-C). As corruption severity increases, layer-11 clusters fragment~(a) and per-class purity falls~(b), and the degradation tracks the accuracy drop~(c). One line/marker per corruption; the clean baseline is dashed.}
  \label{fig:shift_aggregate}
\end{figure*}

\begin{figure}[!htb]
  \centering
  \begin{subfigure}{0.48\linewidth}
    \includegraphics[width=\linewidth]{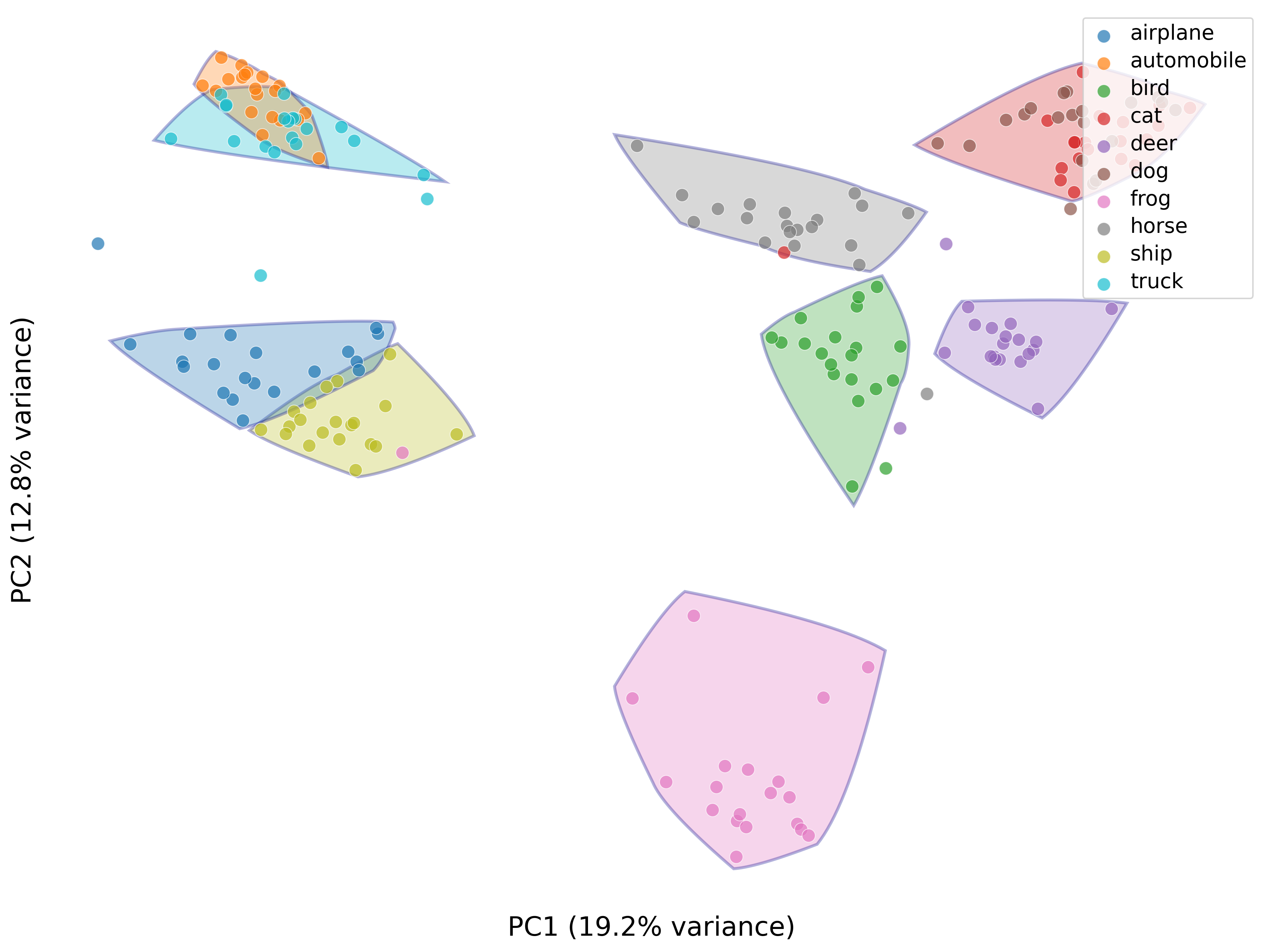}
    \caption{Clean}
  \end{subfigure}\hfill
  \begin{subfigure}{0.48\linewidth}
    \includegraphics[width=\linewidth]{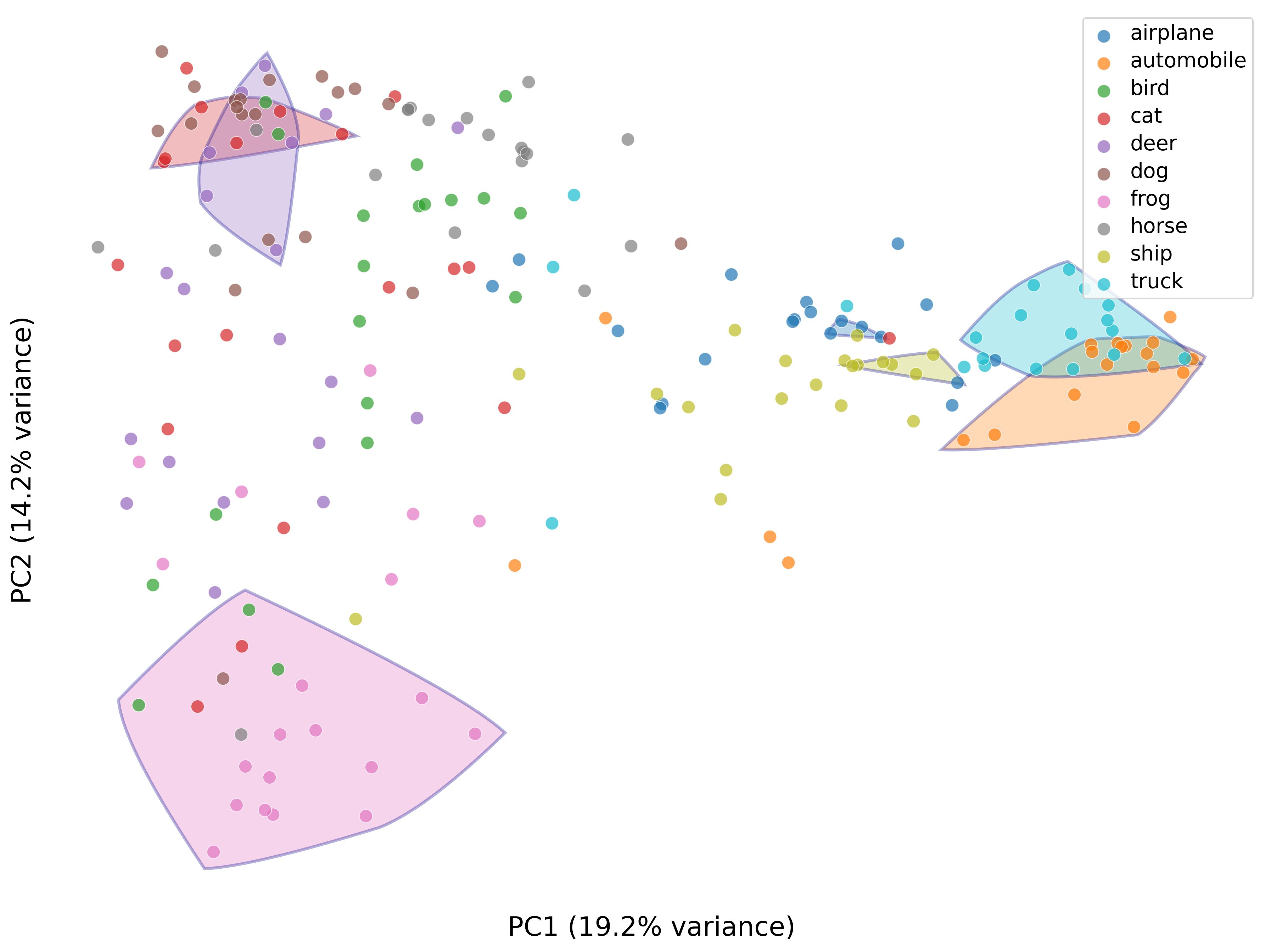}
    \caption{Gaussian noise, severity 5}
  \end{subfigure}
  \caption{Layer-11 \vblob{blob graphs} (unfiltered) for a clean probe set~(a) and under severe Gaussian corruption~(b). Under shift the compact per-class clusters break into many small fragments rather than merging across classes.}
  \label{fig:shift_blobs}
\end{figure}

The results (\cref{tab:shift_summary} and \cref{fig:shift_aggregate}) reveal a consistent signature. As severity increases, \emph{global purity stays high} ($0.92$ to $0.96$), so corruption does not cause different classes to merge into shared clusters. Instead the class structure \emph{shatters}: the number of layer-11 components grows sharply, from $33$ on clean inputs to $128$ under the strongest Gaussian noise, and per-class purity falls correspondingly, from $0.87$ to $0.37$, because each class fragments into several small clusters rather than remaining a single one. \Cref{fig:shift_blobs} shows this at the point-cloud level. Crucially, the topological degradation tracks task performance: per-class purity decreases with the accuracy drop (\cref{fig:shift_aggregate}c), so the topological signature acts as an accuracy-free indicator of representational damage. Gaussian noise is the most disruptive corruption and fog the mildest, mirroring the accuracy ranking.

\begin{table}[!htb]
  \centering
  \caption{HOLE under distribution shift (ViT-B/16, CIFAR-10-C, layer~11, cosine distance, mean over 3 seeds). Global purity stays high while per-class purity falls and the component count grows: class clusters shatter rather than mix as severity increases.}
  \label{tab:shift_summary}
  \resizebox{\linewidth}{!}{%
  \small
  \begin{tabular}{llrrcc}
    \toprule
    \textbf{Corruption} & \textbf{Sev.} & \textbf{Acc.} & \textbf{\#Clus.} & \textbf{Global pur.} & \textbf{Per-class pur.} \\
    \midrule
    (clean)          & 0 & 95.8\% & 33  & 0.96 & 0.87 \\
    \midrule
    Gaussian noise   & 1 & 88.7\% & 58  & 0.96 & 0.69 \\
                     & 3 & 75.2\% & 83  & 0.93 & 0.57 \\
                     & 5 & 68.3\% & 128 & 0.95 & 0.37 \\
    \midrule
    Motion blur      & 1 & 93.8\% & 34  & 0.94 & 0.84 \\
                     & 3 & 85.7\% & 76  & 0.95 & 0.62 \\
                     & 5 & 81.8\% & 94  & 0.95 & 0.52 \\
    \midrule
    Fog              & 1 & 95.2\% & 30  & 0.93 & 0.88 \\
                     & 3 & 92.2\% & 34  & 0.92 & 0.83 \\
                     & 5 & 75.8\% & 94  & 0.92 & 0.53 \\
    \midrule
    JPEG compression & 1 & 89.2\% & 57  & 0.95 & 0.73 \\
                     & 3 & 85.2\% & 77  & 0.96 & 0.64 \\
                     & 5 & 80.0\% & 89  & 0.96 & 0.56 \\
    \bottomrule
  \end{tabular}}
\end{table}

\subsection{Uneven Sampling}

We next relax the class-balanced assumption. Each run draws an \emph{imbalanced} probe set following one of three profiles (mild, long-tailed, and extreme), and the per-class counts are randomly permuted across classes at every seed, so which classes are starved varies from run to run; this separates the effect of under-sampling a class from that class being intrinsically hard to separate.

\begin{figure}[!htb]
  \centering
  \includegraphics[width=0.72\linewidth]{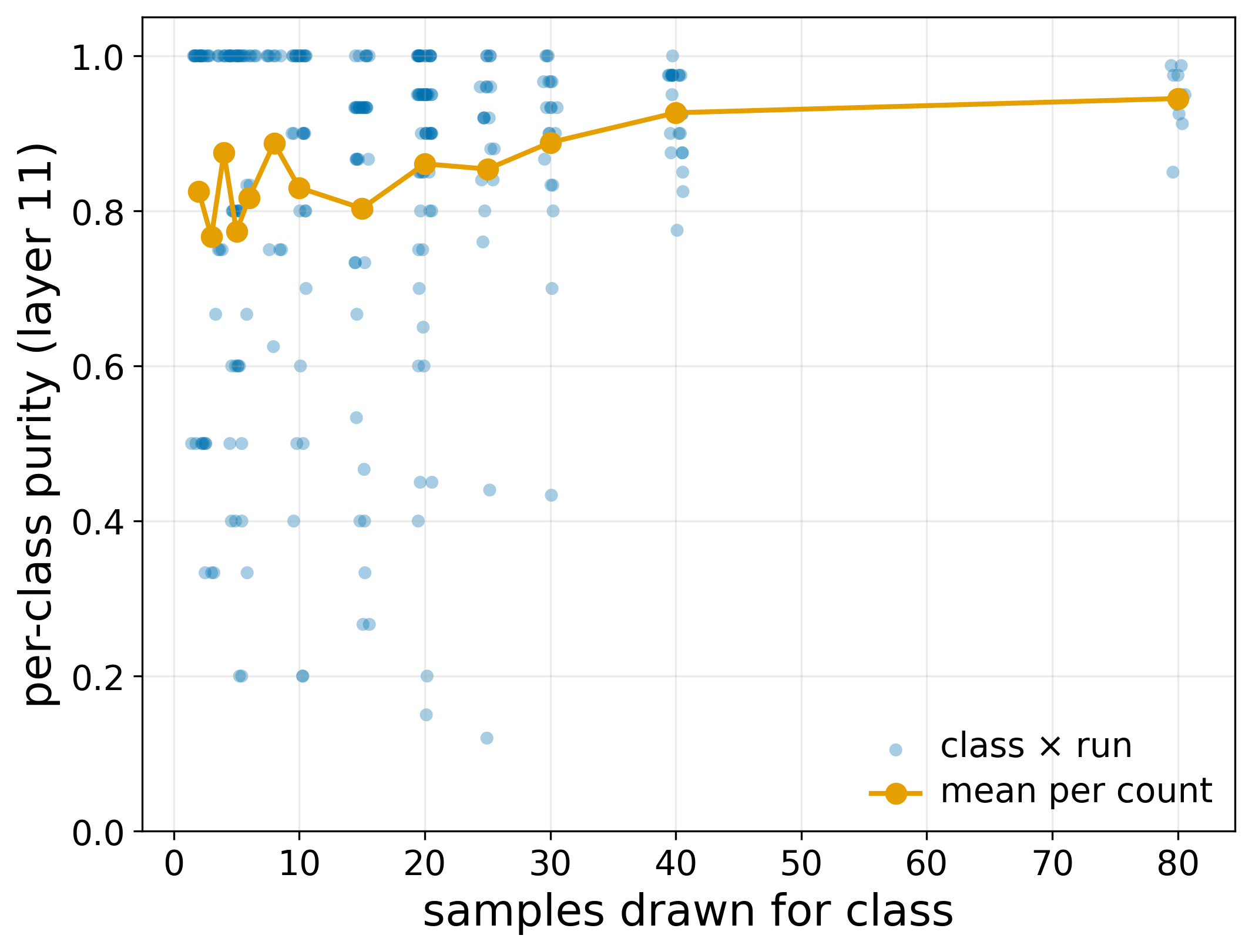}
  \caption{Uneven sampling at ViT-B/16 layer~11: per-class purity against the number of samples drawn for each class, pooled over imbalance profiles and seeds. Well-represented classes remain pure; starved classes lose their cluster structure.}
  \label{fig:imbalance}
\end{figure}

The topological conclusions remain valid for adequately represented classes and degrade gracefully for starved ones (\cref{fig:imbalance}). Averaged over seeds, classes drawn with at least $15$ samples retain high per-class purity ($0.91$ under the long-tailed profile and $0.93$ under the extreme profile), whereas classes drawn with at most $5$ samples fall to $0.77$ and $0.82$ respectively, while probe-set accuracy itself stays near $0.96$ throughout. The effect is therefore a property of how many samples a class contributes to the point cloud, not of the model degrading. Under the minimum-size filter, a class with very few samples can drop out of the diagram entirely, which is the expected and interpretable behavior: a class contributing only a handful of activations cannot form a persistent component.

Together, these experiments show that HOLE's topological readings survive realistic departures from clean, balanced sampling, and that the \emph{manner} of degradation, fragmentation under covariate shift and the drop-out of starved classes under imbalance, is itself an interpretable diagnostic.

\section{Scalability Evaluation}
\label{sec:scalability_eval}

The reviewers ask how HOLE scales for modern networks, both computationally and visually. We evaluate scalability along these two dimensions across three axes: the number of probe samples $N$, the number of classes, and the network width (activation dimension $d$). We profile the full twelve-layer pipeline with cosine distance, and for each configuration report the wall-clock time of distance-matrix construction and $H_0$ cluster evolution (the computational cost) together with the number of components the views must render (the visual cost); all times are seconds summed across the twelve encoder layers. The three axes are reported in \cref{tab:scale_samples,tab:scale_classes,tab:scale_width}.

\begin{table}[!htb]
  \centering
  \caption{Scalability in probe-set size (ViT-B/16, CIFAR-10, cosine; times summed over 12 layers). Distance cost grows quadratically in $N$ and the cluster-evolution cost faster still, which is why HOLE operates at $N \approx 200$ to $250$.}
  \label{tab:scale_samples}
  \resizebox{\linewidth}{!}{%
  \small
  \begin{tabular}{rrrrr}
    \toprule
    \textbf{$N$} & \textbf{Acc.} & \textbf{Dist.\ (s)} & \textbf{Cluster (s)} & \textbf{\#Clus.\ (L11)} \\
    \midrule
    50   & 90.0\% & 0.01  & 0.5    & 13  \\
    100  & 94.0\% & 0.02  & 1.1    & 23  \\
    250  & 95.6\% & 0.28  & 12.1   & 23  \\
    500  & 95.6\% & 1.18  & 63.9   & 138 \\
    1000 & 95.4\% & 5.00  & 358.7  & 260 \\
    1500 & 96.0\% & 13.44 & 1045.5 & 358 \\
    2000 & 96.3\% & 27.62 & 2300.3 & 464 \\
    \bottomrule
  \end{tabular}}
\end{table}

\begin{table}[!htb]
  \centering
  \caption{Scalability in number of classes (ViT-Base, cosine). HOLE runs at the $1000$-class scale of ImageNet-1k; the final-layer component count tracks the class count, not the network size.}
  \label{tab:scale_classes}
  \resizebox{\linewidth}{!}{%
  \small
  \begin{tabular}{lrrrrr}
    \toprule
    \textbf{Dataset} & \textbf{Cls.} & \textbf{$N$} & \textbf{Acc.} & \textbf{Cluster (s)} & \textbf{\#Clus.\ (L11)} \\
    \midrule
    Imagenette  & 10   & 500  & 83.2\% & 75.6  & 24  \\
    CIFAR-10    & 10   & 1000 & 95.4\% & 358.7 & 260 \\
    CIFAR-100   & 100  & 1000 & 91.9\% & 248.4 & 287 \\
    ImageNet-1k & 1000 & 1000 & 81.4\% & 286.4 & 947 \\
    \bottomrule
  \end{tabular}}
\end{table}

\begin{table}[!htb]
  \centering
  \caption{Scalability in network width (CIFAR-10, $N = 1000$, $10$ classes, cosine; one finetuning recipe). Distance cost scales linearly with the activation dimension $d$; even the $1408$-dimensional EVA-Giant is tractable, and the final-layer component count does not grow with width.}
  \label{tab:scale_width}
  \resizebox{\linewidth}{!}{%
  \small
  \begin{tabular}{lrrrrr}
    \toprule
    \textbf{Model} & \textbf{$d$} & \textbf{Acc.} & \textbf{Dist.\ (s)} & \textbf{Cluster (s)} & \textbf{\#Clus.} \\
    \midrule
    DeiT-Tiny  & 192  & 95.1\% & 3.18  & 338.2  & 276 \\
    DeiT-Small & 384  & 97.0\% & 3.68  & 337.6  & 200 \\
    ViT-Base   & 768  & 98.6\% & 5.05  & 367.1  & 37  \\
    ViT-Large  & 1024 & 98.9\% & 11.82 & 762.0  & 56  \\
    ViT-Huge   & 1280 & 97.1\% & 17.79 & 908.5  & 66  \\
    EVA-Giant  & 1408 & 98.9\% & 23.78 & 1150.5 & 27  \\
    \bottomrule
  \end{tabular}}
\end{table}

\paragraph{Computational Scalability}
Runtime is governed by the probe-set size. Along the samples axis (\cref{tab:scale_samples}) the distance-matrix cost grows quadratically in $N$, as expected for an $O(N^2 d)$ computation, and the $H_0$ cluster-evolution cost grows faster still, since more points create more merge events to track; at the operating point used throughout the paper ($N \approx 200$ to $250$) the entire twelve-layer analysis completes in about twelve seconds, whereas $N = 2000$ takes roughly forty minutes. Network width adds only a linear, sub-dominant term: sweeping six architectures at fixed $N = 1000$ (\cref{tab:scale_width}), the distance cost scales approximately linearly with $d$, and even the $1408$-dimensional EVA-Giant, a model with over a billion parameters, completes the full analysis in about twenty minutes. Network depth contributes a linear, embarrassingly parallel factor, since each layer is analyzed independently, and at a fixed point count the number of classes has little effect on runtime (\cref{tab:scale_classes}). The dominant cost is therefore the probe-set size, the very quantity HOLE fixes to a small value, and the topological signal already saturates there (\appRef{sec:stability_analysis}).

\begin{figure*}[!htb]
  \centering
  \setlength{\tabcolsep}{2pt}
  \renewcommand{\arraystretch}{0.6}
  \begin{tabular}{@{}cccc@{}}
     & \small 10 classes (Imagenette) & \small 100 classes (CIFAR-100) & \small 1000 classes (ImageNet-1k) \\
    \rotatebox{90}{\small\quad Cluster flow} &
      \includegraphics[width=0.29\textwidth]{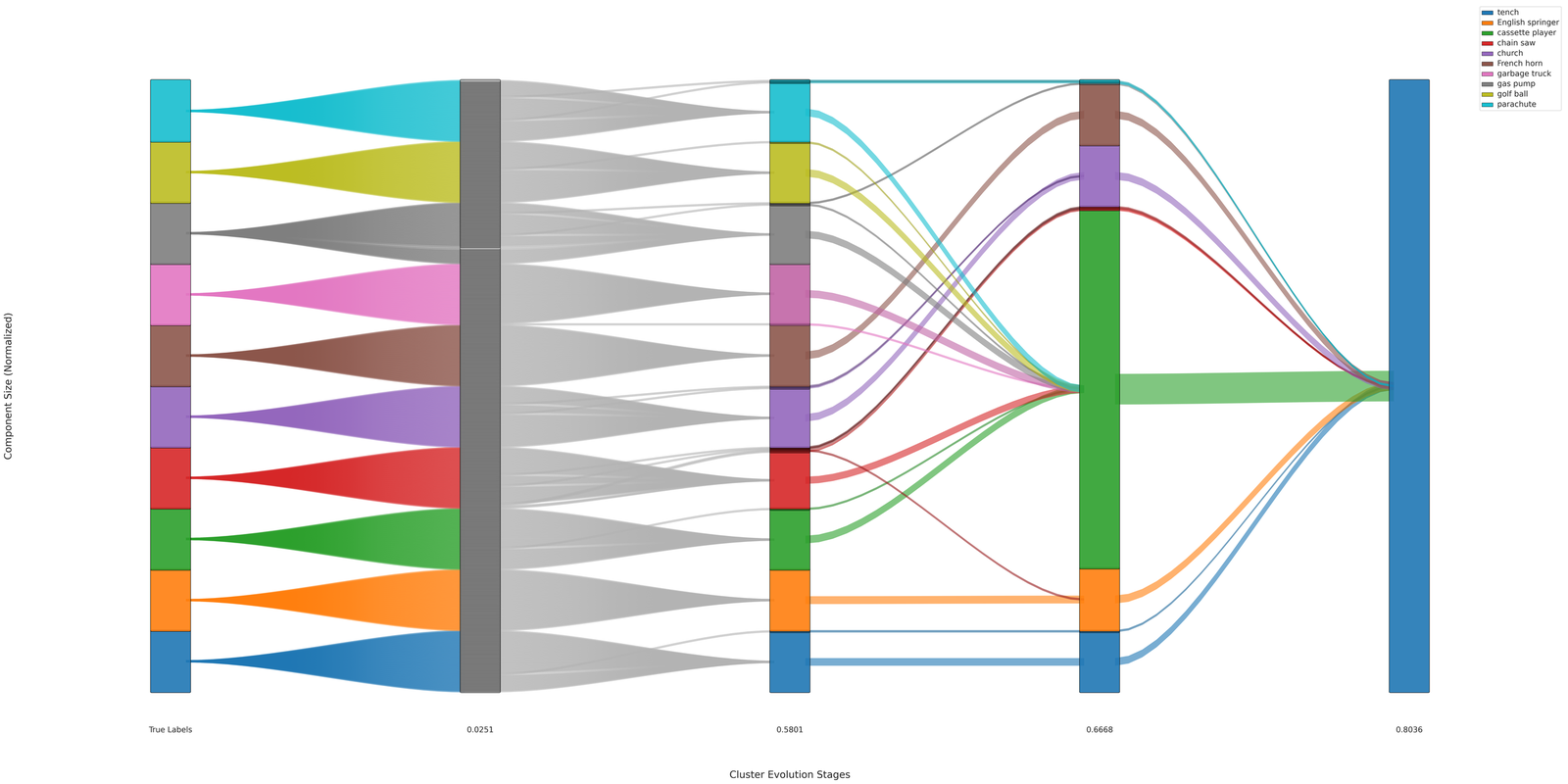} &
      \includegraphics[width=0.29\textwidth]{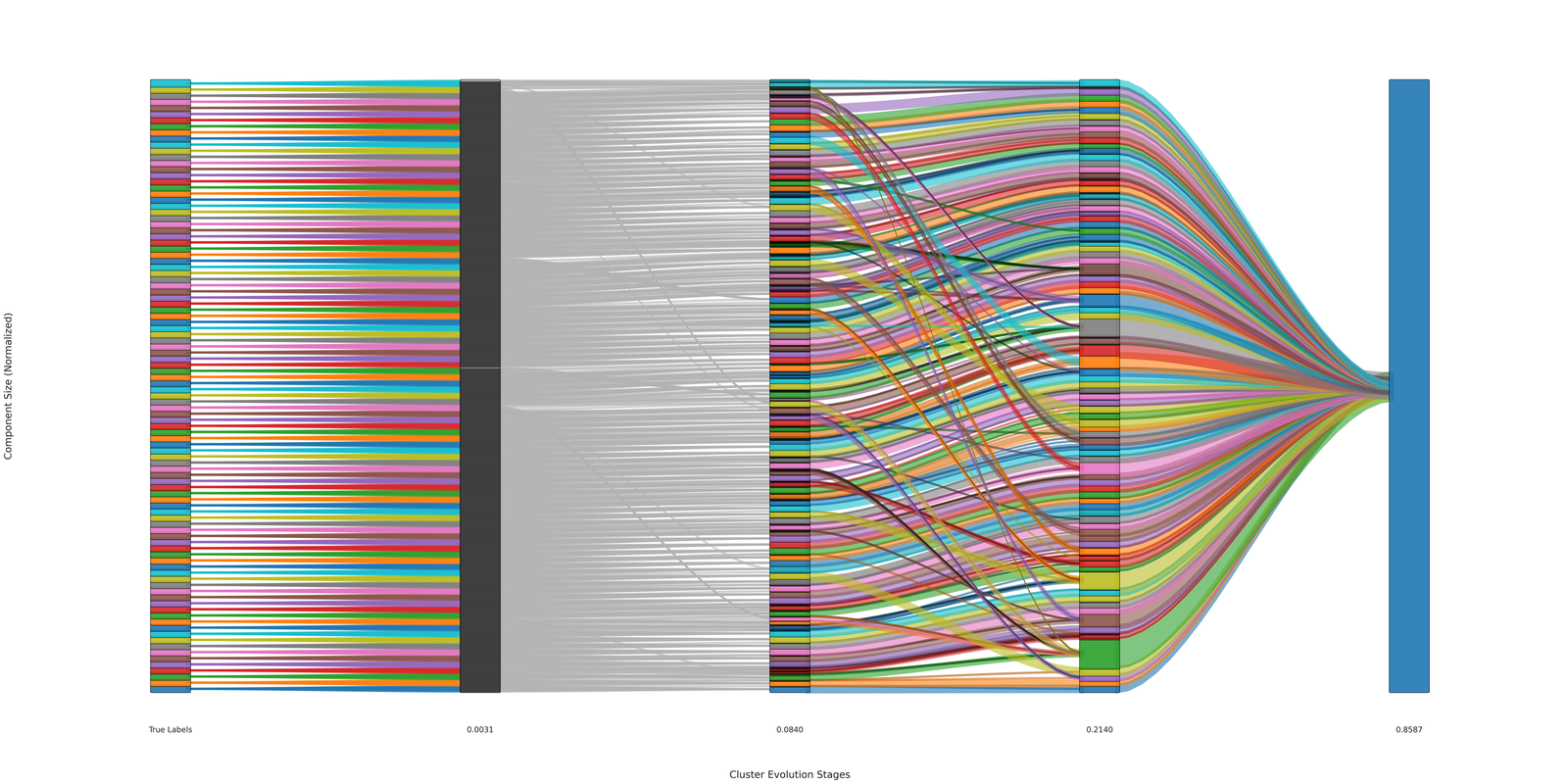} &
      \includegraphics[width=0.29\textwidth]{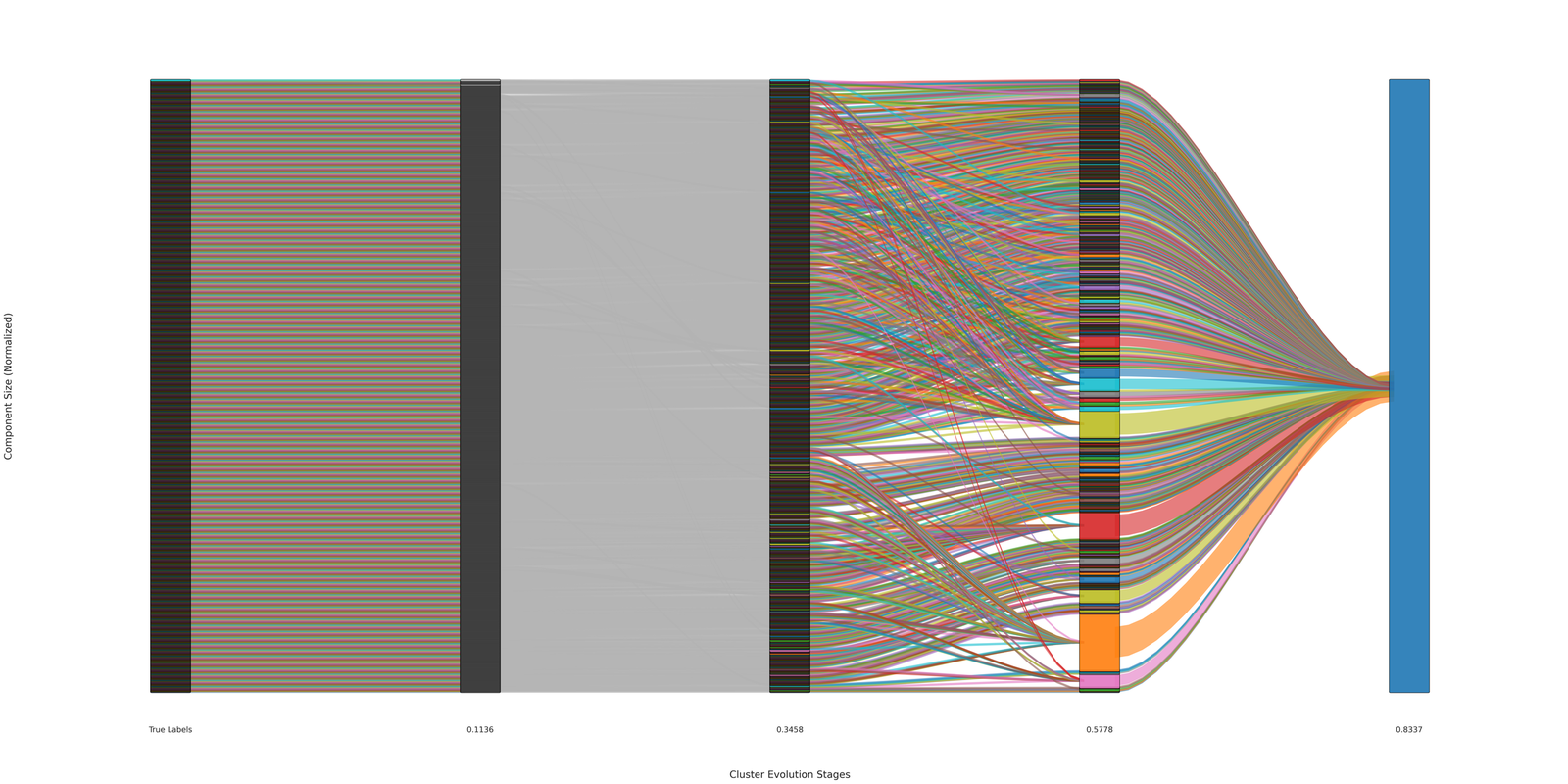} \\
    \rotatebox{90}{\small\quad\ Blob graph} &
      \includegraphics[width=0.29\textwidth]{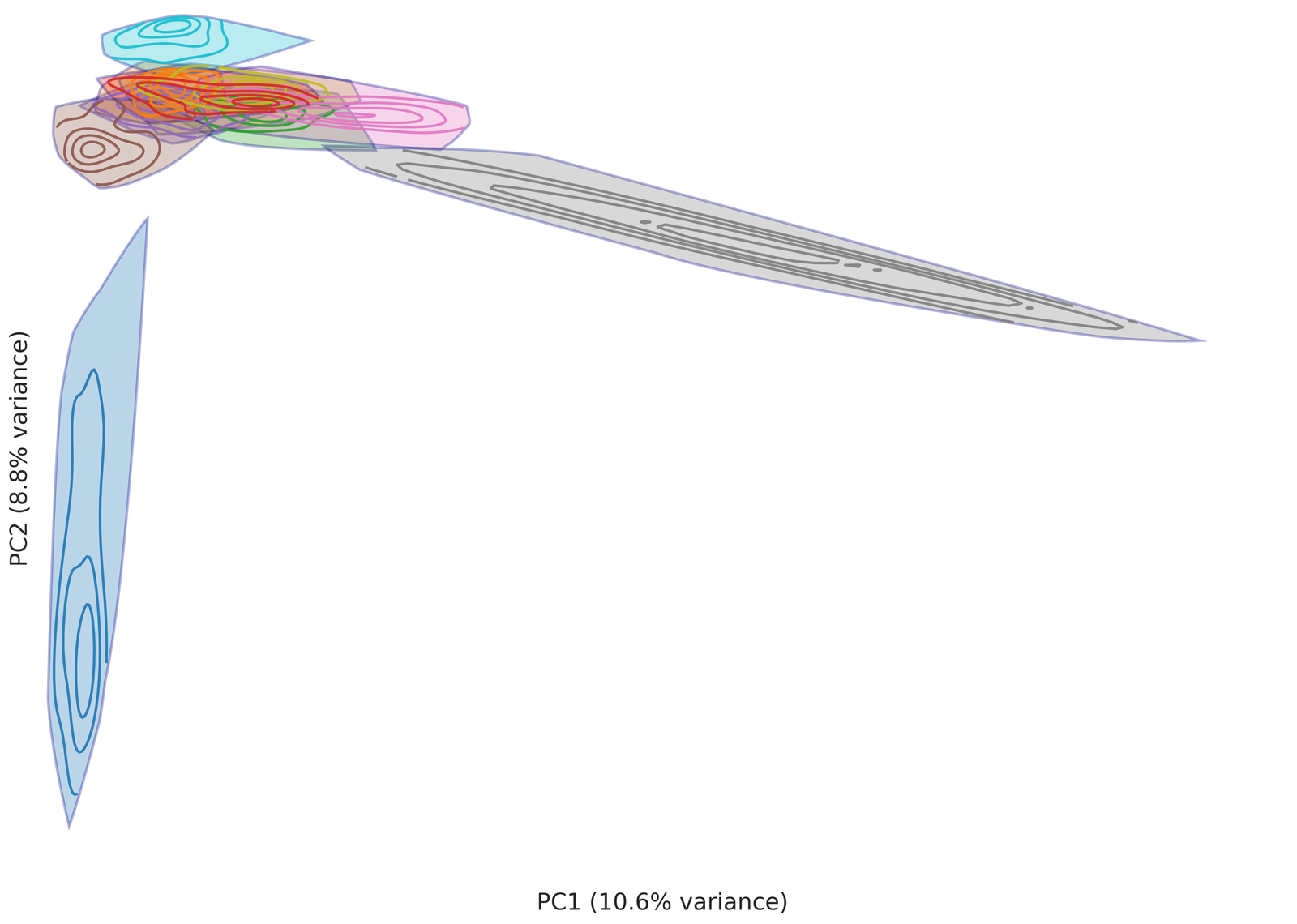} &
      \includegraphics[width=0.29\textwidth]{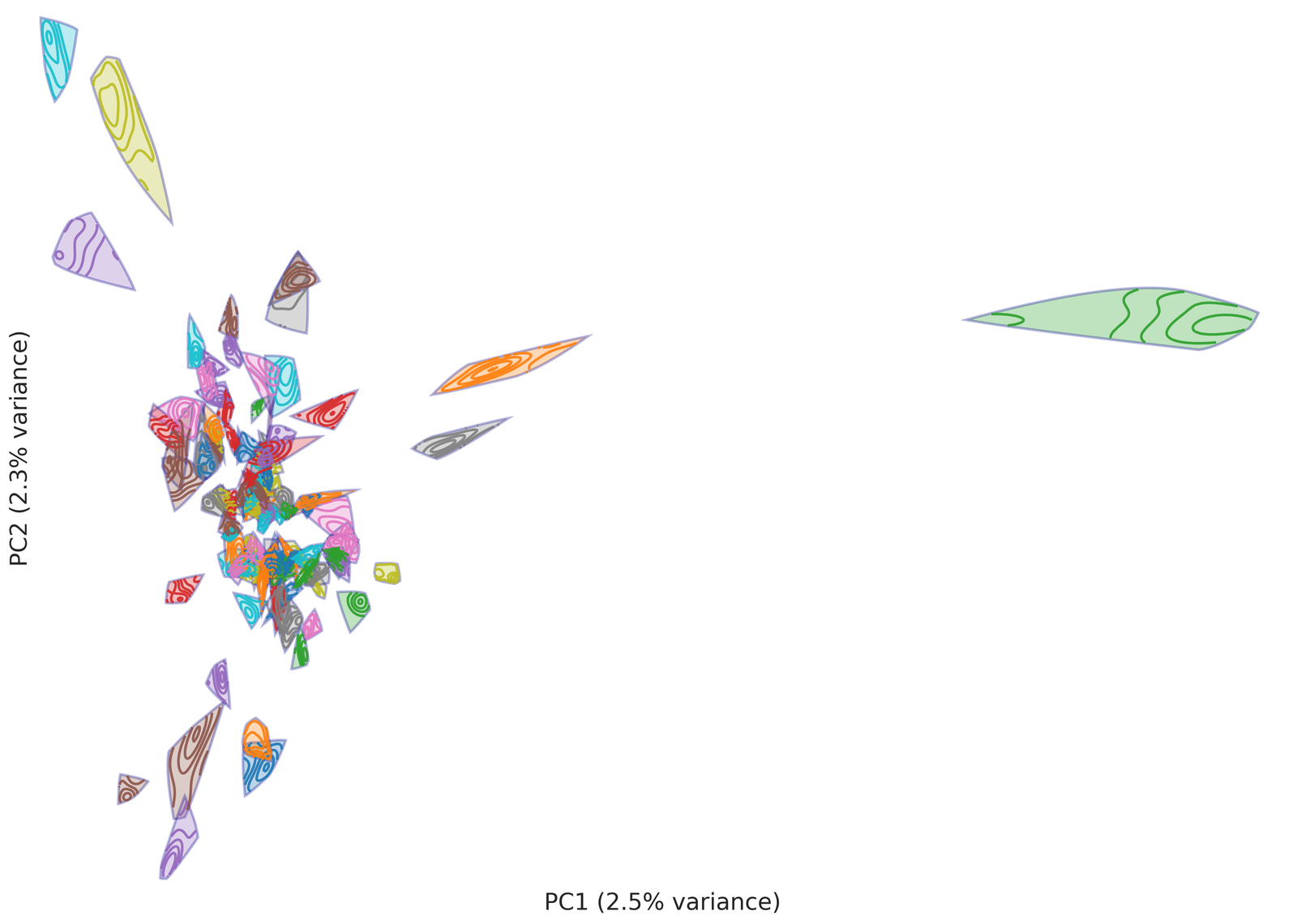} &
      \includegraphics[width=0.29\textwidth]{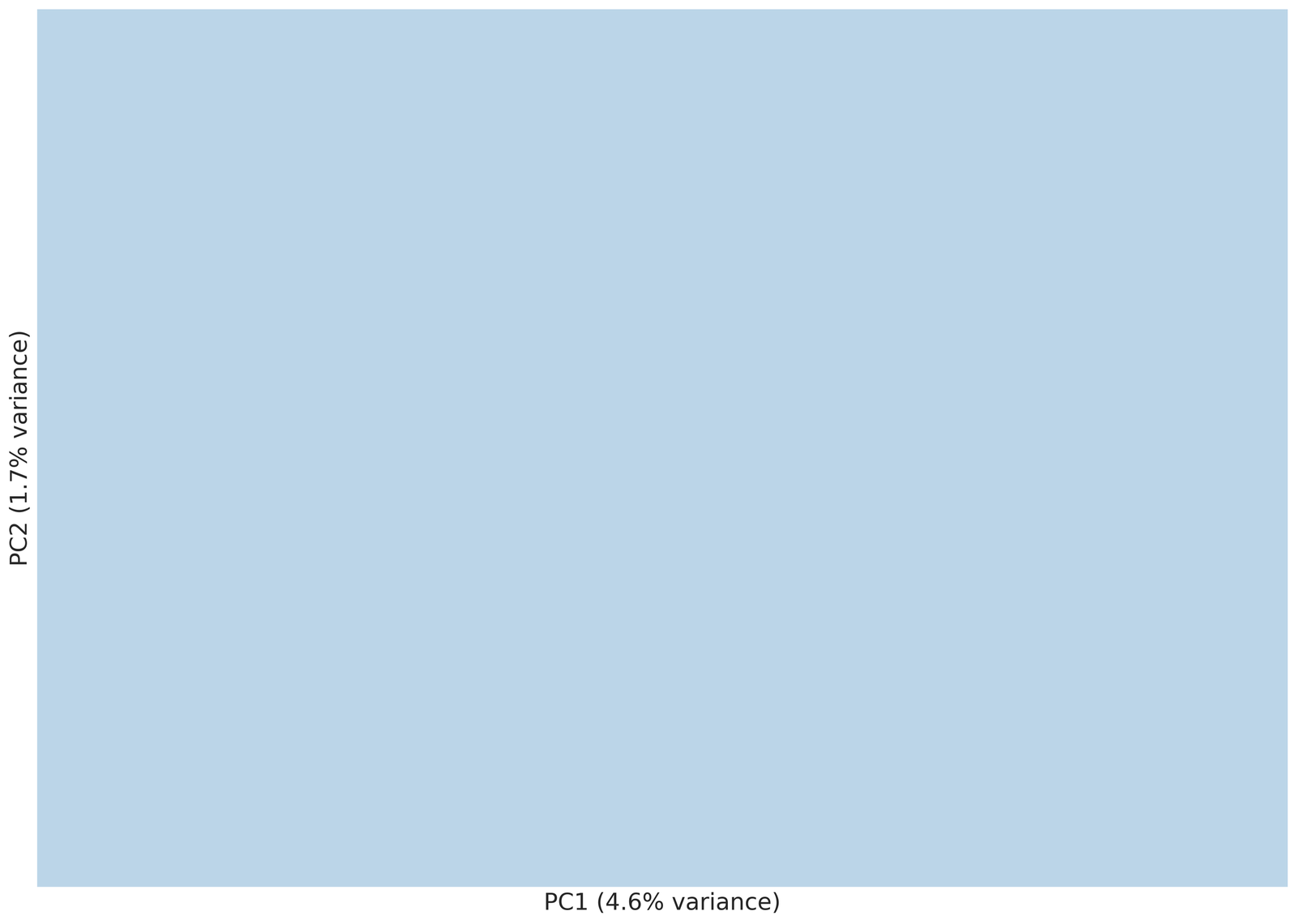} \\
    \rotatebox{90}{\small\ Dendrogram} &
      \includegraphics[width=0.29\textwidth]{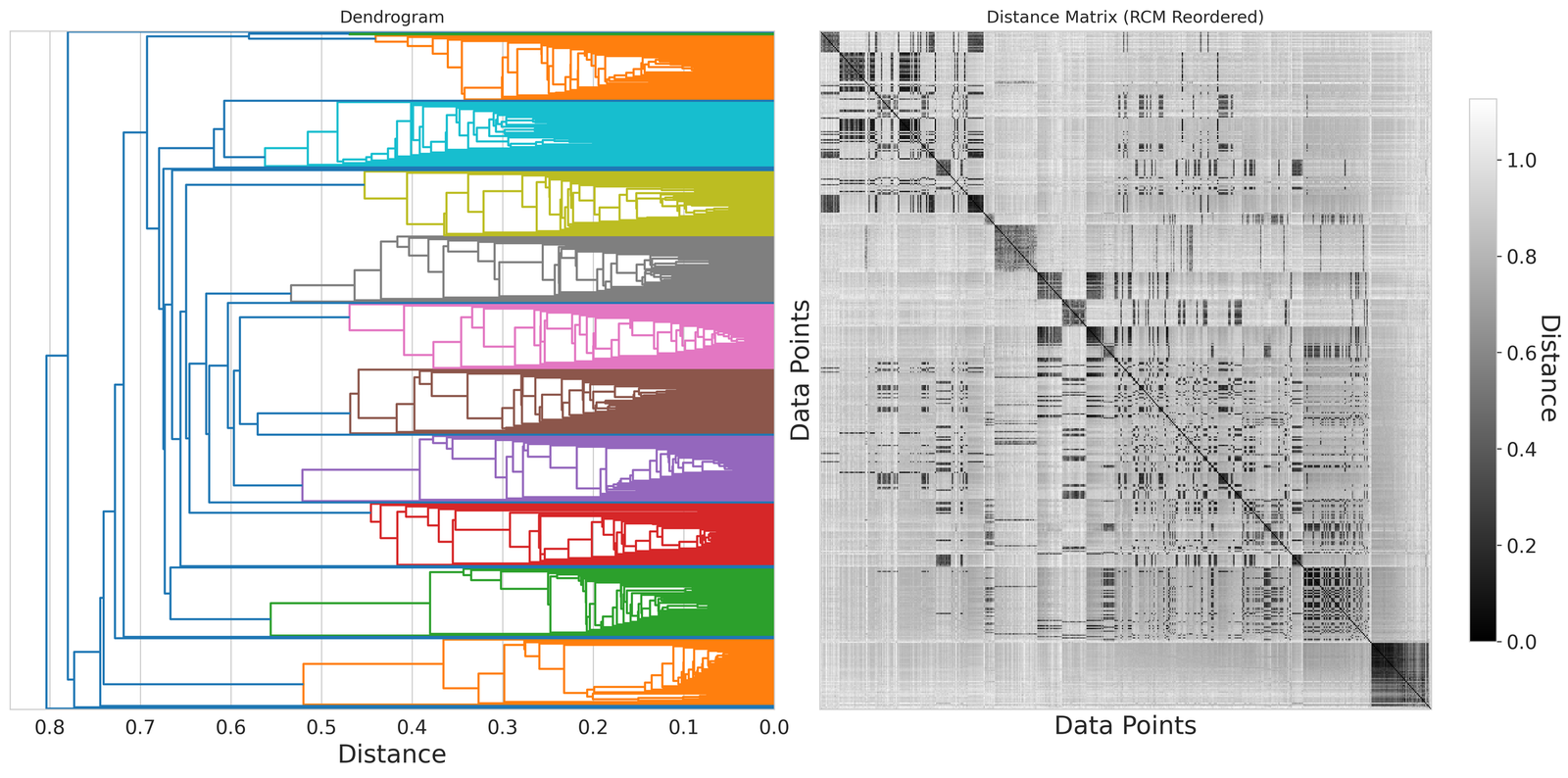} &
      \includegraphics[width=0.29\textwidth]{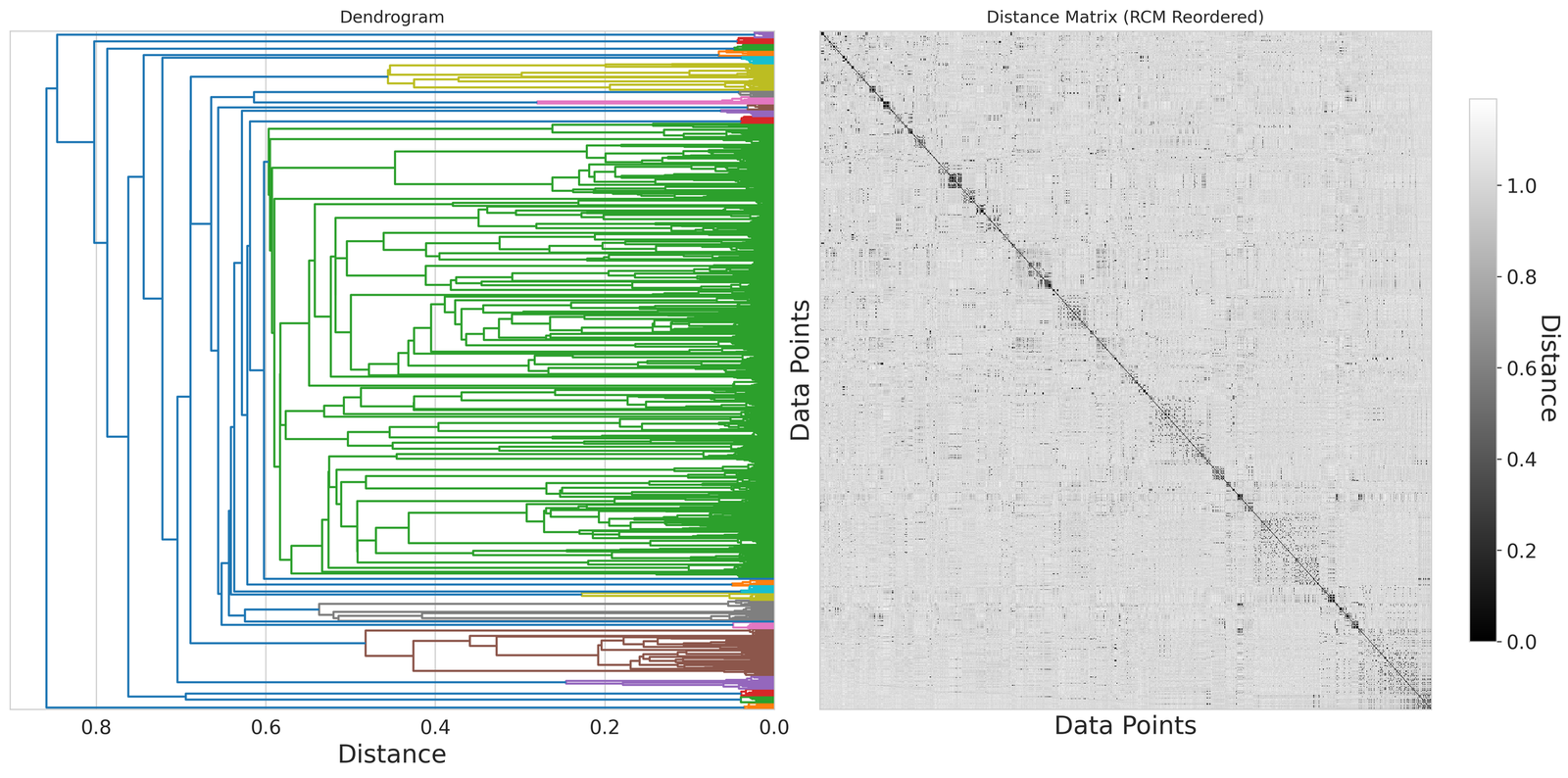} &
      \includegraphics[width=0.29\textwidth]{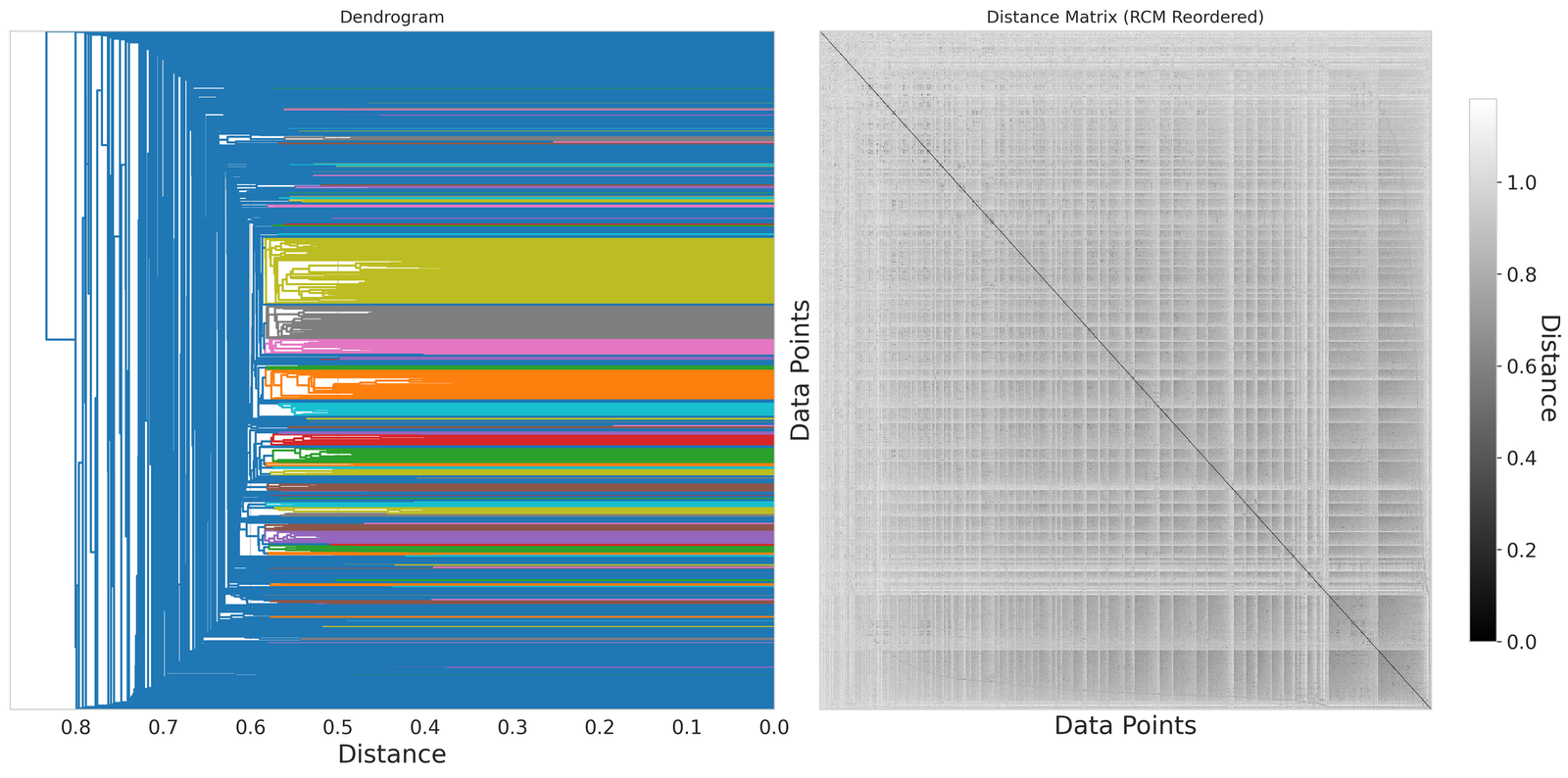} \\
  \end{tabular}
  \caption{Visual scalability of the coordinated views (ViT-B/16 layer~11, unfiltered) as the number of classes grows from $10$ to $1000$. The \vcflow{cluster flow} (top) and \vblob{blob graph} (middle) remain legible because their complexity tracks the number of classes and clusters rather than the network size; the \vhmd{heatmap dendrogram} (bottom) becomes visually occluded at the largest scale, consistent with its quadratic visual cost in the number of points and its role as a complementary rather than primary view.}
  \label{fig:visual_scalability}
\end{figure*}

\paragraph{Visual Scalability}
The coordinated views render \emph{clusters and classes}, not neurons or activations, so their visual complexity is bounded by the semantic content rather than by the network. This is clearest along the width axis: as the architecture grows from DeiT-Tiny ($d = 192$) to EVA-Giant ($d = 1408$), the number of final-layer components the \vcflow{cluster flow} and \vblob{blob} views draw does not grow (\cref{tab:scale_width}), so a wider or deeper modern network does not make the visualizations busier, and per-layer Sankey rendering stays at a few seconds. Along the classes axis the visual complexity scales with the number of classes, as intended: final-layer components grow from $24$ (Imagenette) to $947$ (ImageNet-1k, \cref{tab:scale_classes}), tracking the class count a user reasons about, and the aggregate views remain interpretable at the $1000$-class scale. The dimension that does stress the views is the probe-set size: larger $N$ produces more transient singleton components (\cref{tab:scale_samples}), which we suppress with minimum-size flow filtering (\appRef{sec:sankey_filtering}) so that the dominant class-level flows stay legible. The \vhmd{heatmap dendrogram} is the single view whose visual cost is intrinsically quadratic in $N$; beyond roughly $500$ points it becomes occluded, which is why we treat it as complementary to the aggregate views and recommend coreset approximations for larger probe sets.
\Cref{fig:visual_scalability} shows all three view types at $10$, $100$, and $1000$ classes: the \vcflow{cluster flow} and \vblob{blob graph} stay legible as the class count grows, whereas the \vhmd{heatmap dendrogram} occludes at the largest scale.

Together, these measurements bound HOLE's scalability from both sides. Computationally, cost is dominated by the probe-set size, with width linear and sub-dominant and depth embarrassingly parallel. Visually, complexity is governed by the number of classes and the probe-set size, both small and user-controlled, rather than by the size of the network. HOLE therefore remains tractable and legible for modern networks up to at least $1408$ activation dimensions and $1000$ classes, provided the probe set is kept to the small, fixed size that the stability analysis (\appRef{sec:stability_analysis}) shows is sufficient.

\section{Quantitative Cluster Quality Across Applications}
\label{sec:quantitative_cluster_quality}

The main text summarizes representation quality with the purity and homogeneity of the $H_0$ clustering (\cref{sec:cluster_quality_main}). Here we report the full per-layer values behind those summaries, evaluated at the class-optimal filtration scale selected by the combined score and averaged over three probe-set seeds.

\Cref{tab:quant_representation} gives the learned-representation analysis of Application~1 (clean ViT-B/16, CIFAR-10). Purity is uniformly high, because even a heavily fragmented clustering is locally pure; the discriminating quantity is homogeneity, which stays near its floor ($k/N \approx 0.05$) while classes remain scattered across many components in the early and middle layers, then rises sharply to $0.77$ at layer~10 and $0.87$ at layer~11 as the components collapse from $199$ to $33$. This is the quantitative signature of class-discriminative structure emerging in the deepest layers, and it illustrates why the two metrics are reported together: purity alone would suggest good structure at every layer.

\begin{table}[!htb]
  \centering
  \caption{Application 1 (learned representation): per-layer cluster quality for clean ViT-B/16 on CIFAR-10 (cosine, class-optimal scale, mean over 3 seeds). Homogeneity, not purity, tracks the emergence of class structure across depth.}
  \label{tab:quant_representation}
  \small
  \begin{tabular}{rrrr}
    \toprule
    \textbf{Layer} & \textbf{Purity} & \textbf{Homogeneity} & \textbf{\#Clusters} \\
    \midrule
    0  & 1.00 & 0.05 & 199 \\
    1  & 1.00 & 0.05 & 199 \\
    2  & 1.00 & 0.06 & 198 \\
    3  & 1.00 & 0.06 & 198 \\
    4  & 1.00 & 0.06 & 199 \\
    5  & 1.00 & 0.07 & 194 \\
    6  & 1.00 & 0.06 & 197 \\
    7  & 0.97 & 0.14 & 180 \\
    8  & 0.98 & 0.13 & 181 \\
    9  & 0.96 & 0.23 & 159 \\
    10 & 0.95 & 0.77 & 52  \\
    11 & 0.96 & 0.87 & 33  \\
    \bottomrule
  \end{tabular}
\end{table}

\Cref{tab:quant_compression} gives the compression analysis of Application~3, layer by layer, for the FP32 baseline and its INT8 quantization. In the early and middle layers the two are indistinguishable; in the late layers where class structure forms, INT8 lowers the homogeneity, e.g., from $0.71$ to $0.61$ at layer~10 and from $0.85$ to $0.80$ at layer~11, while purity stays high and top-1 accuracy is nearly unchanged. The metric therefore exposes a fragmentation of the class clusters under quantization that accuracy alone does not register.

\begin{table}[!htb]
  \centering
  \caption{Application 3 (compression): per-layer cluster quality, FP32 versus INT8 ViT-B/16 on CIFAR-10 (cosine, class-optimal scale, mean over 3 seeds). Early layers ($0$--$6$) are indistinguishable (homogeneity $\approx 0.05$--$0.07$ for both) and are omitted; INT8 lowers late-layer homogeneity while purity and accuracy are nearly unchanged.}
  \label{tab:quant_compression}
  \small
  \begin{tabular}{rrrrr}
    \toprule
    & \multicolumn{2}{c}{\textbf{FP32}} & \multicolumn{2}{c}{\textbf{INT8}} \\
    \cmidrule(lr){2-3}\cmidrule(lr){4-5}
    \textbf{Layer} & \textbf{Purity} & \textbf{Homog.} & \textbf{Purity} & \textbf{Homog.} \\
    \midrule
    7  & 0.99 & 0.10 & 0.99 & 0.08 \\
    8  & 0.98 & 0.13 & 0.99 & 0.09 \\
    9  & 0.96 & 0.24 & 0.98 & 0.16 \\
    10 & 0.98 & 0.71 & 0.95 & 0.61 \\
    11 & 0.98 & 0.85 & 0.98 & 0.80 \\
    \bottomrule
  \end{tabular}
\end{table}

\end{document}